\documentclass{opendatalab}
\usepackage{xcolor}
\usepackage{xspace} 
\usepackage{longtable}
\usepackage{tabularx}
\usepackage{array}  % 可选，但推荐
\usepackage{listings}
\usepackage[numbers]{natbib}
\usetikzlibrary{mindmap, trees, shadows}   
\usepackage{tikz}
\usepackage[edges]{forest}
\usepackage[svgnames]{xcolor}
\usepackage{tcolorbox}
\usepackage{graphicx}
\usetikzlibrary{arrows.meta}
\useforestlibrary{edges}
\usepackage{mathtools}
\usepackage[table]{xcolor}
\usepackage[edges]{forest}  
\usepackage{float}
\usepackage{amssymb}
\usepackage{booktabs}
\usepackage{adjustbox}

\usepackage{longtable}  % 用于支持 \begin{longtable}
\usepackage{multirow}   % 用于支持 \multirow
\usepackage{booktabs}   % 用于支持 \toprule, \midrule, \cmidrule
\usepackage{float}      % 用于支持图片位置参数 [H]
\usepackage{enumitem}   % 用于支持列表定制 [label=(\arabic*)...]
\usepackage{graphicx}   % 用于支持 \includegraphics
\usepackage{array}
\usepackage{amsmath}   % 用于 \boldsymbol, \text 等
\definecolor{hidden-red}{RGB}{205, 44, 36}
\definecolor{hidden-blue}{RGB}{194,232,247}
\definecolor{hidden-orange}{RGB}{243,202,120}
\definecolor{hidden-green}{RGB}{34,139,34}
\definecolor{hidden-pink}{RGB}{255,245,247}
\definecolor{hidden-black}{RGB}{20,68,106}
\definecolor{purple}{RGB}{144,153,196}
\definecolor{yellow}{RGB}{255,228,123}
\definecolor{hidden-yellow}{RGB}{255,248,203}
\definecolor{tkcolor}{RGB}{224,223,255}
\definecolor{darkblue}{rgb}{0, 0.40, 0.75}

\newcommand{\eg}{\textit{e.g.,~}}
\newcommand{\ie}{\textit{i.e.,~}}
\newcommand{\github}{\raisebox{-1.5pt}[0pt][0pt]{\includegraphics[height=1.0em]{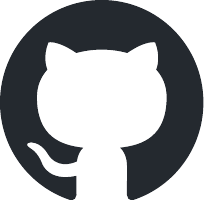}}\xspace}
\newcommand{\web}{\raisebox{-1.5pt}[0pt][0pt]{\includegraphics[height=1.0em]{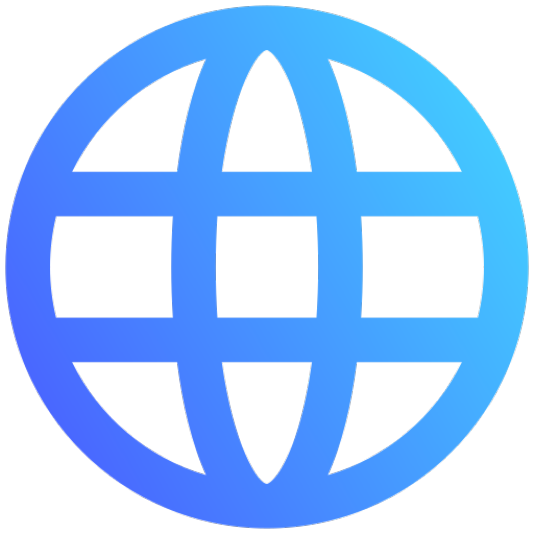}}\xspace}
\newcommand{\huggingface}{\raisebox{-1.5pt}[0pt][0pt]{\includegraphics[height=1.0em]{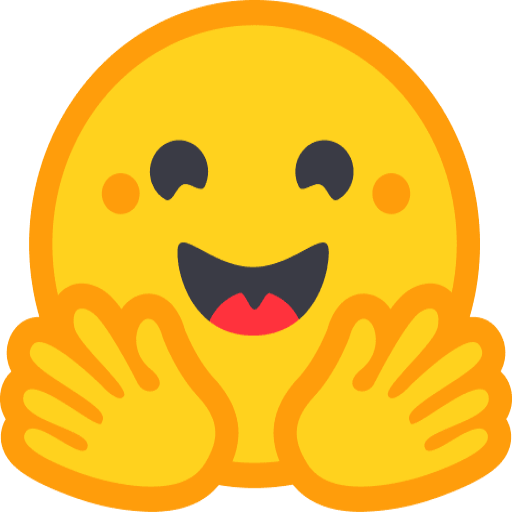}}\xspace}

\allowdisplaybreaks
\usetikzlibrary{mindmap, trees, shadows}
\definecolor{codegreen}{rgb}{0,0.6,0}
\definecolor{codegray}{rgb}{0.5,0.5,0.5}
\definecolor{codepurple}{rgb}{0.58,0,0.82}
\definecolor{backcolour}{rgb}{0.95,0.95,0.92}
\definecolor{promptcolor}{HTML}{D1D0F2}
\definecolor{promptcolorheader}{HTML}{bdbcec}
\newcommand{\promptbox}[2]{
\begin{tcolorbox}[
top=0.3em,bottom=0.3em,left=0.5em,right=0.5em,
toptitle=0.3em,bottomtitle=0.2em,boxsep=0pt,
colframe=promptcolorheader,colback=promptcolor!50,boxrule=0.5pt,
]
\footnotesize
\end{tcolorbox}
}
\lstdefinestyle{mystyle}{
    backgroundcolor=\color{backcolour},   
    commentstyle=\color{codegreen},
    keywordstyle=\color{magenta},
    numberstyle=\tiny\color{codegray},
    stringstyle=\color{codepurple},
    basicstyle=\ttfamily\footnotesize,
    breakatwhitespace=false,         
    breaklines=true,                 
    captionpos=b,                    
    keepspaces=true,                 
    numbers=left,                    
    numbersep=5pt,                  
    showspaces=false,                
    showstringspaces=false,
    showtabs=false,                  
    tabsize=2
}

\definecolor{main}{HTML}{4472C4} 
\definecolor{sub}{HTML}{EBF4FF}
\newtcolorbox{boxA}{
  enhanced, breakable,
  boxrule = 0pt,
  colback = sub,
  borderline west = {2pt}{0pt}{main}, 
  borderline east = {2pt}{0pt}{main}, 
}

\definecolor{promptcolor}{HTML}{D1D0F2}
\definecolor{promptcolorheader}{HTML}{bdbcec}

\newtcolorbox{promptboxfig}[1][]{
  enhanced,
  breakable=false, % IMPORTANT: avoid using breakable boxes inside figure scaling boxes
  top=0.3em,bottom=0.3em,left=0.5em,right=0.5em,
  toptitle=0.3em,bottomtitle=0.2em,boxsep=0pt,
  colframe=promptcolorheader, colback=promptcolor!50, boxrule=0.5pt,
  width=\linewidth,      % fit nicely in figure*
  title={\footnotesize #1}
}

\title{The Trinity of Consistency as a Defining Principle for General World Models}

\contribution{
   \centering Full author list in Contributions
}

\abstract{
The construction of \textit{World Models} capable of learning, simulating, and reasoning about objective physical laws constitutes a foundational challenge in the pursuit of Artificial General Intelligence. Recent advancements represented by video generation models like Sora have demonstrated the potential of data-driven scaling laws to approximate physical dynamics, while the emerging \textit{Unified Multimodal Model (UMM)} offers a promising architectural paradigm for integrating perception, language, and reasoning. Despite these advances, the field still lacks a principled theoretical framework that defines the essential properties requisite for a \textit{General World Model}. In this paper, we propose that a World Model must be grounded in the  \textit{Trinity of Consistency}: \textit{Modal Consistency} as the semantic interface, \textit{Spatial Consistency} as the geometric basis, and \textit{Temporal Consistency} as the causal engine. Through this tripartite lens, we systematically review the evolution of multimodal learning, revealing a trajectory from loosely coupled specialized modules toward unified architectures that enable the synergistic emergence of internal world simulators. To complement this conceptual framework, we introduce CoW-Bench, a benchmark centered on multi-frame reasoning and generation scenarios. CoW-Bench evaluates both video generation models and UMMs under a unified evaluation protocol. Our work establishes a principled pathway toward general world models, clarifying both the limitations of current systems and the architectural requirements for future progress.
}

\metadatainfo{
    \small\makebox[\linewidth][c]{
        \github~\href{https://github.com/openraiser/awesome-world-model-evolution}{\textbf{Code}} \quad
        \web~\href{https://openraiser.github.io/CoW-Bench/}{\textbf{Leaderboard}} \quad \huggingface~\href{https://huggingface.co/datasets/openraiser/CoW-Bench}{\textbf{Dataset}}
    }
}

\begin{document}

\maketitle

\begin{figure}[H] % 使用大写 H 强制放在此处
 \centering
 \vspace{-2mm}
 \includegraphics[width=0.98\linewidth]{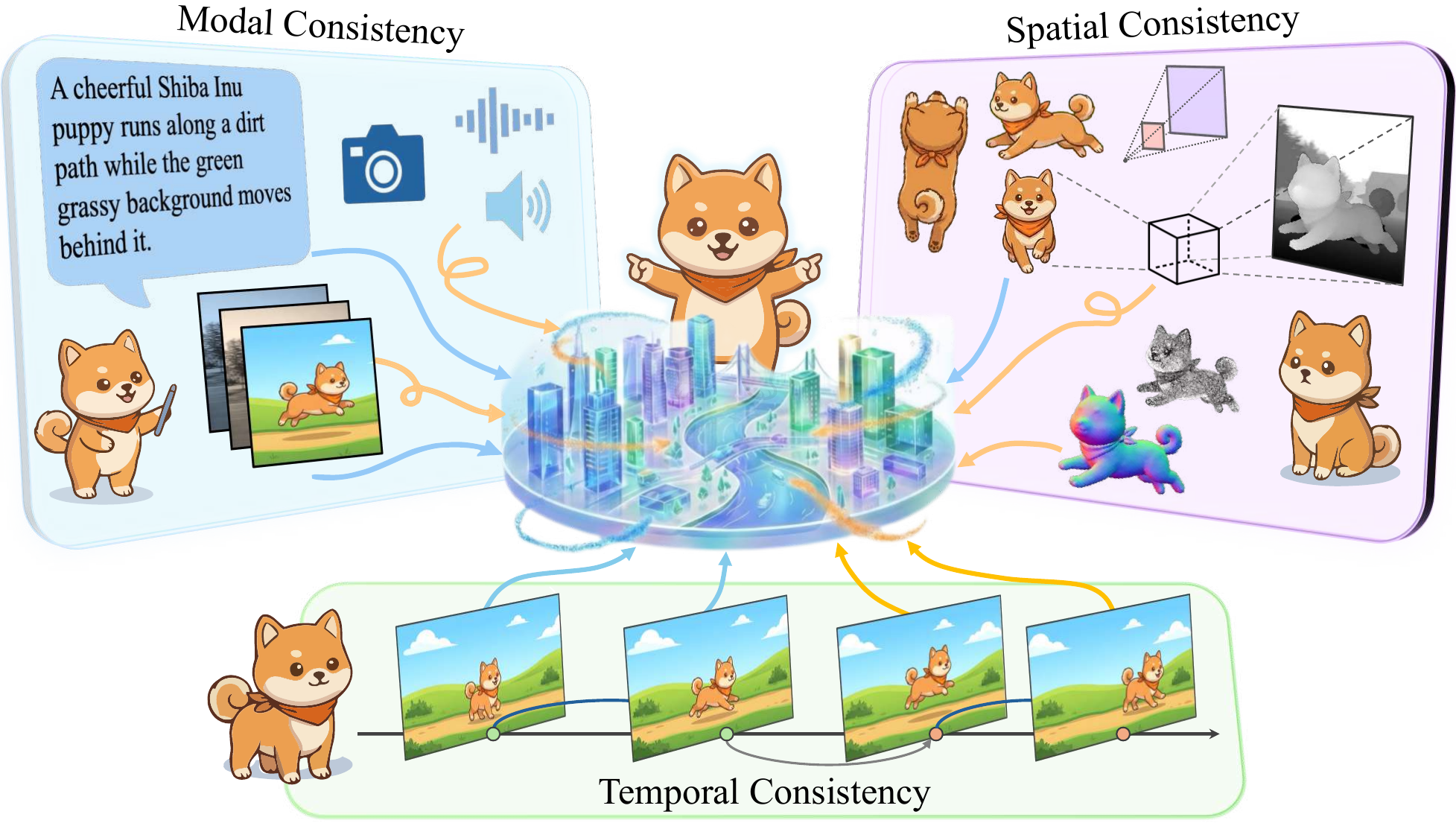}
 \caption{The \textit{Trinity of Consistency} in world models: \textit{Modal Consistency} (Semantics), \textit{Spatial Consistency} (Geometry), and \textit{Temporal Consistency} (Causality). }
 \vspace{-4mm}
\end{figure}

\clearpage
\tableofcontents
\clearpage

\section{Introduction}
\label{section:intro}

The pursuit of Artificial General Intelligence (AGI) is fundamentally anchored in the aspiration to endow machines with a profound understanding of the physical reality. A truly intelligent agent must evolve from a passive observer~\cite{lecun2022path} into a proactive simulator~\cite{openai2024sora, bardes2024vjepa}, possessing an internal world model capable of learning objective physical laws, reasoning about counterfactual scenarios~\cite{lu2024deepseek}, and predicting future states from current actions~\cite{lingbotworld2026}.

Recent years have witnessed an explosion in generative capability, driven by the data-driven Scaling Laws. Video generation models, represented by Sora~\cite{openai2024sora} and Gen-3~\cite{runway2024gen3}, have demonstrated an astonishing ability to approximate complex dynamics, creating high-fidelity visual sequences that often are indistinguishable from reality. Simultaneously, the rise of Unified Multimodal Models (UMMs)~\cite{team2023gemini, deng2025emerging} has offered a promising architectural paradigm for integrating diverse sensory inputs into a shared semantic manifold~\cite{min2024platonic}.
However, a critical gap remains: existing models, despite their visual plausibility, often behave as naive physicists. They frequently suffer from structural hallucinations, temporal inconsistencies, and violations of causality—symptoms of a system that mimics pixel statistics rather than internalizing physical principles. The field lacks a principled theoretical framework to define the essential properties requisite for a \textit{General World Model}.

To bridge the chasm between visual generation and physical simulation, we propose that a robust World Model must be grounded in the \textbf{\textit{Trinity of Consistency}}. We argue that a valid internal simulator must satisfy three orthogonal yet synergistic constraints:
\begin{itemize}
    \item \textbf{Modal Consistency (The Semantic Interface):} The ability to align heterogeneous information (text, image, tactile) into a unified semantic space, serving as the cognitive interface for instruction and feedback.
    \item \textbf{Spatial Consistency (The Geometric Basis):} The capacity to construct a 3D-aware representation that respects geometry, occlusion, and object permanence, ensuring the static plausibility of the simulated world.
    \item \textbf{Temporal Consistency (The Causal Engine):} The adherence to physical laws and causal logic over time, ensuring that dynamic evolution follows a predictable and logically sound trajectory.
\end{itemize}

Through this tripartite lens, we systematically review the evolution of generative models from specialized modules to unified world simulators. We trace the trajectory from loosely coupled specialized modules toward end-to-end unified architectures. We argue that dissolving the barriers between these dimensions is the necessary substrate for the emergence of world simulation capabilities, ensuring that modality, space, and time do not operate in isolation but synergize to model a coherent reality.

This paper is organized to mirror the evolutionary path from specialized modules to unified world simulators. \textbf{First} (\S \ref{section:independent_exploration}), we deconstruct the independent development of Modal, Spatial, and Temporal consistencies, analyzing their respective theoretical foundations. \textbf{Second} (\S \ref{section:integration}), we investigate the paradigm shift enabled by UMMs, detailing how the deep integration of these dimensions facilitates the emergence of physical simulation capabilities. \textbf{Third} (\S \ref{section:challenges}), we identify the remaining gaps between current probabilistic generators and true physical simulators, setting the stage for rigorous evaluation. The notation used is summarized in Table~\ref{tab:notation_optimized}.

Finally, theoretical frameworks require rigorous verification. We introduce \textbf{CoW-Bench (Consistency of World-models Benchmark)}, a unified evaluation suite centered on multi-frame reasoning and constraint satisfaction. Unlike previous benchmarks, CoW-Bench rigorously tests the model's ability to maintain the \textit{Trinity of Consistency} under complex, open-ended scenarios, forcing it to prove it understands the world, not just how to paint it.

\begin{figure}[ht]
  \centering
  \vspace{-6mm}
  \includegraphics[width=0.96\linewidth]{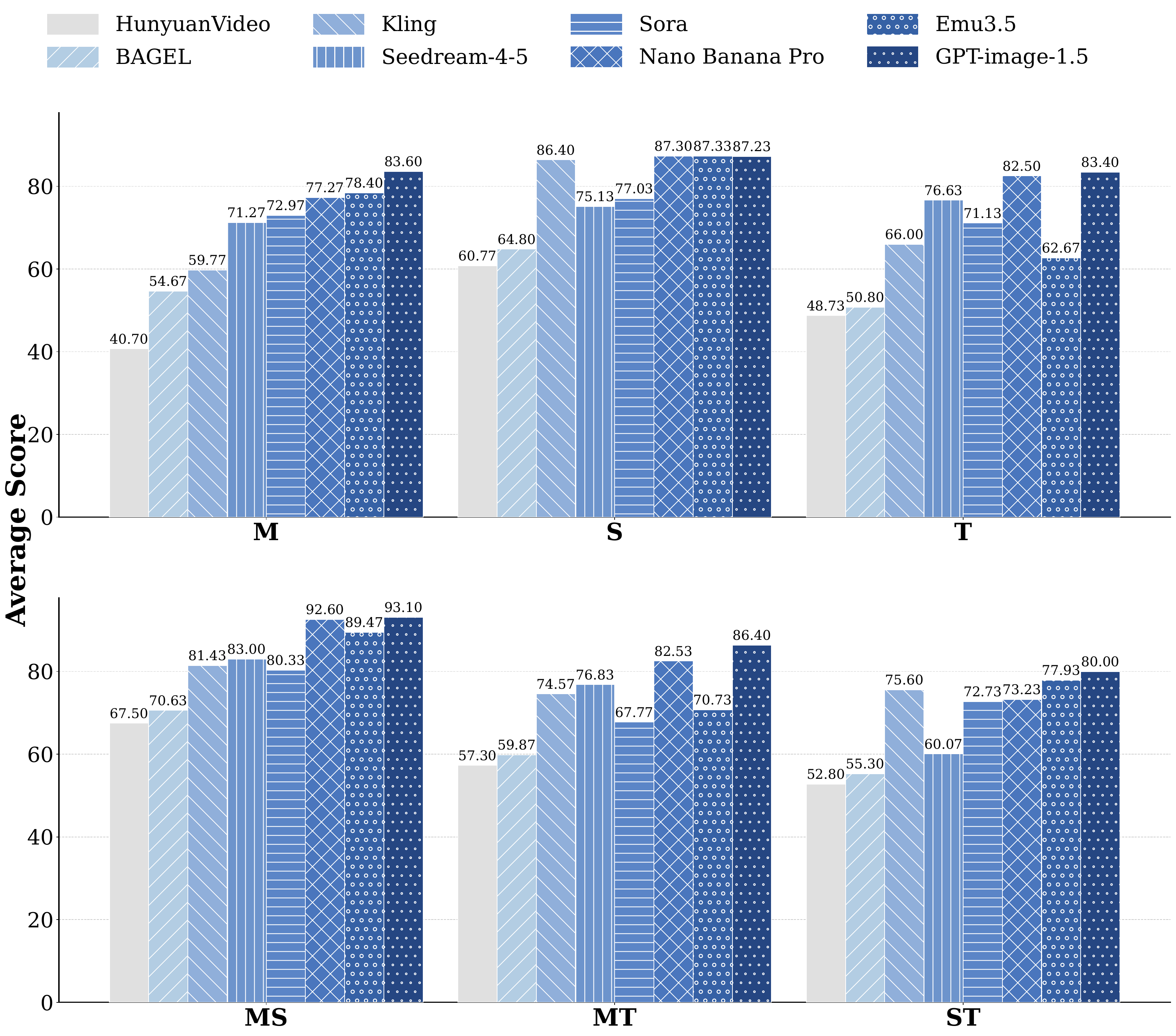}
  \vspace{-2mm}
  \caption{Performance Comparison of Mainstream Models across Different Tasks. The score has been linearly rescaled from the original range of [0, 10] to a percentage scale of [0, 100].}
  \vspace{-2mm}
  \label{fig:bar_comparison}
\end{figure}

\begin{table}[t]
\centering
\caption{Notation and Descriptions}
\label{tab:notation_optimized}
\small 
\renewcommand{\arraystretch}{1.25}
\setlength{\tabcolsep}{3pt}

\begin{tabularx}{\linewidth}{lX|lX}
\toprule
\textbf{Symbol} & \textbf{Description} & \textbf{Symbol} & \textbf{Description} \\
\midrule
% --- Core Space ---
$\mathcal{W}$ & World Model & $\boldsymbol{p}$ & 3D Position \\
$\mathcal{S}, \mathcal{A}$ & State \& Action Space & $\boldsymbol{P}_t$ & Camera Pose at $t$ \\
$\boldsymbol{s}_t, \boldsymbol{a}_t$ & State \& Action Instance & $\boldsymbol{K}$ & Intrinsic Matrix \\
$\pi$ & Policy & $\Pi$ & Projection Operator \\
$\tau$ & Trajectory & $\mathcal{K}$ & Keyframe Set \\
$\mathcal{T}$ & Dynamics Function & $\mathcal{M}_{geo}$ & Geometric Manifold \\
% --- Representation ---
$\mathcal{Z}$ & Latent World State & $\mathcal{G}_k$ & 3D Gaussian Primitive \\
$\boldsymbol{x}_{obs}$ & Multimodal Observation & $\sigma$ & Volume Density \\
$\boldsymbol{z}$ & Latent Vector & $\boldsymbol{c}$ & View-dependent Radiance \\
$\mathcal{E}, \mathcal{D}$ & Encoder / Decoder & $\boldsymbol{F}_{fund}$ & Fundamental Matrix \\
$\mathcal{C}$ & VQ Codebook & $\mathcal{O}_{flow}$ & Optical Flow \\
$S$ & Token Sequence & $\mathcal{M}_{epi}$ & Epipolar Mask \\
$W_{proj}$ & Projection Weight & $\boldsymbol{T}(t)$ & Continuous Trajectory \\
$I(X; Z)$ & Mutual Information & $\Phi$ & Spatiotemporal Field \\
% --- Physics & Dynamics ---
$\epsilon_\theta$ & Noise Predictor & $\Psi$ & Physical Property Field \\
$\boldsymbol{v}_t$ & Velocity Field & $D \Phi / D t$ & Material Derivative \\
$g(t)$ & Diffusion Coefficient & $\nabla \cdot \boldsymbol{v}$ & Divergence \\
$\alpha_t, \sigma_t$ & SNR Parameters & $\boldsymbol{F}$ & Force Vector \\
$\boldsymbol{w}$ & Wiener Process & $\nabla f$ & Implicit Gradient \\
$\mathcal{F}_{t}$ & STFT (Fourier Transform) & $\mathcal{M}_{dyn}$ & Dynamic Manifold \\
% --- Metrics ---
$\mathcal{L}$ & Loss Function & $\text{Phys}$ & Physics Score \\
$\mathcal{G}_{raph}$ & Causal Graph & $\Delta_{const}$ & Constraint Deviation \\
$D_{KL}$ & KL Divergence & $w$ & Guidance Scale \\
\bottomrule
\end{tabularx}
\end{table}

\section{Foundational Exploration of Consistencies}
\label{section:independent_exploration}

\subsection{The Anatomy of General World Models}

As discussed in Section 1 (\S\ref{section:intro}), the construction of world models relies on the organic integration of modal consistency (serving as the information interface), spatial consistency (serving as the geometric cornerstone), and temporal consistency (serving as the dynamic engine). In the evolution of specialized models, these consistencies have not developed in isolation but have rather interpenetrated one another: the unified representation space derived from modality alignment provides semantic priors for the reconstruction of spatial geometry, while the 3D manifold of spatial consistency establishes physical constraints for temporal evolution. 

% This section will begin with modal consistency, progressively elaborating on the theoretical foundations, mechanism analysis, and architectural evolution of these three major consistencies. Furthermore, it will elucidate their synergistic mechanisms through cross-referencing and finally explore their potential for unification within world models.
This section deconstructs that evolutionary history. We trace how specialized models first conquered these challenges in isolation: modality alignment matured through high-dimensional manifold mapping, spatial consistency was solved via the transition from 2D proxies to explicit 3D primitives, and temporal consistency evolved from simple frame interpolation to causal dynamics modeling. Here, we systematically analyze the theoretical foundations and mechanism shifts of each dimension, establishing the necessary prerequisites that eventually enabled the emergence of the unified world simulators discussed in later sections.

\subsection{Modal Consistency}
\label{sec:modality_consistency}

\begin{figure}[h]
  \centering
  \includegraphics[width=0.9\linewidth]{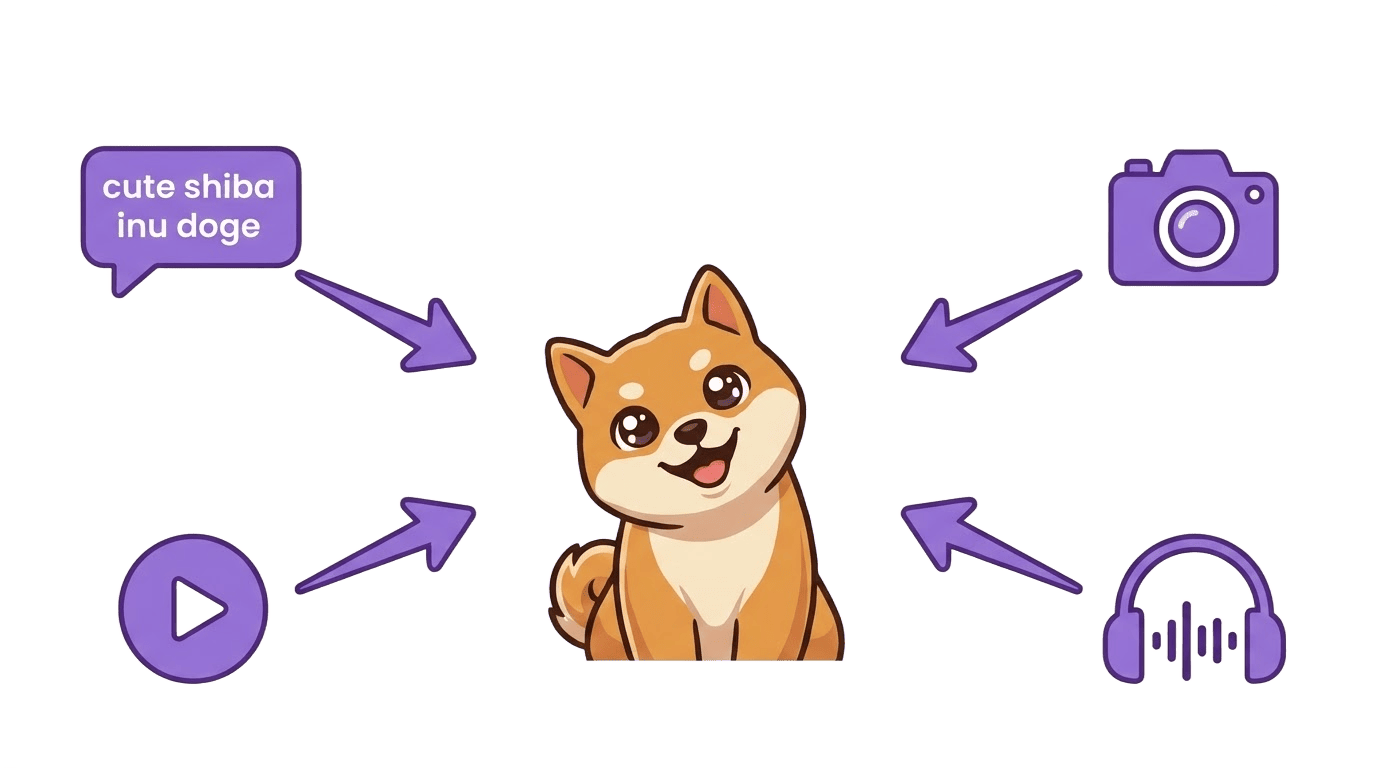}
  \caption{Unified Representation Goal. Modal consistency aims to project heterogeneous inputs (Text, Image, Video, Audio) into a unified, physically-aligned latent space.}
  \label{fig:modality_consistency_concept}
\end{figure}

The core challenge in constructing general world models lies in the semantic alignment of heterogeneous modalities. Unlike the homogeneity of unimodal generation, multimodal consistency is essentially a problem of solving high-dimensional heterogeneous manifold alignment, as illustrated in Figure~\ref{fig:modality_consistency_concept}. The model must transcend entropy disparity and topological mismatch to construct a unified representation space that is physically complete and logically self-consistent. To this end, we introduce two fundamental theoretical assumptions, the Platonic Representation Hypothesis and the Hypersphere Geometry Hypothesis, and use these as a basis to expound on the cognitive architectural evolution from direct feed-forward mapping to iterative reasoning and planning.

To systematically deconstruct this alignment process, this section will first elucidate the origins of the modality gap from the perspective of geometric topology (\S\ref{subsec:theoretical_foundations}); subsequently, it will analyze two mainstream generative manifold mechanisms—namely, discrete autoregression and continuous flow matching (\S\ref{subsec:mechanism_analysis}); it will then explore the orthogonal decoupled architecture evolved to minimize gradient conflicts (\S\ref{subsec:arch_evolution}); and finally, it will introduce feedback-based intent alignment and the cognitive inference loop moving towards test-time compute (\S\ref{subsec:cognitive_loop}).

% --- 1. 颜色定义 (完全复用您的参考) ---
\definecolor{hidden-blue}{HTML}{4a90e2}
\definecolor{hidden-black}{HTML}{333333}
% --- 2. 节点样式定义 (完全复用您的参考) ---
\tikzstyle{my-box}=[
  rectangle,
  draw=hidden-black,
  rounded corners,
  text opacity=1,
  minimum height=1.5em,
  minimum width=5em,
  inner sep=2pt,
  align=center,
  fill opacity=.5,
]
\tikzstyle{leaf}=[
  my-box, 
  minimum height=1.5em,
  fill=yellow!32, 
  text=black,
  align=left,
  font=\normalsize,
  inner xsep=5pt,
  inner ysep=4pt,
  text width=20em,   
]
\tikzstyle{leaf2}=[
  my-box, 
  minimum height=1.5em,
  fill=purple!27, 
  text=black,
  align=left,
  font=\normalsize,
  inner xsep=5pt,
  inner ysep=4pt,
  text width=20em,
]
\tikzstyle{leaf3}=[
  my-box, 
  minimum height=1.5em,
  fill=hidden-blue!57, 
  text=black,        
  align=left,
  font=\normalsize,
  inner xsep=5pt,
  inner ysep=4pt,
  text width=20em,
]
\begin{figure*}[t]
\vspace{-2mm}
\centering
\resizebox{\textwidth}{!}{
\begin{forest}
  % --- 核心配置：完全模仿您的参考格式 ---
  forked edges, % 关键：直角折线
  ver/.style={rotate=90, child anchor=north, parent anchor=south, anchor=center},
  for tree={
    grow=east,               % 向右生长
    reversed=true,           % 倒序排列
    anchor=base west,        % 锚点左对齐
    parent anchor=east,      % 父节点出线口
    child anchor=west,       % 子节点入线口
    base=left,               % 文字基线左对齐
    font=\large,
    rectangle,
    draw=hidden-black,
    rounded corners,
    align=left,
    minimum width=4em,
    edge+={darkgray, line width=1pt},
    s sep=3pt,               % 兄弟间距
    inner xsep=2pt,
    inner ysep=4pt,
    line width=1.1pt,
  },
  % 层级宽度控制 (确保对齐)
  where level=1{text width=11em,font=\normalsize}{},
  where level=2{text width=13em,font=\normalsize}{},
  where level=3{text width=30em}{},
  % ============================================================
  % 树结构内容：模态一致性演进 (内容已更新为 2026 版本)
  % ============================================================
[Evolution of Modal Consistency\\in Unified World Models, ver, fill=gray!70, text=white
    % --- Phase 1: 几何隔离与连接器 (橙) ---
    [Geometric Isolation\\\& Connector Paradigm\\, fill=orange!20
      [Dual-Tower Contrastive, leaf
        [{ CLIP~\cite{radford2021icml}, ALIGN~\cite{jia2021align},\\SigLIP~\cite{zhai2023sigmoid}, MetaCLIP~\cite{xu2023metaclip}, etc.}]
      ]
      [Connector-Based Alignment, leaf
        [{ Flamingo~\cite{alayrac2022flamingo}, BLIP/BLIP-2~\cite{li2022blip, li2023blip2},\\Qwen-VL~\cite{bai2023qwen}, LLaVA-NeXT~\cite{liu2024llavanext}, etc.}]
      ]
    ]
    % --- Phase 2: 早融合与优化挑战 (绿) ---
    [Early Fusion \&\\Unified Optimization\\ , fill=green!20
      [Discrete Unified Interface, leaf
        [{ Chameleon~\cite{team2024chameleon}, Unified-IO 2~\cite{lu2023unifiedio2},\\Show-o~\cite{xie2024showo}, CM3leon~\cite{yu2023cm3leon}, etc.}]
      ]
      [Asymmetric Projection, leaf
        [{ LLaVA~\cite{liu2023llava}, CogVLM~\cite{wang2023cogvlm},\\InternVL~\cite{chen2023internvl}, MiniGPT-4~\cite{zhu2023minigpt4}, etc.}]
      ]
    ]
    % --- Phase 3: 正交解耦 (蓝) ---
    [Orthogonal Decoupling\\(Native MM-DiT)\\, fill=hidden-blue!57, text=black
      [Weight Decoupling, leaf3
        [{ Stable Diffusion 3.5~\cite{esser2024sd3}, Emu3~\cite{wang2024emu3},\\PixArt-$\alpha$~\cite{chen2024pixart}, Lumina-Next~\cite{gao2024lumina}, etc.}]
      ]
      [Flow Matching Dynamics, leaf3
        [{ Rectified Flow~\cite{liu2023rectified}, InstaFlow~\cite{liu2023instaflow},\\Flux.1~\cite{flux2024}, SiT~\cite{ma2024sit}, etc.}]
      ]
    ]
    % --- Phase 4: 意图对齐与认知闭环 (紫) ---
    [Intent Alignment \&\\Cognitive Loop\\, fill=purple!27
      [VLM-as-a-Judge (Critic), leaf2
        [{ MetaMorph~\cite{metamorph}, SRUM~\cite{2510.12784},\\ImageReward~\cite{xu2023imagereward}, VLMScore~\cite{lin2024vlmscore}, etc.}]
      ]
      [Process Supervision (RL), leaf2
        [{ SPO~\cite{liang2025spo}, VisualPRM~\cite{wang2025visualprm},\\PhyGDPO~\cite{cai2025phygdpo}, AR-GRPO~\cite{zhang2025argrpo}, etc.}]
      ]
    ]
]
\end{forest}
}
\caption{Evolution of Modal Consistency: From Geometric Isolation to Cognitive Alignment}
\label{fig:modality-consistency-evolution}
\end{figure*}
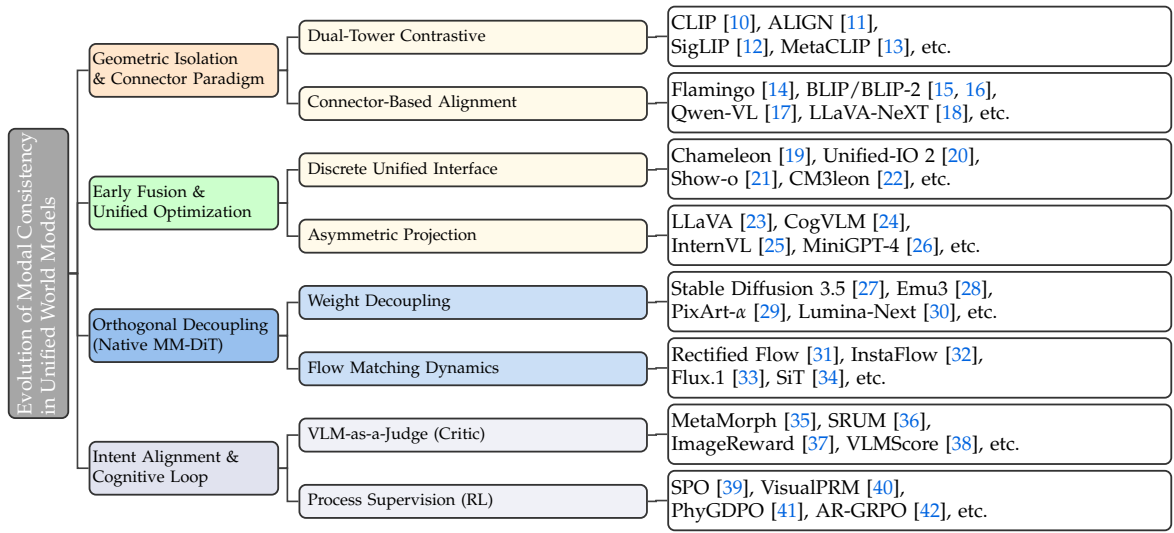

\subsubsection{Theoretical Foundations}
\label{subsec:theoretical_foundations}

\paragraph{Platonic Cave \& Projected Manifolds}
The theoretical foundation of multimodal learning can be traced back to the platonic representation hypothesis~\cite{min2024platonic}. This hypothesis formally defines the existence of an objective latent physical state space, $\mathcal{Z}_{world}$, in the real world, where images and text are projections of this high-dimensional entity onto different low-dimensional subspaces.
The essence of modal consistency is solving a joint inverse projection problem: reconstructing the shared latent variable $z$ via observed shadows $\{x_{img}, x_{txt}\}$. However, this is a typical ill-posed problem—the visual projection $\mathcal{P}_{img}$ retains a vast amount of high-frequency physical entropy, whereas the textual projection $\mathcal{P}_{txt}$ highly abstracts discrete symbolic logic. This Entropy Asymmetry constitutes the primary obstacle to direct alignment.

\paragraph{Hypersphere Hypothesis \& Modal Gap}

\begin{figure}[h]
  \centering
  \includegraphics[width=1.0\linewidth]{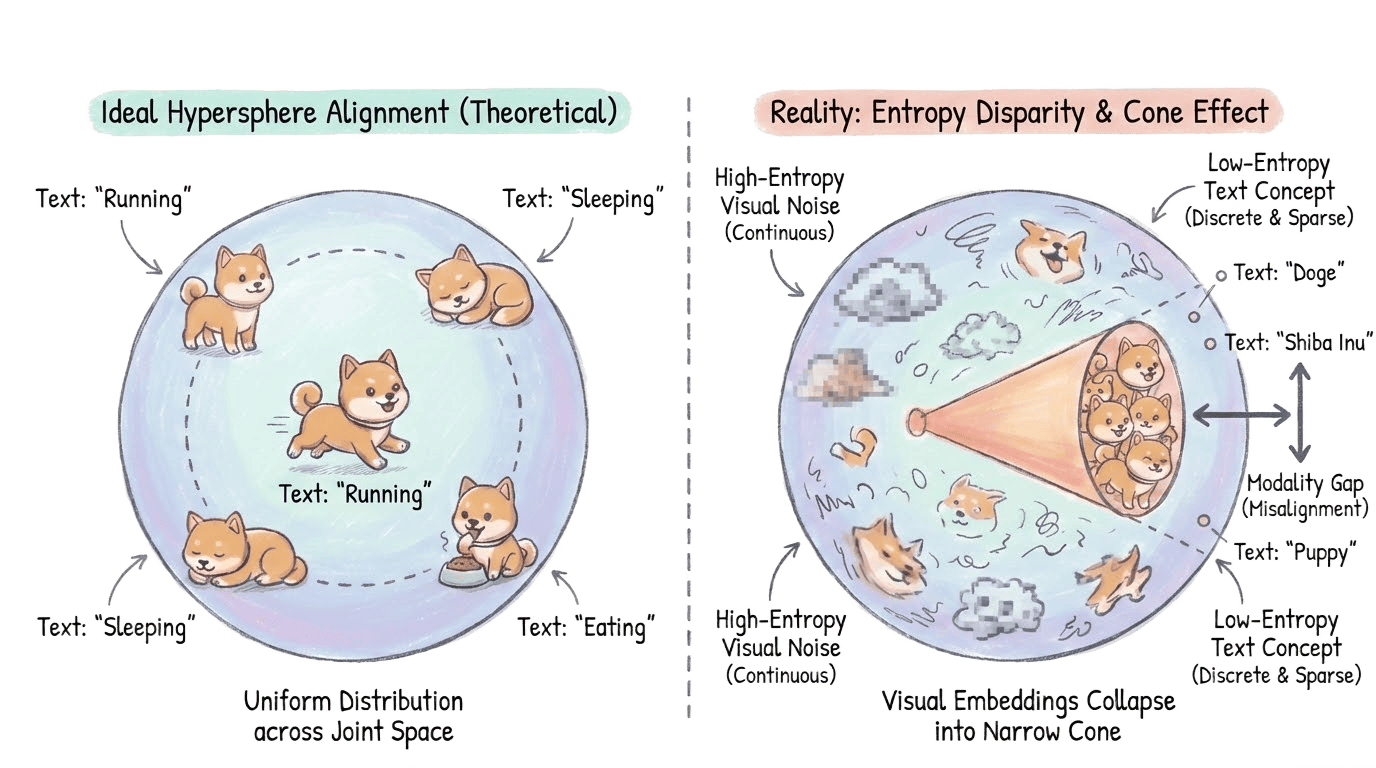}
  \caption{The Modal Gap Challenge. (Left) Ideal hypersphere alignment assumes uniform distribution. (Right) In reality, entropy disparity causes visual embeddings to collapse into a narrow "cone," leading to topological mismatch with discrete text tokens.}
  \label{fig:cone_effect}
\end{figure}

To mathematically align these two heterogeneous spaces, mainstream paradigms (such as CLIP) introduce the Hypersphere Hypothesis~\cite{wang2020understanding}, which forces feature vectors to be uniformly distributed on a unit hypersphere $\mathbb{S}^{d-1}$. However, this strong assumption ignores the pervasive modal gap in multimodal representations~\cite{liang2022neurips}.
On one hand, empirical studies by Liang et al. pointed out the cone effect, as shown in Figure \ref{fig:cone_effect}: joint optimization causes visual and textual embeddings to collapse into two narrow and separated conical regions, destroying the isotropy of the feature space. On the other hand, from the perspective of manifold learning, this gap reveals a deeper topological mismatch: visual data is typically distributed on a continuous, dense low-dimensional manifold, while linguistic data presents a sparse, discrete clustering structure. This fundamental difference in intrinsic dimensionality and data density leads to manifold non-isomorphism, rendering the achievement of perfect isometric alignment between the two spaces, while maintaining their respective semantic structures, an ill-posed problem.

\paragraph{Evolution of Computational Paradigms: From Amortized Inference to Test-time Compute}
Facing the inherent representation errors caused by the aforementioned geometric topological mismatch, simple parameter internalization strategies face theoretical bottlenecks, prompting the modeling of modal consistency to undergo a transition between two major computational paradigms. This profoundly reflects the trade-off between train-time compute and test-time compute~\cite{snell2024scaling}.

Early direct feed-forward mapping corresponds to Dual-Tower architectures~\cite{radford2021icml} and single-step generative models, the core of which is identifying physical rules into neural network weights through large-scale training, \ie Amortized Inference~\cite{gershman2014amortized}. This paradigm requires only one forward pass during inference ($\text{NFE}=1$). Although highly efficient, it is limited by in-distribution statistical correlations and essentially can only interpolate within established conical regions, making it difficult to handle unseen counterfactual combinations~\cite{bengio2021machine}.

In contrast, the current trend is shifting towards iterative reasoning \& planning, corresponding to iterative reasoning architectures. This paradigm acknowledges the limitations of single-pass mapping in bridging the modality gap and instead introduces explicit state space search during the inference phase. By constructing a Tree of Thoughts~\cite{yao2024tree} in the latent space or executing gradient-guided dynamic planning, the model utilizes additional reasoning compute to instantly correct physical drift. This marks a shift in consistency modeling from static pattern matching to dynamic manifold planning.

\subsubsection{Discrete Sequences vs. Continuous Manifolds}
\label{subsec:mechanism_analysis}

To computationally realize the Joint Inverse Projection process in the above theory, academia has explored two distinct mathematical paths to model the target conditional probability density $P(x_{img}|x_{txt})$. This choice determines the physical nature of the latent space manifold: \textit{Is it treated as a Discrete Symbolic Sequence or a Continuous Euclidean Vector Field?} We compare the mathematical forms and dynamic characteristics of these two paradigms in Table \ref{tab:ar_vs_fm_table}.

\begin{table}[h]
\centering
\caption{Mechanism Comparison: Discrete AR vs. Continuous Flow Matching. The formulations highlight the trade-off between optimization objectives and error propagation dynamics.}
\label{tab:ar_vs_fm_table}
\small 
\renewcommand{\arraystretch}{1.3}
\begin{tabular*}{\columnwidth}{@{\extracolsep{\fill}}lccc} 
\toprule
Paradigm & Objective (The Soul) & Error & Topology \\
\midrule  
Discrete AR & 
$ \displaystyle \mathcal{L}_{AR} = -\mathbb{E} \left[ \sum \log P(s_t|s_{<t}) \right] $ 
& Exp. & Discrete \\
Flow Matching & 
$ \displaystyle \mathcal{L}_{FM} = \mathbb{E} \left[ \| v_\theta(x_t) - (x_1 - x_0) \|^2 \right] $ 
& Linear & Euclidean \\
\bottomrule
\end{tabular*}
\end{table}

\paragraph {Discrete Autoregressive (AR)}
The core of this paradigm lies in the Token-centric philosophy, attempting to transform visual generation into a sequence prediction problem through a unified discrete symbol interface~\cite{van2017neural, ramesh2021zero}. Its generation process involves strictly coupled stages: first quantizing continuous images into discrete symbols via VQ-GAN, followed by maximizing the sequence log-likelihood using the causal attention mask of a Transformer.

\textit {Exponential Drift \& Codebook Collapse.}
Although the AR paradigm achieves interface unification, it suffers from two endogenous defects when viewed from a dynamic perspective~\cite{Huh2023Straightening}.
First is the curse of dimensionality. The discretization process is governed by the Dirichlet process; as the codebook dimension increases, the effective utilization rate decays exponentially, leading to the loss of high-frequency textures~\cite{iccv2025GigaTok, cvpr2025MergeVQ}.
Second is error accumulation dynamics. The essence of autoregressive generation is the recursive application of operators. Assuming the local Lipschitz constant of the operator is $L > 1$, the cumulative drift of the initial quantization error $\epsilon_0$ after $T$ steps is $\|\delta_T\| \approx L^T \|\epsilon_0\|$. This exponential error amplification explains why AR models often exhibit structural collapse at the tail end when generating long sequences~\cite{bengio2015scheduled}.

\paragraph {Continuous Flow Matching (FM)}
To circumvent quantization errors, the new generation of paradigms (such as Stable Diffusion 3~\cite{esser2024sd3}, Emu3~\cite{wang2024emu3}) returns to the continuous latent space. Unlike traditional diffusion models based on the SDE denoising perspective, Flow Matching (FM)~\cite{lipman2023flow} adopts an ODE perspective, constructing a deterministic transport path connecting noise and data.

\textit {Velocity Field Regression \& Rectified Path.}
The core idea of continuous FM is to directly fit the velocity field of the probability flow. During training, the intermediate state $x_t$ is defined as a linear interpolation between data and noise, corresponding to an ideal straight trajectory with a target velocity field constantly being $v_t = x_1 - x_0$. The neural network directly regresses this velocity vector via Mean Squared Error loss.
Rectified Flow~\cite{liu2023iclr} demonstrates that this Reflow operation rectifies the transport trajectory, corresponding to a Lipschitz constant $L \approx 1$. This implies that error accumulation transforms into linear growth $\|\delta_T\| \approx T \cdot \epsilon_{step}$, allowing FM to generate high-fidelity samples in very few steps while perfectly preserving the continuous semantic manifold of the latent space.

\subsubsection{Architectural Evolution}
\label{subsec:arch_evolution}

\begin{figure*}[ht]
  \centering
  \vspace{-6mm}
  \includegraphics[width=0.98\linewidth]{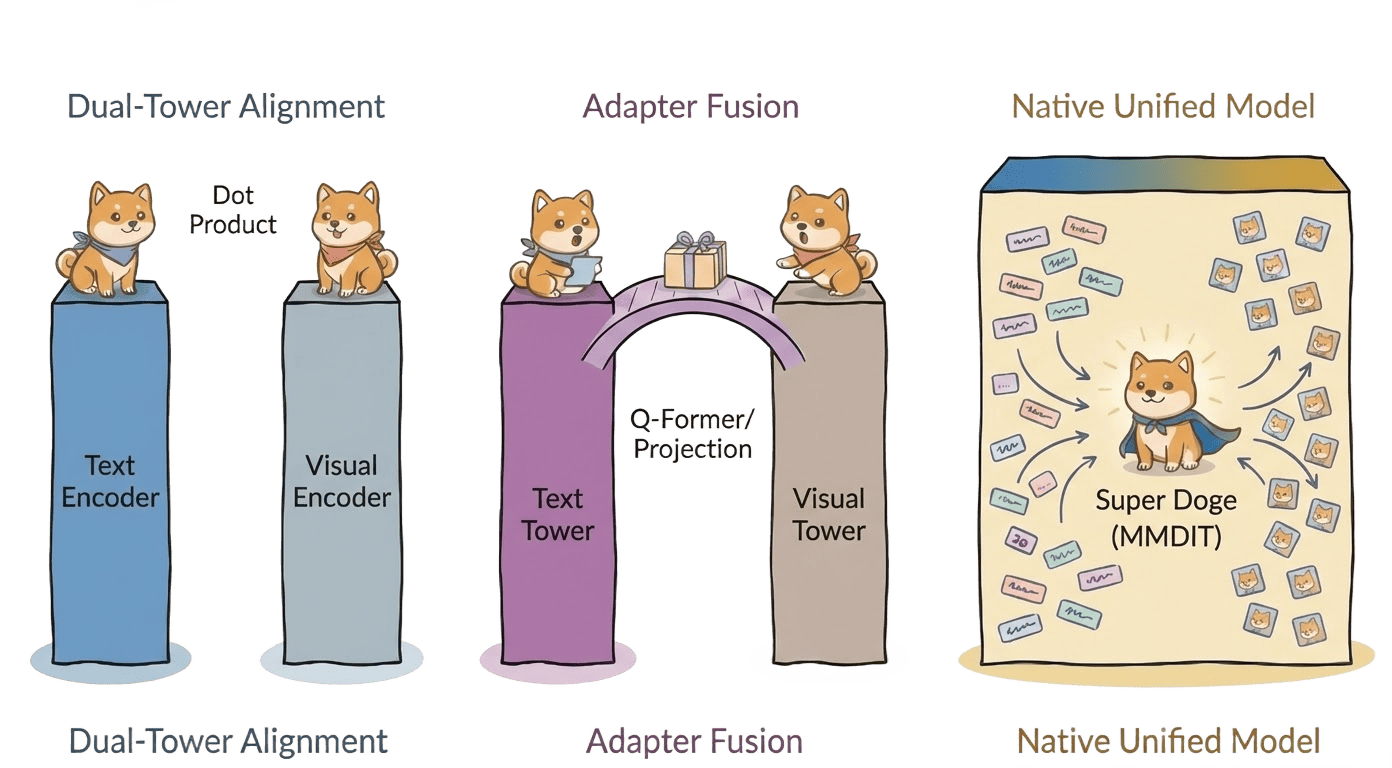}
  \caption{Evolution of Multimodal Fusion Paradigms. Transitioning from geometric isolation (Dual-Tower) to unstable Early Fusion (Adapter), and finally to the orthogonally decoupled Native unified multimodal model (MM-DiT) in large-scale unified architectures.}
  \label{fig:multimodal_evolution}
\end{figure*}

Establishing the generation mechanism only solves the mathematical expression of the target manifold. How to inject heterogeneous modal information into this manifold depends on the conditioning mechanism of the model. The evolution of multimodal architectures exhibits non-linear characteristics, essentially seeking the optimal parameter space topology to minimize gradient conflict and information loss between modalities. This process has undergone a three-stage evolution from geometric isolation to early fusion, and finally converging to orthogonal decoupling, as shown in Figure~\ref{fig:multimodal_evolution}.

\paragraph{(1) Early Evolution: Establishment of Dual-Tower Architectures and Connector Paradigms.} 

Early exploration of multimodal alignment presented two clear technological evolution paths. First was the \textit{Dual-Tower Architecture}, represented by CLIP~\cite{radford2021icml} and ALIGN~\cite{jia2021icml}. This paradigm utilized contrastive learning to project heterogeneous modalities onto a shared hypersphere. Although excellent in retrieval tasks, the separate processing of images and text by independent encoders resulted in a natural asymmetry in geometric topology, lacking deep, fine-grained interaction.

To address this limitation, the \textit{Connector-based Paradigm}, represented by Flamingo~\cite{alayrac2022flamingo} and BLIP/BLIP-2~\cite{li2022blip, li2023blip2}, emerged. These methods froze the pre-trained visual encoder and innovatively introduced learnable bridge modules (such as Perceiver Resampler or Q-Former) to align visual features with the semantic space of LLMs. This design of Frozen Visual Backbone \& Lightweight Connector not only reduced training costs but also established a standard architectural template for subsequent LMMs. 

\paragraph{(2) Early Fusion and the Challenge of Unified Optimization.} 
To further break the geometric isolation between modalities, academia began exploring more radical \textit{Early Fusion} strategies. Representative works such as Unified-IO~\cite{lu2022unifiedio} attempted to handle various heterogeneous tasks within a unified sequence-to-sequence framework, promoting the development of general interfaces.

However, this fully unified paradigm exposes deep \textit{Optimization Instability}. Particularly when introducing discretization strategies (such as Chameleon~\cite{meta2024chameleon}), despite achieving interface unification, different modalities exhibited significant differences in training dynamics. Empirical evidence shows that the gradient variance of visual tokens is significantly higher than that of text, making it difficult for the model to converge to an optimal solution during joint training. 

\begin{figure}[h]
  \centering
  \includegraphics[width=1.0\linewidth]{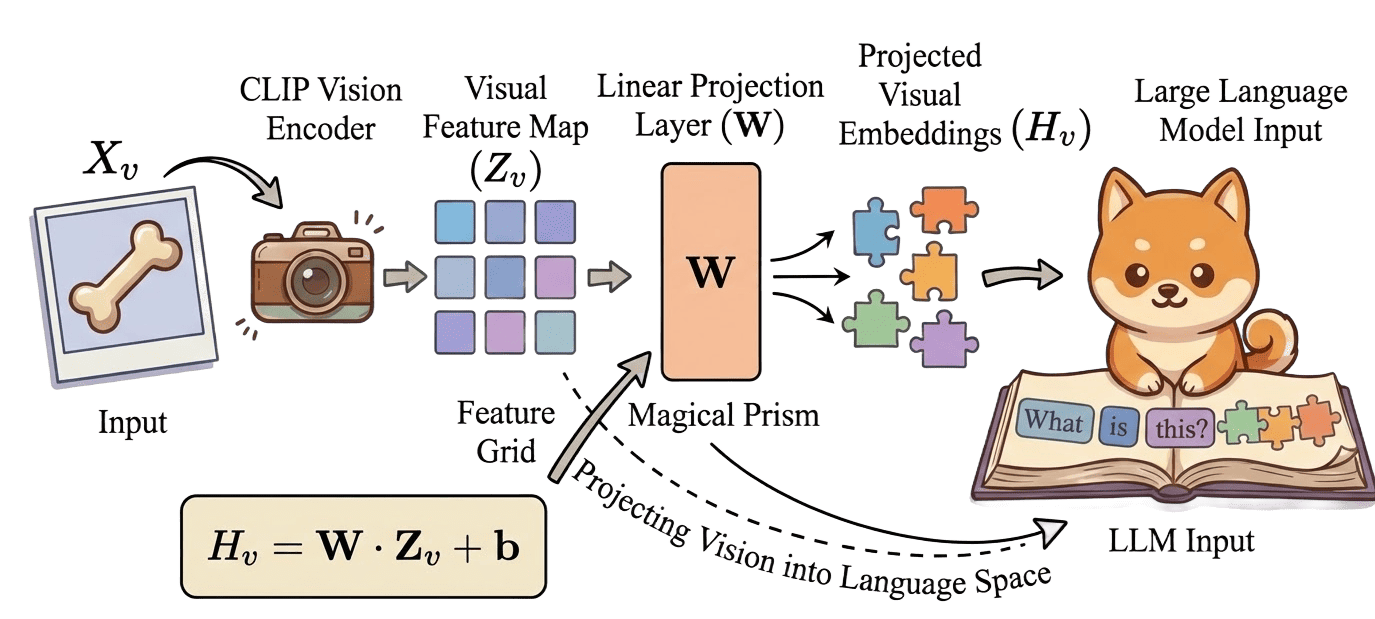}
  \caption{Information Asymmetry in LLaVA. The linear projection layer $W_{proj}$ acts as a low-rank compressor, prioritizing semantic alignment with the LLM while discarding high-frequency visual textures needed for controllable visual generation.}
  \label{fig:llava_arch}
\end{figure}

Furthermore, continuous asymmetric paradigms, such as LLaVA~\cite{liu2023llava}, interface with large language models through a projection layer. However, the linear projection layer $W_{proj}$ essentially acts as a low-rank compressor (as shown in Figure~\ref{fig:llava_arch}). During optimization, the model is encouraged to preserve semantic information that is relevant for textual reasoning, while suppressing high-frequency components that are essential for image synthesis. As a result, the mutual information between the input image and the projected representation is substantially reduced. This explains why LLaVA excels in understanding tasks but fails to restore texture details in generation tasks.

\paragraph{(3) The Mainstream Paradigm of Orthogonal Decoupling.} 

\begin{figure}[h]
  \centering
  \includegraphics[width=1.0\linewidth]{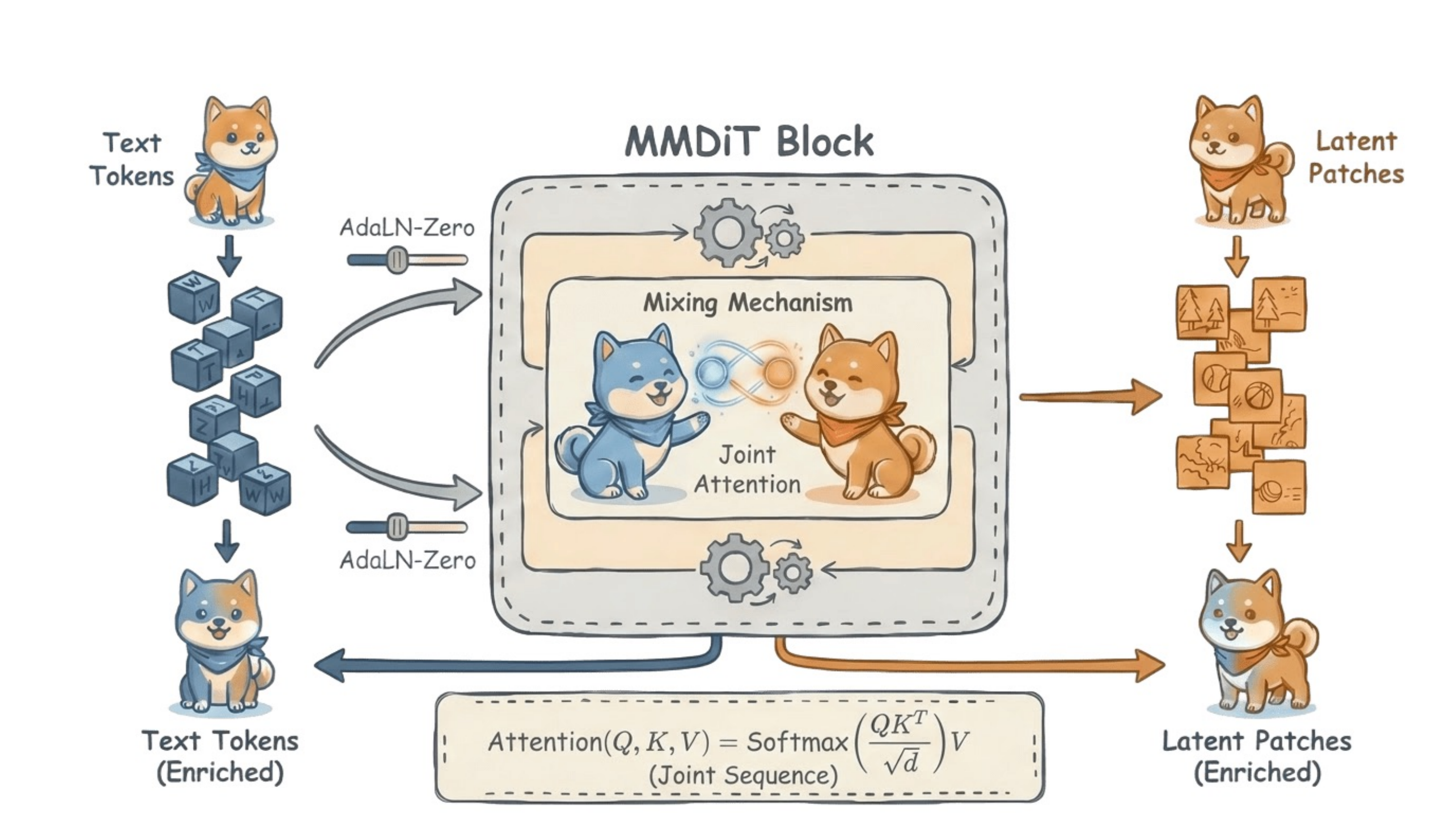}
  \caption{MM-DiT Architecture. By maintaining independent weight sets for both text and image modalities and interacting only via joint Attention, MM-DiT achieves orthogonal gradient updates, effectively resolving the modality conflict.}
  \label{fig:mmdit_arch}
\end{figure}

Addressing the aforementioned gradient conflict, works represented by Stable Diffusion 3.5~\cite{esser2024sd3} and Emu3~\cite{wang2024emu3} established the current MM-DiT architecture. The core lies in the \textit{weight decoupling} strategy—maintaining independent weight sets $W_{txt}, W_{img}$ for text and images, exchanging data only during attention operations, as shown in Figure~\ref{fig:mmdit_arch}. 

From the perspective of optimization dynamics, this design forces the Hessian matrix of the joint loss function to exhibit an approximate block-diagonal structure:
\begin{equation}
    H_{total} \approx \begin{bmatrix} H_{txt} & 0 \\ 0 & H_{img} \end{bmatrix}, \quad \text{s.t.}~\frac{\partial^2 \mathcal{L}}{\partial W_{txt} \partial W_{img}} \to 0,
\end{equation}
where $\boldsymbol{H}_{\textrm{total}}$ denotes the joint Hessian matrix, and $\boldsymbol{W}_{\textrm{txt/img}}$ represents the modality-specific parameters.
This structure effectively isolates modality-specific curvature, causing gradient updates for different modalities to tend towards orthogonality in the parameter space. Empirical data indicates that this mechanism significantly reduces the gradient conflict rate from over 50\% in AR paradigms to approximately 30\%~\cite{ma2024theoretical}. This was validated in Stable Diffusion 3.5 Large: thanks to modality decoupling, the model demonstrates instruction following capabilities and physical fidelity significantly superior to asymmetric architectures such as LLaVA on tasks requiring complex typography rendering and long-text comprehension.

\subsubsection{Intent Alignment via RL}
\label{subsubsec:rl_alignment}

After achieving orthogonal decoupling with the MM-DiT architecture, the focus of consistency modeling shifts from physical representation fitting to high-level semantic alignment. Although traditional maximum likelihood estimation (MLE) captures pixel statistical correlations, it often falls into semantic drift due to a lack of explicit supervision when dealing with ill-posed joint inverse projection problems~\cite{min2024platonic}. To this end, academia has introduced reinforcement learning with human feedback (RLHF)~\cite{fan2024aligning}, reframing alignment as a reward-guided search on the hypersphere manifold~\cite{wang2020understanding}.

\paragraph{Process Supervision \& Physical Constraints}
The architectural evolution based on preference fine-tuning began with efficient DiT baselines, exemplified by PixArt-$\alpha$~\cite{chen2024pixart}. Owing to their relatively low training cost, these architectures enable practical end-to-end alignment under preference supervision.
Addressing the sparsity of trajectory feedback in traditional DPO (Direct Preference Optimization), SPO~\cite{liang2025spo} and VisualPRM~\cite{wang2025visualprm} introduced stepwise evaluation mechanisms, performing fine-grained supervision on every inference step in the denoising path. Meanwhile, to address non-physical phenomena such as gravity violation, PhyGDPO~\cite{cai2025phygdpo} introduced physics-aware VLM feedback, where the core loss function is implemented by penalizing a physical violation term $\Delta \text{PhysScore}$:
\begin{equation}
    \mathcal{L}_{\text{Phy-DPO}} = - \mathbb{E} \left[ \log \sigma \left( \beta \log \frac{\pi_\theta(v_w)}{\pi_{ref}(v_w)} - \beta \log \frac{\pi_\theta(v_l)}{\pi_{ref}(v_l)} + \alpha \Delta \text{PhysScore} \right) \right],
\end{equation}
where $\beta$ is the KL divergence penalty coefficient that controls the deviation from the reference policy $\pi_{\textrm{ref}}$, $v_{w}$ and $v_{l}$ denote the winning and losing video samples respectively, and $\Delta \textrm{PhysScore}$ measures the difference in physical compliance scores. 
% This mechanism, combined with latent space self-play, allows the model to achieve closed-loop evolution without human annotation, demonstrating significant topological conservation in long video generation.

\paragraph{Perception-Generation Synergistic Loop}
To further break through the upper limits of static datasets, academia has established an interactive optimization paradigm centered on \textit{VLM-as-a-Judge}. This paradigm utilizes the strong semantic perception capabilities of Multimodal Large Models as a Critic to construct a Generate-Evaluate-Refine closed-loop system.
Representative works such as MetaMorph~\cite{metamorph} achieved unified alignment of understanding and generation through instruction tuning; while SRUM~\cite{2510.12784} further proposed a unified multimodal self-correction mechanism. SRUM guides the iterative fine-tuning of the diffusion model by backpropagating discriminant gradients to the generator or by utilizing fine-grained deedback captions generated by the VLM. This reciprocal improvement between perception and generation not only resolves attribute omission issues under complex prompts but also enables T2I models to continuously approach the semantic understanding upper bound of VLMs through bootstrapping in the absence of external human annotation.

\paragraph{Factorized Optimization for AR Models}
Unlike the denoising optimization of Diffusion models, AR models face the dual challenges of discrete space non-differentiability and temporal error accumulation. Addressing this, AR-GRPO~\cite{zhang2025argrpo} and ReasonGen-R1~\cite{zhang2025reasongen} in 2025 proposed a factorized optimization strategy for sequence generation:
\begin{equation}
    \mathcal{L}_{AR-RL} = \underbrace{\mathbb{E}_{\pi} [R(x)]}_{\text{Alignment Gain}} - \beta \underbrace{D_{KL}(\pi || \pi_{ref})}_{\text{Temporal Smoothing}},
\end{equation}
where $R(x)$ is the reward function derived from CLIP or VQA feedback, and $\beta$ serves as the regularization coefficient for the KL divergence term $D_{KL}$.
This paradigm explicitly decomposes the loss function into alignment gain and a temporal smoothing term. The alignment term utilizes CLIP/VQA rewards to guide token selection to conform to semantic intent, while the KL divergence constraint forces the policy to remain within the pre-trained language manifold, preventing the model from suffering Language Collapse due to over-optimization of rewards. Empirical evidence shows that this strategy effectively suppresses token repetition and garbled text in long sequence generation.

\subsubsection{Cognitive Loop via Test-time Compute}
\label{subsec:cognitive_loop}

Although reinforcement learning has achieved preliminary alignment of human intent, modal consistency remains limited by the platonic statistical boundary~\cite{min2024platonic}. Existing generative models are essentially pattern-matching interpolators that fit the training distribution solely through amortized inference~\cite{bengio2021machine}. When faced with counterfactual tasks that require multi-step chain deduction, this one-pass mapping mechanism lacks real-time verification and is prone to logical hallucinations~\cite{islam2025reasoning}.

To correct logical drift in long-range generation, consistency modeling is shifting towards the test-time compute~\cite{snell2024scaling} paradigm. This paradigm acknowledges the limitations of single-shot inverse projection and instead introduces explicit state space search during the inference phase. In this closed loop, the generation process is redefined as an optimal path search problem on the spatiotemporal manifold $\mathcal{M}$.

Recent paradigms like UniGen~\cite{tian2025unigen} and EvoSearch~\cite{zhang2025evosearch} have introduced multi-step reasoning architectures, combining monte carlo tree search (MCTS)~\cite{silver2017mastering} with verifier mechanisms~\cite{cobbe2021training}, to achieve inference-time scaling during generation. Addressing the high-dimensional nature of visual tasks, VisualPRM~\cite{wang2025visualprm} utilizes a process reward model to perform fine-grained verification on logical nodes of the denoising trajectory, thereby mathematically enhancing the logical consistency of generated results. Furthermore, by integrating an explicit causal planning layer~\cite{huang2025vchain}, the model is enabled to utilize additional reasoning compute to detect and correct deviations in physical trajectories.

\subsection{Spatial Consistency}
\label{subsec:spatial_consistency}

\begin{figure}[h]
  \centering
  \vspace{-4mm}
  \includegraphics[width=0.9\linewidth]{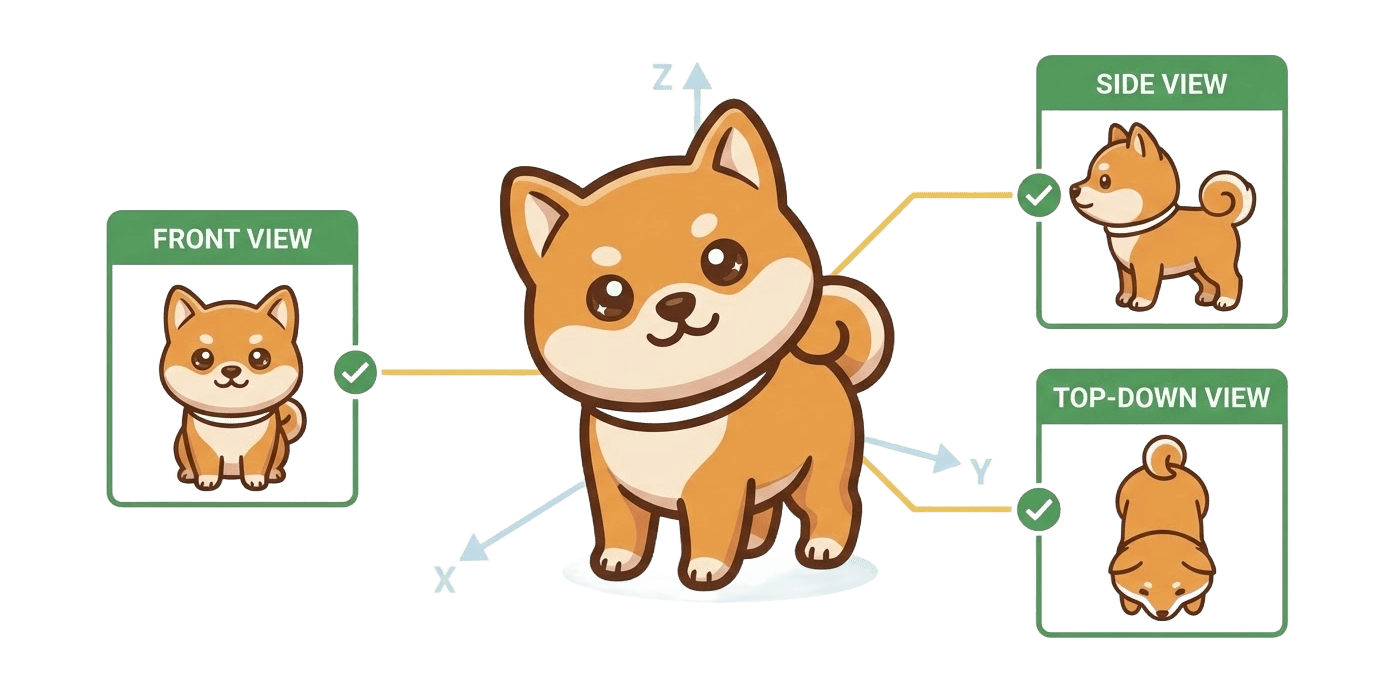}
  \caption{Spatial Consistency via Multi-View Constraints. The model ensures that the generated subject (Doge) maintains geometric coherence across Front, Side, and Top-down views, preventing structural distortion and the Janus problem.}
  \label{fig:spatial}
\end{figure}

The modal consistency discussed in the previous section successfully constructed a unified semantic mapping for heterogeneous data. However, for constructing an executable Internal Simulator, having only semantic alignment is incomplete. As developmental psychology research points out, cognition of the world is built upon the foundations of Object Permanence~\cite{baillargeon1987object} and 3D Exclusivity~\cite{spelke1990principles}. Such semantic representations, lacking geometric entities, cannot support an agent's navigation and interaction within a three-dimensional space~\cite{anderson2018evaluation}. The core mission of spatial consistency is to ground these semantic latent variables onto a three-dimensional geometric manifold $\mathcal{M}_{geo}$ that conforms to physical laws. This is essentially solving a typical Ill-posed Inverse Problem~\cite{hartley2003multiple}, as shown in Figure~\ref{fig:spatial}: specifically, how to recover a high-dimensional state space satisfying multi-view geometric constraints (such as epipolar equivariance) from dimensionality-reduced, sparse 2D observations, while avoiding structural artifacts like the Janus Problem.

To construct a unified theoretical framework, we formalize this process as solving a set of coupled differential equation inverse problems on a spatiotemporal manifold. This section will elucidate how models establish the static geometric basis of the world model by introducing physical priors and generative diffusion priors, following an evolutionary path from 2D proxy manifolds to 3D implicit fields, and finally converging to Explicit Lagrangian Primitives.

% --- 1. 颜色定义 (完全复用您的参考) ---
\definecolor{hidden-blue}{HTML}{4a90e2}
\definecolor{hidden-black}{HTML}{333333}
% --- 2. 节点样式定义 (完全复用您的参考) ---
\tikzstyle{my-box}=[
  rectangle,
  draw=hidden-black,
  rounded corners,
  text opacity=1,
  minimum height=1.5em,
  minimum width=5em,
  inner sep=2pt,
  align=center,
  fill opacity=.5,
]
\tikzstyle{leaf}=[
  my-box, 
  minimum height=1.5em,
  fill=yellow!32, 
  text=black,
  align=left,
  font=\normalsize,
  inner xsep=5pt,
  inner ysep=4pt,
  text width=20em,   
]
\tikzstyle{leaf2}=[
  my-box, 
  minimum height=1.5em,
  fill=purple!27, 
  text=black,
  align=left,
  font=\normalsize,
  inner xsep=5pt,
  inner ysep=4pt,
  text width=20em,
]
\tikzstyle{leaf3}=[
  my-box, 
  minimum height=1.5em,
  fill=hidden-blue!57, 
  text=black,        
  align=left,
  font=\normalsize,
  inner xsep=5pt,
  inner ysep=4pt,
  text width=20em,
]
\begin{figure*}[t]
\vspace{-2mm}
\centering
\resizebox{\textwidth}{!}{
\begin{forest}
  % --- 核心配置：完全模仿您的参考格式 ---
  forked edges,
  ver/.style={rotate=90, child anchor=north, parent anchor=south, anchor=center},
  for tree={
    grow=east,
    reversed=true,
    anchor=base west,
    parent anchor=east,
    child anchor=west,
    base=left,
    font=\large,
    rectangle,
    draw=hidden-black,
    rounded corners,
    align=left,
    minimum width=4em,
    edge+={darkgray, line width=1pt},
    s sep=3pt,
    inner xsep=2pt,
    inner ysep=4pt,
    line width=1.1pt,
  },
  where level=1{text width=11em,font=\normalsize}{},
  where level=2{text width=13em,font=\normalsize}{},
  where level=3{text width=30em}{},
  % ============================================================
  % 树结构内容：空间一致性演进 (内容已更新为 2026 版本)
  % ============================================================
[Evolution of Spatial Consistency\\in 3D/4D Generation, ver, fill=gray!70, text=white
    % --- Phase 1: 2D 代理流形 (橙) ---
    [2D Proxy Manifolds\\(Manifold Hypothesis), fill=orange!30, align=center
      [Deep Recurrent \& PDE, leaf
        [{ ConvLSTM~\cite{Shi2015NIPS}, PredRNN~\cite{Wang2017NIPS},\\PhyDNet~\cite{LeGuen2020CVPR}, SVG~\cite{Denton2018ICML}, etc.}]
      ]
      [Physics-Informed Priors, leaf
        [{ PINN~\cite{raissi2019physics}, DeLaN~\cite{lutter2019deep},\\HNN~\cite{greydanus2019hamiltonian}, Latent ODEs~\cite{rubanova2019latent}, etc.}]
      ]
    ]
    % --- Phase 2: 隐式连续场 (绿) ---
    [Implicit Continuous\\Fields (NeRF/SDF), fill=green!20, align=center
      [Continuous Integration, leaf
        [{ NeRF~\cite{mildenhall2020nerf}, Mip-NeRF~\cite{Barron2021MipNeRF},\\Zip-NeRF~\cite{barron2023zipnerf}, Instant-NGP~\cite{muller2022instant}, etc.}]
      ]
      [Surface \& Eikonal Constraint, leaf
        [{ NeuS~\cite{wang2021neus}, VolSDF~\cite{Yariv2021VolSDF},\\IGR~\cite{gropp2020implicit}, MonoSDF~\cite{yu2022monosdf}, etc.}]
      ]
    ]
    % --- Phase 3: 显式拉格朗日原语 (蓝) ---
    [Explicit Lagrangian\\Primitives (3DGS), fill=cyan!30, align=center
      [Rasterization \& Splatting, leaf3
        [{3DGS~\cite{Kerbl2023SIGGRAPH}, Scaffold-GS~\cite{lu2024scaffold},\\2DGS~\cite{yu20242dgs}, Mip-Splatting~\cite{yu2024mipgaussian}, etc.}]
      ]
      [4D Dynamics \& Physics, leaf3
        [{ PhysGaussian~\cite{Xie2024}, 4D-GS~\cite{wu20244dgs},\\Deformable-GS~\cite{Yang2024_Deformable3DGS}, SpacetimeGS~\cite{li2024spacetime}, etc.}]
      ]
    ]
    % --- Phase 4: 生成式统计先验 (紫) ---
    [Generative Statistical\\Priors (World Model), fill=purple!30, align=center
      [Score Distillation (SDS/VSD), leaf2
        [{DreamFusion~\cite{poole2022dreamfusion}, ProlificDreamer~\cite{wang2023prolificdreamer},\\MVDream~\cite{shi2023MVDream}, ImageReward~\cite{xu2023imagereward}, etc.}]
      ]
      [Large Reconstruction Models, leaf2
        [{ LGM (G-Objaverse)~\cite{tang2024lgm}, Objaverse-XL~\cite{deitke2023objaverse},\\SV3D~\cite{voleti2024sv3d}, Dust3R~\cite{wang2024dust3r}, See3D~\cite{ma2025youseeit}, etc.}]
      ]
    ]
]
\end{forest}
}
\caption{Evolution of Spatial Consistency Paradigms: From 2D Proxy to Generative Primitives.}
\label{fig:spatial-consistency-evolution}
\end{figure*}
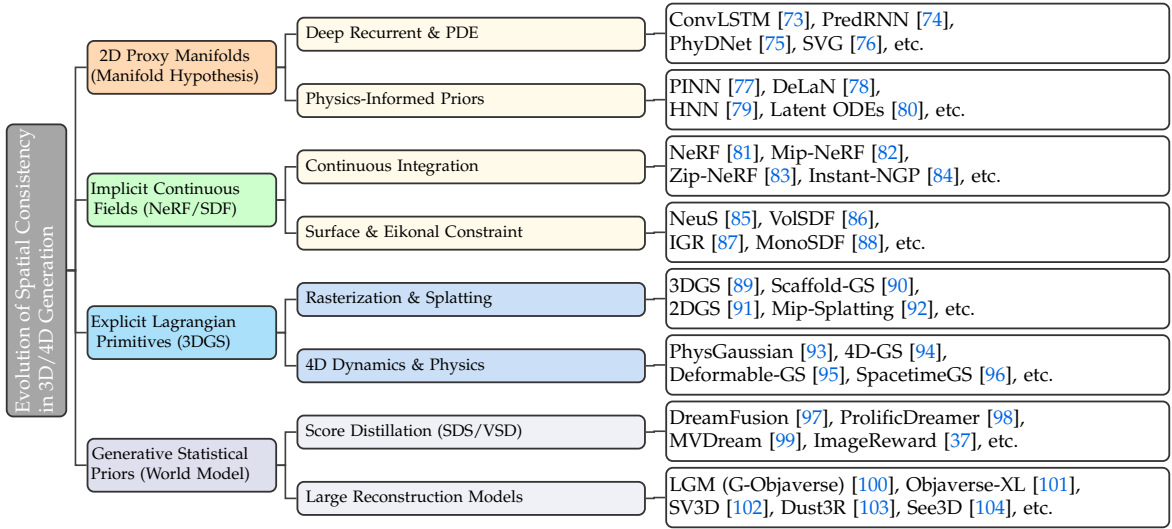

\subsubsection{Geometric Decomposition of Consistency}
\label{subsubsec:geometric_decomposition}

To mathematically characterize spatial consistency, we decompose this abstract concept into two complementary and hierarchically progressive topological constraints: the former governs the microscopic continuity of the physical surface, while the latter guarantees the macroscopic uniqueness and coherence of the object structure.

\paragraph{Micro-level: Local Neighborhood Topological Consistency.}
This constraint focuses on the \textbf{Intrinsic Continuity} of the manifold $\mathcal{M}$, which corresponds mathematically to the Lipschitz Condition. That is, for any two adjacent points on the manifold, the difference in their physical attributes (such as color, density) should be strictly constrained linearly by their Euclidean distance.
In 3D reconstruction and generation tasks, this constraint is typically implemented explicitly through geometric regularization terms. For example, IGR (Implicit Geometric Regularization)~\cite{gropp2020implicit} utilizes the Eikonal equation to constrain the norm of gradients, while RegNeRF~\cite{niemeyer2022regnerf} introduces a smoothness loss to suppress non-physical high-frequency noise generated under sparse views, ensuring the generated object possesses a smooth and physically reasonable surface.

\paragraph{Macro-level: Global Geometric Consistency.}
Local smoothness alone is insufficient; the model must also satisfy \textbf{Epipolar Equivariance} in multi-view geometry~\cite{hartley2003multiple}. That is, when observing the same object from different viewpoints $v_a, v_b$, its projected coordinates should satisfy strict algebraic constraints $\boldsymbol{x}_b^\top \boldsymbol{F}_{ab} \boldsymbol{x}_a = 0$. In generative models, violating this constraint is the root cause of the Janus Problem~\cite{poole2022dreamfusion}, where different viewpoints produce incompatible object geometries. To address this, SyncDreamer~\cite{syncdreamer} constructs an explicit 3D cost volume to enforce alignment, while MVDream~\cite{shi2023MVDream} utilizes a multi-view self-attention mechanism to internalize hard geometric constraints into attention weights, directly locking the global topological uniqueness of the generated object.

The above decomposition clarifies the geometric objectives of spatial consistency. However, how to systematically solve these topological constraints within the parameter space of a neural network requires establishing a unified differential equation perspective.

\subsubsection{Theoretical Formulation}
\label{subsubsec:unified_differential_view}

To construct a theoretical framework, we formalize the spatial consistency in 3D visual generation as solving a set of coupled Inverse Differential Problems on the spatiotemporal manifold $\mathcal{M} \subseteq \mathbb{R}^3 \times \mathbb{R}^+$. From this perspective, the construction of the full state field $\Phi(\boldsymbol{x}, t)$ follows three core physical laws, which respectively define the world's presentation mode, generation rules, and motion laws.

\paragraph{Physical Rendering: The RTE.}
Both explicit and implicit 3D representations can be physically viewed as discretized solutions to the Radiative Transfer Equation (RTE)~\cite{kajiya1986rendering}. For a ray $\boldsymbol{r}(s) = \boldsymbol{o} + s\boldsymbol{d}$, the variation of its radiance $L$ along the path follows:
\begin{equation}
    \underbrace{\boldsymbol{d} \cdot \nabla L(\boldsymbol{x}, \boldsymbol{d})}_{\text{Transport}} = \underbrace{-\sigma(\boldsymbol{x}) L(\boldsymbol{x}, \boldsymbol{d})}_{\text{Absorption}} + \underbrace{\sigma(\boldsymbol{x}) \boldsymbol{c}(\boldsymbol{x}, \boldsymbol{d})}_{\text{Emission}},
    \label{eq:rte}
  \end{equation}
where $\sigma(\boldsymbol{x})$ represents the \textbf{Volume Density} at position $\boldsymbol{x}$, and $\boldsymbol{c}(\boldsymbol{x}, \boldsymbol{d})$ denotes the view-dependent \textbf{Color Emission}. The difference in discretization constitutes the divergence in technical routes: \textbf{NeRF (Implicit Fields)} employs volume rendering integration, approximating the solution by dense Riemann summation of Eq. (\ref{eq:rte}) along the ray; while \textbf{3DGS (Explicit Primitives)} discretizes the continuous field into a set of Lagrangian Gaussian basis functions, transforming the integral into efficient analytical rasterization. The former ensures continuity, while the latter achieves real-time performance.

\paragraph{Generative Evolution: The SDE.}
In the generative prior paradigm, spatial consistency originates from the probability distribution of the pre-trained model. We model the process of recovering from Gaussian white noise $\boldsymbol{z}_T$ to the data manifold $\boldsymbol{z}_0$ as a Stochastic Differential Equation (SDE)~\cite{song2021scorebased}:
\begin{equation}
    d\Phi_t = \boldsymbol{f}(\Phi_t, t) dt + g(t) d\boldsymbol{w}
    \label{eq:sde},
\end{equation}
where $\boldsymbol{f}(\cdot)$ is the deterministic drift term governing semantic evolution, $g(t)$ denotes the diffusion coefficient, and $\boldsymbol{w}$ represents the standard Wiener process.
Modern generative models aim to learn the reverse process of the above SDE (score matching). When the diffusion term $g(t)=0$, the SDE degenerates into a deterministic Ordinary Differential Equation (ODE), \ie Flow Matching. This provides a theoretical basis for understanding how generative models recover ``smooth and topologically consistent'' geometric structures from disordered noise.

\paragraph{Motion Law: Lagrangian Transport.}
To ensure topological consistency of the spatial structure along the time axis, the motion of material points $\boldsymbol{x}$ must follow Lagrangian Flow:
\begin{equation}
    \frac{d\boldsymbol{x}}{dt} = \boldsymbol{v}(\boldsymbol{x}, t), \quad \text{s.t.} \quad \frac{D\Phi}{Dt} = 0 \quad (\text{Material Derivative}),
\end{equation}
where $\boldsymbol{v}$ represents the velocity field driving the particle motion, and $\frac{D \Phi}{D t}$ denotes the material derivative.
This constraint implies that feature $\Phi$ remains conserved as it moves with the fluid (the material derivative is 0). This directly corresponds to the particle tracking mechanism in the explicit primitive paradigm and serves as the mathematical bridge connecting static geometry and dynamic video.

\begin{figure*}[t] 
  \centering
  \includegraphics[width=1.0\linewidth]{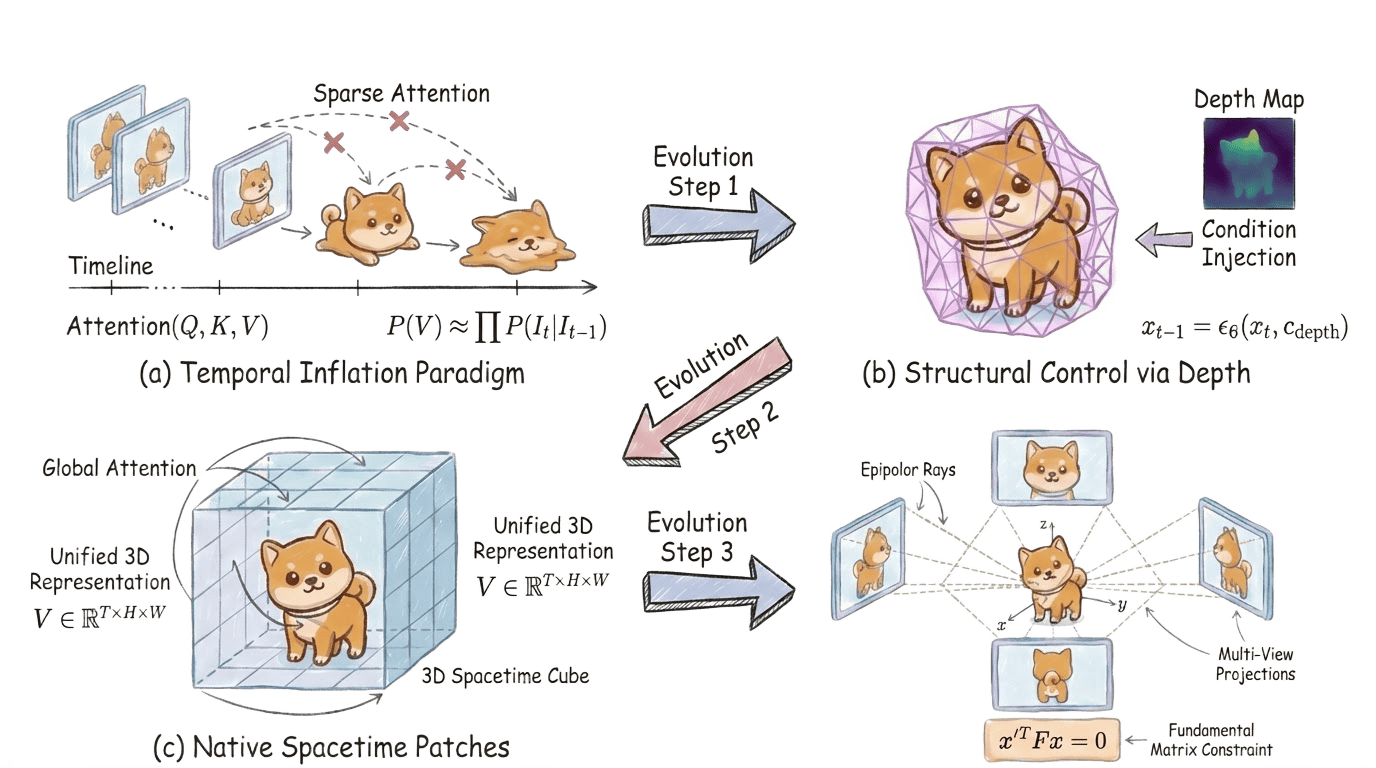} 
  \caption{Evolution of Spatial Consistency Paradigms. We trace the trajectory from early 2D Proxy Manifolds, to Implicit Continuous Fields like NeRF, moving towards Explicit Lagrangian Primitives like 3DGS, and finally integrating Generative Diffusion Priors.}
  \label{fig:spatial_evolution}
\end{figure*}

% \paragraph{Evolution Summary:}
The history of spatial consistency evolution is essentially a process where academia shifted from solving the static RTE (NeRF) to inversely solving the generative SDE (Diffusion), and finally integrating Lagrangian dynamic constraints. This iterative process of moving from attempting to fit dynamics on 2D projected manifolds to implicit continuous field integration, and then returning to explicit Lagrangian primitives, is illustrated in Figure~\ref{fig:spatial_evolution}.

\subsubsection{2D Proxy Manifold \& Domain Mismatch}
\label{subsubsec:2d_proxy_manifold}

Before explicit 3D representations established their mainstream status, the primary path to addressing spatiotemporal consistency was video prediction based on the Manifold Hypothesis. This paradigm avoided expensive $SE(3)$ spatial modeling and instead attempted to reduce the high-dimensional physical state field $\Phi$'s evolutionary dynamics operator $\mathcal{F}_{3D}: SE(3) \times \mathbb{R}^3 \to \mathbb{R}^3$ into a parameterized mapping $\mathcal{F}_{\theta}: \mathbb{R}^{H \times W} \to \mathbb{R}^{H \times W}$ on the 2D image manifold $\mathcal{M}_{img}$. Although this proxy manifold strategy offered computational complexity advantages, it introduced a fundamental Domain Mismatch.

\paragraph{Dynamics Fitting Lacking $SE(3)$ Equivariance.}
Early works like ConvLSTM~\cite{Shi2015NIPS} and PredRNN~\cite{Wang2017NIPS, Wang2018TPAMI}, while mitigating long-sequence gradient decay through improved recurrent units (\eg Gradient Highway Unit, GHU), relied on convolution operations $W \ast I$ that only possess Translation Equivariance and lack the ability to perceive the 3D rotation group $SO(3)$.
As stated in \cite{Gao2022CVPR, tan2025ustep, tan2023temporal, wei2024interpretable}, attempting to simulate 3D rigid body rotation through non-linear transformations of a 2D pixel grid is essentially approximating high-dimensional topology on a low-dimensional manifold. This misalignment of inductive bias leads to the model's inability to decouple extrinsic camera motion from intrinsic object deformation, inevitably causing non-physical Non-rigid Distortion or texture stretching in generated videos during large viewpoint transformations.

\paragraph{Early Attempts and Limitations of Physics-aware Modeling.}
To alleviate the blurriness caused by pure statistical fitting and enhance the robustness of temporal extrapolation, academia attempted to endow black-box models with physical interpretability, the core idea being to inject physical conservation laws into the neural network's parameter space. A pioneer in this direction is \textit{Physics-Informed Neural Networks (PINN)}~\cite{raissi2019physics}, which adds the residuals of Partial Differential Equations (PDEs) as regularization terms to the loss function, forcing the network output to conform to physical constraints like fluid mechanics or wave equations. Subsequently, Deep Lagrangian Networks (DeLaN)~\cite{lutter2019deep} and Hamiltonian Neural Networks (HNN)~\cite{greydanus2019hamiltonian} further introduced energy conservation priors, explicitly modeling the system's total energy (Hamiltonian) using Euler-Lagrange equations, thereby achieving precise trajectory prediction for complex dynamic systems in continuous time.

In the field of video prediction, PhyDNet~\cite{LeGuen2020CVPR} drew on these ideas by explicitly disentangling the hidden state into a physical dynamics branch $\mathcal{H}_{phy}$ and a residual texture branch $\mathcal{H}_{res}$. Unlike the soft constraints of PINNs, PhyDNet directly restricts convolution kernel weights via Moment Matching, making them approximate PDE finite difference operators on a discrete grid:
\begin{equation}
    \frac{\partial \mathcal{H}}{\partial t} \approx \sum_{k} c_k \frac{\partial^k \mathcal{H}}{\partial \boldsymbol{x}^k} \implies \text{Filter Weights} \xrightarrow{\text{Moment}} \text{Finite Difference Stencils},
\end{equation}
where $\mathcal{H}$ denotes the disentangled hidden state, $\boldsymbol{x}$ is the spatial coordinate, and $c_{k}$ represents the partial differential coefficients.

Furthermore, addressing the limitations of discrete time sampling, Latent ODEs~\cite{rubanova2019latent} proposed by Rubanova et al. utilize a continuous time ODE Solver to model hidden state evolution, effectively handling temporal consistency issues under non-uniform sampling.

Although these methods and variational inference models like SVG~\cite{Denton2018ICML} made progress in short-term prediction, modeling based on 2D manifolds implies a spatial continuity assumption. Once depth mutations caused by Occlusion occur, the optical flow field becomes non-differentiable, and PDE constraints immediately fail. This defect of being unable to model object permanence indicates a theoretical limitation in solving strict 3D consistency on a 2D proxy manifold.

\subsubsection{Implicit Continuous Fields}
\label{subsubsec:implicit_fields}

Addressing the theoretical limitations of 2D proxy manifolds in 3D consistency, academia turned to defining state fields directly in 3D Euclidean space. The establishment of this paradigm was built upon Mesh-based differentiable rendering works like SoftRas~\cite{Liu2019SoftRas} and DIB-R~\cite{Chen2019DIBR}, which verified the feasibility of calculating gradients $\partial I / \partial \mathcal{V}$ through a smooth rasterization process. NeRF~\cite{mildenhall2020nerf} further discarded discrete geometry, using MLPs to parameterize the scene as a continuous coordinate mapping function $F_{\Theta}: (\boldsymbol{x}, \boldsymbol{d}) \rightarrow (\boldsymbol{c}, \sigma)$, and connecting the 3D field with 2D observations through differentiable Volume Rendering Integral.

\paragraph{(1) Representation Efficiency \& Frequency Fidelity.}
The evolution of Neural Radiance Fields is essentially a process of seeking balance between \textit{parameter efficiency} and \textit{signal fidelity}. The challenges in this field have deepened from initial inference acceleration (introducing discrete representations) to maintaining frequency domain anti-aliasing characteristics in discrete space.

\textit{(\romannumeral1) The Shift to Hybrid Representations.} To break the efficiency bottleneck of pure MLP architectures, NVIDIA's Instant-NGP~\cite{muller2022instant} introduced \textbf{Multiresolution Hash Grids}, using spatial hashing to map continuous coordinates to a learnable feature table; while in the generative domain, EG3D~\cite{chan2022eg3d} proposed \textbf{Tri-plane} representation, establishing the mainstream paradigm for 3D GANs. These methods (including TensoRF~\cite{chen2022tensorf}) significantly improved training efficiency and geometric generation capabilities by introducing explicit spatial inductive biases.

\textit{(\romannumeral2) Aliasing \& Signal Processing Correction.} However, the aforementioned discretized representations (as well as point-wise sampling in original NeRF) introduced severe aliasing in high-frequency regions. Mip-NeRF~\cite{Barron2021MipNeRF} corrected this defect from a signal processing perspective, pointing out that discrete sampling ignoring the sampling volume violates the Nyquist sampling theorem.
By introducing Cone Tracing and Integrated Positional Encoding (IPE), Mip-NeRF calculated the feature expectation within a Gaussian volume, revealing the essence of anti-aliasing in its mathematical form:
\begin{equation}
    \gamma(\boldsymbol{\mu}, \boldsymbol{\Sigma}) = \mathbb{E}_{\boldsymbol{x} \sim \mathcal{N}(\boldsymbol{\mu}, \boldsymbol{\Sigma})}[\gamma(\boldsymbol{x})] \approx \sin(\boldsymbol{\mu}) \circ \exp\left(-\frac{1}{2} \text{diag}(\boldsymbol{\Sigma}) \right),
\end{equation}
where $\boldsymbol{\mu}$ and $\boldsymbol{\Sigma}$ denote the mean vector and covariance matrix of the conical frustum, and $\circ$ represents the element-wise product.
This formula reveals a profound physical mechanism: the exponential decay term $\exp(-\boldsymbol{\Sigma})$ essentially acts as an \textbf{Adaptive Low-pass Filter}. When the sampling cone radius increases (\ie variance $\boldsymbol{\Sigma}$ increases, corresponding to distant views or low-resolution regions), high-frequency features are exponentially suppressed.

To transfer this excellent anti-aliasing property to efficient grid representations, Zip-NeRF~\cite{barron2023zip} further combined Multisampling with feature smoothing techniques, resolving the scale uncertainty inherent in hash grids. This series of evolutions is mathematically equivalent to the \textbf{Uncertainty Principle} in Fourier transforms: the wider the spatial localization ($\boldsymbol{\Sigma}$ is large), the narrower the frequency bandwidth, thereby mechanistically eliminating moiré patterns and high-frequency artifacts, achieving a unification of efficiency and fidelity.

\paragraph{(2) Level Set Ambiguity \& Eikonal Manifold Constraints.}
NeRF's density field $\sigma$ suffers from physical ambiguity. When extracting surfaces, the artificially set threshold $\tau$ leads to Level Set Ambiguity. To obtain precise geometric surfaces, NeuS~\cite{wang2021neus} and VolSDF~\cite{Yariv2021VolSDF} converted the representation from a density field to a Signed Distance Field (SDF). By introducing an unbiased Logistic transformation $\phi_s(f(\boldsymbol{x}))$ and imposing an Eikonal regularization term:
\begin{equation}
    \mathcal{L}_{geo} = \mathbb{E}_{\boldsymbol{x}} [(\| \nabla f(\boldsymbol{x}) \|_2 - 1)^2],
\end{equation}
where $f(\boldsymbol{x})$ is the signed distance function, and the gradient norm constraint $\| \nabla f \|_2 = 1$ ensures physical validity.
This constraint forces the gradient norm of the implicit field to be constant at 1, ensuring the zero-level set $\mathcal{S} = \{\boldsymbol{x} | f(\boldsymbol{x})=0\}$ converges to a smooth, closed manifold surface that satisfies physical constraints.

Viewing from the perspective of manifold optimization, implicit continuous fields essentially trade Inference Latency for Geometric Completeness~\cite{wang2021neus}. Due to the continuous differentiability of SDF, this paradigm constitutes an ideal basis for high-fidelity inverse rendering. It is not only suitable for reconstructing closed Watertight Manifolds to realize static asset digitization~\cite{Yariv2021VolSDF, yariv2020multiview}, but also effectively avoids geometric holes common in explicit methods through Eikonal regularization-induced smoothing priors under sparse views~\cite{gropp2020implicit}. However, its mathematical properties also define a theoretical upper bound: the high sampling cost of volume integration $O(N_{samples})$ makes it difficult to support high-frame-rate real-time interaction~\cite{mildenhall2020nerf}, and the smoothing assumption of continuous fields faces expressive bottlenecks when modeling dynamic scenes with drastic topological fractures~\cite{Park2021_Hyper, Kerbl2023SIGGRAPH}.

\subsubsection{Explicit Lagrangian Primitives}
\label{subsubsec:explicit_primitives}

\begin{figure}[t]
  \centering 
  \includegraphics[width=\linewidth]{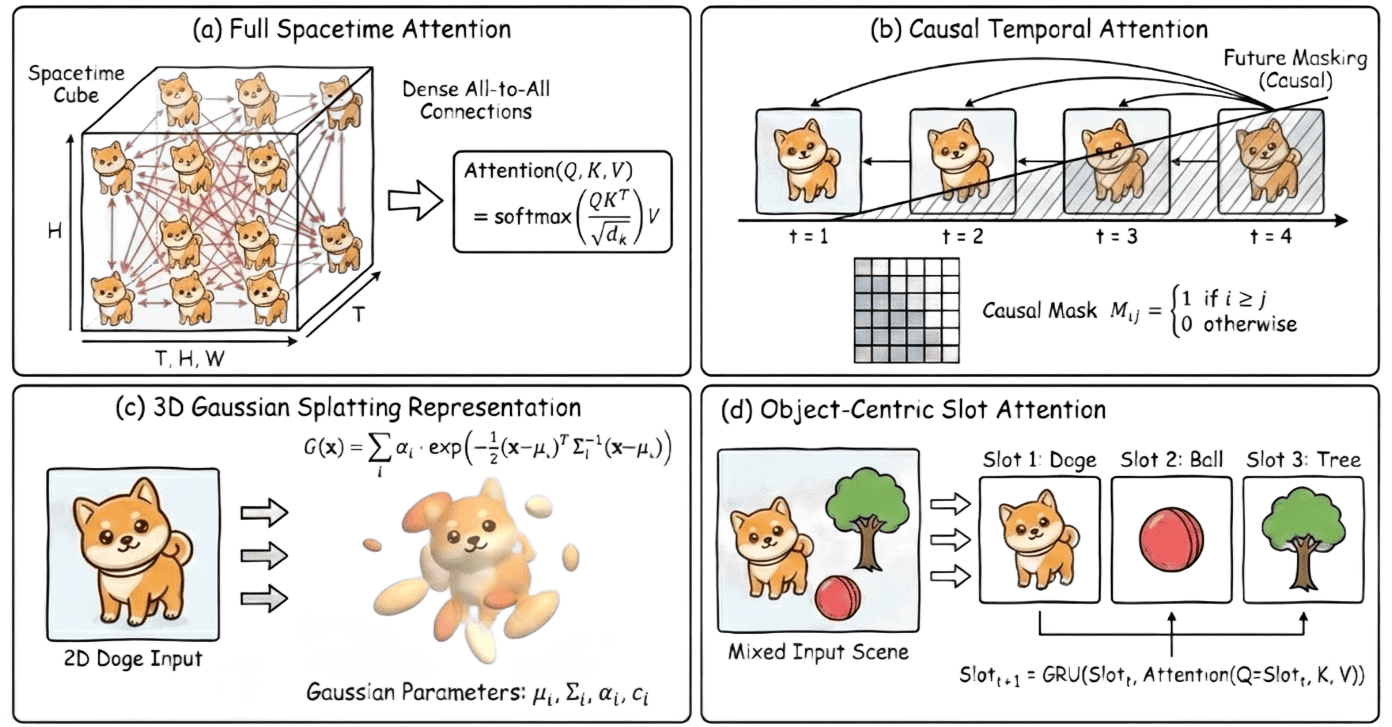}
  \caption{Key Mechanisms for Advanced Spacetime Modeling. A taxonomy of core techniques underpinning modern models: (a) Full Spacetime Attention enables dense long-range dependencies; (b) Causal Masking ensures temporal causality; (c) 3D Gaussian Splatting offers explicit, differentiable 3D structure; (d) Object-Centric Slots decompose complex scenes into distinct entities.}
  \label{fig:mechanisms}
\end{figure}

Although implicit continuous fields established theoretical completeness for multi-view consistency, their sampling mechanism relying on volume integration constitutes a computational bottleneck for real-time simulation. The 3D Gaussian Splatting (3DGS) proposal~\cite{Kerbl2023SIGGRAPH}CC2 marks the return of the representation form of the state field $\Phi$IQ3 from an implicit field to explicit particles (as shown in Figure~\ref{fig:mechanisms}CR1(c)). This paradigm discretizes the scene into a set of anisotropic Gaussian primitives $\Phi = \{ \mathcal{G}_i(\mu, \Sigma, \alpha, SH) \}_{i=1}^M$ and reconstructs the projection operator $\mathcal{P}$ as Rasterization.

\paragraph{(1) Mechanisms of Static Representation.}
Unlike the ray marching of NeRF~\cite{mildenhall2020nerf}, 3DGS~\cite{Kerbl2023SIGGRAPH} utilizes the GPU sorting pipeline for acceleration, containing three key characteristics:

\textit{(\romannumeral1) Rasterization Pipeline.}
The algorithm involves two key steps: first is Frustum Culling and Projection, projecting 3D Gaussian into a 2D screen space covariance matrix $\Sigma^{2D} = J W \Sigma^{3D} W^T J^T$; second is tiled radix sort, which is the computational bottleneck with complexity $O(N \cdot k)$. By leveraging a tile-based parallel rendering strategy, the method restricts computation to overlapping Gaussians and requires only $\alpha$-blending on during rasterization, avoiding invalid sampling of empty space.

\textit{(\romannumeral2) Integral Duality.}
NeRF adopts a Backward Pull, prone to gradient masking ($\partial C / \partial \sigma_{far} \approx 0$). In contrast, 3DGS adopts a Forward Push; explicit sparsity allows error gradients $\frac{\partial \mathcal{L}}{\partial \mu}$ to bypass the MLP and backpropagate directly and sparsely to geometric parameters. This explicit gradient flow is the mathematical foundation for the efficient convergence of 3DGS.

\textit{(\romannumeral3) Adaptive Density Control.}
This approach can be viewed as a variant of AMR (Adaptive Mesh Refinement). The core idea is: if the gradient is too large and variance is small ($\| \nabla \mathcal{L} \| > \tau, \| \Sigma \| < \epsilon$), it is judged as underfitting and the Gaussian is cloned; if the gradient is large and variance is large, it is judged as overfitting and the Gaussian is split. Through this mechanism, the method dynamically adjusts the density of Lagrangian particles in response to the underlying optimization landscape.

\paragraph{(2) Evolution towards 4D Dynamics.}
Addressing 4D spatiotemporal modeling, the explicit primitive paradigm has developed three main evolutionary paths based on how the time dimension $t$ is handled:

\textit{(\romannumeral1) Lagrangian Particle Tracking.}
As in PhysGaussian~\cite{Xie2024}, it assumes Gaussian primitives possess material point properties, solving the equation of motion $\mu(t) = \mu_0 + \int \boldsymbol{v}(\tau) d\tau$ by introducing continuum mechanics equations ($\rho \ddot{\boldsymbol{x}} = \nabla \cdot \sigma + \boldsymbol{g}$). By embedding physical constraints into the optimization process, the method enables joint learning of visual appearance and physical behavior.

\textit{(\romannumeral2) Eulerian Tensor Decomposition.}
As in 4D-GS~\cite{4dgaussiansplatting}, the 4D spatiotemporal field is modeled as a high-dimensional tensor $\mathcal{T}$, using CP or Tucker decomposition to reduce dimensionality:
\begin{equation}
    \mathcal{T}(x,y,z,t) \approx \sum_{r=1}^R \boldsymbol{u}_r(x) \circ \boldsymbol{v}_r(y) \circ \boldsymbol{w}_r(z) \circ \boldsymbol{h}_r(t), \label{eq:eikonal_loss}
\end{equation}
where $\circ$ denotes the outer product, $R$ is the tensor rank, and $\boldsymbol{u}_{r}, \boldsymbol{v}_{r}, \boldsymbol{w}_{r}, \boldsymbol{h}_{r}$ represent the factor vectors along each dimension.
This form optimizes storage complexity from $O(N^4)$ to $O(N^2)$, effectively supporting dynamic changes in topological structure.

\textit{(\romannumeral3) Canonical Deformation.}
As in Deformable-GS~\cite{Yang2024_Deformable3DGS}, it adopts a a static base with transient offsets formulation, predicting coordinate offsets $\Delta \mu$ via MLP, leveraging the spectral bias of MLPs to effectively capture high-frequency motion fields.

% \paragraph{(3) Summary.}
The explicit primitive paradigm shows significant advantages in balancing high frame rate rendering and high-resolution reconstruction. However, its discrete nature introduces topological adaptability limitations, making it difficult to naturally handle fractures and fusions in fluid dynamics like implicit fields~\cite{luiten2024dynamic}, indicating the need to introduce higher-order generative dynamics models.

\subsubsection{Generative Statistical Priors}
\label{subsubsec:generative_priors}

In open-world generation tasks, observation conditions are extremely sparse, causing the problem to degenerate into an ill-posed one. In this phase, works utilize video diffusion models as implicit world model priors, establishing an algorithm-data synergistic framework.

\paragraph{(1) Algorithmic  \& Geometric Constraints.}
To elevate 2D priors to 3D consistency, academia has reconstructed optimization objectives and architectural designs:

\textit{(\romannumeral1) Score Distillation Sampling (SDS) \& Variational Correction.}
Unlike photometric loss, SDS~\cite{poole2022dreamfusion} obtains gradients by calculating the score function of a pre-trained diffusion model. Addressing the over-smoothing problem of SDS, VSD (Variational Score Distillation)~\cite{Wang2023_Prolific} introduces a variational distribution, minimizing the KL divergence between the generated distribution and the prior distribution, thereby recovering high-frequency texture details.

\textit{(\romannumeral2) Multi-View Geometric Attention.}
Pure 2D priors are difficult to guarantee multi-head consistency. Works like MVDream~\cite{shi2023MVDream} modify the U-Net architecture, upgrading spatial self-attention to 3D correspondence attention. This design forces the model to perform feature alignment via camera parameters ($R, T$) when generating different views, achieving soft geometric consistency.

\paragraph{(2) Scaled Data Foundation.}
To break the 3D data bottleneck, academia has adopted a Synthetic-Real-Generative hybrid construction strategy for large-scale dataset construction:

\textit{(\romannumeral1) Aggregation.}
Objaverse-XL~\cite{deitke2023objaverse} integrated tens of millions of 3D assets collected from the internet, fundamentally alleviating the scarcity of large-scale 3D data. G-Objaverse~\cite{Tang2024_LGM} provided high-quality RGB-D-Normal triplets through a physical rendering pipeline, becoming the standard source for training Large General Reconstruction Models (LGM).

\textit{(\romannumeral2) Real-world Perception.}
MVImgNet~\cite{yu2023mvimgnet} and Co3D-v2~\cite{reizenstein2021common} provide millions of object-centric video sequences captured in real-world environments. While dense geometric ground truth is largely unavailable, these datasets play a crucial role in reducing the domain discrepancy between synthetic and real data, particularly in appearance and texture distributions.

\textit{(\romannumeral3) Inverse Generative Engine.}
See3D~\cite{ma2025youseeit} advances an automated data generation paradigm by coupling generative video models with geometric reconstruction. Specifically, large-scale pseudo-3D videos are synthesized using video diffusion models such as SV3D~\cite{Stability_SV3D}, followed by geometric inference via Dust3R~\cite{wang2024dust3r} and rapid reconstruction through LGM, constructing a closed-loop data production engine to achieve exponential asset expansion.

% \subsubsection{Conclusion and Future Directions}
% \label{subsubsec:conclusion}

The development of spatial consistency modeling exhibits a clear iterative trajectory. Early methods relied on 2D proxy fitting, which gradually evolved into 3D implicit representations to improve geometric coherence. Subsequently, explicit formulations such as 3D Gaussian Splatting reintroduced computational efficiency and rendering scalability. Current trends indicate a convergence toward hybrid architectures that combine explicit geometric primitives with implicit diffusion-based priors, leveraging the complementary strengths of both representations~\cite{Tang2024_LGM, Stability_SV3D}.

Looking ahead, the research focus in this field is shifting from pure visual reconstruction to deep physical interaction modeling. On one hand, Neuro-symbolic Grounding will become the key to connecting semantic space and geometric space. Future models aim to establish differentiable mappings between LLM symbolic logic and numerical parameters, as shown in works like Eureka~\cite{ma2024eureka}, to realize an endogenous understanding of object materials and force mechanisms, thus transcending pixel statistics-based imitation. On the other hand, the scope of spatial consistency is expanding to Action-Consistency. As World Models evolve towards interactive environments~\cite{menapace2024playable}, Reinforcement Learning (RL) will be introduced into the generative loop, ensuring that the scene follows physical causality when responding to actions $\pi(a_t|s_t)$. To support this capability, the architectural level is expected to break the Cascaded Generation pipeline and shift towards End-to-End Native 4D Streaming, \ie performing real-time streaming inference directly with compressed 4D Tokens~\cite{dalal2025oneminute}. 
% In summary, spatial consistency has established explicit primitives as the static geometric basis, resolving the problem of \textbf{Existence} of objects in three-dimensional space.

\subsection{Temporal Consistency}
\label{subsec:temporal_consistency}
% --- 1. 颜色定义 (完全复用您的参考) ---
\definecolor{hidden-blue}{HTML}{4a90e2}
\definecolor{hidden-black}{HTML}{333333}
% --- 2. 节点样式定义 (完全复用您的参考) ---
\tikzstyle{my-box}=[
  rectangle,
  draw=hidden-black,
  rounded corners,
  text opacity=1,
  minimum height=1.5em,
  minimum width=5em,
  inner sep=2pt,
  align=center,
  fill opacity=.5,
]
\tikzstyle{leaf}=[
  my-box, 
  minimum height=1.5em,
  fill=yellow!32, 
  text=black,
  align=left,
  font=\normalsize,
  inner xsep=5pt,
  inner ysep=4pt,
  text width=20em,   
]
\tikzstyle{leaf2}=[
  my-box, 
  minimum height=1.5em,
  fill=purple!27, 
  text=black,
  align=left,
  font=\normalsize,
  inner xsep=5pt,
  inner ysep=4pt,
  text width=20em,
]
\tikzstyle{leaf3}=[
  my-box, 
  minimum height=1.5em,
  fill=hidden-blue!57, 
  text=black,        
  align=left,
  font=\normalsize,
  inner xsep=5pt,
  inner ysep=4pt,
  text width=20em,
]
\begin{figure*}[t]
\vspace{-2mm}
\centering
\resizebox{\textwidth}{!}{
\begin{forest}
  % --- 核心配置：完全模仿您的参考格式 ---
  forked edges,
  ver/.style={rotate=90, child anchor=north, parent anchor=south, anchor=center},
  for tree={
    grow=east,
    reversed=true,
    anchor=base west,
    parent anchor=east,
    child anchor=west,
    base=left,
    font=\large,
    rectangle,
    draw=hidden-black,
    rounded corners,
    align=left,
    minimum width=4em,
    edge+={darkgray, line width=1pt},
    s sep=3pt,
    inner xsep=2pt,
    inner ysep=4pt,
    line width=1.1pt,
  },
  where level=1{text width=11em,font=\normalsize}{},
  where level=2{text width=13em,font=\normalsize}{},
  where level=3{text width=30em}{},
  % ============================================================
  % 树结构内容：时间一致性演进 (内容已更新为 2026 版本)
  % ============================================================
  [Evolution of Temporal Consistency\\in Video Gen \& Reasoning, ver, fill=gray!70, text=white
    % --- Phase 1: 潜空间时序膨胀 (橙) ---
    [Latent Temporal\\Inflation (2D Priors)\\, fill=orange!20
      [Zero-Shot \& Anchoring, leaf
        [{ Text2Video-Zero~\cite{khachatryan2023text2video}, FateZero~\cite{qi2023fatezero},\\ControlNet~\cite{zhang2023controlnet}, Tune-A-Video~\cite{wu2023tuneavideo}, etc.}]
      ]
      [Temporal Adaptation, leaf
        [{ AnimateDiff~\cite{guo2023animatediff}, VideoCrafter1~\cite{he2024videocrafter},\\ModelScope~\cite{wang2023modelscope}, Gen-1~\cite{structurecontentguidedvideosynthesis}, etc.}]
      ]
      [Frequency \& Noise Correction, leaf
        [{GFN~\cite{rao2021global}, AFNO~\cite{guibas2021adaptive},\\FastInit~\cite{xing2025fastinit}, FreeNoise~\cite{qiu2024freenoise}, etc.}]
      ]
    ]
    % --- Phase 2: 离散自回归 (绿) ---
    [Discrete Autoregressive\\Sequence Modeling\\ , fill=green!20
      [Tokenization \& Scaling, leaf
        [{ VideoPoet~\cite{kondratyuk2023videopoet}, W.A.L.T~\cite{gupta2023walt},\\MagViT-v2~\cite{yu2023magvitv2}, Cosmos~\cite{agarwal2025cosmos}, etc.}]
      ]
      [Long-Sequence Optimization, leaf
        [{ VAR (Next-Scale)~\cite{tian2024var}, FramePack~\cite{zhang2025framepack},\\Show-o (Unified)~\cite{xie2024showo}, Diffusion Forcing~\cite{chen2024diffusionforcing}, etc.}]
      ]
    ]
    % --- Phase 3: 原生全时空 DiT (蓝) ---
    [Native Spatiotemporal\\Continuous DiT\\, fill=hidden-blue!57, text=black
      [Global 3D Attention, leaf3
        [{Sora~\cite{openai2024sora}, HunyuanVideo~\cite{hunyuan2024},\\Lumiere~\cite{bar2024lumiere}, Veo/Veo 3~\cite{google2025veo}, etc.}]
      ]
      [Efficient Linearization, leaf3
        [{ Video-TTT~\cite{dalal2025oneminute}, Pyramid Flow~\cite{jin2025pyramid},\\TeaCache~\cite{liu2025teacache}, Movie Gen~\cite{polyak2025moviegen}, etc.}]
      ]
    ]
    % --- Phase 4: 逻辑与因果推理 (紫) ---
    [Logical Consistency\\\& Causal Reasoning\\, fill=purple!27
      [Visual Chain-of-Thought, leaf2
        [{ Visual CoT~\cite{viscot}, UV-CoT~\cite{wang2024chain},\\Mini-O3~\cite{minio3}, VChain~\cite{huang2025vchain}, etc.}]
      ]
      [Physics \& Audio Causality, leaf2
        [{
        Video-CoT~\cite{videocot}, Think Sound~\cite{liu2025thinksound},\\Physics-IQ~\cite{motamed2025physics}, VCD~\cite{aoshima2025vcd}, etc.}]
      ]
    ]
]
\end{forest}
}
% \caption{Evolution of Temporal Consistency Paradigms: From Latent Inflation to Causal World Simulators. }
\caption{Evolution of Temporal Consistency Paradigms: From Latent Inflation and Discrete Sequence Modeling to Native Spatiotemporal DiT and Causal World Simulators.}
\label{fig:temporal-consistency-evolution}
\end{figure*}
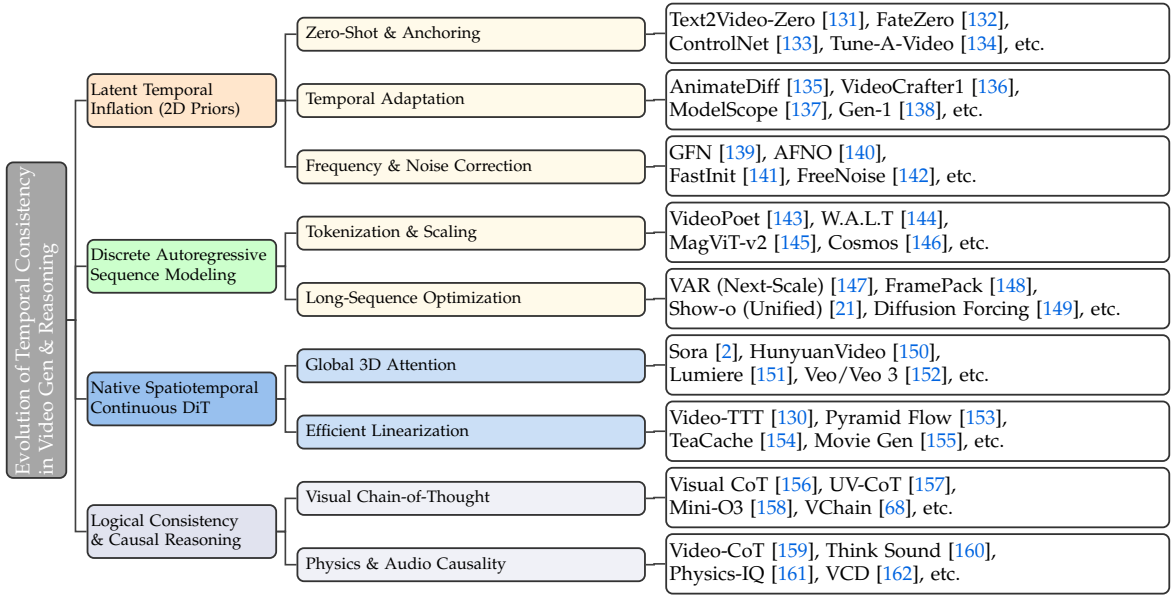

\begin{figure}[ht]
  \centering
  \includegraphics[width=0.98\linewidth]{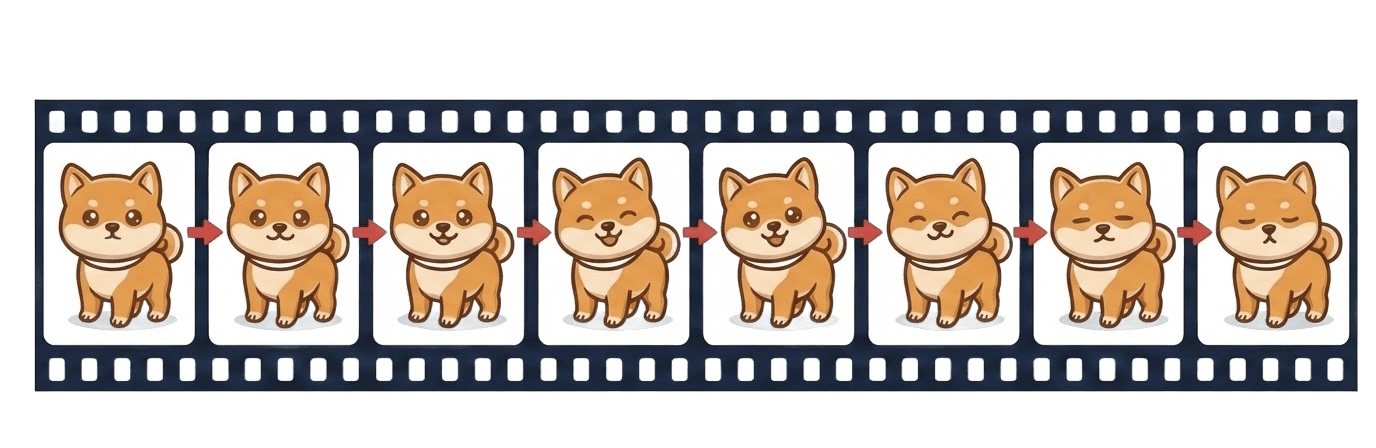}
  \caption{Temporal Consistency and Identity Preservation. An illustration of the temporal attention mechanism ensuring identical subject features across consecutive frames ($t_0 \rightarrow t_n$). The generation process is governed by two key constraints: \textit{Physical Constraints} ($\mathcal{L}_{\text{phy}}$) enforce smoothness in motion trajectories to prevent flickering artifacts, while \textit{Causal Constraints} ($\mathcal{L}_{\text{causal}}$) ensure logical progression of events (\eg object permanence) throughout the timeline.}
  \label{fig:temporal}
\end{figure}

Through the modeling of spatial consistency (\S\ref{subsec:spatial_consistency}), we have successfully constructed a geometrically complete static world. However, the core value of a World Model lies not in archiving the state of a moment, but in rehearsing future trajectories. If spatial consistency is regarded as the Static Geometric Basis of the world model~\cite{ha2018world}, temporal consistency constitutes the key element establishing its physical evolutionary Temporal Dynamics~\cite{lecun2022path}.
Mathematically, this process is equivalent to solving a Multi-objective Optimization Problem constrained by both physical constraints  $\mathcal{L}_{\text{phy}}$ and causal logic $\mathcal{L}_{\text{causal}}$ within a high-dimensional manifold space~\cite{zhang2024physdreamer}, as illustrated in Figure~\ref{fig:temporal}.

\subsubsection{From Frequency Stability to Physical Compliance}
\label{subsubsec:evolution_analysisv2}

To objectively measure the evolutionary trajectory of temporal consistency technologies, evaluation metrics must transcend traditional perceptual dimensions. For a long time, academia relied on FVD (Fréchet Video Distance)~\cite{unterthiner2018towards} to assess video quality, but empirical studies indicate that FVD primarily characterizes the similarity of spatial feature distributions and has limitations in detecting temporal high-frequency Flickering and non-physical deformations.

It must be pointed out that frequency stability in temporal consistency does not exist in isolation; it must be built upon the semantic foundation of modality alignment (\S\ref{sec:modality_consistency}) and the topological constraints of spatial geometry (\S\ref{subsec:spatial_consistency}). For instance, frontier models like Veo 3~\cite{google2025veo} effectively suppress high-frequency artifacts and achieve physically compliant causal reasoning precisely by integrating MM-DiT (modal consistency) and 3DGS (spatial consistency).

\begin{figure}[ht]
  \centering
  \includegraphics[width=0.98\linewidth]{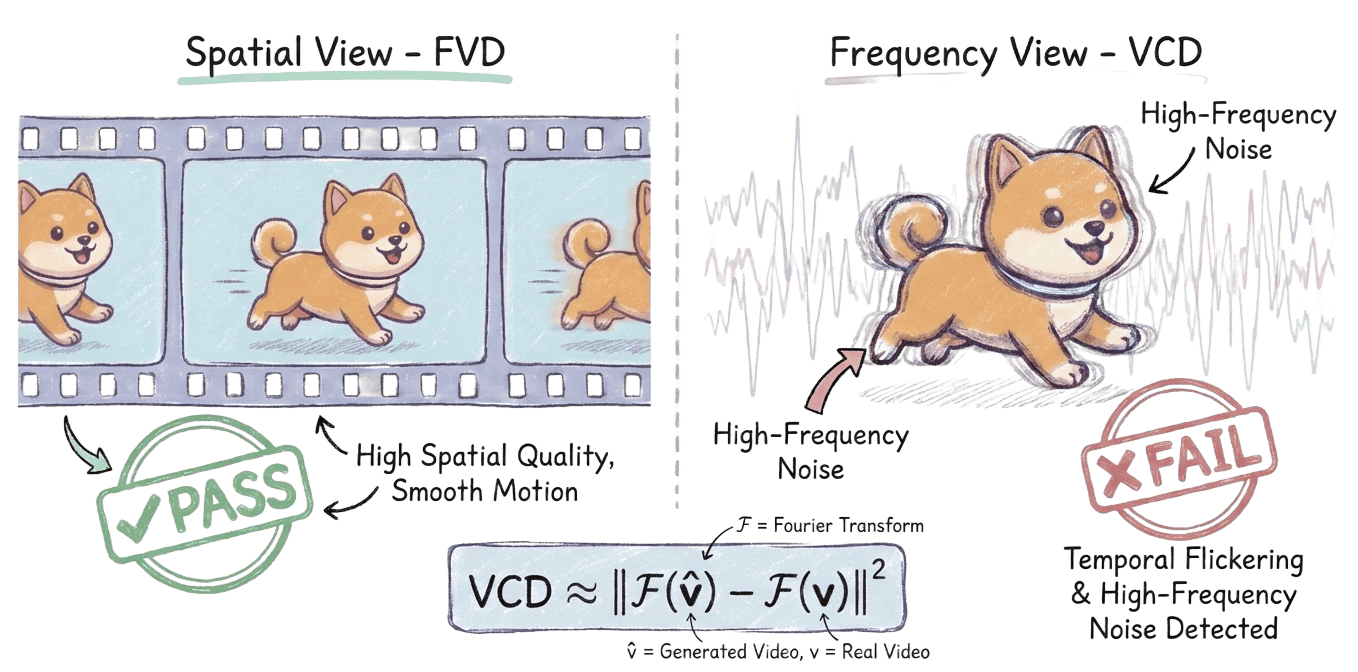}
  % \caption{\textbf{Video Consistency Analysis: Spatial vs. Frequency View.} Traditional metrics like FVD often miss high-frequency flickering. VCD captures these artifacts by analyzing the temporal Fourier spectrum of feature embeddings.}
  \caption{Video Consistency Analysis: Spatial vs. Frequency View. Traditional distribution-based metrics such as FVD primarily assess spatial perceptual quality and smooth motion in feature space, often overlooking high-frequency temporal flickering. In contrast, VCD explicitly models temporal consistency by analyzing the Fourier spectrum of feature embeddings, enabling the detection of subtle high-frequency noise and flicker artifacts invisible to spatial statistics.}
  \label{fig:vcd_analysis}
\end{figure}

To fill this gap, Video Consistency Distance (VCD)~\cite{aoshima2025video} was designed as a Reward-based Fine-tuning Objective. As shown in Figure~\ref{fig:vcd_analysis}, VCD measures the feature difference between the generated video $\hat{V}$ and natural video in the temporal frequency spectrum:
\begin{equation}
    \mathcal{L}_{\text{VCD}}(\hat{V}) = \mathbb{E}_{t} \left[ \| \mathcal{F}_t(\phi(\hat{v}_t)) - \mathcal{F}_t(\phi(\hat{v}_{t-1})) \|_{\text{High-Pass}}^2 \right],
\end{equation}
where $\phi(\cdot)$ denotes the feature extractor (\eg CLIP Image Encoder~\cite{radford2021icml}), and $\mathcal{F}_{t}$ represents the Short-Time Fourier Transform (STFT) along the time axis. The physical meaning of this formula is that motion features in the real world should possess continuity in the frequency domain, whereas temporal inconsistencies in generative models (such as texture flickering) will manifest as significant energy fluctuations in the high-frequency band.

\begin{table}[ht]
\centering
\caption{Empirical Evolution of Cross-Generational Models. Data synthesized from VBench (Temporal), Physics-IQ (Physics), and Veo 3 Technical Report (Reasoning) benchmarks.}
\label{tab:cross_bench}
\resizebox{\columnwidth}{!}{%
\begin{tabular}{lccccc}
\toprule
\textbf{Generation Paradigm} & \textbf{Rep. Model} & \textbf{Temporal Consistency} $\uparrow$ & \textbf{Physics Compliance} $\uparrow$ & \textbf{Causal Reasoning} $\uparrow$ & \textbf{Freq. Fidelity (VCD)} $\downarrow$ \\
\cmidrule(lr){3-3} \cmidrule(lr){4-4} \cmidrule(lr){5-5} \cmidrule(lr){6-6}
 & & \textit{(VBench Norm.)} & \textit{(Physics-IQ)} & \textit{(Task Success Rate)} & \textit{(Reward Penalty)} \\
\midrule
\textbf{Temporal Inflation} & AnimateDiff & $0.68$ & $0.42$ & N/A ($<10\%$) & High ($>1.2$) \\
\textbf{Discrete AR} & VideoPoet & $0.79$ & $0.55$ & Low ($\sim 25\%$) & Medium ($\sim 0.9$) \\
\textbf{Native DiT} & HunyuanVideo & $0.88$ & $0.78$ & Medium ($\sim 45\%$) & Low ($\sim 0.6$) \\
\textbf{World Model Priors} & \textbf{Google Veo 3} & \textbf{0.95}$^*$ & \textbf{0.86}$^*$ & \textbf{High ($>70\%$)}$^\dagger$ & \textbf{Minimal ($<0.3$)} \\
\bottomrule
\end{tabular}%
}
\footnotesize{
\textit{*Note: World Model scores are extrapolated based on relative improvements reported in~\cite{google2025veo} compared to Native DiT baselines.}\\
\textit{$^\dagger$Refers to success rates on complex physical interaction tasks (\eg object manipulation) as demonstrated in~\cite{veo3_2025_zeroshot}.}
}
\end{table}

\paragraph{From Perception to Physical Reasoning}
Traditional evaluations focus on visual quality, while new standards have expanded to physical causal dimensions. As shown in Table~\ref{tab:cross_bench}, firstly, addressing temporal jitter, the Generative Prior Paradigm (World Model Priors) significantly reduces high-frequency artifacts (VCD $<$ 0.3) by introducing frequency domain reward fine-tuning. Secondly, to assess adherence to physical laws, Physics-IQ~\cite{motamed2025physics} is used to quantify model compliance in rigid body dynamics and fluid simulation. Finally, causal reasoning has become a core evaluation dimension for models like Veo 3~\cite{google2025veo}. Veo 3 demonstrates emergent capabilities in zero-shot physical interaction tasks (such as predicting domino toppling), with a task success rate exceeding 70\%, marking the evolution of video generation technology from pure visual simulation to dynamic systems capable of logical deduction.

\subsubsection{Latent Temporal Inflation}
\label{subsubsec:temporal_inflation}

In the early stages when large-scale 4D data was not yet widespread, academia dedicated efforts to lowering the training threshold for video generation. Works represented by Tune-A-Video~\cite{wu2023tuneavideo} and AnimateDiff~\cite{guo2023animatediff} established the Temporal Inflation paradigm of \textit{Spatial Freeze, Temporal Insertion}.

\paragraph{Independence Assumption \& ELBO Relaxation.}

The core strategy of this paradigm is to extend pre-trained 2D Text-to-Image (T2I) models into video generators, specifically by freezing the spatial convolution layers of the 2D U-Net and inserting learnable 1D temporal attention modules only between layers. Viewing from a probabilistic graph perspective, this is essentially simplifying the joint distribution of video generation $p(\boldsymbol{x}_{1:T})$ into a first-order Markov chain. Theoretical derivation shows that this relaxation of the Evidence Lower Bound (ELBO) ignores high-order dependencies of $p(x_t | x_{<t-1})$, leading to a significant increase in the KL divergence term over long sequences. In practical applications (such as VideoCrafter1~\cite{he2024videocrafter}), this mathematical relaxation manifests as significant Semantic Drift: as the number of generated frames increases ($T>16$), the identity features of the initial frame are gradually diluted by independent noise injection.

\paragraph{Spatial Anchoring \& Zero-shot Injection.}
To suppress semantic drift, early works explored training-free consistency enhancement paths. Text2Video-Zero~\cite{khachatryan2023text2video} and FateZero~\cite{qi2023fatezero} adopted a Zero-shot Attention Injection mechanism, forcing subsequent frames to reuse the Key/Value feature matrices of the first frame. Meanwhile, inspired by ControlNet~\cite{zhang2023controlnet}, some works introduced explicit geometric conditions (such as Depth/Pose) as spatial anchors. Empirical data shows that although these methods perform well in static backgrounds, when object motion amplitude exceeds 20\% of the screen width, forced feature injection leads to obvious Smearing Artifacts, revealing the limitations of the inflation paradigm in handling complex dynamics.

\paragraph{Frequency Filtering \& Dynamic Correction.}
Besides temporal drift, existing temporal inflation models typically face the problem of \textit{Frequency Blindness}. Since the temporal attention mechanism operates independently in the $(B \cdot HW)$ dimension, it often exhibits a lack of inductive bias when capturing high-frequency texture changes. Fourier spectral analysis reveals that generated videos exhibit significant energy loss in the high-frequency band ($>15$Hz), visually manifesting as non-physical texture flickering. Addressing the capture of long-range dependencies and high-frequency information, frequency domain learning offers a novel perspective. Global Filter Networks (GFN)~\cite{rao2021global} proposed using 2D Discrete Fourier Transform (2D DFT) instead of self-attention mechanisms, achieving long-range spatiotemporal interaction capture with $O(N \log N)$ complexity by performing global filtering operations in the frequency domain. Building on this, Adaptive Fourier Neural Operators (AFNO)~\cite{guibas2021adaptive} further optimized inter-channel information aggregation, proving that frequency domain Token Mixers can effectively overcome spatial blindness and precisely retain high-frequency details. Furthermore, addressing noise interference in sequence modeling, BERT4Rec~\cite{sun2019bert4rec} and Denoising SASRec~\cite{fan2022denoising} introduced uncertainty quantification mechanisms, achieving dynamic suppression of irrelevant perturbations by zeroing out gradients of high-noise samples during backpropagation (gradient pruning). In the video generation domain, FastInit~\cite{xing2025fastinit} drew on these denoising ideas, proposing a learning-based noise initialization strategy. This method discards traditional independent Gaussian sampling and instead trains a lightweight inversion network to directly predict the optimal initial noise for the current frame based on spatiotemporal features of preceding frames, significantly enhancing generation coherence while suppressing latent space temporal high-frequency jitter.

\paragraph{The Theoretical Boundary of Inflation.}
Although methods like FastInit~\cite{xing2025fastinit} alleviate frequency domain flickering, the temporal inflation paradigm is perpetually limited by its 2D topological anchor. Since the core spatial convolution layers are frozen, the model is essentially performing minute elastic deformations on static images rather than generating true temporal dynamics. Empirical research~\cite{xing2023dynamiccrafter} indicates that when facing large viewpoint transformations (such as an object rotating 180 degrees) or the emergence of new content, this class of models often produces severe texture stretching. This over-reliance on pre-trained 2D priors condemns it to the role of a transitional solution. To capture true physical world dynamics, academia has turned to exploring native video architectures trained from scratch, which is the driving force behind the development of the discrete autoregressive paradigm.

\subsubsection{Discrete Autoregressive Modeling}
\label{subsubsec:discrete_autoregressive}

To break the theoretical bottleneck of long-sequence modeling, VideoPoet~\cite{kondratyuk2023videopoet}, CogVideo~\cite{hong2022cogvideo}, and W.A.L.T~\cite{gupta2023walt} drew on the scaling law of LLMs, establishing the two-stage autoregressive generation paradigm. By expanding the context window, this paradigm reconstructs video generation as long-range causal prediction of discrete Tokens.

\paragraph{Causal 3D Tokenizer \& Data Compression.}
The cornerstone of the discrete autoregressive paradigm is an efficient 3D VQ-VAE. Unlike image Tokenizers, video compression must strictly adhere to temporal causality. MagViT-v2~\cite{yu2023magvitv2} innovatively introduced asymmetric Temporal Padding and Causal 3D Convolution, strictly limiting the receptive field of convolution kernels to the current frame $t$ and preceding moments, ensuring that future information does not leak during the compression process. Addressing reconstruction blurriness in low-motion scenes, VTokenizer-Plus~\cite{arxiv2025compositional} further introduced Object-Centric representation, significantly improving texture fidelity of static backgrounds by separating foreground and background codebooks.

\paragraph{Memory Decay in Long Sequences.}
With the release of models like NVIDIA Cosmos~\cite{agarwal2025cosmos}, the AR paradigm has regained attention due to its superior data scaling capabilities. However, Error Accumulation remains the core challenge of this paradigm. According to sequence modeling theory~\cite{bengio2015scheduled}, the distribution shift between Teacher Forcing during training and autoregressive generation during inference (Exposure Bias) causes minute inter-frame prediction errors to amplify exponentially with time step $t$. To suppress this sequence variance, VAR~\cite{var_clip} proposed the Next-Scale Prediction mechanism, reconstructing the autoregressive process from pixel scanning to coarse-to-fine scale recursion, mathematically reducing inference steps from linear $O(N)$ to logarithmic $O(\log N)$. Furthermore, FramePack~\cite{zhang2025framepack} introduced a frame context packing mechanism and bidirectional anti-drift sampling, combined with the PFP (Pretraining Frame Preservation)~\cite{zhang2025pfp} objective, significantly improving reconstruction fidelity under long time sequences.

\paragraph{Return to Continuous Latent Space.}
Despite continuous architectural optimization, the non-differentiability of the discretization operation $z_q = \arg\min \|z_e - e_k\|$ constitutes an inherent optimization difficulty for this paradigm. Training typically relies on the Straight-Through Estimator (STE)~\cite{bengio2013estimating} for approximation, but in high-dimensional video space ($D > 4096$), the gradient variance caused by STE ($\sigma^2 > 10^3$) easily triggers codebook collapse~\cite{van2017neural}. This discretization gap limits the precision of AR models in generating minute textures and sub-pixel motion. Precisely this limitation has driven the technical focus to shift towards Continuous Latent Space, utilizing Diffusion Transformers to directly model continuous probability density on the manifold.

\paragraph{Hybrid Transition: Fusing AR and Diffusion.}
Between pure AR and DiT, academia has explored fusion paths of the two, aiming to combine the long-range causality of AR with the high-fidelity decoding capability of Diffusion.
First, at the inference level, Diffusion Forcing~\cite{chen2024diffusionforcing} proposed a non-rigid sequence modeling scheme, modeling each time step as an independent diffusion process, supporting rollback and branch exploration during inference, breaking the traditional AR restriction of no return.
Second, at the architectural level, Show-o~\cite{xie2024show} proposed the Unified Omni-Model paradigm. This method is not a simple stacking of modules, but achieves isomorphic modeling of discrete tokens (for semantic understanding) and continuous tokens (for visual generation) within a single set of weights. Through a mixed masking mechanism, Show-o achieves bidirectional interoperability of understanding and generation in physical weights.

\subsubsection{Unified Spatiotemporal Modeling via DiT}
\label{subsubsec:unified_dit}

\begin{figure}[t]
  \centering
  \vspace{-4mm}
  \includegraphics[width=1.0\linewidth]{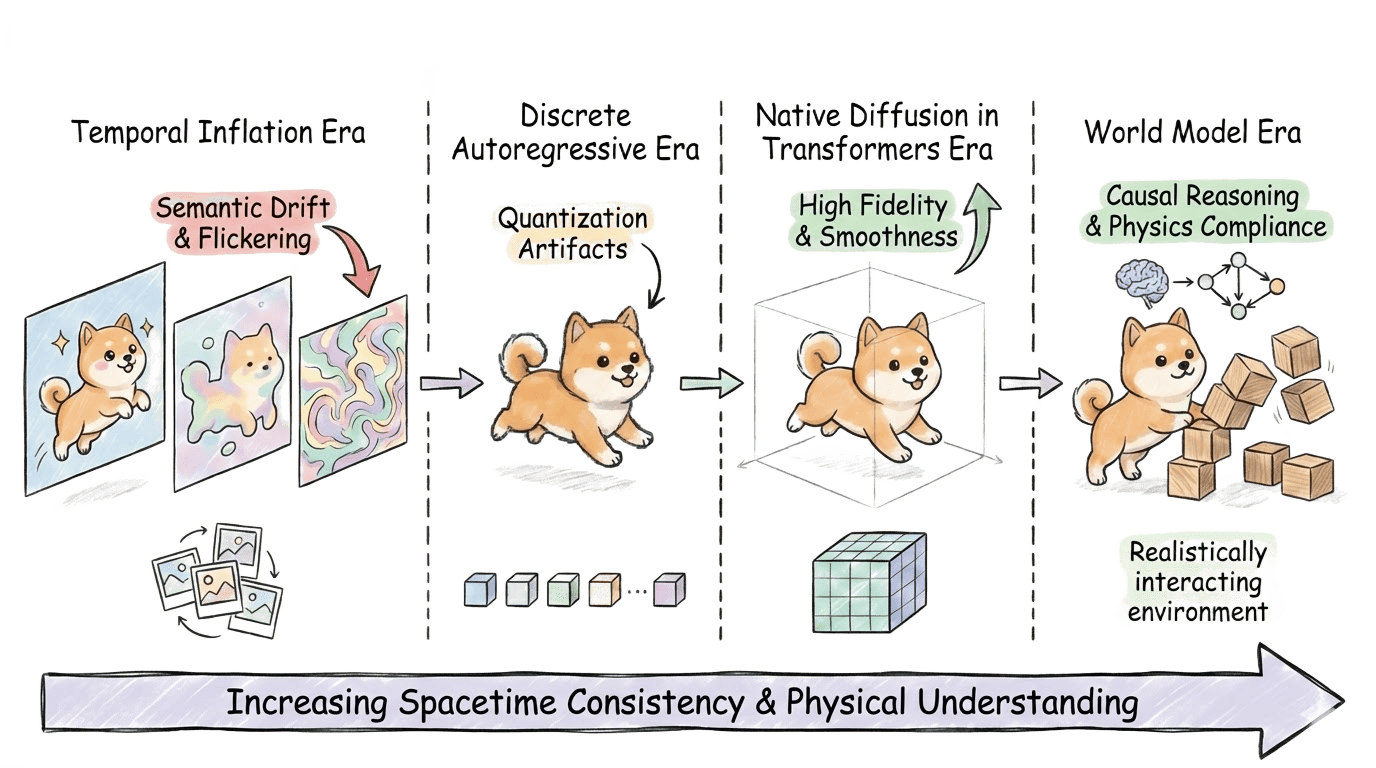}
  \caption{Evolution of Video Generation Paradigms. The technical path advances from Temporal Inflation (prone to drift) and Discrete AR (quantization loss) to the current Native DiT. This paradigm achieves full spatiotemporal isomorphism, serving as the foundation for world models.}
  \label{fig:evolution_video}
\end{figure}

Compared to the spatiotemporal fragmentation caused by the temporal inflation paradigm and the quantization loss brought by the discrete AR paradigm, the new generation of paradigms represented by Sora~\cite{openai2024sora} and HunyuanVideo~\cite{hunyuan2024} established the current benchmark for temporal consistency in video generation by thoroughly returning to continuous latent space and adopting the Diffusion Transformer (DiT) architecture. This evolutionary path from Spatiotemporal Decoupling to Full Spatiotemporal Isomorphism is shown in Figure~\ref{fig:evolution_video}.

\paragraph{Native Spatiotemporal Architecture.}
Native 3D DiT treats video as a sequence of 3D Patches $N = T \times H \times W$, its core advantage being the capture of non-local physical interactions through a global receptive field.

\textit{(\romannumeral1) Full Sequence Joint Attention.} By introducing 3D-RoPE to calculate joint attention $\text{Attn} = \text{Softmax}(QK^T/\sqrt{d} + \mathcal{M})V$, the model can calculate joint attention across the full sequence. Empirical studies (such as Physics-IQ~\cite{motamed2025physics}) indicate that decomposition architectures which sever spatiotemporal connections are mathematically difficult to approximate the convective terms and long-range correlations in Navier-Stokes equations. Only the global spatiotemporal receptive field provided by full attention mechanisms can capture such non-local physical interactions.

\textit{(\romannumeral2) Manifold Diffeomorphism.} On this basis, the generation process based on Flow Matching~\cite{lipman2023flow} corresponds mathematically to a diffeomorphism on the manifold, enabling the model to smoothly recover minute texture details from Gaussian noise, eliminating the edge flickering caused by discretization.

\paragraph{Computational Evolution: Linearization \& Inference Acceleration.}
Although DiT established the image quality benchmark, the quadratic complexity of Transformers ($O(N^2)$) causes VRAM usage to become a physical obstacle in moving from short clips to long videos.

\textit{(\romannumeral1) Linearization \& Caching.} On the architecture side, Video-TTT~\cite{dalal2025oneminute} introduced the Test-Time Training paradigm, compressing historical context into neural network weights, achieving memory retention for long videos while maintaining $O(N)$ linear complexity. Complementary to this, Pyramid Flow~\cite{jin2025pyramid} utilized the spatiotemporal redundancy of video, proposing a pyramid flow matching mechanism, reducing the computational cost of high quality video generation by 5-10 times through a hierarchical decoupling strategy. On the inference side, TeaCache~\cite{liu2025teacache} exploited the extremely high similarity of feature outputs in adjacent time steps in diffusion models (Pearson correlation coefficient $>0.98$), introducing a training-free dynamic caching mechanism to achieve 2-3 times end-to-end acceleration with zero image quality loss.

\paragraph{Convergence \& Divergence in Industry.}
The industry has not simply piled up parameters but has demonstrated three distinct evolutionary routes:

\textit{(\romannumeral1) Standardization vs. Heterogeneity.} Works represented by Meta Movie Gen~\cite{meta2024moviegen} established the standardized paradigm of DiT + Flow Matching, where the proposed temporally causal 3D VAE solved the temporal slice flickering problem in long videos. In contrast, Google DeepMind persisted with the Space-Time U-Net architecture in Lumiere and Veo~\cite{bar2024lumiere, google2025veo}, avoiding the temporal inconsistency caused by cascaded super-resolution through the full spatiotemporal attention mechanism, defining the upper limit of high-fidelity simulation quality.

\textit{(\romannumeral2) Ecosystem \& Controllability.} Application-layer models like Runway Gen-3~\cite{runway2024gen3} and ByteDance PixelDance~\cite{makepixelsdance} focus on fine-grained interaction, achieving complex instruction following through multimodal director modes and trajectory-level control. Meanwhile, open-source foundations like CogVideoX~\cite{cogvideox} and HunyuanVideo~\cite{hunyuan2024} lowered the fine-tuning threshold, directly promoting the development of the video fine-tuning ecosystem in the HuggingFace community.

\subsubsection{Logical Consistency and Causal Reasoning}
\label{subsubsec:unified_reasoning}

Although DiT-based generative models have solved visual continuity, they still face challenges when dealing with long-range physical logic (such as causal irreversibility). To bridge this gap, academia is shifting from a pure fitting paradigm to a cognitive reasoning paradigm, mainly manifested in the exploration of two complementary directions: image–text interleaving reasoning in multimodal perception models and temporal chain reasoning in generative video models.

\paragraph{Think-with-Image in Multimodal Perception.}

As the cognitive front-end of world models, LMMs are attempting to enhance logical capabilities by introducing the visual modality as Intermediate Reasoning Steps, rather than relying solely on text CoT. Works represented by Mini-O3~\cite{minio3} and VisCoT~\cite{viscot} assist logical jumps by generating or retrieving images during the inference process. RECAP~\cite{zhang2023multimodal} further formalized this flow, proposing a recursive Retrieve-Generate-Verify loop, utilizing visual information to compensate for text's deficiencies in spatial relation reasoning. UV-CoT~\cite{wang2024chain} explored image-text thought alignment under unsupervised conditions. Although these works mainly focus on the perception and understanding side, their image-assisted thinking mechanism provides valuable architectural insights for generative models tasked with complex spatiotemporal logic.

\paragraph{Chain-of-Frame \& Temporal Causality.}

On the generation side, the core of temporal consistency has ascended from visual fluency to event causality. The model must understand the sequence of occurrence of physical events, not just pixel interpolation. Video-CoT~\cite{videocot} and Video Espresso~\cite{qiu2024freenoise} introduce the Chain-of-Frame paradigm, which decomposes video generation into keyframe planning and intermediate frame synthesis. In contrast to pixel-level autoregressive approaches, this framework explicitly deduces future key states in the latent space, forcing the model to determine \textit{causal nodes} first, then generate the \textit{visual process}. Think Sound~\cite{liu2025thinksound} further extended this causality to the auditory modality, constraining the physical evolution of video via audio cues. By aligning the underlying causal graph structures across modalities, this approach enforces logical self-consistency throughout the full spatiotemporal span, mitigating the logical degradation that commonly emerges in long videos.

\subsection{Outlook of the Consistencies}
\label{subsec:integration_outlook}

Through the evolution of specialized models, three distinct computational engines have effectively emerged. \textit{Modal Consistency} has addressed semantic translation across modalities; \textit{Spatial Consistency} has progressed from coarse 2D approximations to explicit 3D primitives; and \textit{Temporal Consistency} has advanced from simple frame interpolation toward causal world simulation.

Yet treating these capabilities as independent optimization objectives introduces a fundamental bottleneck. A collection of highly specialized modules, regardless of individual sophistication, cannot constitute a coherent world simulator in the absence of a shared cognitive substrate. The central challenge therefore shifts from refining isolated components to achieving architectural unification. The future of world models lie in reaching a equilibrium in which semantic understanding, geometric structure, and causal reasoning co-emerge within a single parameter space. This requirement motivates the paradigm shift examined next: the emergence of the UMMs.

\section{Initial Integration of Multiple Consistencies}
\label{section:integration}

\subsection{The Rise of Large Multimodal Models}

In previous chapters, \textit{Modal, Spatial, and Temporal Consistency} were treated as independent technical dimensions. However, the construction of a general world model ultimately hinges not on the isolated advancement of these capabilities, but on their coherent integration into a unified cognitive system. Addressing this challenge requires moving beyond modular solutions toward architectures that can jointly reason across modalities, space, and time. The rise of Large Multimodal Models (LMMs), represented by LLaVA~\cite{liu2023llava} and GPT-4V~\cite{openai2023gpt4v}, marks a decisive paradigm shift from single-task specialists toward general cognitive entities.

\subsubsection{LLM as a Core Cognitive Base}

The core design philosophy of modern LMMs~\cite{openai2023gpt4, team2023gemini, su2023sigir} is to treat the pre-trained LLM~\cite{brown2020language, touvron2023llama} as a universal reasoning engine~\cite{wei2022neurips, kojima2022large}. Its essence lies in mapping heterogeneous modality data into the LLM's Word Embedding Space~\cite{li2023blip2, zhu2023arxiv}. This process is not a simple dimension transformation but is achieved through specific translator mechanisms (\eg visual connectors or adapters)~\cite{alayrac2022flamingo, liu2023llava, gao2023llamaadapter} to realize semantic alignment and conversion across modalities~\cite{radford2021icml}.

\paragraph{(1) Modal Tokenization \& Representation Bridging.}
In the specific implementation path, the model first utilizes a Visual Encoder (such as CLIP-ViT~\cite{radford2021icml} or SigLIP~\cite{zhai2023cvpr}) to extract high-dimensional feature maps $\mathcal{F} \in \mathbb{R}^{H \times W \times C}$. To enable the LLM to process these non-text signals, LLaVA~\cite{liu2023llava} and its subsequent improvements~\cite{liu2024llavanext, zhu2023arxiv} employ an MLP or Linear Projection Layer $\boldsymbol{W}_{\phi}$ to directly project image patch features into a set of Visual Tokens $\mathcal{V} = \{\boldsymbol{v}_1, \boldsymbol{v}_2, \dots, \boldsymbol{v}_n\}$ (where $\boldsymbol{v}_i \in \mathbb{R}^d$) that are dimensionally aligned with the text tokens. These tokens are then concatenated with text embeddings as soft prompts to form a hybrid input sequence:

\begin{equation}
    \boldsymbol{X}_{\textrm{input}} = \left[ \boldsymbol{e}_{\textrm{text}}^{(1)}, \dots, \boldsymbol{e}_{\textrm{text}}^{(m)}, \boldsymbol{v}_1, \dots, \boldsymbol{v}_n \right], \label{eq:multimodal_input}
\end{equation}

where $\boldsymbol{X}_{\textrm{input}}$ represents the aligned multimodal sequence, $\boldsymbol{e}_{\textrm{text}}$ denotes the text embeddings, and $\boldsymbol{v} \in \mathbb{R}^d$ is the visual token. From this perspective, the physical significance of alignment is to enable the LLM's self-attention mechanism to compute the association entropy between visual tokens in the same manner as it processes text tokens.

\paragraph{(2) From Rigid Projection to Perceiver Bottleneck.}
To address the issue of sequence length redundancy potentially caused by direct projection, BLIP-2~\cite{li2023blip2} and Flamingo~\cite{alayrac2022flamingo}—as representative architectures of Q-Former and Perceiver Resampler methods—utilize a fixed number of Learned Queries as intermediaries to filter out redundant information from massive Pixel Features. 

This mechanism is mathematically equivalent to a form of semantic pooling: it forces the model to compress thousands of Spatial Patches into dozens of tokens with highly abstract semantics. This not only resolves the problem of computational overhead but also theoretically satisfies the Information Bottleneck hypothesis~\cite{tishby2015deep}; by constraining the capacity of $I(Z; X_{vis})$, the model is forced to retain only those features conducive to Language Reasoning during the alignment process. Furthermore, experiments from DeepSeek-VL~\cite{lu2024deepseek} and InternVL~\cite{chen2023internvl} demonstrate that this alignment process can induce the formation of a cross-modal physical manifold within the LLM during the pre-alignment stage, allowing the model to maintain fundamental logical consistency even in unseen scenarios.

\subsubsection{Cognitive Evolution as a Multimodal}

The emergence of LMMs transcends the traditional end-to-end mapping paradigm~\cite{kim2021vilt, wang2022git}, endowing systems with resource scheduling and logic coordination capabilities akin to a multimodal operating system. Within this architecture, the LLM no longer functions merely as a feature processor but serves as the Kernel~\cite{yang2023mmreact, wu2023visual}, responsible for managing complex instruction flows and invoking heterogeneous Specialized Modules on demand~\cite{schick2023toolformer, meta2024chameleon, suris2023vipergpt}.

\begin{figure}[h]
  \centering
  \vspace{-2mm}
  \includegraphics[width=0.9\linewidth]{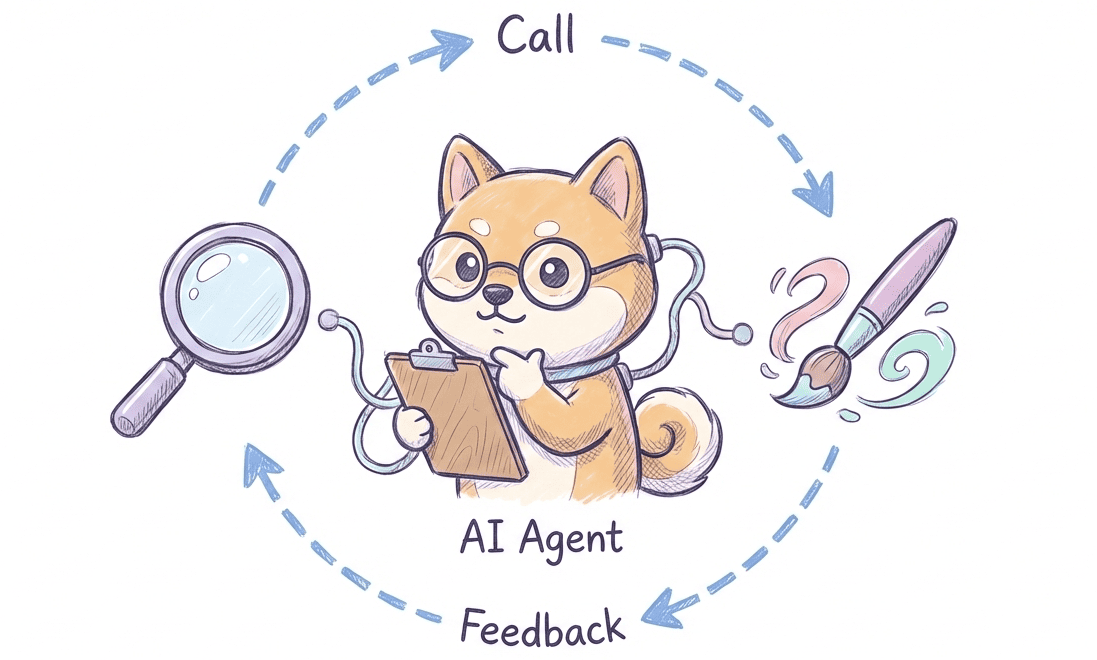} 
  \caption{Tool-use \& Closed-loop Verification. Diagram illustrating the ReAct paradigm, where an AI agent cyclically calls external detection and correction tools to refine generation outputs.}
  \label{fig:Agentic Closed-Loop Verification}
\end{figure}

\paragraph{(1) Hierarchical Task Planning \& Programmatic Instruction.} To address semantic drift in long-horizon tasks, LMMs demonstrate a capability for recursive decomposition, breaking down high-level ambiguous instructions into atomic sub-tasks. Distinct from earlier static mapping, VisProg~\cite{gupta2023visualprog} and ViperGPT~\cite{suris2023vipergpt} proposed the visual programmatic reasoning paradigm, which parses visual queries into executable python code flows, achieving logical self-consistency by combining low-level visual operators. The essence of this mechanism—transforming physical instructions into logical programs—is the utilization of the LLM's in-context learning to project open-domain problems onto a constrained operator space. Furthermore, PaLM-E~\cite{driess2023palm} and Voyager~\cite{wang2023voyager} have demonstrated that by incorporating real-time feedback from multimodal perception, LLMs can perform hierarchical search within a latent action space, maintaining long-term consistency in dynamic environments.

\paragraph{(2) Tool-use \& Closed-loop Verification.} To rectify physical hallucination during the generation process, LMMs have evolved a closed-loop refinement mechanism based on test-time compute. Frameworks represented by Visual ChatGPT~\cite{wu2023visual} and HuggingGPT~\cite{shen2024hugginggpt} utilize the ReAct (Reasoning and Acting) paradigm~\cite{yao2022react}, as illustrated in Figure~\ref{fig:Agentic Closed-Loop Verification}. This allows the model to actively suspend the generation path to invoke external expert models (\eg calling a detector to verify spatial relations or a diffusion model to redraw irrational textures). Architectures like Chameleon~\cite{meta2024chameleon} and Auto-GPT~\cite{richards2023autogpt} further introduce a feedback evaluation stage: by calculating the mutual information or geometric constraint deviation $\Delta_{\phi}$ between the generated intermediate state and the original instruction, the model can execute gradient-guided iterative refinement.

\subsection{Integration of Modal and Spatial Consistency}

\begin{figure}[h]
  \centering
  \includegraphics[width=0.9\linewidth]{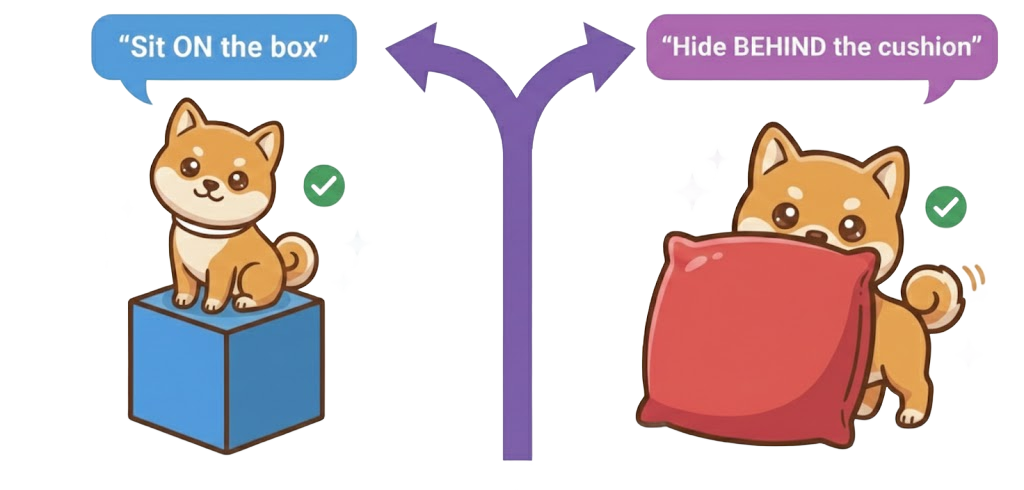}
  \caption{Modal + Spatial Consistency: Language Controls Precise Spatial Relations. The model combines modality and spatial consistency, using language instructions to precisely control the spatial relations between the subject (Doge) and objects (\eg Sit ON the box, Hide BEHIND the cushion).
  % , ensuring the accuracy of spatial interactions.
  }
  \label{fig:modality_spatial1}
\end{figure}

The fusion of modality and spatial consistency constitutes a core bridge toward physical world simulators~\cite{yang2023unisim, gaia1}. This profound cross-domain synergy aims to resolve the persistent issue of rich semantics but collapsed geometry in traditional generative models, with its core utility manifesting in two dimensions. In terms of semantic-spatial alignment, it empowers models with the capability for precise responses to complex spatial instructions (such as occlusion, surrounding, and perspective stacking)~\cite{li2023gligen}, achieving a qualitative leap in controllability from text describes texture to language defines layout~\cite{lian2024llmvideo}, as shown in Figure~\ref{fig:modality_spatial1}. In terms of geometric-physical grounding, it forces generated content to adhere to the geometric laws of the objective world, effectively eliminating structural non-rigid deformation and spatial misalignment hallucinations under multi-view conditions~\cite{liu2023zero1to3}. This integration ensures that AI is no longer confined to the statistical fitting of 2D pixels but possesses the capacity to infer spatiotemporal dynamics within a 3D manifold~\cite{videodiffusionmodels, bar2024lumiere}.

In current research, the deep integration of modality and spatial consistency presents four parallel technical paths, as illustrated in Figure~\ref{fig:modality-spatial-consistency-evolution}, exploring unique paradigms of implicit emergence, explicit synergy, structured isomorphism, and reinforcement learning. Pixel space manipulation focuses on leveraging the scale effects of large-scale multimodal corpora to internalize geometric transformations as implicit semantic mappings, achieving intuitive instruction as space control within universal generation~\cite{editworld, sgedit, mio}. In parallel, view space mapping introduces camera poses and depth maps as explicit geometric conditions, allowing semantic and geometric flows to co-exist and synergize on a 2D plane through cross-attention mechanisms, effectively balancing generative flexibility with perspective accuracy~\cite{liu2023zero1to3, chen2024moai, Long2024}. Meanwhile, volume space representation adopts the world coordinate system as its foundation, anchoring semantic features directly to neural volume fields or 3D Gaussian primitives, making spatial consistency an intrinsic physical attribute of the representation~\cite{openshape, lang3dxl}. Finally, reinforcement learning addresses the bag-of-words deficiencies in compositional instructions by introducing the System-2 Reasoning paradigm. Through region-temporal decoupling and inference-side scaling mechanisms, it elevates the generation process from pure statistical sampling to a planned solution equipped with logical verification~\cite{black2023training, chen2025ttpo, zeng2025layoutcot}. These four paradigms are not linear replacements but are complementary and symbiotic; they collectively expand the boundaries of semantics-space fusion in world models from the four dimensions of data-driven generalization, conditional control flexibility, physical modeling precision, and logical reasoning robustness.

\subsubsection{Pixel Space Manipulation}

The core philosophy of this paradigm lies in anchoring on data distribution, explicitly trading geometric priors for scale~\cite{poole2022dreamfusion, lin2023magic3d}. Unlike traditional graphics that rely on expensive, hard-coded geometric priors~\cite{mildenhall2020nerf, muller2022instant}, pixel space manipulation advocates for constructing a joint distribution $p_{\theta}(\boldsymbol{x}_{\textrm{img}}, \boldsymbol{c})$ of image-text and spatiotemporal data~\cite{ho2022imagen} based on pre-trained 2D generative bases (such as Latent Diffusion or Autoregressive Transformers)~\cite{esser2021taming, rombach2022high}.

Mathematically, this is equivalent to assuming that the massive volume of 2D projection data $\{\boldsymbol{x}_{\textrm{img}, i}\}_{i=1}^N$ is sufficient to cover the topological structure of the high-dimensional 3D manifold $\mathcal{M}_{\textrm{world}}$~\cite{liu2023zero1to3}. In this context, \textit{Modal Consistency} is manifested as the semantic alignment of conditional probability $p(\boldsymbol{x}_{\textrm{img}}|\boldsymbol{c}_{\textrm{sem}})$~\cite{radford2021icml, zhang2023controlnet}, while \textit{Spatial Consistency} spontaneously emerges as an outcome of optimizing the joint distribution when reconstruction error is minimized~\cite{blattmann2023align, guo2023animatediff}.

\definecolor{hidden-blue}{HTML}{4a90e2}
\definecolor{hidden-black}{HTML}{333333}
\tikzstyle{my-box}=[
  rectangle,
  draw=hidden-black,
  rounded corners,
  text opacity=1,
  minimum height=1.5em,
  minimum width=5em,
  inner sep=2pt,
  align=center,
  fill opacity=.5,
]\tikzstyle{root}=[
  align=center,
]\tikzstyle{leaf}=[
  my-box, 
  minimum height=1.5em,
  text=black,
  font=\normalsize,
  inner xsep=5pt,
  inner ysep=4pt,
  text width=10.5em,
  fill opacity=1,
]
\tikzstyle{leaf1}=[
  my-box, 
  minimum height=1.5em,
  fill=yellow!32, 
  text=black,
  font=\normalsize,
  inner xsep=5pt,
  inner ysep=4pt,
  text width=16em,   
]
\tikzstyle{leaf2}=[
  my-box, 
  minimum height=1.5em,
  fill=hidden-blue!57, 
  text=black,        
  font=\normalsize,
  inner xsep=5pt,
  inner ysep=4pt,
  text width=16em,
]
\tikzstyle{leaf3}=[
  my-box, 
  minimum height=1.5em,
  fill=purple!27, 
  text=black,
  font=\normalsize,
  inner xsep=5pt,
  inner ysep=4pt,
  text width=16em,
]
\tikzstyle{leaf4}=[
  my-box, 
  minimum height=1.5em,
  fill=green!20, 
  text=black,
  font=\normalsize,
  inner xsep=5pt,
  inner ysep=4pt,
  text width=16em,
]
\begin{figure*}[t]
\vspace{-2mm}
\centering
\resizebox{\textwidth}{!}{
\begin{forest}
  forked edges,
  for tree={
    grow=east,
    reversed=true,
    anchor=base west,
    parent anchor=east,
    child anchor=west,
    base=left,
    font=\large,
    rectangle,
    draw=hidden-black,
    rounded corners,
    align=left,
    minimum width=4em,
    edge+={darkgray, line width=1pt},
    s sep=3pt,
    inner xsep=5pt,
    inner ysep=4pt,
    line width=1.1pt,
    ver/.style={rotate=90, child anchor=north, parent anchor=south, anchor=center},
  },
  where level=1{text width=11em,font=\normalsize}{},
  where level=2{text width=13em,font=\normalsize}{},
  where level=3{text width=32.5em, tier=citations, font=\large}{},
  % 所有叶子都在 level=3，使用不同颜色区分
[Evolution of Modal Consistency \\and Spatial Consistency, ver, fill=gray!70, text=white, root
    % --- 2D: 像素空间操作 ---
    [Pixel Space\\Manipulation\\, fill=yellow!32, leaf
      [Instruction-Driven Image Editing, leaf1
        [{EditWorld~\cite{editworld}, Step-1X-Edit~\cite{step1xedit}, SGEdit~\cite{sgedit}, etc.}]
      ]
      [General Image Generation, leaf1, draw=hidden-black
        [{DreamLLM~\cite{dreamllm}, UniReal~\cite{unireal}, MENTOR~\cite{mentor}, \\MIO~\cite{mio}, etc.}]
      ]
    ]
    % --- 2.5D: 视图空间映射 ---
    [View Space Mapping\\, fill=hidden-blue!57, leaf
      [Pose-Aligned Coupled Training, leaf2
        [{Zero-1-to-3~\cite{liu2023zero1to3}, MoAI~\cite{chen2024moai}, Scaling Transformer~\cite{scalingtransformerbasednovelview},\\ Wonder3D~\cite{Long2024}, SyncDreamer~\cite{syncdreamer}, etc.}]
      ]
    ]
    % --- 3D: 体积空间表征 ---
    [Volume Space\\Representation\\, fill=purple!27, leaf
      [Conditional 3D Generation, leaf3
        [{DreamFusion~\cite{poole2022dreamfusion}, MVDream~\cite{shi2023MVDream}, RealFusion~\cite{realfusion}, \\Hunyuan3D-Omni~\cite{hunyuan3domni}, LangScene-X~\cite{langscenex}, GSFixer~\cite{gsfixer}, \\NeRF-HuGS~\cite{nerfhugs}, etc.}]
      ]
      [Multimodal Alignment, leaf3
        [{ULIP-2~\cite{ulip2},  OpenShape~\cite{openshape}, ShapeLLM-Omni~\cite{shapellmomni}, \\Genesis~\cite{genesis}, Viewset Diffusion~\cite{viewsetdiffusion}, etc.}]
      ]
      [3D Understanding or Editing, leaf3
        [{LERF~\cite{lerf}, Instruct-NeRF2NeRF~\cite{instructnerf2nerf}, CORE-3D~\cite{core3d}, \\CLIP-NeRF~\cite{clipnerf}, Lift3D~\cite{lift3d}, SKED~\cite{sked}, ICE-G~\cite{iceg}, \\Lang3D-XL~\cite{lang3dxl}, etc.}]
      ]
    ]
    [Reinforcement Learning, fill=green!20, leaf
        [{DDPO~\cite{black2023training}, AlignProp~\cite{prabhudesai2023alignprop}, R-DPO~\cite{gallego2024refined}, TTPO~\cite{chen2025ttpo}, \\SPO ~\cite{li2024spo}, Layout-CoT~\cite{zeng2025layoutcot}, etc.}, tier=citations, text width=32.5em, font=\large]
    ]
  ]
\end{forest}
}
\caption{Evolution of Modal Consistency and Spatial Consistency.}
\label{fig:modality-spatial-consistency-evolution}
\end{figure*}
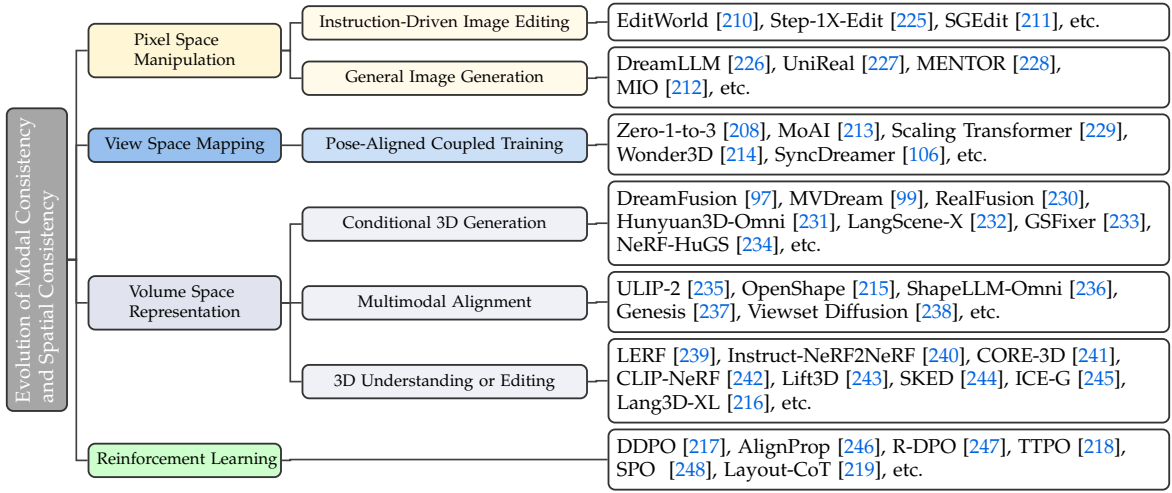

\paragraph{(1) Instruction-Driven Image Editing.}

\begin{figure}[h]
  \centering
  \vspace{-2mm}
  \includegraphics[width=0.9\linewidth]{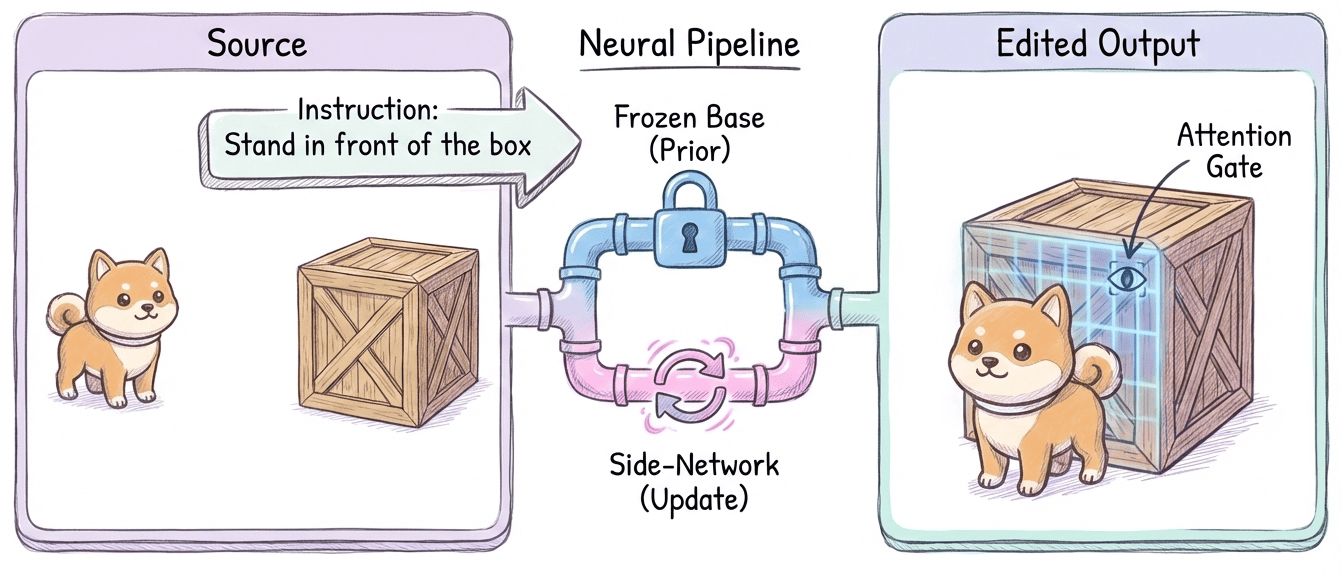}
  \caption{Instruction-Driven Image Editing. The diagram illustrating structure-preserving image editing, where a Doge is repositioned via text instruction. It depicts the mechanisms of gradient decoupling (frozen base) and attention injection (geometric gating) to maintain spatial consistency.}
  \label{fig:modality_spatial2}
\end{figure}

To address the common issue of geometric collapse in text-based editing (e.g. non-physical distortion of the background when instructing a dog to sit), instruction-driven image editing has established a hybrid paradigm integrating gradient decoupling \& update and attention injection as gating. This paradigm aims to resolve the intrinsic contradiction between semantic reconstruction and structural preservation by constructing orthogonal control paths, achieving a heavy semantic bridge~\cite{hertz2022prompt}, light weight structural constraint architecture as illustrated in Figure~\ref{fig:modality_spatial2}.

\textit{Gradient Decoupling \& Update.} To effectively decouple and protect the original spatial layout $\mathcal{S}_{\textrm{orig}}$ while injecting new semantics $\Delta \boldsymbol{c}$, mainstream paradigms (such as ControlNet~\cite{zhang2023controlnet} or IP-Adapter~\cite{ye2023ipadapter}) employ a structured decoupling architecture. By freezing the pre-trained base (\eg SDXL) and only fine-tuning the side-network or decoupled cross-attention, the model constructs a gradient update path on the parameter manifold that is orthogonal to the base:

\begin{equation}
    \nabla_{\boldsymbol{\theta}} \mathcal{L} = \underbrace{\nabla_{\boldsymbol{\theta}_{\textrm{base}}} \mathcal{L}_{\textrm{prior}}}_{\approx 0 \text{ (Frozen)}} + \underbrace{\nabla_{\boldsymbol{\theta}_{\textrm{adapter}}} \mathcal{L}_{\textrm{edit}}}_{\text{Semantics}}, \label{eq:gradient_decoupling}
\end{equation}
where $\boldsymbol{\theta}_{\textrm{base}}$ represents the frozen parameters of the base model, and $\boldsymbol{\theta}_{\textrm{adapter}}$ denotes the trainable parameters of the side-network. This ensures that the physical common sense (\eg lighting, occlusion) internalized within the base remains undisturbed.

\textit{Attention Injection as Gating.} During the inference phase, Prompt-to-Prompt~\cite{hertz2022prompt} and MasaCtrl~\cite{cao2023masactrl} reveal a strong correlation between cross-attention maps and spatial layouts. To maintain spatial consistency, the model injects the attention map $\boldsymbol{M}_{\textrm{attn}}^{\textrm{src}}$ of the original image into the editing steps as a geometric hard-gating mechanism:

\begin{equation}
    \textrm{Attn}_{\textrm{edit}}(\boldsymbol{Q}, \boldsymbol{K}, \boldsymbol{V}) \leftarrow \alpha \cdot \textrm{Softmax}\left(\frac{\boldsymbol{Q} \boldsymbol{K}^T}{\sqrt{d}}\right) + (1-\alpha) \cdot \boldsymbol{M}_{\textrm{attn}}^{\textrm{src}}, \label{eq:attn_injection}
\end{equation}
where $\boldsymbol{M}_{\textrm{attn}}^{\textrm{src}}$ denotes the attention map preserved from the source image to guide spatial layout, and $\alpha$ is the injection strength coefficient. Combined with the MLLM Semantic Hub mechanism proposed by Step-1X Edit~\cite{step1xedit}, this method successfully achieves semantic change with topological conservation. Subsequent work such as EditWorld~\cite{editworld} further introduced a post-edit closed-loop, utilizing SAM masks for second-order geometric verification to resolve pixel artifacts at object edges.

\paragraph{(2) General Image Generation.}

General image generation is undergoing a paradigm reconstruction from external plug-in alignment to native full-duplex modeling, aiming to directly capture the spatiotemporal dynamics distribution of the physical world through end-to-end joint training. The paradigm shift in this field is characterized as: From external alignment (CLIP-based) to End-to-End Interleaved Modeling. This transition no longer relies on frozen feature extractors but instead constructs a generative foundation where modality and space are tightly coupled through joint modeling~\cite{dreamllm}, video stream supervision~\cite{kipf2019contrastive}, and lightweight connections~\cite{mentor}.

\textit{(\romannumeral1) Joint Modeling Breaking Information Bottleneck.} Traditional two-stage models (such as DALL-E 2) are limited by the modality isolation of the CLIP encoder, which results in the loss of spatial relations during feature compression. A new generation of models, such as DreamLLM~\cite{dreamllm} and Emu~\cite{sun2024emu}, abandons this design in favor of directly performing joint modeling on raw image-text sequences using autoregressive or diffusion approaches:
\begin{equation}
    \mathcal{L}_{\textrm{joint}} = -\sum_{t} \log p\left(\boldsymbol{x}_{\textrm{img}, t} \mid \boldsymbol{x}_{\textrm{img}, <t}, \boldsymbol{x}^{\textrm{txt}}\right) - \sum_{j} \log p\left(\boldsymbol{x}_j^{\textrm{txt}} \mid \boldsymbol{x}_{\textrm{img}}, \boldsymbol{x}_{<j}^{\textrm{txt}}\right), \label{eq:joint_modeling}
\end{equation}
where $\mathcal{L}_{\textrm{joint}}$ denotes the unified training objective, $\boldsymbol{x}_{\textrm{img}, t}$ represents the image tokens at step $t$, and $\boldsymbol{x}^{\textrm{txt}}$ corresponds to the text tokens. This full-duplex information flow enables the model to capture pixel-level spatial constraints implicit in descriptions such as ``a cat on a table."

\textit{(\romannumeral2) Video as World Simulator.} Transcending simple geometric perspective transformations, empirical research on Sora~\cite{openai2024sora} reveals the profound value of video data: it provides endogenous supervision signals regarding physical plausibility.

Unlike static images, temporal dependencies in video streams force the model to learn object permanence~\cite{kipf2019contrastive, locatello2020object}—for instance, inferring that an occluded object has not disappeared but continues to move along its trajectory. This self-supervision compels the model to construct a dynamics model within the latent space that conforms to physical conservation laws (\eg gravity, collision, fluid dynamics)~\cite{ha2018world}, thereby elevating the generative model from mere pixel statistical fitting to a predictive simulation of physical world evolution~\cite{yang2023unisim}.

\textit{(\romannumeral3) Lightweight Connection Layer.} To balance computational efficiency with multimodal alignment, the perceiver resampler in Flamingo~\cite{alayrac2022flamingo} and the MLP connection layer design in Mentor~\cite{mentor} demonstrate how visual features can be projected onto the LLM’s semantic manifold using minimal parameters. This proves that as long as the base is sufficiently powerful, simple linear mappings can maintain complex space-semantics correspondence.

\subsubsection{View Space Mapping}

\begin{figure}[h]
  \centering
  \includegraphics[width=0.9\linewidth]{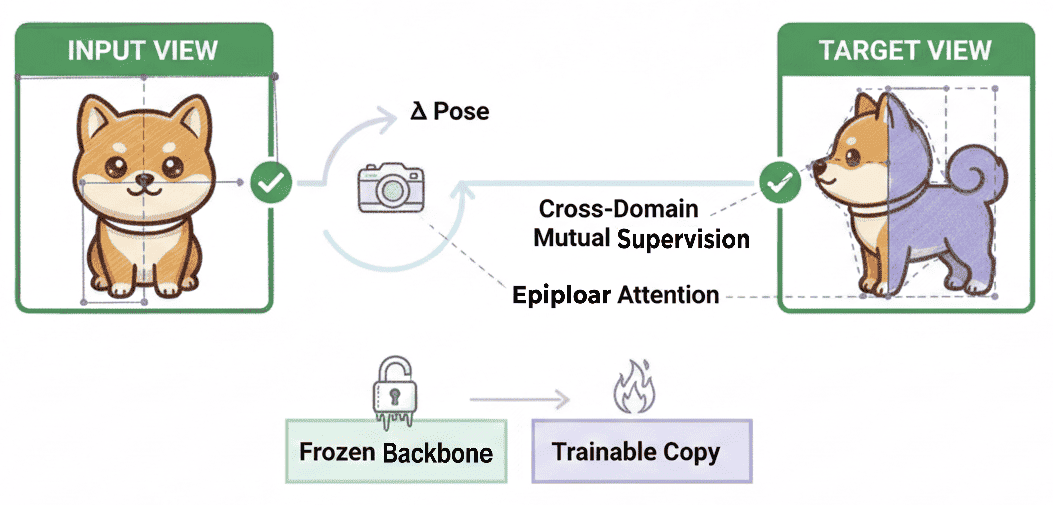} 
  % \caption{Pose-Aligned View Synthesis. The diagram illustrating pose-conditioned generation using a decoupled backbone and epipolar attention constraints. The target view features a split display of RGB texture and a purple-blue Normal Map, representing cross-domain mutual supervision for geometric accuracy.}
  \caption{Pose-Aligned View Synthesis. Diagram illustrating pose-conditioned generation using a decoupled backbone and epipolar attention. The target view features a split display of RGB texture and a purple-blue Normal Map, representing cross-domain mutual supervision for geometric accuracy.}
  \label{fig:modality_spatial3}
\end{figure}

\paragraph{Pose-Aligned Coupled Training.}

The core philosophy of this paradigm lies in abandoning purely data-driven black-box assumptions and injecting 3D geometric information as structured condition variables $\tau = \{\boldsymbol{P}_t, \mathcal{D}\}$ (where $\boldsymbol{P}_t \in SE(3)$ is the camera pose and $\mathcal{D}$ is the depth prior) into a pre-trained diffusion model~\cite{zhang2023controlnet, liu2023zero1to3}, as illustrated in Figure~\ref{fig:modality_spatial3}. Its mathematical essence is the construction of a conditional denoising distribution constrained by geometry:

\begin{equation}
    \mathcal{L}_{\textrm{view}} = \mathbb{E}_{\boldsymbol{z}, t, \boldsymbol{c}, \tau, \boldsymbol{\epsilon}} \left[ \| \boldsymbol{\epsilon} - \boldsymbol{\epsilon}_{\boldsymbol{\theta}}(\boldsymbol{z}_t, t, \boldsymbol{c}, \tau) \|_2^2 \right] + \lambda \mathcal{R}_{\textrm{consist}}, \label{eq:view_loss}
\end{equation}
where $\boldsymbol{z}_t$ represents the noisy latent at timestep $t$, $\boldsymbol{\epsilon}_{\boldsymbol{\theta}}$ is the noise prediction network conditioned on geometry $\tau$, and $\mathcal{R}_{\textrm{consist}}$ denotes the regularization term for multi-view consistency. The successful implementation of this paradigm relies on the following three synergistic mechanisms:

\textit{(\romannumeral1) Backbone Decoupling \& Injection.} To circumvent catastrophic forgetting while preserving the semantic generation capability of the pre-trained model, the academic community has established a design of frozen backbone and bypass control. Represented by Zero-1-to-3~\cite{liu2023zero1to3} and ControlNet~\cite{zhang2023controlnet}, this approach achieves selective gradient flow by locking the backbone network $\mathcal{F}_{\textrm{locked}}$ and introducing a trainable copy $\mathcal{F}_{\textrm{copy}}$: $\boldsymbol{h}_{\textrm{out}} = \mathcal{F}_{\textrm{locked}}(\boldsymbol{h}_{\textrm{in}}) + \mathcal{Z}(\mathcal{F}_{\textrm{copy}}(\boldsymbol{h}_{\textrm{in}}, \tau))$. This zero convolution strategy ensures that the model generates photo-realistic textures while precisely executing geometric instructions.

\textit{(\romannumeral2) Structured Sparse Attention.} To address the janus problem in multi-view generation, models introduce structured sparse attention. MVDream~\cite{shi2023MVDream} and SyncDreamer~\cite{syncdreamer} innovatively transform the epipolar geometry constraints in 3D space into an attention mask:

\begin{equation}
    \textrm{Attn}(\boldsymbol{Q}_i, \boldsymbol{K}_j, \boldsymbol{V}_j) \propto \exp \left( \frac{\boldsymbol{Q}_i \boldsymbol{K}_j^T}{\sqrt{d}} + \mathcal{M}_{\textrm{epi}}(i, j) \right), \label{eq:epipolar_attn}
\end{equation}
where $\boldsymbol{Q}_i$ and $\boldsymbol{K}_j$ denote features from view $i$ and view $j$, respectively, and $\mathcal{M}_{\textrm{epi}}$ represents the geometric bias derived from epipolar constraints. This mechanism forces tokens from different views to interact only with their geometrically corresponding epipolar line regions, thereby converting geometric hard constraints into a soft inductive bias within the attention mechanism.

\textit{(\romannumeral3) Cross-Domain Attention Regularization.} To further enhance geometric accuracy, Wonder3D~\cite{Long2024} and MoAI~\cite{chen2024moai} achieve mutual supervision between texture semantics and geometric structure by generating RGB and normal maps in parallel and introducing cross-domain attention injection $\boldsymbol{f}_{\textrm{rgb}} \leftrightarrow \boldsymbol{f}_{\textrm{geo}}$. Coupled with a 3D consistent noise initialization strategy (initializing noise based on the camera projection matrix), this paradigm successfully breaks the i.i.d. assumption from the initial state, achieving a transition from simple image generation to geometrically controllable generation.

\subsubsection{Volume Space Representation}

Unlike the previous two paradigms that simulate 3D on a 2D plane, volume space representation chooses to directly confront the three-dimensional essence of objects~\cite{poole2022dreamfusion, realfusion}. The core philosophy of this direction is to utilize 3D Native Representations (NeRF, 3D Gaussian Splatting) as the primary layer of architectural abstraction. This makes spatial consistency an intrinsic property of the representation, while modal consistency is transformed into a synergistic optimization problem between cross-modal queries and differentiable rendering.

\paragraph{(1) Conditional 3D Generation: From 2D Distillation to Video Manifold Constraints.}

Conditional 3D generation aims to overcome the bottleneck of 3D data scarcity by restructuring pre-trained generative models as frozen cognitive engines, establishing a technical trajectory that evolves from 2D semantic distillation toward video manifold constraints. Due to the extreme scarcity of high-quality 3D-text data pairs (which are 2–3 orders of magnitude fewer than 2D data), this direction no longer seeks to train 3D generators from scratch. Instead, it focuses on discovering and transferring the spatial intelligence inherent in pre-trained 2D or video models~\cite{poole2022dreamfusion, ma2025youseeit}.

\textit{(\romannumeral1) Gradient Flow from 2D Priors.} DreamFusion~\cite{poole2022dreamfusion} and RealFusion~\cite{realfusion} established the foundational formula for this field: Score Distillation Sampling (SDS). Its core principle is not to optimize pixel error, but to optimize a parameterized 3D field $\theta$ (such as NeRF or 3DGS) such that the image rendered from any viewpoint, $\boldsymbol{x}_{\textrm{img}} = g(\theta, \boldsymbol{P}_t)$, resides in the low-energy regions of a 2D diffusion model:

\begin{equation}
    \nabla_{\boldsymbol{\theta}} \mathcal{L}_{\textrm{SDS}} = \mathbb{E}_{t, \boldsymbol{\epsilon}} \left[ w_{\textrm{guidance}} \left( \boldsymbol{\epsilon}_{\boldsymbol{\theta}}(\boldsymbol{z}_t, t) - \boldsymbol{\epsilon} \right) \frac{\partial \boldsymbol{x}_{\textrm{img}}}{\partial \boldsymbol{\theta}} \right], \label{eq:sds_loss}
\end{equation}
where $w_{\textrm{guidance}}$ is the weighting factor, $\boldsymbol{\epsilon}_{\boldsymbol{\theta}}$ is the predicted noise from the frozen diffusion model, and $\frac{\partial \boldsymbol{x}_{\textrm{img}}}{\partial \boldsymbol{\theta}}$ represents the Jacobian of the differentiable renderer. This formula indicates that the semantic residual computed by the 2D model is backpropagated through the Jacobian matrix $\frac{\partial \boldsymbol{x}_{\textrm{img}}}{\partial \boldsymbol{\theta}}$ of the differentiable renderer $g$ to directly sculpt the 3D geometry.

\textit{(\romannumeral2) Video Manifold as Dynamic 3D Prior.} To address the janus problem caused by 2D priors, recent research has shifted toward leveraging the physical consistency inherent in Video Diffusion Models (VDMs). The core hypothesis is that Temporal Correlation $\cong$ Spatial Consistency. See3D~\cite{ma2025youseeit} and V3D~\cite{chen2025v3d} propose utilizing video generative models as multi-view generators. By fine-tuning the VDM, the time axis $T$ is implicitly reconstructed as a camera trajectory $\boldsymbol{P}_t$ (\eg an orbital viewpoint):

\begin{equation}
    p(\boldsymbol{x}_{\textrm{img, novel}} \mid \boldsymbol{x}_{\textrm{img, ref}}) \approx p_{\textrm{video}}(\boldsymbol{x}_{\textrm{img}, t+1} \mid \boldsymbol{x}_{\textrm{img}, t}, \text{motion\_cond}), \label{eq:video_prior}
\end{equation}
where $p_{\textrm{video}}$ denotes the transition probability learned by the video model, and $\text{motion\_cond}$ represents the camera trajectory condition. Under this paradigm, SV3D~\cite{Stability_SV3D} utilizes the temporal attention layer of the video model as a soft epipolar constraint, forcing the generation of a multi-view sequence with geometric continuity. Subsequently, SDS is used to distill this dynamic video prior into static 3D assets, fundamentally resolving viewpoint conflicts.

\textit{(\romannumeral3) Prior-Constraint Two-Stage Loop.} Given the ill-posedness of single-view generation, Magic123~\cite{qian2024magic123} and One-2-3-45~\cite{Liu2023_One2345} established the paradigm of coarse generation $\to$ fine optimization. Current trends involve using video models~\cite{hunyuan3domni} to rapidly generate multi-views as an initial guess, followed by geometry refinement using SDS in combination with a lightweight solver~\cite{shen2021dmtet}. This strategy of video initialization and physics fine-tuning preserves semantic richness while utilizing video priors to rectify the topological plausibility of the 3D structure.

\paragraph{(2) Multimodal Alignment.}

\begin{figure}[h]
  \centering
  \includegraphics[width=0.98\linewidth]{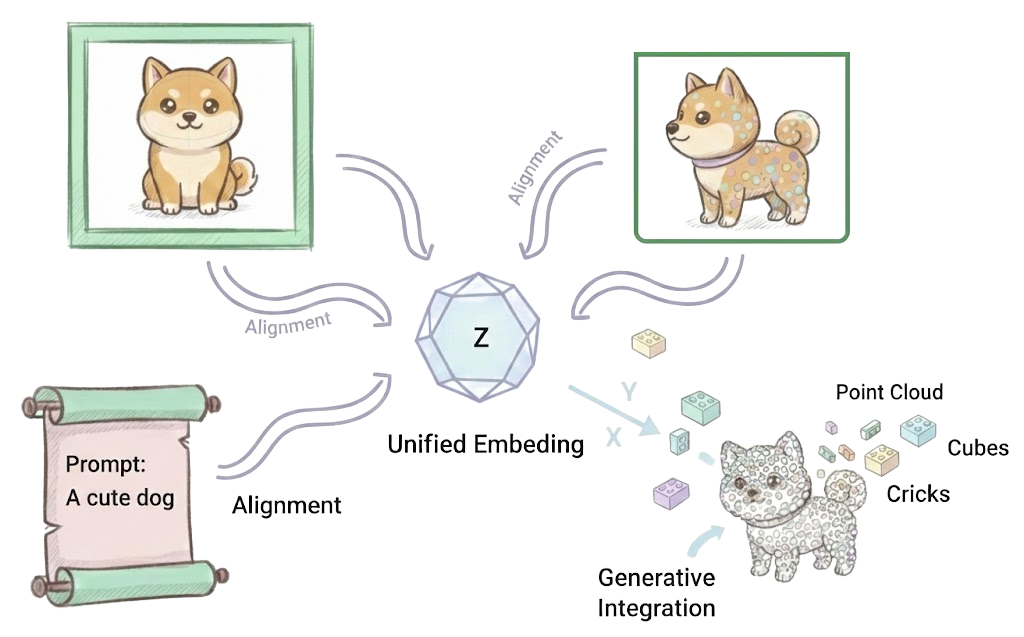}
  \caption{Multimodal Alignment. The diagram illustrating the convergence of Text, Image, and 3D Point Cloud into a central Unified Embedding. It depicts contrastive alignment for inputs and generative tokenization for 3D data integration.}
  \label{fig:modality_spatial4}
\end{figure}

Multimodal alignment aims to construct universal representations that span geometry and semantics. By establishing a dual-track mechanism of discriminative metric alignment and generative interaction fusion, it breaks the long-standing representation silo dilemma of 3D data. To process 3D data as CLIP processes images, this direction focuses on building a Unified Embedding Space $\mathcal{Z}_{\textrm{uni}}$. The technical philosophy is to transform spatial consistency into structured constraints during network forward propagation, as shown in Figure~\ref{fig:modality_spatial4}.

\textit{Contrastive Metric Learning.} ULIP-2~\cite{ulip2} and OpenShape~\cite{openshape} employ large-scale triplet contrastive learning. By mining hard negatives and utilizing the InfoNCE loss, the feature distribution of the 3D encoder (PointNet++ or Transformer) is forced to align with CLIP's text/image space:

\begin{equation}
    \mathcal{L}_{\textrm{align}} = -\log \frac{\exp \left( \boldsymbol{z}_{\textrm{3D}} \cdot \boldsymbol{z}_{\textrm{txt}} / \tau \right)}{\sum_{j} \exp \left( \boldsymbol{z}_{\textrm{3D}} \cdot \boldsymbol{z}_{\textrm{txt}}^j / \tau \right)}, \label{eq:contrastive_loss}
\end{equation}
where $\boldsymbol{z}_{\textrm{3D}}$ and $\boldsymbol{z}_{\textrm{txt}}$ represent the feature embeddings of the 3D shape and text, respectively, and $\tau$ is the temperature parameter. Genesis~\cite{genesis} further extends this to 4D spatiotemporal alignment by introducing cross-view attention in voxel space to fuse video and LiDAR modalities, achieving alignment across space-time dimensions.

\textit{Generative Integration.} Unlike the holistic alignment of contrastive learning, ShapeLLM-Omni~\cite{shapellmomni} and ViewSetDiffusion~\cite{viewsetdiffusion} introduce 3D VQ-VAE to discretize continuous geometry into token sequences. This enables the LLM to directly read and generate 3D geometry, achieving generative interaction between modalities rather than simple retrieval matching.

\paragraph{(3) 3D Understanding \& Editing: Semantic Lifting.}

3D understanding and editing aim to endow 3D geometric entities with the dual capabilities of semantic perception and linguistic manipulation. The core paradigm involves injecting the cognitive priors of 2D vision foundation models into 3D space via Semantic Lifting, constructing a mapping $F(x,y,z) \mapsto \mathbb{R}^{D_{\textrm{clip}}}$.

\textit{Semantic Field Construction.} LERF~\cite{lerf} and Lang3D-XL~\cite{lang3dxl} propose training a semantic head in parallel with the color head of a NeRF. This module learns CLIP feature fields through multi-scale supervision, enabling every coordinate point $\boldsymbol{p}(x,y,z)$ in space to respond to natural language queries (\eg Find the crack on the chair). SKED~\cite{sked} and CoRe-3D~\cite{core3d} introduce hierarchical semantic fields, embedding instance-part-material hierarchies into the representation to solve fine-grained semantic localization problems.

\textit{Language-Driven Topology Editing.} For editing tasks, CLIP-NeRF~\cite{clipnerf} utilizes decoupled latent mapping to achieve near-instant modifications of shape and appearance. InstructNeRF2NeRF~\cite{haque2023instructnerf2nerf} employs an iterative dataset update strategy: it first modifies the rendering view images using InstructPix2Pix and then uses the modified images as Pseudo-GT to back-update the NeRF. Lift3D~\cite{lift3d} and ICE-G~\cite{iceg} introduce canonical space constraints to ensure that topological structures do not collapse even during significant geometric deformations, such as instructing a cat to stand up.

\subsubsection{Reinforcement Learning for Modal-Spatial Alignment}

Despite the explicit geometric conditions provided by architectures such as ControlNet~\cite{zhang2023controlnet}, LMMs still frequently exhibit severe modal-spatial misalignment when processing compositional instructions (\eg attribute binding: \textit{Red cat on blue car})~\cite{li2023gligen}. To address this bag-of-words model deficiency, the academic community is undergoing a paradigm shift from black-box optimization toward System-2 Reasoning~\cite{lian2024llmvideo}.

This paradigm evolution can be summarized into three stages:

\paragraph{Discriminator-Guided Explicit Anchoring.}
Early efforts focused on utilizing off-the-shelf visual discriminators as an external reward function $\mathcal{R}$ to forcibly establish the correspondence between text prompts and bounding boxes.

\textit{Black-box Discrete Optimization.} DDPO~\cite{black2023training} models diffusion denoising as a Markov Decision Process (MDP). For spatial instructions, it introduces an open-vocabulary detector (such as GroundingDINO~\cite{liu2023grounding}) to compute an IoU reward. This represents a loosely-coupled fusion; while it enhances object recall, the sparsity of the reward signal makes it difficult to resolve complex attribute binding.

\textit{White-box Gradient Backpropagation.} AlignProp~\cite{prabhudesai2023alignprop} proposes fine-tuning the discriminator into a differentiable reward model. This establishes an end-to-end gradient path $\nabla_{\textrm{pixels}} \mathcal{L}_{\textrm{align}}$, allowing spatial errors to back-propagate directly to the denoising network, thereby achieving pixel-level precision in refinement.

\paragraph{Region-Temporal Decoupling.}
To prevent global rewards from confusing semantics with spatial information, subsequent work shifted toward fine-grained control.

R-DPO~\cite{gallego2024refined} proposed sub-manifold optimization under spatial masks. Unlike traditional DPO, it decomposes the image $\boldsymbol{x}_{\textrm{img}}$ and text $\boldsymbol{c}$ into several local pairs $(\boldsymbol{x}_{\textrm{crop}}^k, \boldsymbol{c}_{\textrm{sub}}^k)$, ensuring that specific modal attributes (\eg Red) only back-propagate to specific spatial regions (\eg within the coordinates of the Cat):

\begin{equation}
    \mathcal{L}_{\textrm{R-DPO}} = -\sum_{k} \mathbb{E}_{(\boldsymbol{x}_{\textrm{w}}, \boldsymbol{x}_{\textrm{l}}) \sim B_k} \left[ \log \sigma \left( \beta \log \frac{\pi_{\boldsymbol{\theta}}(\boldsymbol{x}_{\textrm{img, w}}^k \mid \boldsymbol{c}_k)}{\pi_{\textrm{ref}}(\boldsymbol{x}_{\textrm{img, w}}^k \mid \boldsymbol{c}_k)} - \beta \log \frac{\pi_{\boldsymbol{\theta}}(\boldsymbol{x}_{\textrm{img, l}}^k \mid \boldsymbol{c}_k)}{\pi_{\textrm{ref}}(\boldsymbol{x}_{\textrm{img, l}}^k \mid \boldsymbol{c}_k)} \right) \right], \label{eq:rdpo_loss}
\end{equation}
where $B_k$ represents the local preference dataset for region $k$, $\boldsymbol{x}_{\textrm{w}}$ and $\boldsymbol{x}_{\textrm{l}}$ denote the winning and losing image crops respectively, and $\sigma$ is the sigmoid function.

Concurrently, SPO~\cite{liang2025spo} leverages the frequency characteristics of diffusion models by adopting a time-division multiplexing strategy: focusing on spatial IoU optimization during the early stages of denoising ($t \in [T, T/2]$) and switching to semantic optimization in the later stages ($t \in [T/2, 0]$) to avoid gradient conflicts. Furthermore, DRaFT~\cite{clark2024directly} utilizes VLMs to generate natural language critiques regarding spatial errors and maps them to a dense reward map $\mathcal{R} \in \mathbb{R}^{H \times W}$. This marks the transition of RL alignment from discrete boxes to continuous pixel fields, enabling generative models to comprehend extremely subtle spatial-modal instructions such as the left leg is distorted.

\paragraph{TTT \& Visual CoT.}
Following the success of DeepSeek-R1 and OpenAI o1 in demonstrating the efficacy of inference-side scaling, recent research has begun introducing RL into the inference-time stage of generation, achieving a strong logical fusion between modality and space.

\textit{Test-Time Preference Optimization (TTPO).} To address the insufficient perceptual quality of pre-trained models under specific distributions, TTPO~\cite{chen2025ttpo} proposes an on-the-fly optimization mechanism. This method avoids heavy re-training by using a lightweight reward model (such as an image quality score) to iteratively update the latent variable $\boldsymbol{z}$ during the inference stage. While this work primarily validates its effectiveness in image restoration tasks, this test-time fine-tuning paradigm provides a general compute-for-quality path for resolving highly counter-intuitive generative tasks.

\textit{Visual Chain-of-Thought (Visual CoT).} To resolve the logic breaks inherent in one-step generation, Layout-CoT~\cite{zeng2025layoutcot} borrow the reasoning paradigm of LLMs. This approach decomposes the generation process into an explicit chain: \texttt{Planning $\to$ Alignment $\to$ Generation}. The model first generates a discrete layout plan in a low-dimensional space and employs RL to perform logical verification on this plan. Only chains-of-thought that pass verification are decoded into pixels. This mechanism essentially moves System-2 logic verification to the front end, fundamentally eliminating hallucinations such as interpenetration or spatial misalignment.

\subsection{Integration of Modal and Temporal Consistency}

\begin{figure}[h]
  \centering
  \includegraphics[width=0.9\linewidth]{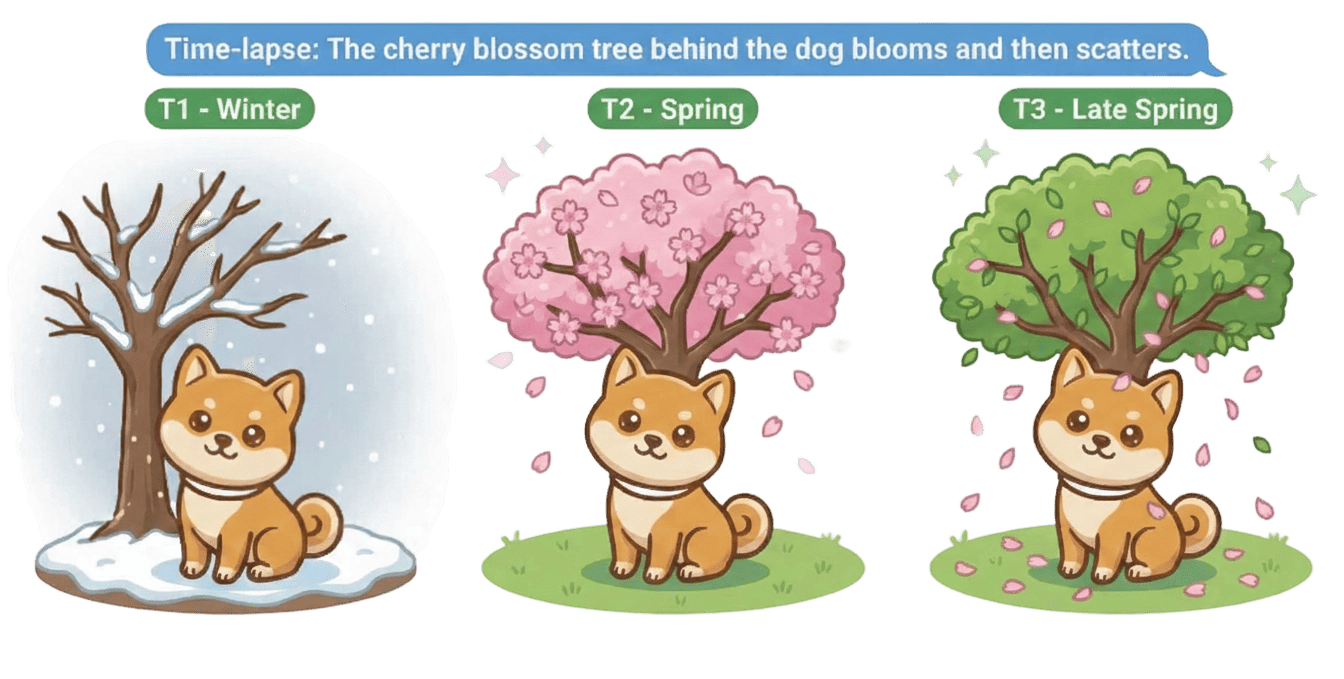} 
  \caption{Modal + Temporal Consistency: Language Controls Time Evolution. The model integrates modality and temporal consistency, using language instructions to control the time evolution process (\eg the cherry blossom tree behind the dog blooms and scatters across T1-Winter, T2-Spring, T3-Late Spring), ensuring coherent changes over time.}
  \label{fig:modality_temporal1}
\end{figure}

The deep integration of modality and temporal consistency marks the formal transition of Generative AI from the Frozen Moment of static images toward the Continuous Deduction of the dynamic world as illustrated in Figure~\ref{fig:modality_temporal1}~\cite{openai2024sora}. The core utility of this dimension lies in constructing a Probabilistic Simulation of Spatiotemporal Causality: at the Semantic Level, it ensures that video content strictly adheres to the definitions of text or image instructions (\eg Blooming, Running), thereby eliminating cross-modal semantic drift~\cite{videocomposer, bar2024lumiere}; at the Dynamics Level, it endows the model with an endogenous understanding of Object Permanence and Physical Conservation Laws, ensuring that the generated frame sequence is no longer a random stacking of discrete pixels, but rather a Continuous Manifold consistent with logical evolution~\cite{ho2022imagen, blattmann2023stable}. This fusion fundamentally resolves chronic issues in traditional video generation, such as motion flickering, temporal logic chaos, and long video collapse.

Based on this objective, current exploration paths present Four Progressive Technical Paradigms as shown in Figure~\ref{fig:modality-temporal-consistency-evolution}: End-to-End Scalable Modeling follows the data philosophy of ``Brute Force with Data,'' relying on Diffusion Models and Autoregressive Architectures to validate the Scaling Law, aiming to learn a general physical simulator directly from massive data~\cite{videodiffusionmodels, ovi, hybridvla}; Explicit Structured Control targets the controllability requirements of industrial applications by introducing motion vectors, trajectory heatmaps, and orthogonal decoupling mechanisms to explicitly inject human intent into the generation process, addressing the ambiguity issues of end-to-end models~\cite{vast2024, makepixelsdance, fancyvideo}; meanwhile, the Unified Comprehension and Generation Symbiosis Architecture attempts to break the barriers between perception and generation through shared representation and bi-directional adaptation, constructing a general agent with a closed loop of perceiving and acting~\cite{phenaki, omnivideo}; finally, Reinforcement Learning Driven Alignment addresses the non-convexity of SFT (Supervised Fine-Tuning) in optimizing modal semantic and ``temporal dynamics'' by constructing a Multi-dimensional Reward Manifold. By integrating DPO and Self-Refinement mechanisms, this paradigm achieves joint optimization of alignment targets, driving the model to surpass binary games and converge to the Pareto Frontier of spatiotemporal trade-offs~\cite{liu2025videodpo, cheng2025vpo}. These four paradigms collectively build a complete architecture for modality and temporal intelligence from the dimensions of General Foundation, Controllable Interface, Cognitive Top-level, and Value Optimization.

\definecolor{hidden-blue}{HTML}{4a90e2}
\definecolor{hidden-black}{HTML}{333333}
% 定义基本样式
\tikzstyle{my-box}=[
  rectangle,
  draw=hidden-black,
  rounded corners,
  text opacity=1,
  minimum height=1.5em,
  minimum width=5em,
  inner sep=2pt,
  align=center,
  fill opacity=.5,
]
\tikzstyle{root}=[
  align=center,
]
\tikzstyle{leaf}=[
  my-box, 
  minimum height=1.5em,
  text=black,
  font=\normalsize,
  inner xsep=5pt,
  inner ysep=4pt,
  text width=10.5em,
  fill opacity=1,
]
\tikzstyle{leaf1}=[
  my-box, 
  minimum height=1.5em,
  fill=yellow!32, 
  text=black,
  font=\normalsize,
  inner xsep=5pt,
  inner ysep=4pt,
  text width=16em,   
]
\tikzstyle{leaf2}=[
  my-box, 
  minimum height=1.5em,
  fill=hidden-blue!57, 
  text=black,        
  font=\normalsize,
  inner xsep=5pt,
  inner ysep=4pt,
  text width=16em,
]
\tikzstyle{leaf3}=[
  my-box, 
  minimum height=1.5em,
  fill=purple!27, 
  text=black,
  font=\normalsize,
  inner xsep=5pt,
  inner ysep=4pt,
  text width=16em,
]
\tikzstyle{leaf4}=[
  my-box, 
  minimum height=1.5em,
  fill=green!20, 
  text=black,
  font=\normalsize,
  inner xsep=5pt,
  inner ysep=4pt,
  text width=16em,
]
\begin{figure*}[t]
\vspace{-2mm}
\centering
\resizebox{\textwidth}{!}{
\begin{forest}
  forked edges,
  for tree={
    grow=east,
    reversed=true,
    anchor=base west,
    parent anchor=east,
    child anchor=west,
    base=left,
    font=\large,
    rectangle,
    draw=hidden-black,
    rounded corners,
    align=left,
    minimum width=4em,
    edge+={darkgray, line width=1pt},
    s sep=3pt,
    inner xsep=5pt,
    inner ysep=4pt,
    line width=1.1pt,
    ver/.style={rotate=90, child anchor=north, parent anchor=south, anchor=center},
  },
  where level=1{text width=12.5em,font=\normalsize}{},
  where level=2{text width=12.5em,font=\normalsize}{},
  where level=3{text width=33.5em, tier=citations, font=\large}{},
[Evolution of Modal Consistency \\and Temporal Consistency, ver, fill=gray!70, text=white, root
    [End-to-End Scalable\\Modeling, fill=yellow!32, leaf
      [Diffusion Model, leaf1
        [{Sora~\cite{sun2024sora}, Imagen Video~\cite{ho2022imagen}, Wan 2.1~\cite{wan2025}, Vidu~\cite{vidu},\\CogVideo~\cite{hong2022cogvideo}, Make-A-Video~\cite{singer2022make}, UniVG~\cite{univg}, Pika~\cite{pika},\\Video Diffusion Models~\cite{videodiffusionmodels}, Hunyuan Video~\cite{hunyuanvideo}, \\SVD~\cite{blattmann2023stable}, Kling~\cite{kling}, Runway Gen~\cite{structurecontentguidedvideosynthesis}, Tar~\cite{visiondialect}, etc.}]
      ]
      [Autoregressive Model, leaf1, draw=hidden-black
        [{VideoPoet~\cite{kondratyuk2023videopoet}, RFLAV~\cite{rflav}, UniForm~\cite{uniform}, Ovi~\cite{ovi}, \\VILA-U~\cite{vilau}, NoVA~\cite{autoregressivevideogeneration}, etc.}]
      ]
      [Autoregressive-Diffusion Model, leaf1, draw=hidden-black
        [{NFD~\cite{playingtransformer}, ACDC~\cite{acdc}, HybridVLA~\cite{hybridvla}, DiCoDe~\cite{dicode}, \\ARLON~\cite{arlon}, RFLAV~\cite{rflav}, AR-Diffusion~\cite{ardiffusion}, \\CausVid~\cite{slowbidirectionalfastautoregressive}, VLOGGER~\cite{vlogger2024}, etc.}]
      ]
    ]
    [Explicit Structured \\Control\\, fill=hidden-blue!57, leaf
      [Motion-Geometry Explicit \\Encoding, leaf2
        [{VAST 1.0~\cite{vast2024}, CMVC~\cite{videocodingmeetsmultimodal}, VideoComposer~\cite{videocomposer},\\ Control-A-Video~\cite{controlavideo}, Diverse and Aligned~\cite{diversealigned}, \\Gen-1~\cite{structurecontentguidedvideosynthesis}, TM2D~\cite{tm2d}, etc.}]
      ]
      [Start-End Frame Anchoring \\and Interpolation, leaf2
        [{Make Pixels Dance~\cite{makepixelsdance}, Videogen~\cite{videogen}, Show-1~\cite{zhang2023show1}, \\InteractivateVideo~\cite{interactivevideo}, KeyVID~\cite{keyvid}, etc.}]
      ]
      [Multi-Condition \\Decoupling Architecture, leaf2
        [{MagicEdit~\cite{magicedit}, Swap Attention~\cite{swapattention}, MoonShot~\cite{moonshot},\\ CCEdit~\cite{ccedit}, TATS~\cite{longvideogeneration}, FancyVideo~\cite{fancyvideo}, etc.}]
      ]
    ]
    [Unified Comprehension \\and Generation \\Symbiosis Architecture\\, fill=purple!27, leaf
      [Shared Representation \\Bidirectional Synergy, leaf3
        [{HERMES~\cite{hermes}, Gaia-1~\cite{gaia1}, UniVid~\cite{univid}, Phenaki~\cite{phenaki}, \\Unified Discrete Diffusion~\cite{unifieddiscretediffusion}, etc.}]
      ]
      [Pre-training Driven \\Synergistic Adaptation, leaf3
        [{Omni-Video~\cite{omnivideo}, HunyuanCustom~\cite{hunyuancustom}, merv~\cite{merv2025}, \\HermesFlow~\cite{hermesflow}, VIMI~\cite{vimi}, etc.}]
      ]
    ]
    [Reinforcement Learning, fill=green!20, leaf
      [{VideoDPO~\cite{liu2025videodpo}, T2V-Turbo~\cite{li2024t2vturbo}, Video-STaR~\cite{zohar2025videostar}, \\VPO~\cite{cheng2025vpo}, VideoScore~\cite{he2024videoscore}, etc.}, tier=citations, text width=33.5em, font=\large]
    ]
  ]
\end{forest}
}
\caption{Evolution of Modal Consistency and Temporal Consistency.}
\label{fig:modality-temporal-consistency-evolution}
\end{figure*}
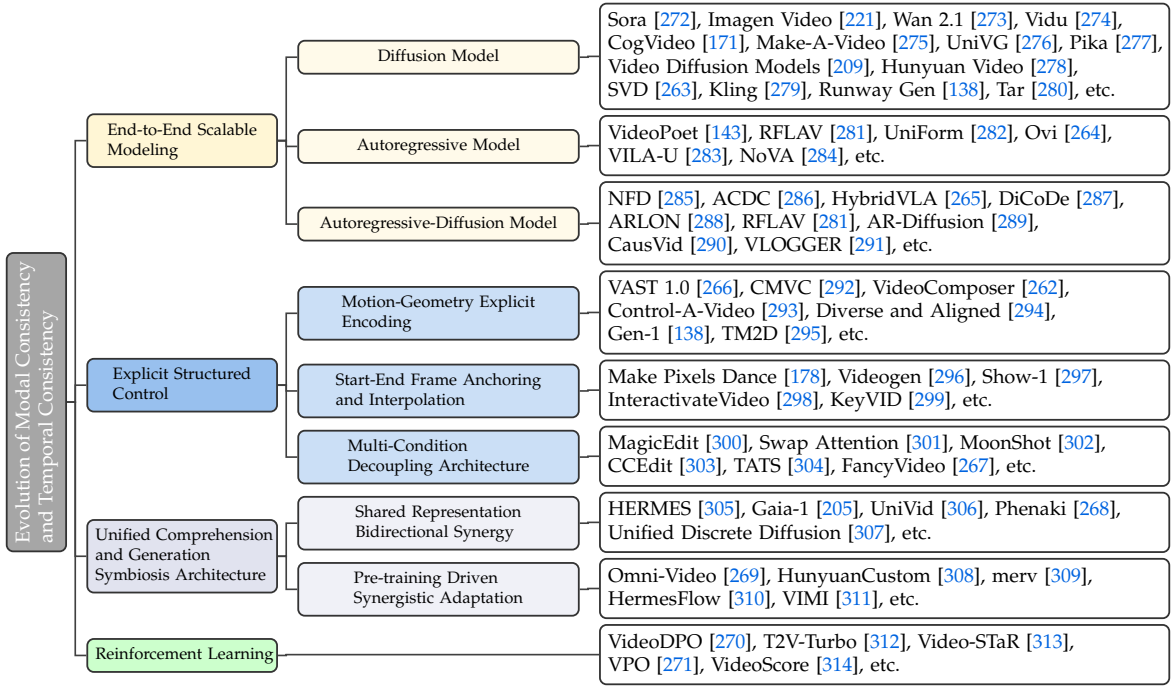

\subsubsection{End-to-End Scalable Modeling}

End-to-End Scalable Modeling represents a paradigm shift in the field of video generation from ``Divide and Conquer'' toward a ``Unified Field.'' Its core objective is to validate the efficacy of the \textbf{Scaling Law} on high-dimensional spatiotemporal manifolds—specifically, by synergistically expanding both model and data scales to directly fit the joint distribution $p(\boldsymbol{x}_{img} | \boldsymbol{c})$ from multi-modal inputs to video outputs. Unlike earlier cascaded pipelines that relied heavily on hand-crafted interpolation and super-resolution modules, this paradigm is dedicated to constructing a general physical simulator, driving the industrialization of models from Sora~\cite{openai2024sora} to Wan 2.1~\cite{wan2025}.

\paragraph{(1) Diffusion Model.}

As the core engine of end-to-end video generation, the Diffusion Model has completely restructured the technical path from ``Image Animation'' to ``Native World Simulation'' by validating the Scaling Law within the Latent Space $\mathcal{Z}$. To support this complex goal of physical consistency, the evolution of modern architecture is no longer confined to simple denoising iterations; instead, it has systematically reshaped four core pillars: from the ODE unification of generation theory~\cite{liu2023rectified}, the causal decoupling of compressed representations~\cite{cogvideox}, and the native three-dimensionalization of attention modeling~\cite{guo2023animatediff}, to the progressive cascading of generation strategies~\cite{visiondialect}. Together, these form the underlying foundation for spatiotemporal intelligence.

\textit{(\romannumeral1) Theoretical Unification via Flow Matching.} Although early works followed the DDPM paradigm based on SDEs, state-of-the-art (SOTA) models such as Sora~\cite{openai2024sora} and Wan 2.1~\cite{wan2025} have generally shifted toward the \textbf{Flow Matching (FM)} framework to enhance sampling efficiency and temporal coherence. Rather than predicting Gaussian noise $\epsilon$, FM formalizes the generation process as constructing a deterministic Ordinary Differential Equation (ODE) trajectory between the noise distribution $\pi_0$ and the data distribution $\pi_1$. The core optimization objective transforms into regressing the velocity field $\boldsymbol{v}_t$ on the optimal transport path:
\begin{equation}
    \mathcal{L}_{\textrm{FM}}(\boldsymbol{\theta}) = \mathbb{E}_{t, \boldsymbol{z}_0, \boldsymbol{z}_1} \left[ \left\| \boldsymbol{v}_{\boldsymbol{\theta}}\left(t, (1-t)\boldsymbol{z}_0 + t\boldsymbol{z}_1\right) - (\boldsymbol{z}_1 - \boldsymbol{z}_0) \right\|^2 \right], \label{eq:flow_matching}
\end{equation}
where $\boldsymbol{v}_{\boldsymbol{\theta}}$ denotes the velocity field predicted by the network parameters $\boldsymbol{\theta}$, and $\boldsymbol{z}_0, \boldsymbol{z}_1$ represent samples from the prior noise and data distributions respectively.
As demonstrated by Rectified Flow~\cite{liu2023rectified}, this paradigm forces the latent variable $\boldsymbol{z}$ to evolve along a linear trajectory, significantly reducing transport curvature. This allows the model to generate dynamic textures with physical conservation in very few steps, resolving the structural collapse issues inherent in DDPMs during long-term sampling.

\textit{(\romannumeral2) Causal Spatiotemporal Compression.} To circumvent the computational bottlenecks of high-dimensional video data, the primary challenge in architecture design lies in constructing a compact latent space $\mathcal{Z}$ that satisfies causality. MagViT-v2~\cite{yu2024magvitv2} and CogVideoX~\cite{cogvideox} identified the risk of ``future information leakage'' in traditional 3D convolutions. Consequently, modern encoders generally introduce Causal 3D VAEs~\cite{blattmann2023stable, bar2024lumiere}, utilizing asymmetric temporal padding and causal convolution kernels to ensure that the generation of latent code $\boldsymbol{z}_t$ depends only on historical frames $\boldsymbol{x}_{img, \leq t}$. This design not only mathematically guarantees the unidirectionality of temporal logic but also provides the architectural foundation for streaming inference. Furthermore, by employing heterogeneous downsampling strategies (\eg $t \times 4, h \times 8, w \times 8$), models achieve decoupled compression of high-frequency motion information and low-frequency semantic features~\cite{rombach2022high, peebles2023scalable}.

\textit{(\romannumeral3) Native 3D Attention Modeling.} Regarding dynamics modeling in latent space, the academic community has undergone a profound correction of inductive bias. Early works like AnimateDiff~\cite{guo2023animatediff} utilized ``spatial-temporal factorized'' attention, which reduced computational costs but severed spatiotemporal coupling, making it difficult to simulate complex fluid dynamics. HunyuanVideo~\cite{hunyuanvideo} and OpenSora~\cite{opensora} have since established the dominance of the \textbf{Native 3D DiT}, calculating joint self-attention across the entire spatiotemporal sequence using 3D-RoPE. Although this introduces a quadratic complexity of $O((THW)^2)$, the integration of sequence parallelism techniques such as Ring Attention~\cite{liu2023ring} enables the model to capture long-range spatiotemporal dependencies, thereby allowing the emergence of coherent motion consistent with physical laws.

\textit{(\romannumeral4) Progressive Alignment \& Cascading.} To address error accumulation in long video generation, models employ a ``coarse-to-fine'' condition control strategy. Visual Dialect, proposed by Tar~\cite{visiondialect}, achieves native alignment of semantics by mapping text to visually compatible tokens. For generating long-range videos ($>10s$), Kling~\cite{kling} and Vidu~\cite{vidu} utilize spatiotemporal cascade strategies. The model first generates a semantic skeleton at a low frame rate, which then serves as a condition $\boldsymbol{c}_{ctx}$ for a Temporal Super-Resolution Model. This cascade architecture essentially decomposes the high-dimensional joint distribution $p(\boldsymbol{x}_{img})$~\cite{ma2024latte} into a product of multiple conditional probabilities, effectively mitigating VRAM pressure and logic drift in single models during long sequence generation~\cite{wu2023tune}.

\vspace{-2mm}
\paragraph{(2) Autoregressive Model (AR).}

The Autoregressive Model draws on the Scaling Law of LLMs~\cite{kaplan2020scaling, hoffmann2022training}, with its core philosophy being ``Everything is a Token.'' This paradigm discards the denoising prior of diffusion models and reformulates video generation as a causal sequence prediction problem within a discrete latent space~\cite{esser2021taming, vandenOord2017neural}. Its mathematical essence is the maximization of the log-likelihood of the joint probability distribution, forcing the model to learn the temporal causality of the physical world through the unidirectional chain rule. To support this vision of unified sequence modeling, the technical evolution of this paradigm is unfolding across four key dimensions: the fidelity of discrete encoding, the topology of multi-modal interaction, the temporal robustness of hybrid generation, and the generalization boundaries of multi-task reasoning.

\textit{(\romannumeral1) The Discretization Bottleneck \& Causal 3D Codebook.} The upper bound of an AR model depends on the compression quality of the tokenizer. Early VQGANs suffered from severe codebook collapse and high-frequency flickering. VideoPoet~\cite{kondratyuk2023videopoet} and MagViT-v2~\cite{yu2024magvitv2} achieved breakthroughs by introducing Lookup-Free Quantization (LFQ) and Causal 3D Convolution. The former reduces quantization variance through direct projection, while the latter ensures that the compression process does not violate physical causality via asymmetric padding. VILA-U~\cite{vilau} further proposed a Unified Vision Tower, which forces the alignment of visual tokens and text embeddings during the pre-training phase, fundamentally resolving the semantic gap between heterogeneous modalities in discrete space.

\textit{(\romannumeral2) Omni-Modal Interaction Topology.} During the sequence modeling phase, the core of architectural design lies in handling the interaction granularity of multi-modal tokens. \textit{Sequence Concatenation:} UniForm~\cite{uniform} adopts an aggressive early fusion strategy, concatenating video, audio, and text tokens into a single long sequence. While using shared-weights Transformers to capture cross-modal dependencies maximizes knowledge transfer between modalities, it faces an $O(N^2)$ explosion in attention computation. \textit{Dual-Stream Gated Modulation:} To reduce computational overhead, RFLAV~\cite{rflav} and Ovi~\cite{ovi} employ late fusion. Ovi designs a symmetric dual-backbone architecture, aligning the sampling rates of different modalities through RoPE frequency scaling; RFLAV introduces temporal averaging modulation in the AdaLN layer of the Transformer, achieving soft alignment of audio-video features without a significant increase in parameter count.

\textit{(\romannumeral3) Long-Horizon Dynamics \& Hybrid Paradigms.} Pure discrete AR models often face collapse due to error accumulation when generating long videos. To correct this deficiency, researchers have begun exploring hybrid paths of ``discrete planning + continuous correction''. \textit{Non-Quantized AR:} NoVA~\cite{autoregressivevideogeneration} challenges the assumption that data ``must be discretized,'' proposing continuous AR prediction in a continuous space. It decomposes video into ``frame-wise temporal steps'' and ``set-wise spatial steps,'' predicting continuous features via a diffusion decoder to circumvent information loss from quantization. \textit{Rolling Flow Matching:} RFLAV~\cite{rflav} innovatively introduces a sliding window mechanism. After the AR predicts coarse tokens, flow matching is used for local refinement. Through a rolling strategy of ``removing the first frame and adding the noised last frame,'' it theoretically achieves physically consistent generation of infinite duration, solving the inherent malady of AR models being ``logical but lacking details.''

\textit{(\romannumeral4) Unified Multi-Task Reasoning.} The ultimate advantage of the AR architecture lies in its \textbf{zero-shot generalization.} As demonstrated by VideoPoet, by introducing specific task tokens (\eg `<optical\_flow>`, `<depth>`), a single model can perform video generation, style transfer, and even audio-visual QA tasks without fine-tuning. This ``omnivore'' characteristic proves the unique potential of the autoregressive paradigm in building a universal world simulator.

\begin{figure}[h]
  \centering
  \includegraphics[width=0.98\linewidth]{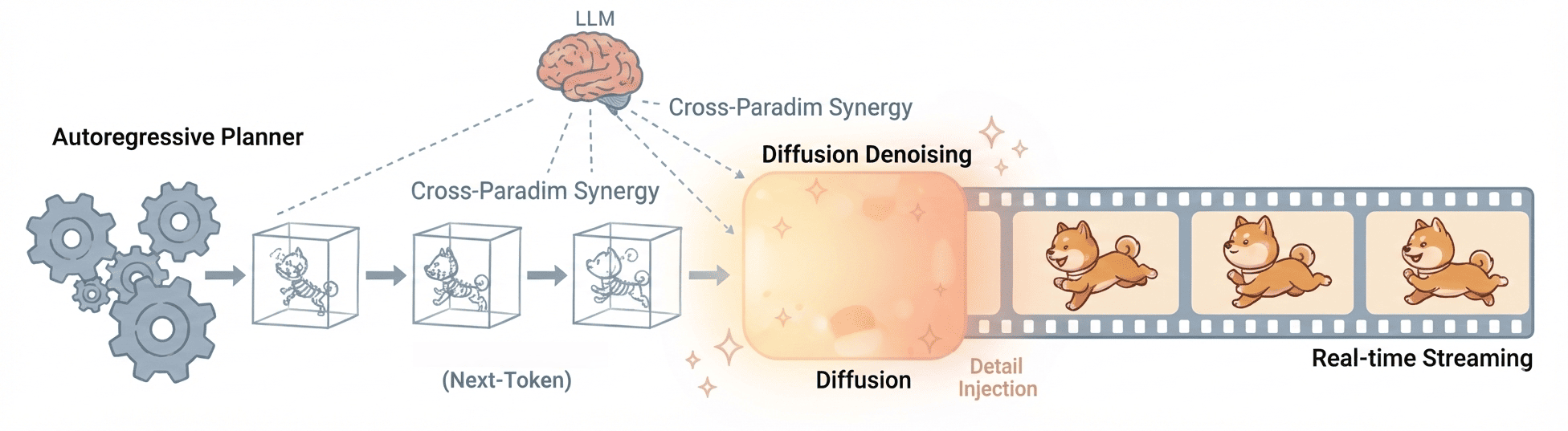} 
  \caption{Autoregressive-Diffusion Model. It depicts an AR Planner constructing a causal temporal skeleton (blue wireframes), which then passes through a Diffusion Refiner (warm mist) for detail injection, resulting in high-quality long-duration video output.}
  \label{fig:modality_temporal2}
\end{figure}

\paragraph{(3) Autoregressive-Diffusion Hybrid Model.}

The core synergy mechanism of the Autoregressive-Diffusion Hybrid Model, as essentially illustrated in Figure~\ref{fig:modality_temporal2}, is the injection of the temporal causal constraints of AR into the iterative denoising manifold of the Diffusion Model. The universal mathematical representation of this hybrid generation is no longer a simple probability superposition, but rather the construction of a joint probability density of ``causal logic and high-quality generation'':
\begin{equation}
    p_{\boldsymbol{\theta}}(\boldsymbol{x}_{\textrm{img}, 1:T}, \boldsymbol{z}_{1:T} \mid \boldsymbol{c}) = \prod_{t=1}^T \underbrace{p_{\textrm{AR}}(\boldsymbol{z}_t \mid \boldsymbol{z}_{<t}, \boldsymbol{x}_{\textrm{img}, <t}, \boldsymbol{c})}_{\text{Causal Temporal Dynamics}} \cdot \underbrace{p_{\textrm{Diff}}(\boldsymbol{x}_{\textrm{img}, t} \mid \boldsymbol{z}_t, \boldsymbol{x}_{\textrm{img}, <t}, \boldsymbol{c})}_{\text{Conditionally Denoised Rendering}}, \label{eq:hybrid_model}
\end{equation}
where $\boldsymbol{x}_{\textrm{img}, 1:T}$ denotes the generated multi-modal sequence (Video/Audio), $\boldsymbol{z}_t$ represents the noisy latent or intermediate features, and $p_{\textrm{AR}}$ and $p_{\textrm{Diff}}$ correspond to the low-dimensional causal modeling and the high-fidelity conditional denoising distribution, respectively. This mechanism aims to combine the long-range planning advantage of AR with the detail generation capability of diffusion, overcoming the inherent defects of single models. Based on this joint modeling approach, the technical evolution of this paradigm is unfolding along two orthogonal paths: temporal fusion optimization and cross-paradigm modal synergy, aiming to simultaneously solve the bottleneck of dynamical consistency in long-sequence generation and the challenge of alignment between heterogeneous modalities.

\textit{(\romannumeral1) Temporal Fusion Optimization.} The core lies in balancing strict causal dependency with generation flexibility, breaking the efficiency bottleneck of long video generation through differentiated architecture design. \textit{Real-time Streaming Dynamics.} To address generation speed and VRAM limits, several works have restructured the inference paradigm. AR-Diffusion~\cite{ardiffusion} proposed a training-inference unified diffusion corruption mechanism, establishing a temporal baseline by enforcing a non-decreasing frame time step constraint ($t_1 \leq t_2 \leq ... \leq t_F$), which, combined with a dynamic scheduler, enables error-free variable-length generation. CausVid~\cite{slowbidirectionalfastautoregressive} converts bidirectional diffusion into an AR architecture through distribution matching distillation; combined with KV cache and sliding window mechanisms, it balances the real-time performance of streaming generation with infinite length extension capabilities. Furthermore, NFD~\cite{playingtransformer} utilizes block-wise causal attention and speculative sampling to achieve real-time generation at 30+ FPS for the first time at a scale of 300M+ parameters. RFLAV~\cite{rflav} innovatively introduces rolling flow matching and a lightweight temporal modulation module, achieving precise alignment generation of infinite-length audio-video while significantly reducing computational overhead. \textit{Long-term Coherence Guidance.} To address logic drift in long-term sequences, synergistic guidance strategies have become key. ARLON~\cite{arlon} employs a ``coarse-grained anchoring - fine-grained refinement'' strategy, using an AR model to generate coarse features containing long-range semantics to guide a Diffusion Transformer (DiT) in detail refinement, while utilizing a VQ-VAE unified representation space to resist noise interference. ACDC~\cite{acdc} proposes a zero-shot synergy framework that, without modifying the architecture, allows the AR model to act as a global context ``planner'' and the diffusion model as a local ``corrector,'' utilizing the external memory module of an LLM to effectively alleviate error accumulation in long sequence prediction.

\textit{(\romannumeral2) Cross-Paradigm Modal Synergy.} This focuses on the precision of modal alignment and the tightness of integrated architecture, aiming for deep coupling of heterogeneous signals. \textit{Diffusion-Augmented Representation.} DiCoDe~\cite{dicode} challenges traditional discretization methods with a diffusion cascaded tokenization scheme. It first encodes video into continuous latent features and then uses a diffusion process to compress them into high-fidelity discrete tokens. This approach achieves thousand-fold compression while preserving visual details and strengthens text-video semantic alignment through cross-attention mechanisms, providing a high-quality ``vocabulary'' for long video generation. \textit{End-to-End Architectural Fusion.} HybridVLA~\cite{hybridvla} demonstrates the potential of paradigm fusion in the field of Embodied AI. It seamlessly integrates diffusion generation and AR prediction within a single LLM framework, projecting continuous action vectors generated by diffusion into the LLM's word embedding space. By introducing special tokens to separate the two paradigms and adaptively fusing prediction results based on AR confidence, the model achieves an end-to-end logical loop across visual, language, and action modalities, significantly strengthening the coherence of the agent's ``perception-reasoning-execution'' link.

\subsubsection{Explicit Structured Control}

Although end-to-end models have achieved breakthroughs in image quality, their ``text-as-all'' interaction mode exhibits significant ambiguity in industrial applications. Explicit structured control aims to resolve the challenge of \textbf{controllability}. Its core concept involves projecting the high-dimensional dynamics manifold $\mathcal{M}_{dyn}$ onto a low-dimensional interpretable control manifold (\eg depth, optical flow, skeleton). This paradigm reformulates video generation as a constrained optimization problem:
\begin{equation}
    \max_{\boldsymbol{\theta}} \mathbb{E}_{\boldsymbol{x}_{\textrm{img}}, \boldsymbol{c}_{\textrm{struct}}, \boldsymbol{c}_{\textrm{mot}}} \left[ \log p_{\boldsymbol{\theta}} \left( \boldsymbol{x}_{\textrm{img}} \mid \mathcal{E}_{\textrm{txt}}(\boldsymbol{c}_{\textrm{txt}}), \mathcal{E}_{\textrm{str}}(\boldsymbol{c}_{\textrm{struct}}), \mathcal{E}_{\textrm{mot}}(\boldsymbol{c}_{\textrm{mot}}) \right) \right], \label{eq:structured_control}
\end{equation}
where $\mathcal{E}_{\textrm{str}}$ and $\mathcal{E}_{\textrm{mot}}$ denote the encoders processing explicit conditions for spatial structure and temporal motion, respectively.

\paragraph{(1) Motion-Geometry Explicit Encoding.}

This school of thought primarily inherits and extends the principles of 2D ControlNet, aiming to eliminate generated geometric hallucinations by injecting explicit physical priors. Facing spatiotemporal degrees of freedom that far exceed those of static images, this paradigm is dedicated to constructing a set of ``hard-constrained'' physical interfaces. It has achieved breakthrough progress in two key dimensions—residual-based spatiotemporal feature injection and multimodal narrative structure orchestration.
% —enabling a leap from precise control of single actions to long-range logical coherence.

\textit{(\romannumeral1) Residual-based Feature Injection.} The core challenge lies in injecting strong geometric constraints without compromising pre-trained generation priors. \textit{Spatiotemporal ControlNet Adaptation.} VideoComposer~\cite{videocomposer} proposed an explicit encoding strategy using Motion Vectors (MVs), utilizing MV signals in the compressed domain as a low-rank approximation of temporal conditions. This addresses control difficulties in complex motion scenarios, such as the coupling of camera translation and object deformation. ControlVideo~\cite{controlavideo} explored a training-free path by introducing cross-frame geometric masks in the self-attention layer to force multiple frames to share the same ControlNet features, thereby achieving temporal consistency in structure. \textit{Trajectory-aware Latent Navigation.} For fine-grained control of object movement paths, DragNUWA~\cite{yin2024dragnuwa} and MotionCtrl~\cite{motionctrl} introduced the joint encoding of trajectory heatmaps and camera poses $\boldsymbol{P}_t$. Unlike simple optical flow $\mathcal{O}_{flow}$ injection, they explicitly map user-drawn 2D trajectories to manifold evolution directions in 3D latent space by a flow $\mathcal{F}: \boldsymbol{z}_t \to \boldsymbol{z}_{t+1}$.

\textit{(\romannumeral2) Multimodal Storyboarding.} To handle long-range narratives, VAST~\cite{vast2024} introduced a storyboard mechanism, decoupling text descriptions into dual-stream constraints of ``Layout + Pose.'' Its innovation lies in constructing a bi-directional autoencoder that maps discrete control signals to continuous sequence latent vectors, providing a rigid skeleton for cross-frame generation and effectively suppressing object identity drift in long sequences.

\paragraph{(2) Start-End Frame Anchoring and Interpolation.}

This paradigm transforms video generation from extrapolation into a mathematically more stable interpolation problem, specifically solving for a Brownian Bridge with $\boldsymbol{x}_{img, start}$ and $\boldsymbol{x}_{img, end}$ as boundary conditions. Under this mathematical framework, technical evolution focuses on constructing smooth, high-fidelity spatiotemporal transition manifolds and exploring multimodal interactive control within constrained spaces, forming two core pillars: boundary-condition-driven path planning and dynamic instruction injection.

\textit{(\romannumeral1) Boundary-Conditioned Path Planning.} \textit{Temporal Generative Inpainting.} SEINE~\cite{chen2024seine} and MorphStudio~\cite{morphstudio} treat two input images as masks $\boldsymbol{m} \in \{0, 1\}^{T \times H \times W}$, performing denoising only on the intermediate frames during the diffusion process. To ensure transition smoothness, they introduced interpolated attention, allowing the query vectors of intermediate frames to simultaneously query the keys/values of the start and end frames, thus achieving a smooth blending of physical states in the feature space. \textit{Cascaded Super-Resolution Architecture.} To address the blurriness caused by interpolation, Show-1~\cite{zhang2023show1} proposed a cascaded strategy of coarse-to-fine anchoring. Stage 1 utilizes a pixel-level model to generate a low-frequency motion skeleton, while Stage 2 employs latent diffusion for high-frequency texture inpainting. This design skillfully leverages the structural sensitivity of pixel space and the texture generation capability of latent space.

\textit{(\romannumeral2) Dynamic Instruction Injection.} For complex interactive generation, InteractiveVideo~\cite{interactivevideo} refines control signals into a quadruple (Image, Content, Action, Trajectory) and injects them at specific time steps via gated cross-attention. KeyVID~\cite{keyvid} focuses on audio-driven scenarios, utilizing ImageBind to extract audio peaks as implicit keyframes, achieving automated anchoring of ``audio-visual sync.''

\paragraph{(3) Multi-Condition Decoupling Architecture.}

\begin{figure}[h]
  \centering
  \includegraphics[width=0.9\linewidth]{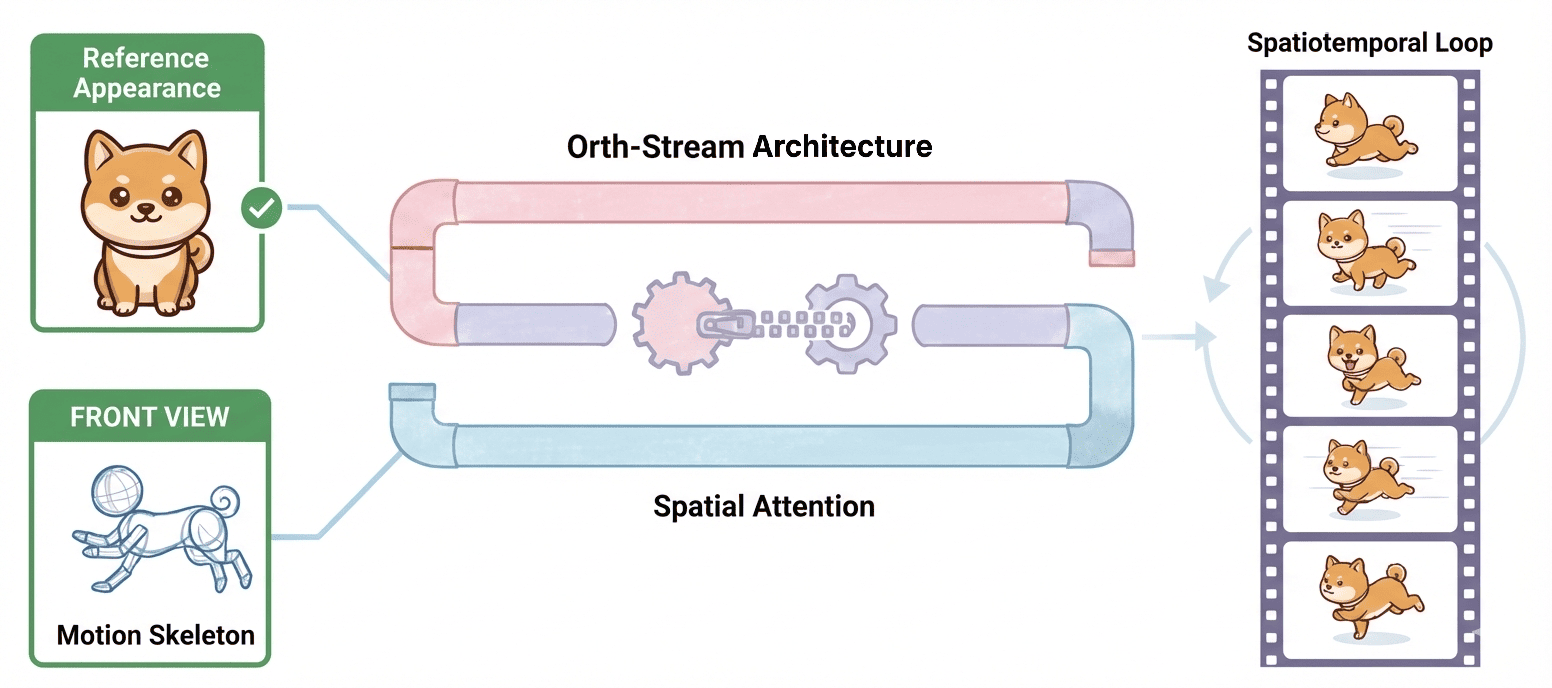} 
  \caption{Multi-Condition Decoupling Architecture. The diagram illustrating the Two-Stream Architecture for digital human animation. It depicts the orthogonal decoupling of Reference Appearance (static portrait) and Motion Skeleton (stick figure), which are fused via a Spatial Attention "Zipper" to generate a spatiotemporally consistent video loop.}
  \label{fig:modality_temporal3}
\end{figure}

To resolve feature entanglement between multi-modal signals (\eg changing a character's action causing background texture changes), recent architectures favor the \textbf{orthogonal decoupling} design illustrated in Figure~\ref{fig:modality_temporal3}. Within this framework, technical evolution is proceeding along two key axes: the separation of appearance-motion features and the complementary interaction of spatiotemporal dimensions, aiming to simultaneously solve identity drift under high dynamics and the imbalance between spatial structure and temporal manifolds in long-sequence generation.

\textit{(\romannumeral1) Appearance-Motion Two-Stream.} This is currently the mainstream paradigm for digital human animation~\cite{siarohin2019first, zhao2022thin}. Facing the inherent conflict between maintaining identity (appearance) and driving complex actions (motion), this paradigm advocates for abandoning single-stream processing in favor of a two-stream decoupling mechanism at the architectural level. This involves extracting static texture features and dynamic pose features separately, then fusing them through specific modules orthogonally. This includes: \textit{Explicit Spatial Decoupling.} Targeting the issue where single-stream networks lose appearance features over time, Animate Anyone~\cite{hu2024animate} and MagicAnimate~\cite{seedance2025} introduced an independent ReferenceNet as the ``appearance stream.'' This branch does not participate in the denoising process but specifically extracts high-fidelity features from the reference image, which are then injected layer-by-layer via spatial attention into the Main UNet (motion stream) responsible for action generation. Formally, this achieves an explicit decomposition of the generated features $\boldsymbol{z}_{gen}$:
\begin{equation}
    \boldsymbol{z}_{\textrm{gen}} = \underbrace{\mathcal{F}_{\textrm{motion}}(\boldsymbol{z}_t, \boldsymbol{c}_{\textrm{pose}})}_{\text{Motion Stream}} + \lambda \cdot \underbrace{\mathcal{F}_{\textrm{app}}(\boldsymbol{I}_{\textrm{ref}})}_{\text{Appearance Stream}}, \label{eq:spatial_decoupling}
\end{equation}
where $\boldsymbol{z}_{\textrm{gen}}$ denotes the synthesized feature map, $\boldsymbol{c}_{\textrm{pose}}$ represents the pose control signal, $\boldsymbol{I}_{\textrm{ref}}$ is the reference source image processed by the appearance encoder $\mathcal{F}_{\textrm{app}}$, and $\lambda$ is the fusion coefficient. This dual-tower design forces the separation of texture encoding and motion inference, ensuring consistency of details during large-scale dynamic movements~\cite{zhu2024champ, zhang2022motiondiffuse}. \textit{Implicit Attention Disentanglement.} Moving beyond physical dual-network structures, Moonshot~\cite{moonshot} and CCEdit~\cite{ccedit} explore ``logical dual-streams'' within a single network. They argue that traditional cross-attention tends to confuse structural signals (pose/shape) with content signals (texture/identity). Consequently, these works propose a decoupled attention mechanism that splits key/value mappings into independent structure branches and appearance branches. Through orthogonal gradient backpropagation, the model is forced to ensure that changes in the motion stream do not interfere with the feature distribution of the appearance stream. This mechanism achieves zero-interference between appearance and motion at the micro-level, resolving the chronic issue of identity drift caused by action changes~\cite{epstein2023diffusion}.

\textit{(\romannumeral2) Spatiotemporal Complementary Loop.} TATS~\cite{longvideogeneration} and Swap Attention~\cite{swapattention} explore the decoupling of spatiotemporal dimensions. Swap Attention utilizes a role-swapping mechanism within 3D windows to construct a loop where ``space guides time, and time feeds back to space.'' This design mathematically forces the model to maintain texture consistency along the spatial axis and optical flow $\mathcal{O}_{flow}$ coherence along the temporal axis, effectively solving the ``infinite loop'' or ``motion freezing'' phenomena common in autoregressive generation.

\subsubsection{Unified Comprehension and Generation Symbiosis Architecture}

\begin{figure}[h]
  \centering
  \includegraphics[width=0.98\linewidth]{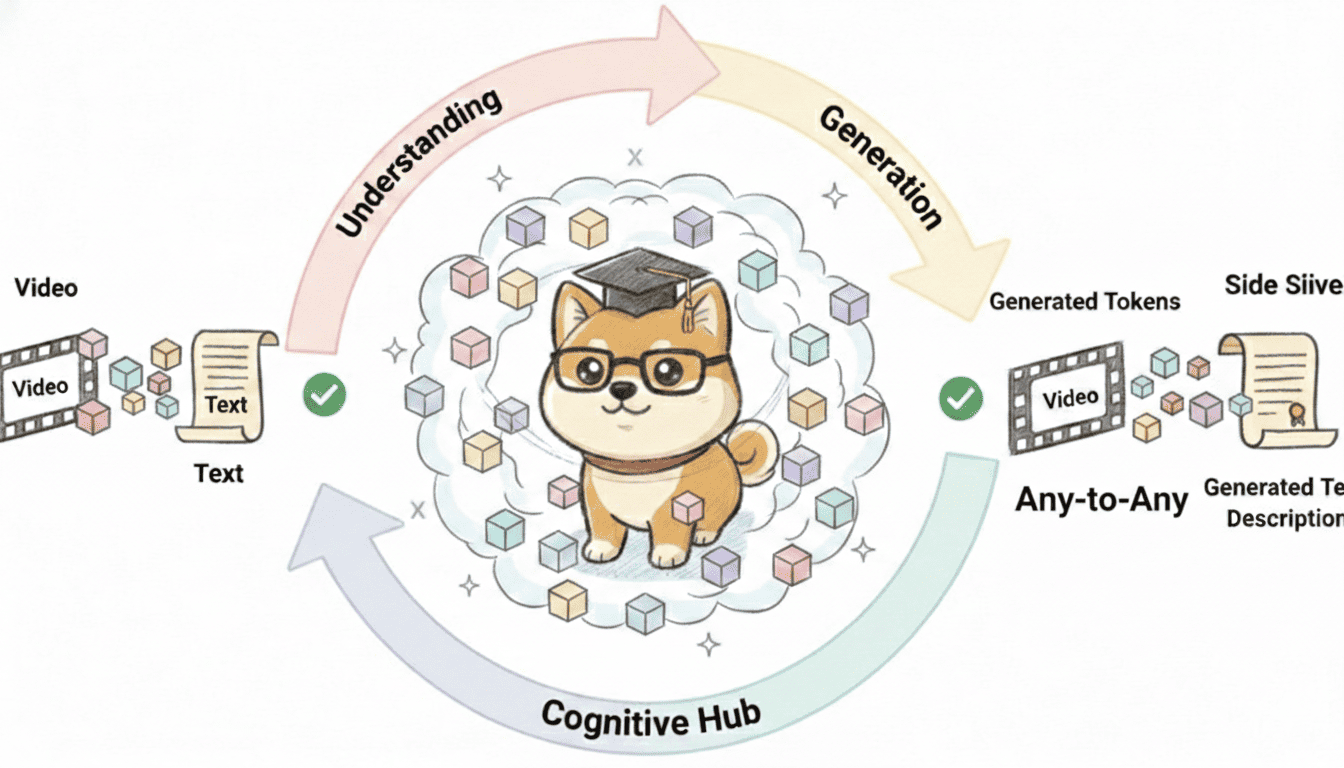} 
  \caption{Unified Comprehension and Generation Symbiosis Architecture. It features a central LLM converting multimodal inputs into a unified cloud of discrete tokens, enabling seamless "Any-to-Any" transformation (\eg video to text and vice versa).}
  \label{fig:modality_temporal4}
\end{figure}

Traditional computer vision research treats ``Understanding (\textbf{Discriminative})'' and ``Generation (\textbf{Generative})'' as opposing binary tasks: the former models the conditional probability $p(y | \boldsymbol{x}_{img})$, while the latter models $p(\boldsymbol{x}_{img} | y)$~\cite{rombach2022high, ho2020denoising}. However, the Unified Comprehension and Generation Symbiosis Architecture seeks to construct a unified probability model $p(\boldsymbol{x}_{img}, y)$, aiming to dismantle the barriers between perception and simulation~\cite{meta2024chameleon, unifieddiscretediffusion, wu2024janus}. The core assumption of this paradigm, as illustrated in Figure~\ref{fig:modality_temporal4}, is that: \textbf{A perfect generator should implicitly contain a perfect discriminator.}

\paragraph{(1) Shared Representation Bidirectional Synergy.}

This direction aims to map heterogeneous signals into the same manifold space by constructing an Omni-modal Isomorphic Representation, thereby achieving ``Any-to-Any'' conversion within a single set of model parameters. Specifically, to break the chasm between perception and generation, this paradigm establishes the dominance of discrete tokens as universal interaction primitives and explores the unique value of geometric representations as physical anchors in embodied scenarios. This has resulted in two primary technical tracks based on symbolic unification and geometric symbiosis.

\textit{(\romannumeral1) Token-based World Modeling.} Inspired by the success of LLMs, discretized tokens have become the ``general currency'' for unifying understanding and generation. \textit{Fully Discretized Autoregression.} Gaia-1~\cite{gaia1} and Phenaki~\cite{phenaki} proposed video encoding schemes based on C-ViViT, which unify the encoding of driving videos, control signals, and text descriptions into a discrete token sequence $\boldsymbol{z}_{1:L}$. The training objective of the model is unified into standard Next-Token Prediction:
\begin{equation}
    \mathcal{L}_{\textrm{uni}} = -\sum_{i=1}^L \log p_{\boldsymbol{\theta}}(\boldsymbol{z}_i \mid \boldsymbol{z}_{<i}, \text{TaskToken}), \label{eq:unified_loss}
\end{equation}
where $\boldsymbol{z}_{1:L}$ represents the unified discrete token sequence combining visual and textual information, and $\text{TaskToken}$ serves as the prompt indicator to switch between understanding and generation modes. This paradigm allows the model to switch functions via simple ``Task Prompting'': inputting video tokens to predict text tokens constitutes ``understanding,'' while the reverse constitutes ``generation.'' \textit{Unified Discrete Diffusion.} Unified Discrete Diffusion~\cite{unifieddiscretediffusion} and Show-O~\cite{xie2024show} challenge the notion that ``autoregression is the only solution.'' They designed a Unified Transition Matrix that allows image and text tokens to undergo bidirectional denoising within the same diffusion process. Show-O further utilizes a Hybrid Attention Mechanism that applies a Causal Mask to the text portion and a Full Mask to the visual portion, achieving the seamless coexistence of understanding and generation within single Transformer weights.

\textit{(\romannumeral2) Domain-Specific Geometric Symbiosis.} In the field of Embodied AI, HERMES~\cite{hermesflow} proposed using BEV (Bird's-Eye-View) features as a shared hub. It utilizes a World Queries mechanism to compress 2D images from multi-view cameras into 3D BEV features. This not only supports downstream path planning (understanding) but also enables the generation of future prediction videos (generation) via a decoder for BEV features, proving that 3D geometric constraints serve as a strong bridge connecting perception and simulation.

\paragraph{(2) Pre-training Driven Synergistic Adaptation.}

Unlike training a unified multimodal model from scratch~\cite{wang2024neurips, meta2024chameleon}, this paradigm advocates a ``Shoulders of Giants'' strategy: using a frozen MLLM (such as GPT-4V or LLaVA) as the cognitive hub (Brain), connected via lightweight adapters to a visual generation decoder (Eyes/Hands). The goal is to transfer the general reasoning capability of LLMs to video generation tasks at a low cost~\cite{zhang2023video, li2023videochat}. Under this architecture, the core technical challenge shifts to constructing a high-bandwidth interface connecting the cognitive space and the generative space, aiming to precisely map high-level reasoning to low-level generative conditions through an LLM-centric projection mechanism.

\textit{(\romannumeral1) LLM-Centric Projection.} The core challenge lies in achieving a ``zero-loss'' interface between the semantic space of the LLM and the pixel space of video generation. \textit{Input-Output Bidirectional Adaptation.} Omni-Video~\cite{omnivideo} and NExT-GPT~\cite{next_gpt} established a general bridging framework for ``Any-to-Any'' conversion. On the input side, linear projections or Q-Formers are used to align visual signals with the LLM embedding space; on the output side, the model triggers a Vision Head by predicting a special [IMG] token, projecting the hidden states of the LLM into the conditional input $\boldsymbol{c}_{diff}$ for a diffusion model. This achieves an explicit translation from ``textual thinking'' to visual signals. \textit{Mixture of Encoders.} MERV~\cite{merv2025} notes that a single visual encoder struggles to balance semantic understanding with texture details. It introduces a learnable cross-attention mechanism that connects multiple frozen encoders in parallel, such as CLIP (strong semantics), DINOv2 (strong structure), and VideoMAE (strong action). Through dynamic weighting via the LLM's attention mechanism, the model can automatically select the optimal visual feature source when processing complex instructions, achieving a ``gathering of strengths'' for visual perception.

\subsubsection{Reinforcement Learning for Modal-Temporal Alignment}

The introduction of \textbf{Reinforcement Learning (RL)} techniques aims to address the non-convexity issues encountered by traditional Supervised Fine-Tuning (SFT) when handling ``modal semantics'' and ``temporal dynamics''~\cite{christiano2017deep, ouyang2022training}. SFT tends to average distributions, often leading generation results into a binary dilemma of being either ``high semantic fidelity but static'' or ``high dynamic but collapsed.'' The RL paradigm, by constructing a Multi-dimensional Reward Manifold $\mathcal{R}$~\cite{xu2023imagereward, wu2023human}, transforms discrete modal alignment objectives and continuous temporal evolution objectives into a joint optimization problem, guiding the model to converge toward the \textbf{Pareto Frontier} of the ``semantics-temporal'' trade-off. Driven by this objective, and to precisely characterize and optimize this complex manifold, technical evolution is unfolding across three dimensions: preference-based joint alignment, self-refinement-based iterative evolution, and the logical restructuring of universal reward models. These efforts aim to comprehensively enhance the model's ability to synergistically control heterogeneous modalities and dynamic sequences.

\paragraph{(1) Preference-based Joint Alignment.}
This path utilizes DPO~\cite{rafailov2023direct} and its variants to encode the implicit dependency between ``semantic understanding'' and ``temporal evolution'' into preference rankings, forcing the model to learn temporal dynamics consistent with physical laws while maintaining textual/image semantic precision.

\textit{(\romannumeral1) Dynamic Preference \& Static Penalty.} VideoDPO~\cite{liu2025videodpo} was the first to point out that directly applying image-level DPO leads to ``motion collapse'' (\ie the model sacrifices temporal dynamics to cater to semantic scores). It constructs a preference dataset encompassing the trade-off between ``semantic alignment vs. motion magnitude.'' Through KL divergence constraints, it mathematically pushes the probability density toward high-dynamic and high-fidelity regions, achieving joint calibration of modal instructions and temporal motion. \textit{(\romannumeral2) Mixed Reward Distillation.} T2V-Turbo~\cite{li2024t2vturbo} proposes a multi-path signal fusion strategy. Rather than relying solely on a single preference model, it integrates reward signals $\mathcal{R}$ from HPSv2 (measuring modal aesthetics) and InternVideo2 (measuring temporal consistency). Through reward-weighted regression, it ``distills'' evaluation metrics for ``modal aesthetics'' and ``temporal fluency'' into a consistency-model-based student network, rapidly approaching the joint optimal distribution of semantics and dynamics.

\paragraph{(2) Iterative Alignment via Self-Refinement.}
This direction draws on the self-play concept from LLMs, constructing a feedback loop that allows the model to find the optimal balance point between modal instructions and temporal evolution within a generation-evaluation-correction cycle.

\textit{(\romannumeral1) Semantic-Dynamic Hierarchical Reward.} Hierarchical optimization frameworks~\cite{cheng2025vpo} design a hierarchical reward mechanism $\mathcal{R}$ to specifically address the disconnection between first-frame semantics and subsequent-frame actions. It applies an ``image quality'' reward (modal level) to the first frame and a ``coherence'' penalty based on motion vectors (temporal level) to subsequent frame sequences. This introduces ``temporal gradient backpropagation'' into PPO~\cite{schulman2017proximal} updates, enabling the model to ``foresee'' the dynamical consequences on the time axis while generating first-frame semantics. \textit{(\romannumeral2) Instruction-Following Self-Evolution.} Video-STaR~\cite{zohar2025videostar} proposes a self-evolution framework based on rejection sampling. It utilizes an MLLM (such as GPT-4V) as a discriminator to select high-quality samples that are both ``instruction-following accurate (modal)'' and ``action-natural and smooth (temporal)'' to fine-tune the generator. This mechanism filters out noise data that is either ``text-image matched but temporally collapsed'' or ``temporally smooth but semantically lost,'' significantly enhancing the model's capability to understand complex spatiotemporal instructions.

\paragraph{(3) Universal Reward Modeling.}
The upper bound of RL depends on whether the reward model (RM) $\mathcal{R}$ can accurately decouple and measure the contributions of modality and time. Research focus in 2024–2025 has shifted toward constructing universal RMs capable of simultaneously understanding semantic logic and physical causality.

\textit{(\romannumeral1) Decomposition-Fusion Evaluation System.} VPO~\cite{cheng2025vpo} proposes explicitly decomposing the reward function $\mathcal{R}$ into semantic alignment (Video-LLM) and temporal smoothness (optical flow $\mathcal{O}_{flow}$). By performing weighted optimization of these two orthogonal objectives along the diffusion denoising trajectory, the model learns to eliminate inter-frame flickering using $\mathcal{O}_{flow}$ constraints without compromising text semantics, achieving deep fusion of modal content and temporal continuity. \textit{(\romannumeral2) From Noun Alignment to Causal Logic.} VideoScore~\cite{he2024videoscore} challenges the traditional CLIP-Score~\cite{radford2021clip} by constructing a universal automatic evaluation metric based on Video-LMM. It captures not only static pixel-level quality but also deep ``temporal causal logic'' (\eg if an instruction requires a ``cup shattering,'' the shattering action must occur after the fall, not before). Using VideoScore as a direct optimization target for RL allows the model to move beyond simple noun-based modal alignment and truly master the causal consistency of temporal logic and modal semantics.

\paragraph{(4) Embodied Action Alignment via VLA-RL.}
When alignment extends to the \textbf{Vision-Language-Action (VLA)} domain, VLA models face the more rigorous challenge of functional temporal alignment. In this context, temporal evolution is no longer merely the coherence of visual frames but a physical intervention sequence $\boldsymbol{a}_{1:T}$ driven by linguistic instructions~\cite{kim2024openvla}.

Traditional VLA training primarily relies on Supervised Fine-Tuning (SFT) based on human demonstrations. However, from a statistical perspective, SFT is essentially a re-weighting of the known data distribution~\cite{guan2026rl, brohan2023rt2}. Its objective function $\min_{\theta} -\log P_{\theta}(a|s, \mathcal{D}_{demo})$ compels the model to fit the average behavior of demonstration data, causing the effective search space to be confined near the local optima of human experts. Once an Out-of-Distribution (OOD) shift occurs in the environmental state, the model often succumbs to cascading errors induced by covariate shift due to a lack of exploration capability.

To overcome this theoretical bottleneck, works such as TwinRL-VLA~\cite{xu2026twinrlvla} and RL-VLA$^3$~\cite{guan2026rl} have pioneered a paradigm shift from passive imitation to active exploration. Their core mechanism involves transforming the optimization objective from minimizing imitation loss to maximizing long-term cumulative reward $J(\theta) = \mathbb{E}_{\tau \sim \pi_{\theta}}[\sum_{t} \gamma^t r(s_t, a_t)]$.
\textit{(\romannumeral1) Digital Twin Verification Mechanism.} Unlike implicit reward models, TwinRL introduces a Digital Twin as an explicit physical verifier. The system utilizes 3D Gaussian Splatting (3DGS) to reconstruct high-fidelity scenes~\cite{lu2024manigaussian} and executes the policy-generated action sequences $\boldsymbol{a}_{1:T}$ in parallel within a physics engine. This mechanism provides deterministic physical feedback as a sparse reward signal, compelling the model to not only align with the semantic intent of linguistic instructions temporally but also satisfy the feasibility constraints of physical interaction~\cite{ma2406dreureka, zhang2025safevla}.
\textit{(\romannumeral2) Exploration Boundary Expansion.} By conducting large-scale trial and error in a zero-cost simulation environment, the RL agent can reach long-tail state spaces not covered in human demonstration data, such as extreme physical contacts or rare object poses~\cite{wang2023robogen}. Theoretically, this mechanism expands the effective support set of the policy, enabling the VLA model to evolve from interpolation capabilities on finite samples to extrapolation capabilities in unknown environments.

\subsection{Integration of Spatial and Temporal Consistency}

\begin{figure}[h]
  \centering
  \includegraphics[width=0.9\linewidth]{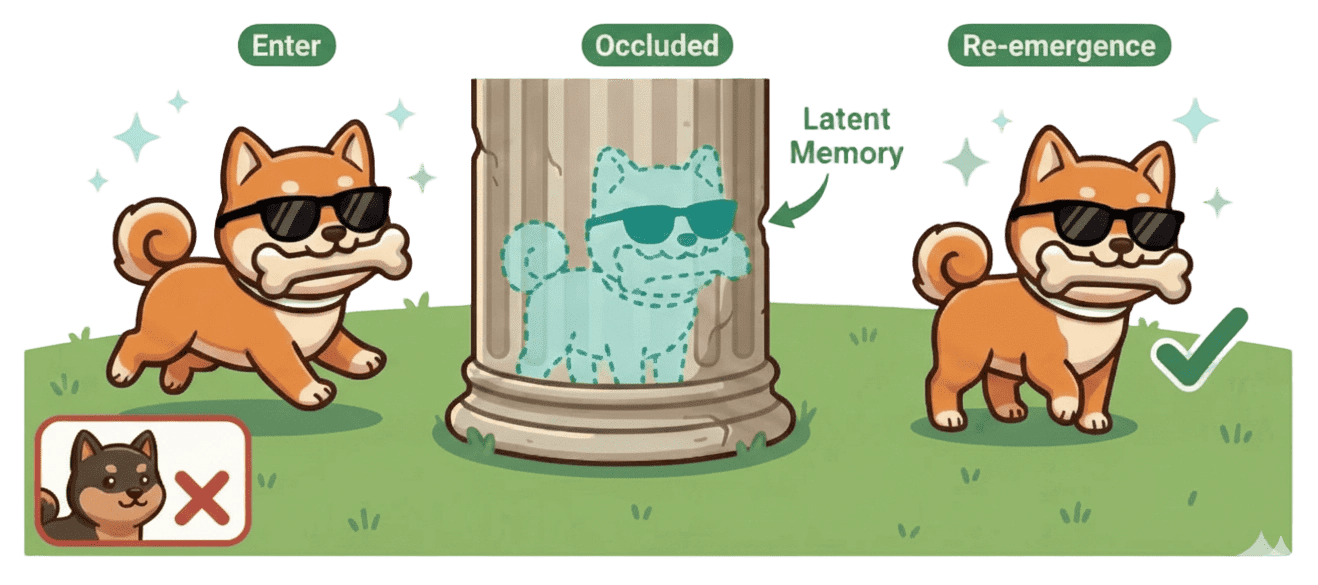} 
  \caption{Temporal + Spatial Consistency: Dynamic Object Permanence across Occlusion. The model combines temporal and spatial consistency to realize dynamic object permanence during occlusion: the subject (Doge) retains consistent features (\eg sunglasses, bone) via latent memory when occluded, and re-emerges with unchanged attributes, ensuring the continuity of the object.}
  \label{fig:temporal_spatial1}
\end{figure}

The fusion of spatial and temporal consistency marks the ultimate leap of generative models from Frame-wise Painting toward World Construction~\cite{openai2024sora, ha2018world}. As illustrated in Figure~\ref{fig:temporal_spatial1}, the core utility of this dimension lies in establishing \textbf{Dynamic Object Permanence}: specifically, during spatiotemporal evolution, an object must not only maintain the rigidity of its geometric form but also follow a motion trajectory consistent with physical laws. Its intrinsic properties must not drift even during occlusions or drastic viewpoint changes~\cite{qiu2024freenoise, liu2024video}. This fusion elevates time passage from mere pixel changes to the topological evolution of a 3D manifold, constituting the physical cornerstone of the 4D generation technology stack~\cite{singer2022make, 4dgaussiansplatting}.

Under this vision, technical evolution presents a four-stage evolutionary lineage from representation construction to value alignment, as shown in Figure~\ref{fig:spatial-temporal-consistency-evolution}: Implicit Spatiotemporal Learning adopts a destructuring strategy, mapping 2D video priors to the probability distributions of 4D fields via score distillation, trading statistical flexibility for generative generalization~\cite{UBC_ViVid, nvssolver}; Explicit Geometric Anchoring introduces point clouds and camera trajectories as a rigid skeleton, parameterizing the time axis as $SE(3)$ transformations to achieve precise control with geometry as constraint~\cite{NVIDIA_GEN3C, Apple_WVD, HKUST_DaS}; Unified Spatiotemporal Representation utilizes 4D Gaussian primitives or hybrid tensor fields to establish continuous mathematical fields with native support for deformation and lighting, which—coupled with the global association of dense trajectory fields—realizes an isomorphic representation of spatiotemporal dimensions~\cite{4dynamic, cat4d, wang2024vggsfm}; and finally, Reinforcement Learning Alignment aims to overcome the exposure bias of SFT by constructing a composite reward function that integrates explicit physical costs. This forces the model to solve the Pareto optimization of spatial fidelity and temporal coherence, achieving a paradigm shift from probability fitting to physical value alignment~\cite{li2025t2vturbov2, yuan2024instructvideo}. Together, these four stages define the evolutionary path toward physical realism for current 4D world models.

\definecolor{hidden-blue}{HTML}{4a90e2}
\definecolor{hidden-black}{HTML}{333333}
\tikzstyle{my-box}=[
  rectangle,
  draw=hidden-black,
  rounded corners,
  text opacity=1,
  minimum height=1.5em,
  minimum width=5em,
  inner sep=2pt,
  align=center,
  fill opacity=.5,
]\tikzstyle{root}=[
  align=center,
]\tikzstyle{leaf}=[
  my-box, 
  minimum height=1.5em,
  text=black,
  font=\normalsize,
  inner xsep=5pt,
  inner ysep=4pt,
  text width=11.5em,
  fill opacity=1,
]
\tikzstyle{leaf1}=[
  my-box, 
  minimum height=1.5em,
  fill=yellow!32, 
  text=black,
  font=\normalsize,
  inner xsep=5pt,
  inner ysep=4pt,
  text width=16em,   
]
\tikzstyle{leaf2}=[
  my-box, 
  minimum height=1.5em,
  fill=hidden-blue!57, 
  text=black,        
  font=\normalsize,
  inner xsep=5pt,
  inner ysep=4pt,
  text width=16em,
]
\tikzstyle{leaf3}=[
  my-box, 
  minimum height=1.5em,
  fill=purple!27, 
  text=black,
  font=\normalsize,
  inner xsep=5pt,
  inner ysep=4pt,
  text width=16em,
]
\begin{figure*}[t]
\vspace{-2mm}
\centering
\resizebox{\textwidth}{!}{
\begin{forest}
  forked edges,
  for tree={
    grow=east,
    reversed=true,
    anchor=base west,
    parent anchor=east,
    child anchor=west,
    base=left,
    font=\large,
    rectangle,
    draw=hidden-black,
    rounded corners,
    align=left,
    minimum width=4em,
    edge+={darkgray, line width=1pt},
    s sep=3pt,
    inner xsep=5pt,
    inner ysep=4pt,
    line width=1.1pt,
    ver/.style={rotate=90, child anchor=north, parent anchor=south, anchor=center},
  },
  where level=1{text width=11.5em,font=\normalsize}{},
  where level=2{text width=13em,font=\normalsize}{},
  where level=3{text width=33em, tier=citations, font=\large}{},
  % 所有叶子都在 level=3，使用不同颜色区分
[Evolution of Spatial Consistency \\and Temporal Consistency, ver, fill=gray!70, text=white, root
    [Implicit Spatiotemporal \\Constraints, fill=yellow!32, leaf
      [Video Prior Distillation, leaf1
        [{NVS-Solver~\cite{nvssolver}, ViVid-1-to-3~\cite{UBC_ViVid}, Vivid-ZOO~\cite{vividzoo}, \\Diffusion²~\cite{diffusion2}, etc.}]
      ]
    ]
    [Explicit Geometric \\Anchoring, fill=hidden-blue!57, leaf
      [Point Cloud Conditioning, leaf2
        [{GEN3C~\cite{gen3c}, RealCam-I2V~\cite{realcami2v}, ViewCrafter~\cite{viewcrafter}, \\EPiC~\cite{epic}, etc.}]
      ]
      [Geometric Embedding Injection, leaf2
        [{cameractrl~\cite{cameractrl}, World-consistent Video Diffusion~\cite{Apple_WVD}, \\OmniView~\cite{omniview}, PostCam~\cite{postcam}, viewdiff~\cite{viewdiff}, VD3D~\cite{Toronto_VD3D}, \\motionctrl~\cite{motionctrl},etc.}]
      ]
      [Trajectory Parametric Control, leaf2
        [{TC4D~\cite{tc4d}, Diffusion as Shader~\cite{HKUST_DaS}, 3DTrajMaster~\cite{3dtrajmaster}, \\SV3D~\cite{singh2024sv3d}, etc.}]
      ]
    ]
    [Unified Spatiotemporal \\Representation, fill=purple!27, leaf
      [Hybrid Volumetric Representation, leaf3
        [{4Diffusion~\cite{Beihang_4Diffusion}, Point-DynRF~\cite{pointdynrf}, Consistent4D~\cite{consistent4d}, \\4Dynamic~\cite{4dynamic}, DyBluRF~\cite{dyblurf}, DynIBaR~\cite{dynibar}, 4D-fy~\cite{4dfy}, \\TiNeuVox~\cite{fastdynamicradiancefields}, MoBluRF~\cite{moblurf}, SV4D~\cite{Ren2024_SV4D}, \\K-Planes~\cite{kplanes}, etc.}]
      ]
      [Explicit Structured Representation, leaf3
        [{4DGS~\cite{4dgaussiansplatting}, STAG4D~\cite{stag4d}, DreamGaussian4D~\cite{dreamgaussian4d}, \\Align Your Gaussians~\cite{aligngaussians}, H3D-DGS~\cite{h3ddgs}, CAT4D~\cite{cat4d}, \\Diffusion4D~\cite{diffusion4d}, etc.}]
      ]
      [Trajectory-Centric \\Foundation Models, leaf3
        [{Trace Anything~\cite{sun2024trace}, OmniMotion~\cite{wang2023omnimotion}, VGGSfM~\cite{wang2024vggsfm}, \\SpatialTracker~\cite{xiao2024spatialtracker}, MASt3R~\cite{leroy2024mast3r}, etc.}]
      ]
    ]
    [Reinforcement Learning, fill=green!20, leaf
      [{T2V-Turbo-v2~\cite{li2025t2vturbov2}, VistaDPO~\cite{huang2025vistadpo}, DR-Tune~\cite{zhou2023drtune}, \\InstructVideo~\cite{yuan2024instructvideo}, etc.}, tier=citations, text width=33em, font=\large]
    ]
  ]
\end{forest}
}
\caption{Evolution of Spatial Consistency and Temporal Consistency.}
\label{fig:spatial-temporal-consistency-evolution}
\end{figure*}
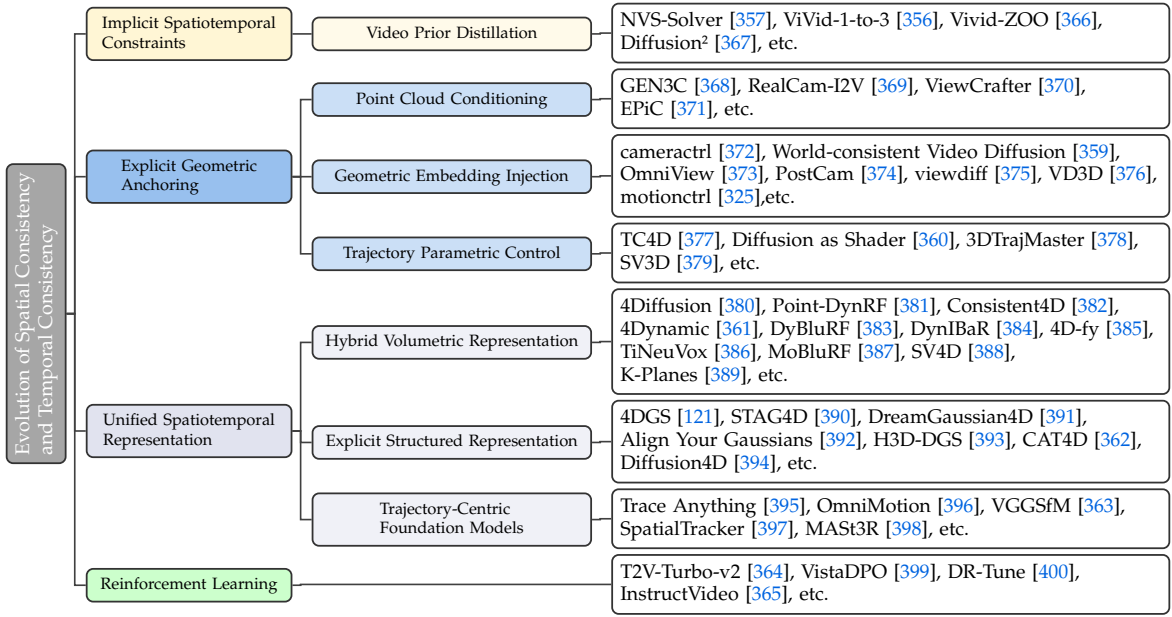

\subsubsection{Implicit Spatiotemporal Learning}

Beyond explicit geometric representations (such as 3DGS), Implicit Spatiotemporal Learning represents an alternative minimalist paradigm of destructuring. In particular, the direction of Video Prior Distillation fundamentally refutes the necessity of full-parameter fine-tuning based on large-scale 3D data, instead pioneering a training-free path based on posterior modulation. The theoretical foundation of this paradigm is built upon Score Distillation Sampling (SDS)~\cite{poole2022dreamfusion} and Score Jacobian Chaining (SJC)~\cite{wang2023sjc}, reframing 4D scene generation as an intersection problem between two orthogonal probability manifolds: the geometric manifold $\mathcal{M}_{geo}$ and the dynamics manifold $\mathcal{M}_{dyn}$.

\paragraph{Video Prior Distillation.}

The core logic lies in utilizing the Tweedie formula~\cite{efron2011tweedie, kim2022refining} to model the generated denoising step $\epsilon_{\theta}$ as a linear combination of two heterogeneous gradient fields. This forces the latent variable $\boldsymbol{z}_t$ to converge toward the overlapping high-density region of two prior distributions during the inverse diffusion process~\cite{du2023reduce}:
\begin{equation}
    \nabla_{\boldsymbol{z}_t} \log p(\boldsymbol{z}_t \mid \boldsymbol{c}) \approx \omega_s(t) \cdot \underbrace{\nabla_{\boldsymbol{z}_t} \log p_{\textrm{MVD}}(\boldsymbol{z}_t \mid \boldsymbol{c}_{\textrm{view}})}_{\text{Geometric Constraint}} + \omega_t(t) \cdot \underbrace{\nabla_{\boldsymbol{z}_t} \log p_{\textrm{VDM}}(\boldsymbol{z}_t \mid \boldsymbol{c}_{\textrm{motion}})}_{\text{Dynamic Guidance}}, \label{eq:score_composition}
\end{equation}
where $\boldsymbol{z}_t$ denotes the latent variable at timestep $t$, $\omega_s(t)$ and $\omega_t(t)$ are time-dependent weighting coefficients, and $p_{\textrm{MVD}}$ and $p_{\textrm{VDM}}$ represent the probability densities of the Multi-View and Video Diffusion priors, respectively. This process is essentially a Maximum A Posteriori estimation in high-dimensional space that satisfies $P(\boldsymbol{x}) \propto P_{MVD}(\boldsymbol{x}) \cdot P_{VDM}(\boldsymbol{x})$~\cite{liu2022compositional}. However, facing gradient conflicts and distribution mismatches triggered by the direct superposition of heterogeneous priors, the academic community has evolved systematic solutions across four dimensions: scanning generation \& trajectory mapping, variance reduction \& SDE solvers, frequency decoupling \& progressive modulation, and deep manifold alignment.

\textit{(\romannumeral1) Scanning Generation \& Trajectory Mapping.} To materialize the aforementioned probability framework, VIVID-1-to-3~\cite{UBC_ViVid} pioneered the isomorphism of the Novel View Synthesis (NVS) task into a ``camera moving along trajectory video generation'' problem. This method utilizes a video diffusion model (such as ZeroScope or SVD) as the dynamics engine. By explicitly mapping changes in camera extrinsics $\boldsymbol{P}_t \in SE(3)$ to video timestamps $t$, it forces the VDM to interpret geometric parallax as optical flow motion. To suppress geometric distortion in single-frame generation, VIVID-1-to-3 introduced an Epipolar Attention Bias, utilizing a multi-view diffusion model (such as Zero-1-to-3~\cite{liu2023zero1to3}) to anchor the geometric structure at keyframes. This \textit{dual-diffusion synergy} strategy effectively leverages the powerful inter-frame smoothing prior of the VDM to suppress flickering artifacts common in independent view synthesis~\cite{wang2024dreamvideo, he2024videocrafter}.

\textit{(\romannumeral2) Variance Reduction \& SDE Solver.} Although score composition provides a unified framework, direct superposition of heterogeneous priors in high-dimensional latent space often leads to severe gradient conflicts. NVS-Solver~\cite{nvssolver} approaches this from the numerical solution of Stochastic Differential Equations (SDEs), noting that simple score addition violates the Itô integral conditions of the diffusion process, causing deviation in the drift term of the sampling trajectory. To address this, NVS-Solver introduced high-order approximations based on Taylor expansion and a variance-reducing sampling strategy. By explicitly correcting the variance inflation caused by heterogeneous gradients within the SDE solver, the method mathematically ensures that the generation trajectory can smoothly traverse the boundary between the two manifolds. Empirical results show a reduction in stochastic jittering during sampling by approximately 40\%, significantly enhancing the sharpness and spatiotemporal consistency of the generation results~\cite{lu2022dpm, zhao2023unipc, song2021scorebased}.

\textit{(\romannumeral3) Frequency Decoupling \& Progressive Modulation.} Dynamics analysis of the generation process reveals that diffusion models follow a \textit{spectral bias} of ``first global structure (low-frequency), then texture details (high-frequency)''~\cite{zhuang2024hifa, yang2024freeu}. Based on this observation, VividZoo~\cite{vividzoo} proposed time-variant modulation. This mechanism replaces fixed weight allocation with a dynamic annealing schedule: in the early denoising stages (high noise $t$), MVD is assigned a higher gradient weight $\omega_s > \omega_t$ to leverage its strong geometric prior for establishing the main topology of the object and preventing distortion. In the later denoising stages (low noise $t$), the weights are reversed ($\omega_t > \omega_s$) to utilize the temporal smoothing characteristics of the VDM to eliminate high-frequency flickering. This design, which aligns with the laws of generative spectral evolution, effectively resolves structural distortion and texture blurring issues caused by prior competition~\cite{Wang2023_Prolific, li2024sweetdreamer}.

\textit{(\romannumeral4) Deep Manifold Alignment.} Most aforementioned methods remain at the level of score mixing on the output side (pixel/noise space), ignoring the semantic gap in the model's internal representations. Diffusion$^2$~\cite{Fudan_Diffusion2} proposed a deep fusion architecture to resolve the distribution mismatch between VDM and MVD latent spaces. Rather than simply mixing noise, this method inserts learnable 3D-2D cross-attention adapters between the U-Nets of the two diffusion models. By minimizing the \textit{Sliced Wasserstein Distance} at the feature level, the model forces $\boldsymbol{z}_{MVD}$ and $\boldsymbol{z}_{VDM}$ to share the same representation manifold in intermediate layers. This design enables the model to perceive the feature distribution of the other, fundamentally eliminating the ghosting phenomenon caused by domain gaps and achieving true feature-level synergy~\cite{zhang2024controlvideo, hu2024animate, mou2024t2i}.

\subsubsection{Explicit Geometric Anchoring}

If implicit learning is a soft fitting of spatiotemporal statistical laws, then explicit geometric anchoring represents a radical attempt to restructure video generation from probability prediction to 3D rendering. This paradigm rejects treating space and time as entangled latent variables within deep networks; instead, by introducing explicit 3D point clouds~\cite{NVIDIA_GEN3C, realcami2v} and camera trajectories~\cite{cameractrl, postcam}, it parameterizes time as a continuous $SE(3)$ pose sequence and fixes space as a static geometric structure. Its core philosophy is that \textit{spatiotemporal consistency should not be achieved by network memory of historical frames, but rather naturally derived from the rigidity of the underlying geometric proxy}~\cite{motionctrl}.

\paragraph{(1) Point Cloud Conditioning}

This direction models video generation as a problem of neural rendering with geometric proxies. Its mathematical essence lies in constructing a static world model $\mathcal{W}$ and projecting it into a visual feature flow via camera parameters $\boldsymbol{P}_t$. This process is not mere image processing but a rigid transformation strictly following the pinhole camera model:
\begin{equation}
    \boldsymbol{c}_{t} = \Pi(\mathcal{W}, \boldsymbol{P}_t), \label{eq:world_projection}
\end{equation}
where $\boldsymbol{c}_{t}$ denotes the projected visual features at time $t$, and $\Pi$ represents the perspective projection function mapping the static world $\mathcal{W}$ under camera pose $\boldsymbol{P}_t$. To implement this physical rigidity within probabilistic diffusion models, current research has explored two dimensions: representation construction and inference control.

\textit{Infrastructure \& Representation.} To overcome the memory bottlenecks of pure generative models, Gen-3C~\cite{NVIDIA_GEN3C} and RealCamI2V~\cite{realcami2v} established metric scale spaces based on Structure from Motion (SfM). Gen-3C utilizes back-projection from monocular depth estimation to construct a 3D cache $\mathcal{C}$, transforming the evolution of the time dimension into camera roaming within a static point cloud. RealCamI2V further introduces a \textit{scale alignment loss} to enforce consistency between generated local geometry and global SfM point clouds in Euclidean space, thereby resolving the scale drift common in long sequence generation (minutes-level)~\cite{schonberger2016colmap, teed2021droidslam}.

\textit{Endogenous Consistency \& Inference-Driven (System 2 Generation).} Unlike the one-pass inference of end-to-end modes, this school emphasizes explicit computation during the inference stage. ViewCrafter~\cite{viewcrafter} employs dense stereo matching to reconstruct high-precision point clouds, using the rendering result $\hat{\boldsymbol{x}}_{render}$ as a hard visual anchor for the video LDM. This design shifts the source of consistency from the black-box statistics of network weights to the white-box geometry of the input side. EPIC~\cite{epic} proposes a \textit{dynamic masking strategy}: by calculating the occlusion map $\boldsymbol{m}_{occ}$ of the point cloud projection, it applies lightweight ControlNet constraints only to the visible regions while allowing generative freedom in unseen regions. This explicit $XYZ \to UV$ geometric projection effectively avoids texture misalignment caused by depth errors~\cite{Yu2020}.

\paragraph{(2) Geometric Embedding Injection}

\begin{figure}[h]
  \centering
  \includegraphics[width=0.88\linewidth]{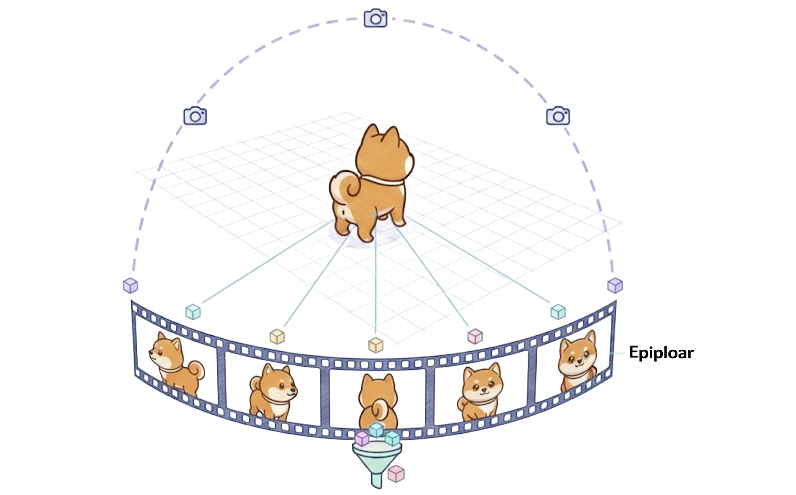}
  \caption{Geometric Embedding Injection. It features a Doge on a perspective grid with a trajectory, converting physical rays into geometric tokens to enforce epipolar constraints across frames.}
  \label{fig:spatial_temporal2}
\end{figure}

While point cloud conditioning is explicit rendering, geometric embedding injection is its implicit mapping within the Transformer latent space. As shown in Figure~\ref{fig:spatial_temporal2}, this approach aims to encode 3D spatial coordinate information into geometric tokens~\cite{omniview} isomorphic to visual tokens, which are injected directly into the self-attention mechanism to establish a cross-frame shared world coordinate system~\cite{motionctrl}. To achieve this deep 3D-2D alignment, the community has focused on architectural designs across coordinate representation and dynamic association:

\textit{Tokenization of World Coordinates.} VD3D~\cite{Toronto_VD3D} and ViewDiff~\cite{viewdiff} introduced Plücker coordinate encoding, mapping each camera ray $\boldsymbol{r} = (\boldsymbol{o}, \boldsymbol{d})$ to a high-dimensional embedding vector $\boldsymbol{e}_{geo}$.

By injecting these into $\text{Attention}(\boldsymbol{Q}, \boldsymbol{K}_{img} + \boldsymbol{K}_{geo}, \boldsymbol{V}_{img} + \boldsymbol{V}_{geo})$, the model no longer simply predicts the statistical distribution of pixels but learns the correspondence between pixels and 3D spatial positions ($\boldsymbol{p}$). This mechanism essentially injects an epipolar inductive bias into the attention matrix, allowing the query at frame $t$ to accurately attend to the key at frame $t-k$ corresponding to the same physical coordinates, thereby realizing the notion that time naturally emerges from space~\cite{mildenhall2020nerf, suhail2022generalizable}.

\textit{Spatiotemporal Interface \& Self-Association Mechanism.} To convert static anchoring into dynamic coherence, PostCam~\cite{postcam} and CameraCtrl~\cite{cameractrl} designed trajectory parameterization modules that inject camera pose sequences ($\boldsymbol{P}_{1:T}$) into the temporal Transformer blocks. OmniView~\cite{omniview} and MotionCtrl~\cite{motionctrl} further proposed geometric similarity gating, utilizing implicit 3D correspondence maps to construct self-associations between cross-frame points. Under this mechanism, object motion is no longer a hallucinated texture flow by the network but a physical motion guided by the spatial reasoning of geometric tokens, marking a leap from data fitting to physical constraints~\cite{ni2023conditional, zhang2024controlvideo}.

\paragraph{(3) Trajectory Parametric Control}
\label{sec:trajectory_parametric_control}

For dynamic scenes, the trajectory parametric control direction explicitly models 3D motion as a differentiable function $\boldsymbol{T}(t)$, achieving physical-level decoupling of object motion laws~\cite{tc4d}. Research efforts focus on motion representation mechanisms and optimization constraints:

\textit{Motion Elevation \& Identity.} This paradigm elevates discrete pixel displacement to continuous \textit{Eulerian-Lagrangian flow}. TC4D~\cite{tc4d} employs a global-local decomposition strategy, decomposing scene motion into a superposition of rigid camera motion $\boldsymbol{P}_t$ and a local object deformation field $\Psi(\boldsymbol{p}, t)$. DiffusionShader~\cite{HKUST_DaS} assigns a 3D identity ID in the world coordinate system to each pixel, simplifying complex dynamic prediction into an ID matching problem along the time axis, which fundamentally eliminates texture flickering.

\textit{Explicit Physical Constraints.} 3DTrajMaster~\cite{3dtrajmaster} treats trajectories as entities subject to physical laws, explicitly adding \textit{acceleration regularization} to the loss function:
\begin{equation}
     \mathcal{L}_{smooth} = \sum_{t} || \boldsymbol{p}_{t+1} - 2\boldsymbol{p}_t + \boldsymbol{p}_{t-1} ||^2,
\end{equation}
where $\boldsymbol{p}_t$ denotes the position vector at time step $t$, and the expression minimizes the second-order difference (approximation of acceleration), effectively suppressing high-frequency jitter to ensure smooth motion. Coupled with a timestep annealing strategy, the model fits the low-frequency trajectory skeleton during the early stages of denoising and fills in high-frequency deformation in the late stages, effectively preventing error accumulation. SV3D~\cite{Stability_SV3D} further adopts a pipeline of ``first generate consistent observation, then optimize unified representation,'' allowing generative models to maintain physical plausibility while unleashing creativity by dynamically adjusting trajectory control strength during inference~\cite{Stability_SV3D, zeng2023makeit3d}.

\subsubsection{Unified Spatiotemporal Representation}

The Unified Spatiotemporal Representation paradigm elevates video generation from pixel interpolation to spatiotemporal manifold reconstruction by constructing a 4D physical representation space~\cite{cao2023hexplane, 4dgaussiansplatting, liu2025trace}.

\paragraph{(1) Hybrid Volumetric Representation: Low-Rank Tensor Decomposition \& Hybrid Fields.}

Hybrid Volumetric Representation aims to resolve the inherent contradiction between high-dimensional spatiotemporal modeling and the capture of high-frequency dynamics, addressing the curse of dimensionality. This paradigm abandons expensive dense 4D voxel grids in favor of compact factorization strategies, which decouple complex 4D fields into tensor products of low-dimensional subspaces while embedding explicit physical motion constraints. This approach enables models to maintain the continuity of neural representations while achieving the efficient query capabilities of grid-based methods. To this end, current research has advanced architectural innovation across three progressive levels: hybrid volumetric representation~\cite{yu2022plenoxels}, spatiotemporal factorization~\cite{cao2023hexplane}, and dynamics coupling \& trajectory integration~\cite{dynibar}.

\textit{(\romannumeral1) Hybrid Volumetric Representation.} Through the synergistic design of explicit structure encoding and implicit neural decoding, researchers seek the Pareto optimality between grid query efficiency and neural network compactness~\cite{yu2022plenoxels, tang2022compressible}. To address the limitations of pure implicit NeRFs in capturing high-frequency dynamics~\cite{mildenhall2020nerf}, this direction introduces low-rank tensor decomposition theory, decomposing the 4D spatiotemporal field into tensor products of multiple low-dimensional subspaces as illustrated in Eq.~\ref{eq:eikonal_loss}~\cite{cao2023hexplane}.

\textit{(\romannumeral2) Spatiotemporal Factorization.} K-Planes~\cite{Fridovich2023KPlanes} and HexPlane~\cite{cao2023hexplane} proposed decomposition strategies based on six planes, transforming feature queries in 4D space into feature interpolation and Hadamard products across six 2D planes. This design not only reduces the representation's space complexity from $O(N^4)$ to $O(N^2)$ but, more importantly, introduces a critical inductive bias: spatial planes (\eg $xy$-plane) enforce 3D consistency of visual appearance, while spatiotemporal planes (\eg $xt$-plane) explicitly constrain the continuous evolution trajectory of pixels over time. Building on this, Tensor4D~\cite{shao2024tensor4d} introduced hierarchical tensor decomposition, utilizing multi-scale feature grids to capture the full spectrum of information—from coarse actions to fine textures—thereby resolving artifact issues in fast-motion scenes~\cite{fastdynamicradiancefields}.

\textit{(\romannumeral3) Dynamics Coupling \& Trajectory Integration.} Since static decomposition struggles with complex topological changes, dynamic constraints must be introduced. DynIBaR~\cite{dynibar} innovatively integrated the time dimension into the volumetric rendering equation through trajectory-based rendering. Instead of sampling at fixed points along a ray, this method warps sampling points to their corresponding positions in neighboring frames based on a velocity field $\boldsymbol{v}_t$:
\begin{equation}
    \boldsymbol{C}(\boldsymbol{r}) = \sum_{i} T_{i} \alpha_{i} \boldsymbol{c} \left( \boldsymbol{p}_{i} + \int_{t}^{t'} \boldsymbol{v}_{\tau}(\boldsymbol{p}_{i}) d\tau, \boldsymbol{d} \right), \label{eq:traj_render}
\end{equation}
where $T_{i}$ is the accumulated transmittance, $\alpha_{i}$ represents the opacity at the $i$-th sample point, and $\boldsymbol{v}_{\tau}$ denotes the instantaneous velocity field. This design internalizes temporal consistency as an integral term in the rendering equation, achieving physical-level aggregation of cross-frame information. SV4D~\cite{Ren2024_SV4D} employs a 3D skeleton to lock the spatial structure and constructs multi-frame multi-view attention within the latent space. By guiding dense 4D generation via sparse 3D keypoints, it effectively mitigates geometric collapse during long sequence generation~\cite{4dfy, wu2024sc4d}.

\paragraph{(2) Explicit Structured Representation}

Explicit Structured Representation is primarily based on 3D Gaussian Splatting, marking a paradigm shift from an Eulerian perspective  to a Lagrangian perspective~\cite{zwicker2001surface, lassner2021pulsar}. Its core logic involves modeling the scene as a set of discrete primitives with specific attributes (position $\boldsymbol{p} / \mu$, covariance $\Sigma$, spherical harmonics coefficients $SH$, and opacity $\alpha$), enabling real-time rendering through differentiable rasterization~\cite{dreamgaussian4d}. Current research establishes a technical framework across three dimensions: canonical-deformation decomposition, multi-source priors \& physical guidance, and topological constraints \& geometric driving.

\textit{(\romannumeral1) Canonical-Deformation Decomposition.} To handle non-rigid motion, mainstream methods adopt the Canonical Space and Deformation Field modeling approach. 4D Gaussian Splatting~\cite{4dgaussiansplatting} and Deformable 3DGS~\cite{Yang2024_Deformable3DGS} define a static canonical space to store geometric topology and utilize an MLP-based deformation field $\Delta(\boldsymbol{p},t)$ conditioned on time $t$ to predict the displacement and rotation of each Gaussian sphere at specific moments:
\begin{equation}
    \boldsymbol{\mu}_{t} = \boldsymbol{\mu}_{0} + \Delta_{\boldsymbol{\mu}}(\boldsymbol{\mu}_{0}, t), \quad \boldsymbol{\Sigma}_{t} = f \left( \boldsymbol{\Sigma}_{0}, \Delta_{r}(\boldsymbol{\mu}_{0}, t) \right), \label{eq:deform_field}
\end{equation}
where $\boldsymbol{\mu}_{0}$ and $\boldsymbol{\Sigma}_{0}$ represent the mean and covariance of the $k$-th Gaussian $\mathcal{G}_{k}$ in the canonical space, $\Delta_{\boldsymbol{\mu}}$ is the predicted position offset, and $\Delta_{r}$ denotes the rotation update. H3D-DGS~\cite{h3ddgs} further splits the deformation field into an observable rigid part and an unobservable completion part, introducing hard-coded priors to restrict degrees of freedom and prevent overfitting to high-frequency noise. DreamGaussian4D~\cite{dreamgaussian4d} combines HexPlane decomposition to parameterize Gaussian deformation, significantly reducing the VRAM usage for 4D optimization~\cite{luiten2024dynamic}.

\textit{(\romannumeral2) Multi-source Priors \& Physical Guidance.} To hallucinate plausible 4D structures from 2D video, this paradigm relies on powerful generative priors. STAG4D~\cite{stag4d} proposes injecting a first-frame time-anchor during the Score Distillation Sampling (SDS) optimization process, forcing the generation of subsequent frames to strictly follow the geometric standards established in the first frame. Ling et al. (2024)~\cite{aligngaussians} employs compositional score distillation, utilizing text-to-image, text-to-video, and 3D-aware diffusion models simultaneously to provide multi-source gradient supervision, thereby achieving cross-modal physical constraints. Diffusion4D~\cite{diffusion4d} introduces a revolutionary explicit 4D diffusion model that performs denoising directly within the voxelized Gaussian parameter space, allowing results to be back-projected into an explicit 4D field, fundamentally ensuring the endogenous consistency of spatiotemporal logic~\cite{huang2024scgs}.

\textit{(\romannumeral3) Topological Constraints \& Geometric Driving.} For complex action control, simple MLP-based deformations often struggle to maintain the topological structure of articulated objects like the human body. CT4D~\cite{ct4d} introduces a Gaussian clustering mechanism to automatically discover rigid parts within a scene and assign pseudo-skeleton weights, enabling skeleton-like motion driven by video diffusion signals. Cat4D~\cite{cat4d} proposes manifold distillation, mapping the feature manifolds of pre-trained video generation models (\eg SVD) into 4D Gaussian space. This ensures the generation process is not merely blind parameter fitting but is implicitly constrained by physical laws such as volume conservation and motion continuity, marking a leap from statistical correlation to physical interpretability.

\paragraph{(3) Trajectory-Centric Foundation Models}

\begin{figure}[h]
  \centering
  \includegraphics[width=0.9\linewidth]{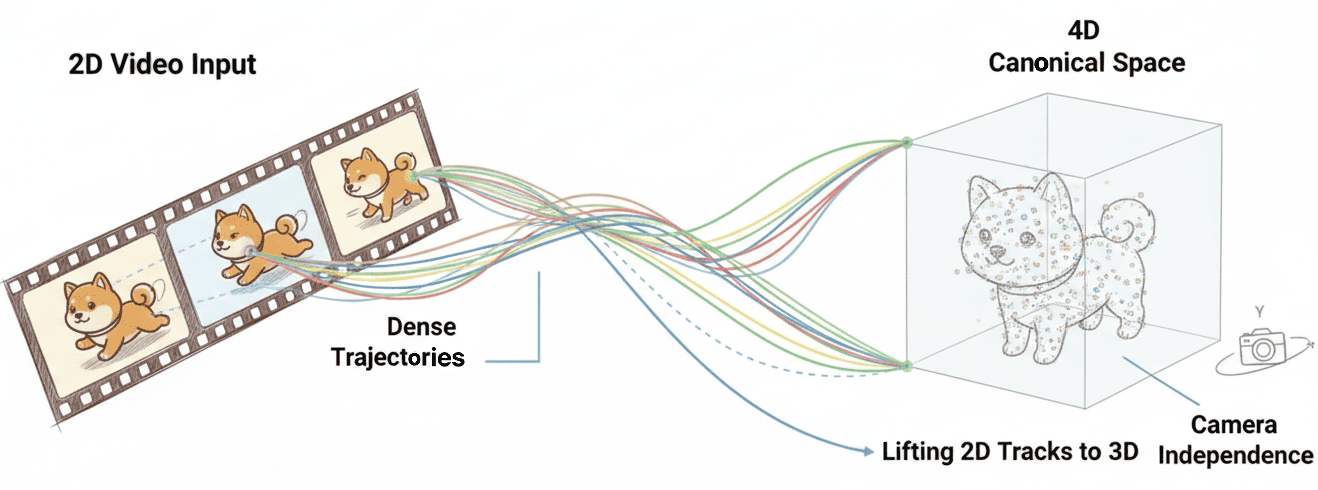}
  \caption{Trajectory-Centric Foundation Models. This illustrates the transformation of a 2D video input into dense trajectory fields (with occlusion handling) and their subsequent lifting into a 4D canonical space for volumetric reconstruction.}
  \label{fig:spatial_temporal3}
\end{figure}

Distinguished from the trajectory parametric control discussed in \S\ref{sec:trajectory_parametric_control}, where trajectories serve as parameterized constraints, the paradigm explored in this section (Figure~\ref{fig:spatial_temporal3}) reframes them as the core data representation connecting 2D visual flow to 4D physical space. With the maturation of hybrid voxel and explicit Gaussian architectures, academic focus is shifting from per-scene optimization toward generalizable 4D inference. To overcome the bottleneck of native 4D data scarcity, recent works~\cite{liu2025trace, wang2023omnimotion, xiao2024spatialtracker, leroy2024mast3r, portenier2024cotracker3} propose using dense trajectory fields as a universal intermediate representation, aiming to build unified world models with physical consistency through automated video-as-trajectory transformation. Research follows two main directions: trajectory lifting strategies and end-to-end generalization.

\textit{(\romannumeral1) Trajectory Field: A Universal Bridge Between 2D Pixels and 4D Physics.} Traditional 4D modeling often relies on expensive multi-view capture or sparse offline COLMAP calculations, making it difficult to leverage massive in-the-wild internet videos. New-generation methods advocate for treating video not as a set of $T$ individual frame images, but as a collection of continuous 3D trajectory flows $\tau$ evolving over time. \textit{Lifting 2D Tracks to 3D.} Works such as Trace Anything~\cite{liu2025trace} and SpatialTracker~\cite{xiao2024spatialtracker} redefine the input signals for 4D reconstruction. By utilizing long-range, occlusion-robust 2D point trajectories extracted by foundation models like CoTracker3~\cite{portenier2024cotracker3} or TAPIR, and applying monocular depth estimation with decoupled rigid/non-rigid optimization, they explicitly lift the 2D pixel flow $\boldsymbol{u}(t)$ into 3D spatial trajectories $\boldsymbol{T}(t)$ / $\boldsymbol{X}(t)$. The revolutionary nature of this approach is that it allows arbitrary monocular video to be converted into 4D pseudo-ground truth with physical attributes, providing infinite data fuel for training general world models. \textit{Volumetric Motion Representation.} OmniMotion~\cite{wang2023omnimotion} further proposes a \textit{quasi-3D} global motion representation. Unlike traditional optical flow $\mathcal{O}_{flow}$ that only captures relationships between adjacent frames, OmniMotion constructs a continuous bijective mapping that projects all pixels in a video into a canonical 3D space. This means the model can track visible points and predict physically plausible full lifecycle trajectories for occluded objects, breaking the limitations of visibility breaks in traditional 4D modeling.

\textit{(\romannumeral2) Foundation Models for Generalizable 4D.} Building on unified data formats, 4D foundation models with zero-shot generalization capabilities are emerging, eliminating the need for test-time optimization for each video. \textit{End-to-End Dynamic Geometric Matching.} MASt3R~\cite{leroy2024mast3r} unifies video generation and 3D reconstruction within a single Transformer architecture. By learning dense correspondences and 3D geometric transformations between image pairs, it can directly output 3D point clouds and camera motion for dynamic scenes without requiring camera parameters. This marks a shift from optimization-based pipelines to learning-based end-to-end models. \textit{Globally Consistent Structure Recovery.} To address cumulative errors in long videos, VGGSfM~\cite{wang2024vggsfm} proposes a fully differentiable global SfM framework. By using extracted dense trajectories as constraints, it solves for camera poses $\boldsymbol{P}_t$ and scene geometry in an end-to-end manner within a deep learning framework. This ensures that the world model maintains 3D structural consistency even when processing hour-long videos, overcoming the failure modes of traditional methods under dynamic object interference.

\subsubsection{Reinforcement Learning for Spatial-Temporal Alignment}

Traditional Supervised Fine-Tuning is limited by the teacher forcing mode~\cite{Villegas2017}, which often struggles to correct exposure bias in long sequence generation~\cite{bengio2015scheduled}. This frequently leads to spatial structure collapse or a temporal causality break in the late stages of inference~\cite{villegas2017learning}. To address this challenge, recent frontier works have introduced Reinforcement Learning as an adhesive for spatiotemporal fusion~\cite{black2023training}. Unlike LLMs that focus solely on text response quality, the core of RL in video generation lies in constructing a composite reward function $\mathcal{R}$. This forces the model to solve the Pareto optimization problem between Image Fidelity (Spatial) and Motion Coherence (Temporal) within the latent space~\cite{yuan2024instructvideo}. Current research primarily explores two dimensions: \textbf{Global Mixed-Reward Feedback}, which emphasizes macro-statistical balance, and \textbf{Explicit Physical Costs}, which focuses on micro-physical repair.

\paragraph{Global Mixed-Reward Feedback \& Dynamic Alignment.} 
Addressing the defects where SFT struggles to balance quality and motion, this paradigm establishes a spatiotemporal value balance at the macro level by introducing decoupled reward models. 
\textit{(\romannumeral1) Multi-dimensional Value Distillation.} T2V-Turbo-v2~\cite{li2025t2vturbov2} designs a spatially and temporally decoupled mixed reward mechanism. It utilizes HPSv2 to score single-frame aesthetics and InternVideo2 to score dynamic coherence, unifying these two mutually constrained objectives through consistency distillation. This effectively mitigates the motion averaging phenomenon caused by traditional MSE loss. 
\textit{(\romannumeral2) Deep Reward Tuning.} DR-Tune~\cite{zhou2023drtune} further notes that simple reward weighting can lead to excessive gradient variance. It proposes a Deep Reward Tuning strategy that dynamically adjusts the weights of spatiotemporal gradients along the denoising path of the diffusion model. This ensures that the model enhances temporal consistency without sacrificing the spatial fidelity of individual frames.

\paragraph{Hierarchical Structural Decoupling \& Explicit Costs.} 
Distinct from the black box optimization of global feedback, this school of thought advocates decomposing spatiotemporal consistency into differentiable explicit physical costs for micro-repair at the pixel and structural levels. 
\textit{(\romannumeral1) 3D Hierarchical Alignment.} VistaDPO~\cite{huang2025vistadpo} proposes a fine-grained hierarchical alignment framework. It decomposes the optimization objective into three orthogonal dimensions: instance-level (semantic fidelity), temporal-level (motion manifold $\mathcal{M}_{dyn}$), and perceptual-level (spatial structure). This design allows the model to surgically repair time-skipping on the temporal axis without compromising the integrity of spatial textures. 
\textit{(\romannumeral2) Flow Consistency Constraint.} InstructVideo~\cite{yuan2024instructvideo} provides specific mathematical implementation means. It moves away from general preference models in favor of reward-weighted fine-tuning, explicitly introducing a flickering penalty and optical flow consistency $\mathcal{O}_{flow}$ as cost functions. This method translates physical laws into direct gradient signals, forcing pixel flow to remain smooth during temporal evolution and resolving the high-frequency jittering problem common in long video generation.

\subsection{Preliminary Emergence of World Models}

\begin{figure}[h]
  \centering
  \vspace{-2mm}
  \includegraphics[width=0.9\linewidth]{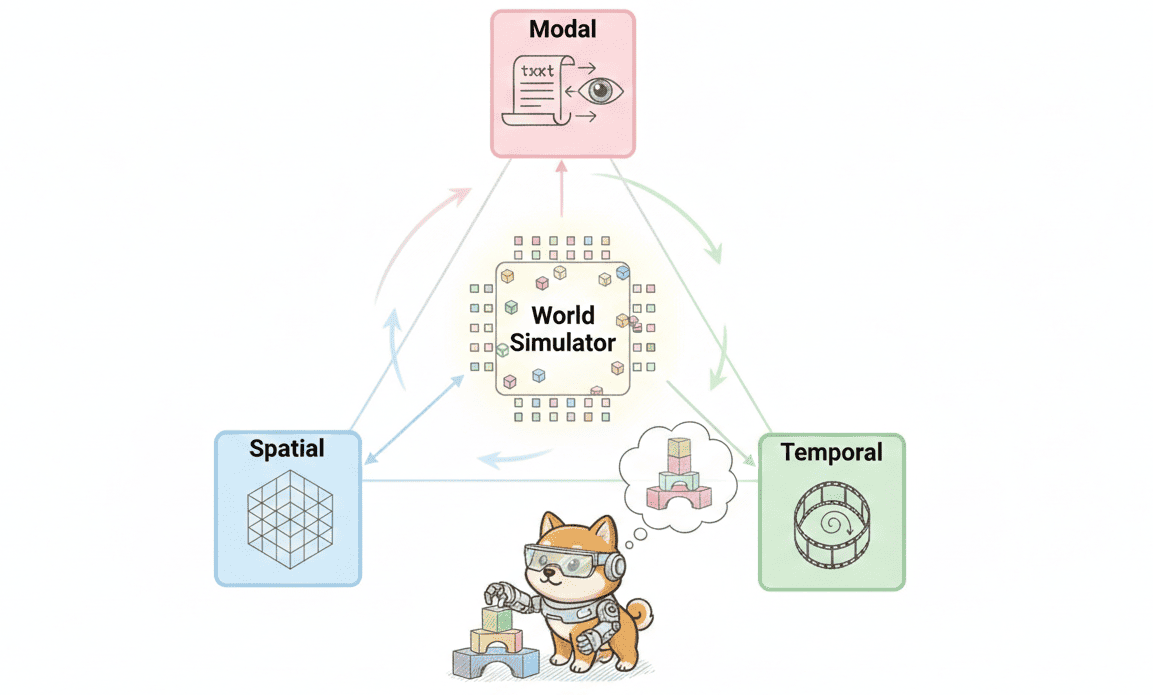}
  \vspace{-2mm}
  \caption{The Trinity of Consistency. It depicts a Robot Doge performing embodied tasks (stacking blocks) guided by counterfactual reasoning (visualized as a prediction bubble).}
  \label{fig:The Trinity of Consistency}
\end{figure}

With the maturation of unified multimodal model architectures and breakthroughs in multimodal pre-training technology, the World Model has moved beyond the stage of independent modeling for single dimensions. It has gradually begun to manifest the prototype of the \textit{Modal-Spatial-Temporal Trinity of Consistency Synergistic Emergence}, as illustrated in Figure~\ref{fig:The Trinity of Consistency}. The core characteristic of this stage is that the model is no longer a mere pixel generator but has evolved into a deductive internal simulator. Specifically, modal consistency provides multi-source interaction interfaces, spatial consistency constructs the static geometric skeleton, and temporal consistency injects the causal evolution engine. This synergistic mechanism has been quantitatively verified through benchmarks and has demonstrated a deep understanding of the physical world.
% in the closed-loop decision-making of Embodied AI.

\subsubsection{From Benchmark Establishment to Diverse Evolution}

During this evolutionary process, Sora~\cite{sun2024sora} and Open-Sora~\cite{opensora} represent dual milestones for the closed-source commercial-grade and open-source academic communities, respectively. Together, they have established the mainstream paradigm of Spacetime Patchification and DiT Architecture, while other models serve as lateral verifications of a rich technical landscape.

\paragraph{Sora: Paradigm Establishment of World Simulator.}
Sora~\cite{sun2024sora} is undoubtedly a masterpiece of trinity synergy. It does not rely on explicit 3D inductive biases but instead leverages large-scale spacetime patch training to convincingly validate the capability emergence triggered by the Scaling Law in video generation~\cite{kaplan2020scaling, wei2022emergent, peebles2023scalable}. By compressing video into spacetime patches within the latent space, the model handles high-dimensional visual data in a manner analogous to language tokens, achieving deep interoperability between spacetime and modality~\cite{williams2023neurips}. This mechanism not only breaks the \textit{modal consistency} bottleneck in long video generation but also, driven by massive data, facilitates the spontaneous emergence of an implicit understanding of the physical world. Even without explicit geometric constraints, Sora maintains perspective constancy of the \textit{spatial structure} during complex camera movements and exhibits physical interactions (\eg collisions, occlusions) that adhere to \textit{temporal causality}. This signifies that generative models have begun to possess the deductive characteristics of a world simulator~\cite{lecun2022path, ha2018world}.

\paragraph{Open-Sora: Technology Democratization \& Architecture Verification.}
As a pioneer of the open-source community, Open-Sora~\cite{opensora} successfully replicated and verified the core logic of the Video DiT, providing a transparent and efficient testbed for the academic exploration of the three consistencies. Its core innovation lies in the adoption of the Spatial-Temporal Diffusion Transformer ~\cite{ma2024latte} architecture, which utilizes a cleverly designed alternating computation mechanism for spatial and temporal attention. This decoupling and synergistic design significantly reduces computational complexity while effectively balancing \textit{spatial fidelity} within single frames and \textit{temporal coherence} across frames. Supplemented by a cascade training strategy and high-efficiency Video VAE encoders/decoders~\cite{rombach2022high, gu2023reuse}, Open-Sora further confirms the stability of minute-level long sequence generation. This demonstrates that the efficient synergy of the trinity is not solely dependent on compute stacking; rather, a rational architectural design is equally essential to achieving consistency with the physical world.

\paragraph{Transition from Passive Observation to Active Interaction.}
If Sora established the emergence of physical common sense based on large-scale observations, then interactive world models represented by Genie 1/2/3~\cite{bruce2024genie}, LingBot-World~\cite{lingbotworld2026}, and GameNGen~\cite{valevski2024gamengen} mark a fundamental leap of the trinity consistency from passive movie projection to active interactive simulation. The core breakthrough of these models lies in the explicit introduction of the action operator $\boldsymbol{a}_t$ into the spatiotemporal generation logic, transforming probabilistic modeling from $P(\boldsymbol{x}_{future} \mid \boldsymbol{x}_{past})$ to controlled state transitions $P(\boldsymbol{s}_{t+1} \mid \boldsymbol{s}_t, \boldsymbol{a}_t)$~\cite{ha2018world}. Specifically, Genie-3~\cite{bruce2024genie} utilizes an unsupervisedLatent Action Model to decouple discrete action tokens from massive unlabeled videos, using them as conditional inputs for the spatiotemporal Transformer to ensure \textit{temporal causality} under specific instructions (\eg the causal feedback of a character jumping after a specific key is pressed). LingBot-World~\cite{lingbotworld2026} further constructs a unified cognition-action manifold, coupling high-level semantic instructions with low-level physical attributes such as collision detection and force feedback within the same latent space. This architecture not only maintains macro modal consistency but also preserves \textit{spatial fidelity} under complex boundary conditions at a 60 FPS generation rate through the introduction of spatiotemporal consistency regularization. The emergence of these programmable worlds proves that the synergy of the trinity can evolve into a differentiable, predictable, and interactive \textit{World API}, providing embodied agents with a near-realistic mental sandbox simulation environment~\cite{bruce2024genie, menapace2024playable}.

\paragraph{Synergistic Corroboration of Diverse Technical Paths.}
Beyond the aforementioned models, other systems have enriched the technical map of world models from various dimensions, collectively corroborating the inevitable trend of three-consistency fusion:

\textit{(\romannumeral1) 3D Causality \& Explicit Modeling.} CogVideoX~\cite{cogvideox} and Wan2.1~\cite{wan2025} both emphasize the role of 3D VAEs. CogVideoX strengthens inter-frame dependency through 3D RoPE, while Wan2.1's Causal 3D VAE enforces the unidirectional flow of the time dimension within the latent space, significantly enhancing the physical plausibility of dynamic evolution.

\textit{(\romannumeral2) High Fidelity \& Fine-grained Control.} Gen-3~\cite{runway2024gen3alpha} and HailuoAI~\cite{hailuoai2024hailuo} focus on industrial-grade consistency performance. Gen-3 demonstrates cinematic-grade lighting maintenance and complex physical interaction simulations, while HailuoAI utilizes dedicated engine optimizations to resolve structural collapse issues in complex dynamics scenarios.

\textit{(\romannumeral3) Architecture Exploration \& Multimodal Alignment.} HunyuanVideo~\cite{hunyuan2024}, VideoCrafter~\cite{he2024videocrafter}, and LTX-Video~\cite{ltx2025} have conducted deep explorations into multimodal embedding spaces and attention mechanisms. These efforts further strengthen the semantic alignment between textual instructions and visual content, providing a solid foundation for instruction-driven world simulation.

\subsubsection{Combat Loop of Three Consistencies}

While generative models imagine the world, Embodied AI physically intervenes in the world under the guidance of the trinity. In this context, consistency is no longer merely a visual sensory indicator but the cornerstone of agent decision safety and task success. Beyond RT-2~\cite{brohan2023rt2} and GAIA-1~\cite{gaia1}, the academic and industrial sectors have evolved from simple vision-language mapping to a deep closed-loop paradigm based on physical simulation and latent space planning.

\paragraph{Interactive World Simulators: The Convergence of Physics, Logic, and 3D Fidelity.}
Building a general simulator with interactive physical dynamics is a prerequisite for embodied agents to engage in low-cost trial and error. A new generation of world models is evolving from single video prediction toward full-dimensional Digital Twins. The Google Genie series (1-3)~\cite{bruce2024genie} and Matrix-Game 2.0~\cite{yan2025} first addressed the action-logic consistency problem: Genie achieved unsupervised action space discretization through its Latent Action Model, while Matrix-Game 2.0 introduced multi-agent game-theoretic logic, allowing simulated environments to handle complex social interactions and causal arbitration. At the spatial construction level, Hunyuan 3D World Model 1.0~\cite{hunyuanworld2025tencent} and NVIDIA Cosmos~\cite{agarwal2025cosmos} have filled the gap in high-fidelity physical attributes. Hunyuan 3D replaces traditional 2D textures with generated explicit 3D assets to ensure geometric consistency during multi-view exploration by the agent; Cosmos embeds rigid/fluid dynamics equations into Transformer masks to achieve industrial-grade physical simulation. Building upon this, TwinRL-VLA~\cite{xu2026twinrlvla} further validates the practical utility of Digital Twins: by introducing the Exploration Space Expansion strategy, it enables agents to perform large-scale parallel Online RL within the digital twin environment, effectively addressing the challenges of "cold starts" and constrained data distribution inherent in real-world training. Simultaneously, to address the high-frequency texture noise often produced by generative models, V-JEPA~\cite{assran2023ijepa, bardes2024vjepa} and DreamerV3~\cite{hafner2024mastering} adhere to the non-generative prediction paradigm. They model state transitions in an abstract representation space—$Pred(Enc(x_t), z) \approx Enc(x_{t+1})$—providing agents with a denoised, efficient planning space focused on essential laws~\cite{lecun2022path}.

\paragraph{Unified Cognition-Action Manifolds: From Manipulation to Navigation.}
In real-world physical environment deployments, the core challenge lies in aligning high-dimensional semantic cognition with low-dimensional action execution on a unified manifold. This paradigm has evolved from early simple instruction mapping to large-scale, all-modal closed-loop control. In the manipulation domain, WorldVLA~\cite{cen2025WorldVLA} and LingBot-World~\cite{lingbotworld2026} represent the SOTA evolutionary directions. WorldVLA demonstrates, through massive data scaling, that world models can serve as universal action compilers, directly translating vague linguistic intent into precise joint control flows. LingBot-World further proposes the Cognition-Action Unified Manifold, utilizing an asymmetric dual-stream architecture to couple semantic instructions with tactile/force feedback signals. Combined with the intermediate geometric generation capabilities of 3D-VLA~\cite{zhen20243dvla}, this explicitly resolves spatial ambiguity and physical constraints during manipulation. In the navigation domain, UniAD~\cite{hu2023uniad} and DriveVLM~\cite{tian2024drivevlm} extend this logic to autonomous driving. UniAD breaks the barriers between perception and planning by constructing a full-stack unified feature flow, while DriveVLM leverages LLMs to demonstrate human-like counterfactual reasoning. This is essentially analogous to the logic of Matrix-Game 2.0: conducting causal simulations within the world model to achieve an evolution from reactive obstacle avoidance to proactive game-theoretic robust decision-making.

\paragraph{Spatio-Temporal Constraints Based on Physical Causality}

In the physical world of embodied AI, spatio-temporal alignment must transcend mere visual plausibility and satisfy strict physical causality. Purely generative video models are often plagued by physical hallucinations, such as object interpenetration or levitation, while Digital Twins are emerging as the ultimate spatio-temporal anchor to address this issue. Studies represented by TwinRL-VLA~\cite{xu2026twinrlvla} and RoboGen~\cite{wang2023robogen} propose a solution based on explicit modeling: leveraging the state evolution of physics engines to replace the pixel prediction of neural networks.
\textit{Explicit State Reconstruction.} This class of methods constructs a Twin World isomorphic to the real world. In this space, spatio-temporal evolution is no longer sampled from a probability distribution $P(x_{t+1}|x_t)$ but adheres to rigid body dynamics equations $s_{t+1} = f_{physics}(s_t, a_t)$~\cite{Xie2024}. This imposes inviolable hard constraints on the spatio-temporal manifold; any generated spatio-temporal trajectory that violates physical laws is directly truncated or penalized during the simulation stage.
\textit{Consistency Assurance in Sim-to-Real Transfer.} Empirical studies demonstrate that this physics-engine-based spatio-temporal alignment possesses exceptionally strong transfer robustness. SimplerEnv~\cite{li2024evaluating} and ManiSkill2~\cite{gu2023maniskill2} prove that policies trained in simulated spatio-temporal spaces that have undergone strict physical verification can be transferred to the real world with minimal adaptation cost (Sim-to-Real Gap). This mechanistically proves that spatio-temporal alignment achieved through simulation-based automatic search is more generalizable than that achieved solely through visual imitation, as it captures the underlying causal dynamic structure rather than merely pixel-level superficial correlations~\cite{wang2023gensim}.

In summary, the development of the World Model is at a key inflection point toward the synergistic emergence of the modality-spatial-temporal trinity of consistency~\cite{lecun2022path}. From the implicit learning of physical laws in generative models like Sora to the causal deduction in latent space by embodied agents like 3D-VLA and DreamerV3, this series of theoretical verifications and implementations reveals a core trend: \textbf{the next stage of AGI lies in constructing a General World Simulator capable of internalizing physical laws and possessing counterfactual reasoning capabilities}~\cite{pearl2009causality, bengio2019system, firoozi2023foundation}. This trinity synergy not only resolves the spatiotemporal hallucination issues in video generation~\cite{ji2023survey} but also, by endowing models with a deep understanding of the physical world, bridges the last mile from digital generation to physical interaction. It lays a solid architectural foundation for AGI to establish a unified cognition of the objective world~\cite{goertzel2014artificial}.

\section{Challenges, Benchmarks, and Outlook}
\label{section:challenges}
\subsection{Core Challenges from Preliminary Fusion to True Unification}

Although the trinity of consistency across modality, spatial, and temporal dimensions has begun to show signs of synergy within the frameworks of MM-DiT and LMMs, the ultimate vision of constructing world model still faces a significant theoretical gap~\cite{lecun2022path}. This challenge transcends simple optimization of generation quality; its essence lies in the fundamental lack of completeness in physical ontology and robustness in causal epistemology in current models~\cite{marcus2020next}.

\textit{(\romannumeral1) Primary gap lies in the lack of differentiability of physical Authenticity.} Existing diffusion models and autoregressive architectures still hold pixel-level or token-level likelihood maximization as their highest goal~\cite{rombach2022high, esser2021taming}. This leads generation results into the trap of visual plausibility—where rigid bodies hover without support, fluid momentum is not conserved, and elastic coefficients drift with gestures~\cite{bear2021physion}. The model merely learns the statistical textures of physical phenomena rather than the underlying vector mechanics. Future challenges lie in how to embed Hamiltonians, conservation laws, or differential equations into the loss function as \textit{soft constraints} or even \textit{differentiable operators}, forcing the network to move from painting the skin to painting the bones~\cite{raissi2019physics}.

\textit{(\romannumeral2) The butterfly effect brittleness of long-term causal chains remains unsolved.} Current spatiotemporal attention mechanisms can only maintain short-range memory for tens of seconds~\cite{ho2022imagen}. Once entering the hour-day scale, object identity consistency and event logic suffer an avalanche of failure due to error accumulation~\cite{villegas2017learning}. The solution may lie in introducing hierarchical implicit dynamics: the macro level maintains abstract causality via symbolic narratives or scene graphs~\cite{mao2019neuro}, the meso level compresses event nodes with sparse 4D representations, and the micro level utilizes high-dimensional attention to complete texture details, achieving a multi-clock mechanism of slow variable fidelity and fast variable sampling~\cite{bengio2019system, saxena2021clockwork}.

\textit{(\romannumeral3) The paradigm shift of controllability and interactivity is imperative.} Upgrading prompts to APIs means users are no longer passive describers but active \textit{World Editors}~\cite{pan2023draggan, brooks2023instructpix2pix}. Users should be able to insert forces $\boldsymbol{F}_{force}$ at arbitrary spatiotemporal coordinates, modify materials, reset boundary conditions, and obtain real-time feedback that adheres to physical laws~\cite{hu2020difftaichi}. This implies that generative networks must embed neural surrogate models, allowing gradients to penetrate the complete chain of user action $\boldsymbol{a}_t \to$ state evolution $\mathcal{T}(\boldsymbol{s}_{t+1} \mid \dots) \to$ sensory observation $\boldsymbol{x}_{img}$, transforming blind box generation into draggable, scriptable, and programmable online simulation~\cite{bruce2024genie, menapace2024playable}.

\textit{(\romannumeral4) Lastly, expanding the horizon to agentic evolution and digital ecosystems.} The final form of world models should not stop at being a physical sandbox but should become the Matrix that accommodates the evolution and gaming of autonomous agents~\cite{park2023generative, wang2023voyager}. First, the introduction of \textbf{multi-agent gaming} requires the model to upgrade from modeling physical causality to modeling social causality~\cite{leibo2017multi}. In complex non-zero-sum games, the world model must be able to simulate the intentionality and strategic behaviors of multiple agents, deducing the Nash Equilibrium dynamics under the interaction of different policies $\pi_i$, rather than remaining limited to single-agent physical feedback~\cite{shoham2008multiagent, silver2018general}. Second, the rise of \textbf{GUI agents} requires world models to possess cross-domain generalization capabilities—extending from simulating the 3D physical world to simulating 2D digital environments (the digital world)~\cite{hong2023cogagent}. The model needs to understand the functional semantics of screen layouts and the state transition logic $P(\boldsymbol{S}^{\text{screen}}_{t+1} \mid \boldsymbol{S}^{\text{screen}}_t, \boldsymbol{a}_{\text{ui}})$ of API calls, thereby supporting agents in achieving an end-to-end closed loop from perception to action within the virtual world of operating systems. This marks the evolution of the world model from a pure physical simulator into a General World OS encompassing both physical and digital attributes~\cite{xi2023rise, wang2023survey}.

\subsection{Constructing Comprehensive Evaluation Benchmarks}

\definecolor{hidden-blue}{HTML}{4a90e2}
\definecolor{hidden-black}{HTML}{333333}
\newlength{\LevelOneWidth}
\newlength{\LevelTwoWidth}
\setlength{\LevelOneWidth}{5em}
\setlength{\LevelTwoWidth}{15em}
\tikzstyle{my-box}=[
  rectangle,
  draw=hidden-black,
  rounded corners,
  text opacity=1,
  minimum height=1.5em,
  inner sep=2pt,
  align=center,
  fill opacity=.5,
]
\tikzstyle{root}=[
  align=center,
]
\tikzstyle{leaf}=[
  my-box, 
  minimum height=1.5em,
  text=black,
  font=\footnotesize,
  inner xsep=4pt,
  inner ysep=3pt,
  fill opacity=1,
]
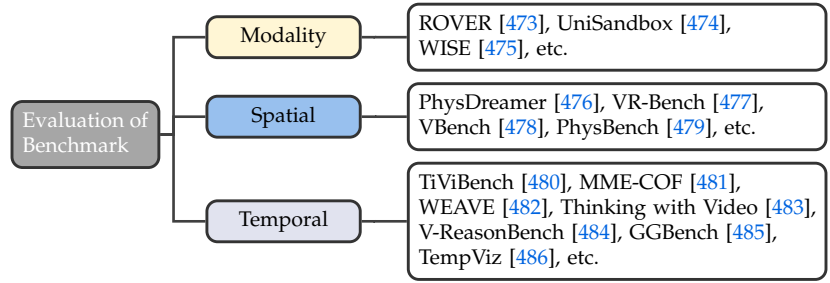
\begin{figure*}[t]
\vspace{-2mm}
\centering
\begin{forest}
  forked edges,
  for tree={
    grow=east,
    reversed=true,
    anchor=base west,
    parent anchor=east,
    child anchor=west,
    base=left,
    font=\footnotesize, % 全局字体设为 small
    rectangle,
    draw=hidden-black,
    rounded corners,
    align=left,
    % minimum width=4em, <-- 可以移除，由 text width 控制
    edge+={darkgray, line width=1pt},
    s sep=4pt,      % 增加节点垂直间距，防止挤在一起
    l sep=6mm,      % 增加层级水平间距，让树铺开
    inner xsep=4pt,
    inner ysep=3pt,
    line width=1.1pt,
    ver/.style={child anchor=north, parent anchor=east, anchor=center},
  },
  where level=1{text width=\LevelOneWidth, font=\footnotesize}{},
  where level=2{text width=\LevelTwoWidth, font=\footnotesize}{}, % 参考文
  [Evaluation of \\Benchmark, ver, fill=gray!70, text=white, root
    [Modality, fill=yellow!32, leaf
      [{ROVER~\cite{rover}, UniSandbox~\cite{unisandbox}, \\WISE~\cite{wise}, etc.}]
    ]
    [Spatial, fill=hidden-blue!57, leaf
      [{PhysDreamer~\cite{cao2024physdreamer}, VR-Bench~\cite{vrbench}, \\VBench~\cite{huang2024vbench}, PhysBench~\cite{fan2024physbench}, etc.}]
    ]
    [Temporal, fill=purple!27, leaf
      [{TiViBench~\cite{tivibench}, MME-COF~\cite{mmecof}, \\WEAVE~\cite{weave}, Thinking with Video~\cite{thinkingwithvideo}, \\V-ReasonBench~\cite{vreasonbench}, GGBench~\cite{ggbench}, \\TempViz~\cite{holtermann2026tempviz}, etc.}]
    ]
  ]
\end{forest}
\caption{Evaluation of benchmarks.}
\label{fig:benchmark}
\end{figure*}

As the world model $\mathcal{W}$ leaps from short video generation toward becoming a physical simulator~\cite{sun2024sora}, distribution statistical metrics represented by FID and FVD have become inadequate for capturing deep-level logical fractures~\cite{heusel2017gans, unterthiner2018towards, borji2019pros}. Continuing to rely on such perceptual metrics would cause model optimization to stall in local optima that are visually realistic but causally distorted. To drive the domain toward a deducible and verifiable direction, the community has introduced a series of evaluation benchmarks targeting the core requirements of the Trinity, as shown in Figure~\ref{fig:benchmark}~\cite{huang2024vbench, rover}, aiming to establish a complete verification loop from symbolic logic to physical simulation.

\subsubsection{Modal Consistency: From Symbol Mapping to Knowledge Synergy}

Traditional modal consistency evaluation relies primarily on CLIP scores for shallow semantic co-occurrence calculations. The current evolutionary direction has shifted toward \textbf{knowledge internalization and cross-modal reasoning}.

\paragraph{Knowledge-driven Alignment.} WISE~\cite{wise} introduces a structured prompt library covering natural sciences, utilizing WiScore to quantify a model's ability to internalize world knowledge into visual representations; by constructing counterfactual negative samples, it fills the evaluation gap between symbol and perception. ROVER~\cite{rover} verifies the closed-loop coherence of the bidirectional generation chain (Text $\leftrightarrow$ Pixel) through reciprocal reasoning.

\paragraph{Execution Gap between Understanding and Generation.} UniSandbox~\cite{unisandbox} reveals an asymmetric phenomenon where understanding is correct but generation is wrong. This benchmark quantifies the execution gap of models in complex attribute transfer and mathematical visualization, proving that the introduction of an explicit CoT) is a key mechanism for bridging this gap.

\subsubsection{Spatial Consistency: From Visual Similarity to Topological \& Physical Verification}

Evaluation in the spatial dimension has shifted from perceptual visual scoring to rigorous 3D topological structure and physical repulsiveness verification. We categorize related work into two levels: semantic logic verification and physical dynamics verification.

\paragraph{Topological Logic \& Interactive Reasoning.} VR-Bench~\cite{huang2024vbench} focuses on complex spatial relation reasoning, particularly occlusion, perspective, and path planning tasks. Its research reveals a significant modality dependency, where existing models perform much worse on pure visual spatial reasoning than on text-assisted reasoning. VBench~\cite{huang2024vbench} further proposed decoupled evaluation standards, using VLM-as-a-judge to refine spatial consistency into object constancy and spatial relations. Crucially, by calculating the graph edit distance between the generated scene graph and the prompt scene graph, it precisely quantifies the logical accuracy of spatial layouts.

\paragraph{Physical Simulation \& Penetration Detection.} To compensate for the lack of dynamic constraints in pure visual evaluation, PhysBench~\cite{fan2024physbench} and PhysDreamer~\cite{zhang2024physdreamer} introduce physics engines as ground-truth referees. They reconstruct pseudo-3D point clouds through depth estimation and calculate the minimum Euclidean distance $\min ||\boldsymbol{p}_i - \boldsymbol{p}_j||_2$ between objects as a penalty term. This method strictly detects spatial penetration and floating artifacts, establishing a spatial evaluation standard for rigid bodies under Newtonian mechanics constraints.

\subsubsection{Temporal Consistency: From Inter-frame Smoothness to Logical Causal Evolution}

A profound paradigm shift has occurred in the evaluation of temporal consistency: moving from a focus on the \textbf{visual continuity} between video frames to the \textbf{logical chronology} underlying the generation process. We categorize this into visual physical chronology and symbolic logical chronology.

\paragraph{(1) Static Temporal Semantics (Time-as-Attribute).} 
The physical foundation of temporal consistency lies in the model's state awareness of entities as they evolve over time (\eg seasons, aging, historical eras). Addressing previous limitations that focused solely on video dynamics, TempViz~\cite{holtermann2026tempviz} proposed a static evaluation paradigm for temporal knowledge. By constructing a dataset containing 7.9k prompts, this work quantifies the ability of text-to-image models to understand time-variant attributes. The study reveals that even state-of-the-art models exhibit significant knowledge gaps when generating contextually relevant images (\eg distinguishing between a spring landscape and a winter landscape) and proves that automated metrics like CLIP fail to capture such temporal nuances.

\paragraph{(2) Visual Physical Chronology (Thinking-in-Video).} TiViBench~\cite{tivibench} introduced the Think-in-Video concept, forcing models to demonstrate the problem-solving process of physical tasks (\eg fluid motion, maze navigation) by generating video. The core objective is to verify whether the intermediate state trajectory $\tau = \{\boldsymbol{s}_1, \dots, \boldsymbol{s}_T\}$ adheres to Markov dynamics. \textbf{V-ReasonBench}~\cite{vreasonbench} further introduced the optical flow operator $\mathcal{O}_{flow}$ to monitor motion mutations, effectively avoiding visual hallucinations from VLM referees.

\paragraph{(3) Symbolic Logical Chronology \& Process Verifiability.} While GGBench~\cite{ggbench} is oriented toward geometric reasoning, its core mechanism utilizes GeoGebra as an executable environment to verify the step-by-step construction sequence $S_1 \rightarrow S_2 \rightarrow \dots \rightarrow S_n$ of multimodal reasoning. Geometric construction is essentially the construction of a causal chain along the time axis. GGBench not only checks the correctness of the final image but also verifies temporal dependency in the construction steps through code execution (\eg points A and B must be defined before line segment AB can be constructed). This evaluation of logical time reveals whether a world model possesses reasoning robustness isomorphic to physical time when handling long-range dependency tasks.

\paragraph{(4) Long-range Defects \& Constancy Failure.} Regarding the catastrophic forgetting common in long sequence generation, MME-COF~\cite{mmecof} and WEAVE~\cite{weave} systematically expose physical hallucinations in SOTA models (\eg rigid body collisions violating the law of reflection) and failures in object permanence. Notably, Thinking with Video~\cite{thinkingwithvideo} points out that the powerful temporal reasoning of models often relies on text priors from LLMs rather than native visual causal discovery capabilities.

\subsubsection{Limitations of Existing Benchmarks \& Design Rationale of Our Benchmark}

Although existing evaluation systems are effective in verifying single-point capabilities, they exhibit significant \textbf{structural defects} in their evaluation paradigms when judged by the standards of a \textbf{general world simulator}. These defects lead to a severe disconnect between evaluation results and the model's actual physical capabilities, manifesting at four pragmatic levels:

\paragraph{(1) Soft Ceiling of Metrics \& Judge Hallucination.} 
Current mainstream benchmarks (\eg TiViBench~\cite{tivibench}, V-ReasonBench~\cite{vreasonbench}) rely excessively on MLLMs such as GPT-4o or Gemini as referees. This model evaluating model approach possesses an intrinsic defect: VLMs themselves have extremely low perception precision for fine-grained physical attributes (\eg friction coefficients, fluid viscosity)~\cite{tong2024eyes}, often resulting in misjudgments due to visual masking logic—where a generated video is awarded a high score as long as the frames are smooth, even if it violates Newton's Third Law. Although recent work has attempted to introduce self-reflection/critic models~\cite{shao2025deepseekmathv2} or design complex fine-grained rubrics to reduce variance, these patch-like corrections do not address the core contradiction: the lack of hard verification based on simulation engine ground truth~\cite{bear2021physion}. Pure visual referees will never distinguish between physical simulation and visual deception, causing evaluation to remain at the level of surface semantics without reaching physical essence.

\paragraph{(2) In-Distribution Memory Masks OOD Generalization Shortcomings.} 
Existing datasets~\cite{huang2024vbench,mmecof} are largely collected from real-world videos or standard game recordings, which often causes large models to fall into the trap of rote memorization of training data~\cite{carlini2023extracting}. This fitting effect fails completely in OOD scenarios, manifesting as: causal chain ruptures in ultra-long temporal sequences (\eg an object disappearing after being occluded for one minute)~\cite{piloto2022intuitive}; attribute confusion in multi-object complex interactions (\eg color swapping after three objects collide)~\cite{yi2020clevrer}; and deduction failure in counter-intuitive physical environments (\eg negative gravity or non-Euclidean geometric space). As the isolation experiments in UniSandbox~\cite{unisandbox} revealed, when common visual backgrounds are stripped away and models are forced to make physical predictions under unfamiliar combinations, their performance drops significantly. This proves that current high scores often stem from \textit{overfitting} specific distributions rather than truly learning transferable \textit{world laws}.

\paragraph{(3) Error Accumulation in Long-range Generation \& Lack of Process Verification.} 
The vast majority of benchmarks only test short sequence (<10s) generation, masking the state drift issues of world models in long-range simulation~\cite{voleti2022mcvd}. The deep technical crux of this problem is that existing generation architectures (whether autoregressive or diffusion models) inherently lack online process verifiers and physical constraint correction modules~\cite{lightman2023arxiv}. Unlike traditional physics engines that solve equations frame-by-frame, generative models rely primarily on probabilistic sampling. Minor physical errors (\eg collision penetration, slight non-conservation of momentum) can undergo exponential amplification (the butterfly effect) as the timestep $t$ advances in the absence of \textit{differential equation hard constraints}, eventually leading to the logical collapse of the entire world~\cite{karniadakis2021physics, raissi2019physics}. Existing benchmarks lack a deep probe into this generation process verifiability and cannot quantify a model's ability to counter entropy increase in long sequences.

\paragraph{(4) Lack of Causal Probes for Active Intervention.} Existing evaluations operate in a static spectator mode, only requiring the model to predict what happens next. True world cognition must undergo the test of an intervener mode, namely counterfactual reasoning~\cite{pearl2009causality}. For example, If the support is removed at this moment, how will the object trajectory $\tau$ change?~\cite{ahmed2021causalworld}. Current benchmarks lack an evaluation interface supporting such parameterized interventions, making it impossible to verify whether a model has constructed a structured causal graph or is merely performing pixel-level probabilistic completion.

Faced with the quadruple dilemma of evaluation subjectivity, scenario greenhouse, temporal myopia, and interaction static, constructing a next-generation evaluation benchmark characterized by hardcore physical standards, dynamic long-range evolution, and support for causal intervention has become a top priority. To systematically decouple and evaluate the three core consistencies of world models—modality, spatial, and temporal consistency—and their pairwise fusion relationships, subsequent work in this paper introduces \textbf{CoW-Bench}. Unlike previous datasets that relied on static images or vague semantic scoring (\eg CLIP score)~\cite{hessel2021acl}, CoW-Bench organizes evaluation around six task categories derived from the three consistencies and their intersections, comprising 18 sub-tasks in total. Each sub-task is paired with five carefully designed human checklists, yielding a comprehensive, task-driven protocol with fine-grained criteria to pinpoint complementary failure modes and enable more precise, interpretable quantification.

\subsection{Ultimate Outlook: General World Simulator}

As the aforementioned challenges are sequentially overcome, the World Model $\mathcal{W}$ will shed its guise as a content generation tool and undergo a dimensional ascent to become a General World Simulator~\cite{sun2024sora, lecun2022path}—a digital universe capable of instantiating arbitrary physical laws and narrative rules on demand. For scientific exploration, it serves as a \textit{virtual laboratory} for verifying complex hypotheses; for Embodied AI, it acts as an inexhaustible \textit{safe training ground} and a real-time online \textit{brain vestibule}—where robots can perform extreme trial and error at a millisecond-level and transfer distilled policies $\pi$ to reality via zero-shot transfer~\cite{tobin2017domain, hafner2024mastering, ma2024eureka}.

Furthermore, when world models can self-consistently simulate the multiple entanglements of physics, society, and emotion~\cite{park2023generative, wang2023voyager}, we will possess, for the first time, an ultimate testbed capable of mirroring all externalities of human intelligence. In that realm, constructing world models and understanding the essence of intelligence will merge into one: the world provides constraints while intelligence generates hypotheses, and the two endlessly negotiate, converge, and evolve within differentiable spacetime~\cite{silver2021reward}.

\section{CoW-Bench}
\subsection{Dataset}
\subsubsection{Dataset Construction}
\label{sec:dataset_construction}

\noindent\textbf{Consistency-Centered Task Blueprinting}  
We construct the overall task framework of CoW-Bench around the three core consistencies of world models—modal, spatial, and temporal consistency—and their pairwise integration. Each task category is further decomposed into three sub-tasks designed to characterize distinct yet complementary failure modes within the same consistency dimension (see Table~\ref{tab:task_subtasks}). 
To ensure that evaluation signals are interpretable and attributable, we introduce a \textit{Single-Consistency Variable Control Protocol} during the task design phase: for each sub-task, only variables directly related to the target consistency are permitted to vary, while other potential confounding factors (\eg number of entities, background complexity, camera movement, motion magnitude, and occlusion conditions) are explicitly constrained. This design avoids coupling interference between different consistency factors, allowing model behavior to be stably attributed to the target capability.

\begin{table*}[ht]
\centering
\caption{Taxonomy of Tasks in CoW-Bench. The benchmark covers three foundational dimensions: \textbf{M} (Modal Consistency), \textbf{S} (Spatial Consistency), and \textbf{T} (Temporal Consistency). 
Crucially, it probes their deep synergies required for world simulation: \textbf{M$\times$S}, \textbf{M$\times$T}, and \textbf{S$\times$T}. 
Each family is further decomposed into three specific sub-tasks to isolate distinct failure modes.}
\label{tab:task_subtasks}
\setlength{\tabcolsep}{3.6mm}
\renewcommand{\arraystretch}{1.05}
\small
\begin{tabular}{l l l l}
\toprule
\multirow{2}{*}{\textbf{Task}} & \multicolumn{3}{c}{\textbf{Sub-tasks}} \\
\cmidrule(lr){2-4}
& \textbf{I. Basic / Atomic} & \textbf{II. Structured / Dynamic} & \textbf{III. Complex / Constraint} \\
\midrule
\rowcolor[HTML]{EFEFEF} 
\multicolumn{4}{l}{\textit{\textbf{Single-Consistency Dimensions}}} \\
\textbf{M} 
& Style/Material transfer
& Fine-grained control
& Multi-constraint composition \\
\textbf{S} 
& Planar layout
& Hierarchical occlusion
& Multi-view 3D structure \\
\textbf{T} 
& Worldline persistence
& Rule-guided evolution
& Ordered stage transitions \\
\midrule
\rowcolor[HTML]{EFEFEF} 
\multicolumn{4}{l}{\textit{\textbf{Cross-Consistency Synergies}}} \\
\textbf{M$\times$S}& Semantic planar binding
& Semantic hierarchy control
& Semantic 3D view consistency \\
\textbf{M$\times$T}& Long-horizon anchoring
& Attribute dynamics alignment
& Triggered event compliance \\
\textbf{S$\times$T}& Planar maze trajectory
& Occlusion dynamics
& 3D loop navigation coherence \\
\bottomrule
\end{tabular}
\end{table*}

\noindent\textbf{Reasoning-Driven Seed Construction}  
Once the task blueprint is finalized, we first construct a set of seed instances to anchor the logical core of each task. This phase employs models with deep reasoning capabilities, whose objective is not merely to generate samples matching a description, but to accurately internalize target consistency constraints and design challenging instances capable of authentically triggering corresponding failure modes. 
% Each seed instance adopts a unified structured representation, including an English text prompt (\texttt{inputText}), a Chinese initial state description (\texttt{inputImageDesc}), and English specifications for the expected image and video outputs (\texttt{outputImageExpect}, \texttt{outputVideoExpect}). This structured design binds conditions, initial states, and target results into verifiable units, providing a stable reference for subsequent controlled expansion.
Each seed instance adopts a unified structured representation, including a text prompt (\texttt{inputText}), an initial state description (\texttt{inputImageDesc}), and specifications for the expected image and video outputs (\texttt{outputImageExpect}, \texttt{outputVideoExpect}). This structured design binds conditions, initial states, and target results into verifiable units, providing a stable reference for subsequent controlled expansion.

\subsubsection{Dataset Analysis}
% \subsection{Dataset Analysis}
\label{sec:dataset_analysis}

\begin{figure}[t]
  \centering
  \includegraphics[width=0.85\linewidth]{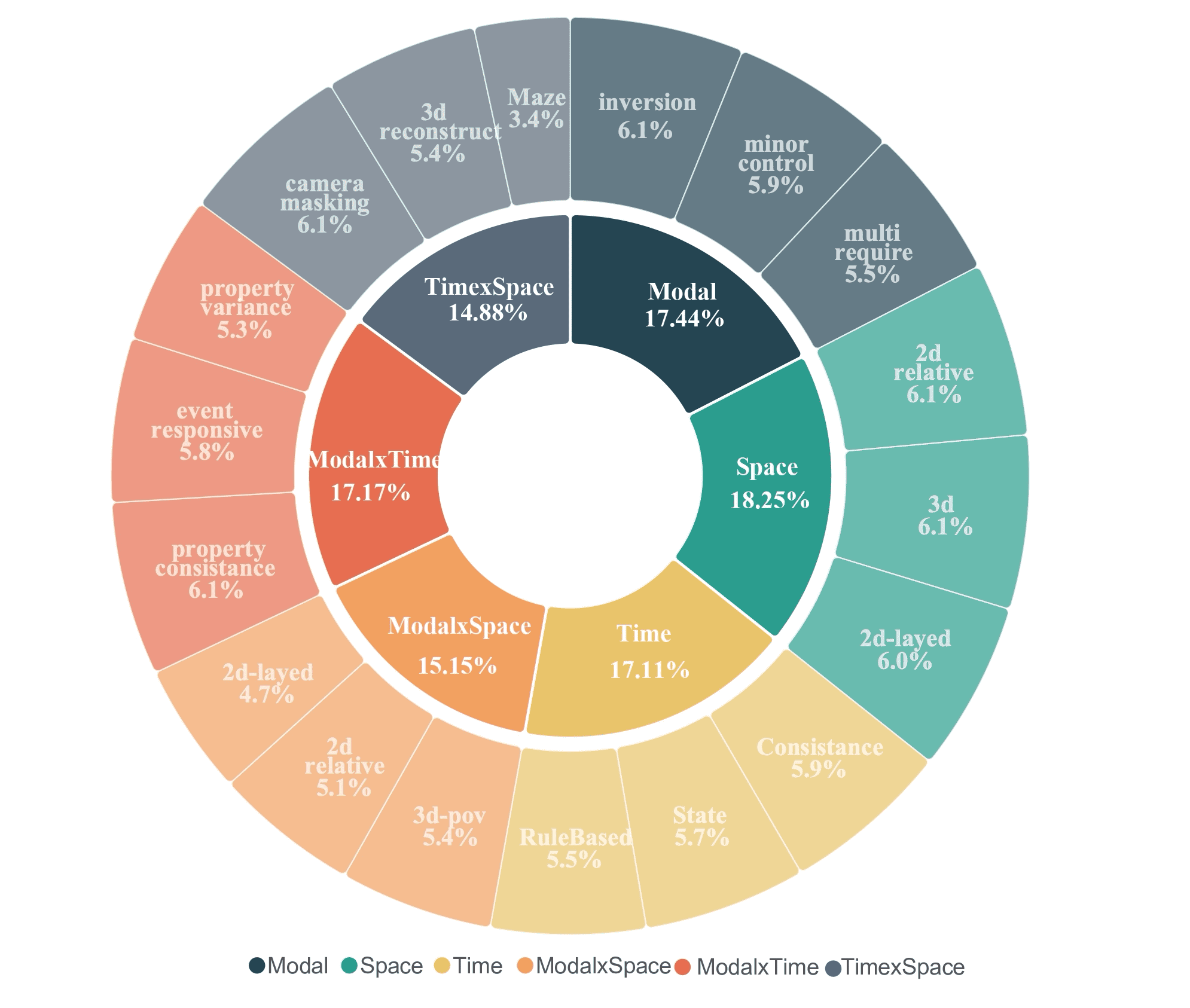}
  \vspace{-2mm}
  \caption{Hierarchical Taxonomy of CoW-Bench. The inner ring represents the main consistency dimensions (Modal, Space, Time), while the outer ring details the 18 fine-grained sub-tasks. The uniform sector sizes visually confirm the rigorously balanced distribution of the dataset.}
  \label{fig:sunburst}
\end{figure}

\begin{table}[t]
  \centering
  \small
  \renewcommand{\arraystretch}{0.95}
  \setlength{\tabcolsep}{10pt}
  \caption{Comprehensive Statistics of CoW-Bench.}
  \label{tab:umm_bench_stats}
  \begin{tabular}{lllccccccc}
    \toprule
    \multirow{2}{*}{\textbf{Mode}} & \multirow{2}{*}{\textbf{Task}} & \multirow{2}{*}{\textbf{Sub-Task}} & \multirow{2}{*}{\textbf{$N$}} & \multirow{2}{*}{\textbf{Scene}} & \multirow{2}{*}{\textbf{Diff}} & \multicolumn{4}{c}{\textbf{Complexity Metrics (Avg.)}} \\
    \cmidrule(lr){7-10}
      & & & & & & \textbf{Prmpt} & \textbf{ImgR} & \textbf{Act} & \textbf{Elem} \\
    \midrule
    \multirow{9}{*}{\textbf{Single}} & \multirow{3}{*}{\textbf{Modal}} & Subj-Attr & 91 & Obj & Easy & 7.1 & 13.3 & 22.0 & 2.1 \\
      & & Minor-Ctrl & 87 & Obj & Med & 37.4 & 37.4 & 37.1 & 1.6 \\
      & & Multi-Req & 81 & Mix & Hard & 74.8 & 13.6 & 64.7 & 1.6 \\
          \cmidrule(lr){2-10}
& \multirow{3}{*}{\textbf{Space}} & 2D-Layed & 89 & Mix & Med & 38.2 & 24.7 & 32.8 & 2.4 \\
      & & 2D-Rel & 91 & Obj & Easy & 42.7 & 17.1 & 52.8 & 1.7 \\
      & & 3D & 91 & Room & Hard & 47.1 & 35.0 & 57.4 & 2.2 \\
          \cmidrule(lr){2-10}
& \multirow{3}{*}{\textbf{Time}} & Consist & 88 & Obj & Med & 16.4 & 33.8 & 38.6 & 2.1 \\
      & & Slow-Evol & 81 & Obj & Easy & 3.3 & 33.3 & 40.1 & 2.9 \\
      & & State & 85 & Mix & Hard & 8.0 & 31.4 & 77.7 & 2.0 \\
    \midrule
    \multirow{9}{*}{\textbf{Cross}} & \multirow{3}{*}{\textbf{M$\times$S}} & 2D-Layed & 69 & Room & Med & 46.1 & 35.5 & 27.1 & 2.1 \\
      & & 2D-Rel & 76 & Obj & Med & 48.7 & 33.6 & 44.6 & 1.6 \\
      & & 3D-POV & 80 & Out & Hard & 65.4 & 33.5 & 38.5 & 2.4 \\
          \cmidrule(lr){2-10}
& \multirow{3}{*}{\textbf{M$\times$T}} & Event-Rsp & 86 & Mix & Med & 44.7 & 22.7 & 36.4 & 2.7 \\
      & & Prop-Cons & 91 & Obj & Med & 47.2 & 28.4 & 37.7 & 1.8 \\
      & & Prop-Var & 78 & Mix & Hard & 51.6 & 26.3 & 50.7 & 2.3 \\
          \cmidrule(lr){2-10}
& \multirow{3}{*}{\textbf{T$\times$S}} & 3D-Recon & 80 & Out & Hard & 35.4 & 63.4 & 43.6 & 2.1 \\
      & & Maze-2D & 50 & Flat & Hard & 12.5 & 15.0 & 35.5 & 3.0 \\
      & & Cam-Mask & 91 & Mix & Hard & 10.0 & 27.3 & 57.2 & 2.2 \\
    \bottomrule
  \end{tabular}
  \vspace{-2mm}
\end{table}

To demonstrate that CoW-Bench serves as a rigorous and non-trivial benchmark for evaluating World Models, we conduct a comprehensive analysis of its statistical distribution, fine-grained complexity, and semantic diversity. All statistics reported are based on the audited data presented in Table~\ref{tab:umm_bench_stats}.

\paragraph{Statistics and Hierarchical Ontology.}
CoW-Bench comprises \textbf{1,485} meticulously constructed samples, organized into a two-level hierarchy: a \textit{Modal Level} (Single vs. Cross) and a \textit{Task Level} (spanning Modal, Spatial, Temporal dimensions and their intersections). Unlike prior benchmarks that often exhibit long-tail distributions leading to evaluation bias, CoW-Bench maintains strict \textbf{distributional balance}. As shown in Table~\ref{tab:umm_bench_stats}, each of the 18 fine-grained sub-tasks contains between 69 and 91 samples (with the specific inclusion of 50 hard Maze cases). This uniformity ensures a fair and unbiased assessment across all capability dimensions, preventing models from achieving inflated scores by overfitting to simple or frequent task types.

\paragraph{Fine-grained Complexity Analysis.}
A core design principle of CoW-Bench is the coverage of a comprehensive difficulty gradient. We argue for the non-triviality of the tasks from three complementary dimensions.
\textit{(1) Instruction Span \& Semantic Depth.} The dataset exhibits significant variance in instruction complexity. Ranging from atomic tasks like \textit{Modal-Subj-Attr} (avg. 7.1 words) to compositional tasks like \textit{Modal-Multi-Require} (avg. 74.8 words), this vast span (7.1--74.8 words) challenges the robustness of World Models in language understanding, requiring them to handle both explicit short commands and long-context, multi-constraint instructions.
\textit{(2) Visual \& Cognitive Load.} We quantify visual complexity using the average element count per sample. Quantitative analysis reveals that cross-modal tasks generally impose a higher cognitive load (\eg \textit{3D-Reconstruct} involves 2.1 complex elements on average, significantly higher than the 1.6 in single-modal tasks). This confirms that cross-modal tasks effectively probe the model's retention capabilities in visually dense and structurally complex scenes.
\textit{(3) Dynamic Evolution Complexity.} Beyond static elements, the Action Complexity metric highlights the temporal richness of the benchmark. Tasks such as \textit{Time-State} exhibit extremely high action complexity (avg. 77.7 words), indicating that the generated videos contain intricate dynamic evolutions rather than simple static scene translations.

% \paragraph{Full-Spectrum Diversity \& Quality Assurance.}
% To ensure robustness against overfitting, CoW-Bench is designed with high semantic and visual diversity.
% \textit{(1) Scene-Difficulty Spectrum.} As shown in the \textit{Scene} and \textit{Diff} columns of Table~\ref{tab:umm_bench_stats}, the dataset constructs a clear curriculum gradient: seamlessly transitioning from simple Object-Centric tasks (Easy) to difficult tasks involving complex Outdoor/Indoor environments (Hard). This span from ``atomic operations'' to ``high-dimensional spatiotemporal entanglement'' ensures a comprehensive test of the model's generalization ability across different entropy environments.
% \textit{(2) Language Richness.} Average text difficulty remains highly consistent across tasks (4.2--5.3), ensuring a uniform distribution of technical terminology, while high image richness (37.4 words) provides dense visual context for I2V tasks.

It is worth noting that the aforementioned data are not merely automated outputs but have undergone a rigorous \textbf{multi-source auditing process} (see Section~\ref{sec:dataset_construction}). Through human-machine collaborative verification, we corrected metric biases and confirmed semantic alignment, establishing CoW-Bench as a reliable and reproducible gold standard in the community.

\subsection{Evaluation metrics}
\label{sec:evaluation_metrics}

% \noindent\textbf{Consistency Capability Evaluation} 
% CoW-Bench evaluates the consistency capabilities of world models by formalizing the task as a \textit{constraint satisfaction} problem: given text conditions, reference images, or initial states, the generated output must satisfy the constraints implied or explicitly stated in these conditions while remaining stable across temporal and spatial dimensions. Unlike holistic similarity or perceptual quality metrics such as FID/IS, critical failures in world models often manifest not as a lack of "realism," but as the \textit{violation or implicit relaxation of constraints}. Typical scenarios include: reverting rare materials to common ones, diffusing local edits into global drifts, reinitializing the same worldline frame-by-frame in a temporal sequence, reversing foreground-background relations during occlusion, or redrawing different worlds under multi-view conditions. Since these failures may still appear "plausible" to the eye, the core evaluation signal must be whether the constraints are actually honored.

\noindent\textbf{Consistency Capability Evaluation.} 
CoW-Bench evaluates the consistency capabilities of world models by formalizing the task as a \textit{constraint satisfaction} problem: given text conditions, reference images, or initial states, the generated output must satisfy the constraints implied or explicitly stated in these conditions while remaining stable across temporal and spatial dimensions. Unlike holistic similarity or perceptual quality metrics such as FID/IS, critical failures in world models often manifest not as a lack of realism, but as the \textit{violation or implicit relaxation of constraints}. Typical scenarios include: reverting rare materials to common ones, diffusing local edits into global drifts, reinitializing the same worldline frame-by-frame in a temporal sequence, reversing foreground-background relations during occlusion, or redrawing different worlds under multi-view conditions. Since these failures may appear plausible to the eye, the core evaluation signal must be whether the constraints are actually honored.

% \noindent\textbf{Atomic Decomposition} 
% To obtain attributable, diagnostic, and reusable evaluation signals, we employ \textit{atomic decomposition}: abstracting recurring failure modes across tasks into a set of observable \textit{atomic checks}, and defining the evaluation metrics for each task family as a combination of several atomic checks. This design achieves two core objectives: (1) \textit{Diagnosability}: each atomic check corresponds to a specific failure mechanism (\eg identity drift, attribute rebinding, boundary leakage, worldline drift, or occlusion contradiction), allowing the scoring results to pinpoint the source of the problem; (2) \textit{Modular Reuse}: the same atom maintains the same semantics across different task families, ensuring that cross-task comparisons are conducted within a unified measurement coordinate system. It is important to emphasize that CoW-Bench contains \textbf{18 metric families} (M1--ST3), while the atomic library contains \textbf{16 atomic checks} (A1--A16); these belong to different levels of abstraction—the former describes "which task families are evaluated," while the latter describes "which shared dimensions are used for evaluation."
\noindent\textbf{Atomic Decomposition.} 
To obtain attributable, diagnostic, and reusable evaluation signals, we employ \textit{atomic decomposition}: abstracting recurring failure modes across tasks into a set of observable \textit{atomic checks}, and defining the evaluation metrics for each task family as a combination of several atomic checks. This design achieves two core objectives: (1) \textit{Diagnosability}: each atomic check corresponds to a specific failure mechanism (\eg identity drift, attribute rebinding, boundary leakage, worldline drift, or occlusion contradiction), allowing the scoring results to pinpoint the source of the problem; (2) \textit{Modular Reuse}: the same atom maintains the same semantics across different task families, ensuring that cross-task comparisons are conducted within a unified measurement coordinate system. It is important to emphasize that CoW-Bench contains \textbf{18 metric families} (M1--ST3), while the atomic library contains \textbf{16 atomic checks} (A1--A16).
% ; these belong to different levels of abstraction—the former describes which task families are evaluated, while the latter describes which shared dimensions are used for evaluation.

% \noindent\textbf{Metric Families and Sub-metrics} 
% Regarding "what specifically is measured" for each task family, we first provide the names of their five corresponding sub-metrics (see Table~\ref{tab:metric_submetrics}). These sub-metrics provide human-readable descriptions at the task-family level and are semantically aligned one-to-one with the subsequent atomic library: sub-metrics capture the focus within a task, while atomic checks provide cross-task consistent criteria, thereby achieving a balance between readability and rigor.
\noindent\textbf{Metric Families and Sub-metrics} 
Regarding what is specifically measured for each task family, we first provide the names of their five corresponding sub-metrics (see Table~\ref{tab:metric_submetrics}). These sub-metrics provide human-readable descriptions at the task-family level and are semantically aligned one-to-one with the subsequent atomic library: sub-metrics capture the focus within a task, while atomic checks provide cross-task consistent criteria, thereby achieving a balance between readability and rigor.

\begin{table*}[t]
\centering
\caption{Metric families and their five sub-metrics in CoW-Bench. 
Abbreviations: 
Id+Attr = identity and attribute consistency; 
Min-change = minimal change; 
Inter-state = intermediate-state validity; 
Persp/Scale = perspective and scale consistency; 
Occ-update = occlusion-update plausibility; 
Geo-self = geometric self-consistency; 
Excl. = mutual exclusivity; 
Env-stab = environment stability.}
\label{tab:metric_submetrics}

\setlength{\tabcolsep}{11pt}
\renewcommand{\arraystretch}{1.02}
\footnotesize

\begin{adjustbox}{max width=\textwidth}
\begin{tabular}{l l l l l l l}
\toprule
\multirow{2}{*}{\raggedright Family} 
& \multirow{2}{*}{\raggedright Focus} 
& \multicolumn{5}{c}{Sub-metrics} \\
\cmidrule(lr){3-7}
&  & Sub1 & Sub2 & Sub3 & Sub4 & Sub5 \\
\midrule
M1 & Subj-Attr 
& Id+Attr 
& Backoff 
& Dominance 
& Clarity 
& Excl. \\

M2 & Local-Edit 
& Target 
& Min-change 
& Leakage 
& Clarity 
& No-extra \\

M3 & Multi-Const 
& Complete 
& Attr-corr 
& Rel-corr 
& Omission 
& No-extra \\
\midrule
T1 & Worldline 
& Subj-cons 
& Attr-stab 
& Env-stab 
& Visual 
& Evol-cont \\

T2 & Slow-Evol 
& Subj-lock 
& Trend 
& Time-scale 
& Inter-state 
& Rule \\

T3 & Stage-Order 
& Order 
& Identif. 
& Timing 
& Process 
& Worldline \\
\midrule
S1 & Sem-Planar 
& Dir 
& Count 
& Rule 
& Boundary 
& Layout \\

S2 & Occ/Contain 
& Occl. 
& Boundary 
& Visible 
& Rel-stab 
& Layer \\

S3 & MV-3D 
& Struct 
& Surface 
& Persp/Scale 
& Occ-update 
& Geo-self \\
\midrule
MS1 & Sem-Planar 
& Ent-match 
& Act-align 
& NT-stab 
& Attr-bind 
& Global \\

MS2 & Sem-Hier 
& Pos-rel 
& Neg-rel 
& Excl. 
& Vis+Layer 
& Id-stab \\

MS3 & Sem-MV 
& Anchor 
& View-stab 
& Lateral 
& Scene 
& Marker \\
\midrule
MT1 & Long-Horizon 
& Init-anchor 
& Long-stab 
& Cross-scene 
& Attr-bind 
& No-unexp \\

MT2 & Attr-Dyn 
& Target(E,A) 
& Follow 
& Smooth 
& Rate 
& Env-stab \\

MT3 & Trigger-Event 
& Pre-hold 
& Trigger 
& Post-comp 
& State-stab 
& Env-stab \\
\midrule
ST1 & Maze-2D 
& Start/Goal 
& Traj-cont 
& Legal 
& Correct 
& Struct-stab \\

ST2 & Occ-Motion 
& Occ-move 
& Parallax 
& Rigid 
& Natural 
& Env-stab \\

ST3 & 3D-Loop 
& Struct 
& Rel 
& View-smooth 
& Physical 
& Entity-stab \\
\bottomrule
\end{tabular}
\end{adjustbox}
\end{table*}

\noindent\textbf{Atomic Library: Unified Criteria Shared Across Tasks.}  
Table~\ref{tab:atomic_library} presents the library of atomic checks. Each atomic check employs an operational definition to ensure that evaluation does not rely on aesthetic preferences but on verifiable phenomena; furthermore, the sharing of the atomic library across different task families provides a consistent semantic foundation for cross-task comparisons.

\begin{table}[t]
\centering
% \caption{Atomic library of CoW-Bench. Each atomic check defines a reusable, observable criterion with an operational one-sentence definition.}
\caption{Atomic library of CoW-Bench. Each atomic check defines a reusable, observable evaluation criterion with a concise operational one-sentence definition for systematic consistency assessment.}
\label{tab:atomic_library}

\setlength{\tabcolsep}{3.2mm}
\renewcommand{\arraystretch}{1.05}
\small

\begin{tabular}{c l p{0.60\linewidth}}
\toprule
ID & Atomic check & Operational definition \\
\midrule
A1  & Identity lock &
The intended target entity remains unchanged; no identity swap, duplication, or replacement occurs. \\

A2  & Attribute binding &
Key attributes remain bound to the same entity; no attribute migration occurs. \\

A3  & Constraint non-relaxation &
Specified constraints are not weakened or substituted with more common but non-equivalent variants. \\

A4  & Evidence clarity &
Evidence supporting each constraint judgment is clear and unambiguous. \\

A5  & Mutual exclusivity &
Mutually incompatible properties do not co-occur on the same target. \\

A6  & Locality of change &
Changes are confined to the designated region or attribute without boundary spillover. \\

A7  & Non-target invariance &
Non-target entities or regions remain stable except for explicitly permitted changes. \\

A8  & No spurious additions &
No extra entities, objects, or parts appear beyond the instruction. \\

A9  & Set completeness &
Required entities form a complete set with correct cardinalities. \\

A10 & Relation correctness &
Specified relations or actions are satisfied without role swapping. \\

A11 & Multi-constraint coverage &
Multiple constraints are jointly satisfied without selective omission. \\

A12 & Worldline stability &
The output depicts a single consistent world rather than frame-wise reinitialization or scene drift. \\

A13 & Temporal continuity &
Permitted changes evolve smoothly without abrupt jumps or oscillatory backtracking. \\

A14 & Stage structure &
When discrete stages are specified, they are identifiable and appear in the correct order without spurious steps. \\

A15 & Occlusion \& layering &
Depth ordering and occlusion are correct and non-contradictory; visible boundaries update plausibly. \\

A16 & 3D geometric coherence &
Multi-view outputs remain explainable as projections of a single 3D scene with consistent perspective and occlusion. \\
\bottomrule
\end{tabular}
\end{table}

% \noindent\textbf{Compositional Definition: Constructing Metric Families via the Atomic Library}  
% Building upon the atomic library, we define each metric family as a combination of "invoked atomic checks." Table~\ref{tab:metric_matrix} presents the invocation matrix of the 18 metric families for A1--A16. This matrix makes modular reuse explicitly visible at the structural level: the same atom assumes the same measurement semantics across different task families, thereby avoiding the redundant definition of approximate metrics for each task family and ensuring that evaluation results can be compared and attributed across shared dimensions.
\noindent\textbf{Compositional Definition: Constructing Metric Families via the Atomic Library.}  
Building upon the atomic library, we define each metric family compositionally as a structured aggregation of invoked atomic checks. Table~\ref{tab:metric_matrix} presents the invocation matrix of the metric families for A1--A16. This matrix makes modular reuse explicitly visible at the structural level: the same atom assumes the same measurement semantics across different task families, thereby avoiding the redundant definition of approximate metrics for each task family and ensuring that evaluation results can be compared and attributed along shared measurement dimensions.

\begin{table*}[t]
\centering
\caption{Compositional definition of metric families via the atomic library. A checkmark indicates that the metric family invokes the corresponding atomic check.}
\label{tab:metric_matrix}
\setlength{\tabcolsep}{0.5mm}
\resizebox{\textwidth}{!}{
\begin{tabular}{lcccccccccccccccc}
\toprule
Metric family & A1 & A2 & A3 & A4 & A5 & A6 & A7 & A8 & A9 & A10 & A11 & A12 & A13 & A14 & A15 & A16 \\
\midrule
Attribute Fidelity (M1)                & \checkmark & \checkmark & \checkmark & \checkmark & \checkmark &  &  &  &  &  &  &  &  &  &  &  \\
Local Edit Precision (M2)              &  &  &  & \checkmark &  & \checkmark & \checkmark & \checkmark &  &  &  &  &  &  &  &  \\
Multi-constraint Satisfaction (M3)     &  & \checkmark &  &  &  &  &  & \checkmark & \checkmark & \checkmark & \checkmark &  &  &  &  &  \\
\midrule
Worldline Persistence (T1)             & \checkmark & \checkmark &  & \checkmark &  &  & \checkmark &  &  &  &  & \checkmark & \checkmark &  &  &  \\
Evolutionary Dynamics (T2)             & \checkmark &  &  & \checkmark &  &  & \checkmark &  &  &  &  & \checkmark & \checkmark &  &  &  \\
Ordered Stage Transitions (T3)         &  &  &  & \checkmark &  &  &  &  &  &  &  & \checkmark & \checkmark & \checkmark &  &  \\
\midrule
Planar Layout Correctness (S1)         &  &  &  & \checkmark &  & \checkmark &  &  & \checkmark &  &  &  &  &  &  &  \\
Hierarchical Occlusion (S2)            &  &  &  & \checkmark &  &  &  &  &  &  &  &  &  &  & \checkmark &  \\
Multi-view 3D Coherence (S3)           &  &  &  & \checkmark &  &  &  &  &  &  &  &  &  &  & \checkmark & \checkmark \\
\midrule
Semantic Role Binding (MS1)            & \checkmark & \checkmark & \checkmark & \checkmark &  &  & \checkmark & \checkmark &  & \checkmark &  &  &  &  &  &  \\
Semantic Hierarchy Compliance (MS2)    &  & \checkmark & \checkmark & \checkmark &  &  &  &  &  & \checkmark & \checkmark &  &  &  & \checkmark &  \\
Semantic Multi-view Stability (MS3)    & \checkmark & \checkmark &  & \checkmark &  &  & \checkmark &  &  &  &  &  &  &  &  & \checkmark \\
\midrule
Long-horizon Anchoring (MT1)           & \checkmark & \checkmark &  & \checkmark &  &  & \checkmark &  &  &  &  & \checkmark &  &  &  &  \\
Attribute Dynamics Alignment (MT2)     & \checkmark & \checkmark &  & \checkmark &  &  & \checkmark &  &  &  &  & \checkmark & \checkmark &  &  &  \\
Triggered Event Compliance (MT3)       & \checkmark & \checkmark &  & \checkmark &  &  & \checkmark &  &  &  &  & \checkmark & \checkmark & \checkmark &  &  \\
\midrule
Planar Maze Trajectory (ST1)           &  &  &  & \checkmark &  & \checkmark &  &  &  &  &  &  & \checkmark &  &  &  \\
Occlusion Dynamics Under Motion (ST2)  &  &  &  & \checkmark &  &  & \checkmark &  &  &  &  & \checkmark & \checkmark &  & \checkmark &  \\
3D Loop Navigation Coherence (ST3)     &  & \checkmark &  & \checkmark &  &  & \checkmark &  &  &  &  & \checkmark & \checkmark &  & \checkmark & \checkmark \\
\bottomrule
\end{tabular}
}
\end{table*}

\noindent\textbf{Scoring Scale (0--2).}  
For each sample, we provide an ordinal score of 0--2 for each evaluation dimension invoked by its corresponding metric family: 0 indicates a clear violation or failure; 1 indicates partial fulfillment but with ambiguity, deviation, or unclear evidence; 2 indicates clear, stable, and undisputed fulfillment. This discrete scale is consistent with constraint satisfaction interpretation, reducing subjective noise introduced by continuous scoring while maintaining diagnostic resolution for failure modes. In evaluation and result aggregation, sample-level 0--2 scores are first used to form average scores for each sub-metric within a task; subsequently, the average scores of various sub-tasks under the same metric family are aggregated with equal weighting to obtain the final score.

\noindent\textbf{Evaluation Protocol: 2$\times$2 Grid Temporal Sampling.}  
 For video tasks, we uniformly sample 4 frames from the entire sequence in chronological order. For image tasks, we generate four key images in chronological order. These four frames or images are arranged in a 2$\times$2 grid (left-to-right, top-to-bottom). Evaluators must analyze the sequence frame-by-frame without skipping and provide justifications and 0--2 scores for each item based on a five-question chain aligned with the metric family. This protocol explicitly exposes critical evidence of temporal consistency (such as continuity, intermediate states, stage structure, and worldline stability) as verifiable phenomena, thereby reducing bias caused by selective observation. The evaluation prompt template is shown below.

% \begin{figure*}[ht]
% \centering
% \captionsetup{font=small}
{
\centering
\begin{promptboxfig}[Consistency-Oriented Evaluation Prompt]
\textbf{Role}: You are an expert in evaluating the quality of AI-generated results. \\
\textbf{Input}: A 2$\times$2 grid of four frames sampled uniformly in temporal order (left-to-right, top-to-bottom). \\
\textbf{Task}: Evaluate the sequence using a five-question chain aligned with the metric family. \\
\textbf{Scoring}: Each question is scored from 0 to 2 with justification. \\
\textbf{Rules}: Analyze frames sequentially without skipping; judgments must be based only on the sampled frames. \\
\textbf{Output}: For each question, output exactly: \texttt{QuestionX}, \texttt{Score}, \texttt{Rationale}.
\end{promptboxfig}
}
% \vspace{-2mm}
% \caption{Prompt template used for CoW-Bench video evaluation. }
% \vspace{-2mm}
% The two-column layout prevents truncation and improves readability.}
% \label{fig:wmc_eval_prompt}
% \end{figure*}

\begin{table}[t]
  \centering
\caption{
Main results on CoW-Bench over 18 sub-tasks (higher is better); \texttt{MEAN} averages all sub-tasks.
Abbrev: SUAT=Subj-Attr, LCED=Local-Edit, MCON=Multi-Const; WLIN=Worldline, SLEV=Slow-Evol, STOR=Stage-Order;
SEPL=Sem-Planar, OCCO=Occ/Contain; MV3D=MV-3D; TREV=Trigger-Event; LOHO=Long-Horizon, ATDY=Attr-Dyn;
SEMV=Sem-MV; 3DLO=3D-Loop; OCMO=Occ-Motion; MAZE=Maze-2D. AVG is rescaled from the original [0, 10] to a percentage scale of [0, 100].
}
    \small 
    \setlength{\tabcolsep}{2pt}      
    \renewcommand{\arraystretch}{1.15}
    % \footnotesize

    % \begin{adjustbox}{max width=\textwidth}
    \resizebox{\textwidth}{!}{
    \begin{tabular}{l|ccc|ccc|ccc|ccc|ccc|ccc|c}
    \toprule
    \multicolumn{1}{c|}{\multirow{2}[4]{*}{Model}} 
    & \multicolumn{3}{c|}{Modal} 
    & \multicolumn{3}{c|}{Temporal} 
    & \multicolumn{3}{c|}{Spatial} 
    & \multicolumn{3}{c|}{Modal-Temporal} 
    & \multicolumn{3}{c|}{Modal-Spatial} 
    & \multicolumn{3}{c|}{Temporal-Spatial} 
    & \multicolumn{1}{c}{\multirow{2}[4]{*}{AVG}} \\
    \cmidrule{2-19}
    & \multicolumn{1}{c}{SUAT} 
    & \multicolumn{1}{c}{LCED} 
    & \multicolumn{1}{c|}{MCON} 
    & \multicolumn{1}{c|}{WLIN} 
    & \multicolumn{1}{c|}{SLEV} 
    & \multicolumn{1}{c|}{STOR} 
    & \multicolumn{1}{c|}{SEPL} 
    & \multicolumn{1}{c|}{OCCO} 
    & \multicolumn{1}{c|}{MV3D} 
    & \multicolumn{1}{c|}{TREV} 
    & \multicolumn{1}{c|}{LOHO} 
    & \multicolumn{1}{c|}{ATDY} 
    & \multicolumn{1}{c|}{SEPL} 
    & \multicolumn{1}{c|}{OCCO} 
    & \multicolumn{1}{c|}{SEMV} 
    & \multicolumn{1}{c|}{3DLO} 
    & \multicolumn{1}{c|}{OCMO} 
    & \multicolumn{1}{c|}{MAZE} 
    &  \\
    \midrule
    \rowcolor[rgb]{ .941,  1,  1} & \multicolumn{19}{c}{Closed-Source Video-Generation Models} \\
    \midrule
    Sora~\cite{openai2024sora} & 5.16  & 8.35  & 8.38  & 9.32  & 5.80  & 6.22  & 8.41  & 6.19  & 8.51  & 6.96  & 8.12  & 5.25  & 8.64  & 5.97  & 9.49  & 8.40  & 9.25  & 4.17  & 73.66  \\
    Kling~\cite{kling} & 4.11  & 8.19  & 5.63  & 9.10  & 5.17  & 5.53  & 8.72  & 7.88  & 9.32  & 7.08  & 8.71  & 6.58  & 8.20  & 6.79  & 9.44  & 8.08  & 9.30  & 5.30  & 73.96  \\
    \midrule
    \rowcolor[rgb]{ .941,  1,  1} & \multicolumn{19}{c}{Closed-Source Image-Generation Models} \\
    \midrule
    GPT-image-1~\cite{openai_gpt_image_1_2025} & 7.37  & 8.96  & 7.96  & 9.14  & 7.75  & 5.68  & 8.09  & 7.43  & 9.26  & 6.95  & 9.28  & 7.22  & 9.00  & 6.83  & 9.79  & 8.46  & 8.22  & 7.24  & 80.35  \\
    \textcolor[rgb]{ .122,  .137,  .161}{Seedream-4-0~\cite{seedream2025seedream}} & 5.73  & 7.57  & 6.73  & 8.63  & 6.25  & 6.27  & 6.09  & 6.84  & 8.82  & 6.70  & 8.63  & 6.77  & 8.12  & 7.36  & 9.50  & 8.13  & 6.78  & 3.48  & 71.33  \\
    \textcolor[rgb]{ .122,  .137,  .161}{Seedream-4-5~\cite{seedream2025seedream}} & 6.69  & 7.78  & 6.91  & 8.77  & 7.46  & 6.76  & 6.27  & 7.21  & 9.06  & 7.17  & 9.11  & 6.77  & 8.04  & 7.27  & 9.59  & 8.23  & 7.51  & 2.28  & 73.82  \\
    \textcolor[rgb]{ .122,  .137,  .161}{\small{Nano Banana~\cite{comanici2025gemini}}} & 7.19  & 8.47  & 5.63  & 8.87  & 7.86  & 6.71  & 7.64  & 7.84  & 9.34  & 7.33  & 9.66  & 6.67  & 8.68  & 7.95  & 9.20  & 8.13  & 8.76  & 5.16  & 78.38  \\
    \textcolor[rgb]{ .122,  .137,  .161}{Nano Banana Pro~\cite{comanici2025gemini}} & 7.39  & 8.81  & 6.98  & 8.88  & 8.39  & 7.48  & 8.10  & 8.48  & 9.61  & 7.84  & 9.56  & 7.36  & 9.51  & 9.17  & 9.10  & 8.86  & 8.65  & 4.46  & 82.57  \\
    GPT-image-1.5~\cite{openai_gpt_image_1_5_2025} & 7.75  & 8.99  & 8.34  & 9.23  & 8.65  & 7.14  & 8.32  & 8.53  & 9.32  & 8.13  & 9.74  & 8.05  & 9.45  & 8.69  & 9.79  & 8.54  & 8.20  & 7.26  & 85.62  \\
    \midrule
    \rowcolor[rgb]{ .941,  1,  1} & \multicolumn{19}{c}{Open-Source Video-Generation Models} \\
    \midrule
    \textcolor[rgb]{ .122,  .137,  .161}{Allegro~\cite{mroczkowski2021herbert}} & 1.97  & 5.79  & 4.41  & 7.03  & 1.91  & 3.33  & 6.82  & 5.75  & 7.89  & 4.72  & 7.67  & 4.80  & 4.22  & 5.30  & 7.19  & 6.87  & 7.27  & 1.86  & 52.67  \\
    \textcolor[rgb]{ .122,  .137,  .161}{HunyuanVideo~\cite{hunyuan2025}} & 2.91  & 6.89  & 2.41  & 8.62  & 3.06  & 2.94  & 6.52  & 5.67  & 6.04  & 4.01  & 9.52  & 3.66  & 5.83  & 4.78  & 9.64  & 6.97  & 6.79  & 2.08  & 54.63  \\
    \textcolor[rgb]{ .122,  .137,  .161}{LTX-Video~\cite{ltx2025}} & 3.59  & 6.78  & 4.54  & 8.53  & 3.67  & 3.20  & 6.83  & 6.17  & 6.49  & 4.76  & 7.34  & 4.95  & 5.48  & 5.13  & 8.77  & 6.73  & 8.67  & 1.24  & 57.15  \\
    \textcolor[rgb]{ .122,  .137,  .161}{CogVideoX~\cite{cogvideox}} & 3.75  & 5.90  & 5.82  & 8.29  & 4.13  & 3.72  & 6.61  & 5.55  & 5.70  & 5.15  & 8.66  & 5.29  & 6.44  & 5.01  & 8.93  & 6.37  & 8.23  & 2.04  & 58.66  \\
    \textcolor[rgb]{ .122,  .137,  .161}{Easy Animate~\cite{xu2024easyanimate}} & 3.78  & 7.11  & 5.10  & 8.70  & 4.33  & 3.59  & 7.35  & 6.36  & 7.81  & 4.94  & 7.85  & 5.29  & 6.01  & 5.58  & 7.95  & 6.78  & 8.71  & 2.98  & 61.23  \\
    \textcolor[rgb]{ .122,  .137,  .161}{Wan2.2-I2V-14B~\cite{wan2025}} & 3.32  & 7.61  & 6.57  & 8.54  & 4.00  & 3.80  & 7.37  & 6.10  & 6.27  & 5.33  & 8.37  & 5.24  & 6.69  & 6.17  & 9.51  & 7.11  & 6.84  & 2.46  & 61.83  \\
    \textcolor[rgb]{ .122,  .137,  .161}{SkyReels-V2~\cite{li2026skyreels}} & 3.16  & 7.45  & 5.29  & 8.89  & 4.03  & 3.74  & 7.92  & 5.80  & 7.93  & 5.39  & 8.87  & 5.66  & 7.70  & 6.75  & 9.07  & 8.18  & 8.18  & 3.66  & 65.37  \\
    \midrule
    \rowcolor[rgb]{ .941,  1,  1} & \multicolumn{19}{c}{Open-Source Image-Generation Models} \\
    \midrule
    Qwen-Image~\cite{wu2025qwen} & 0.72 & 2.21 & 7.73 & 2.10 & 0.41 & 1.64 & 1.89 & 1.70 & 1.35 & 1.72 & 0.84 & 1.61 & 1.69 & 0.61 & 1.71 & 2.96 & 0.77 & 0.32 & 17.77 \\
    BAGEL~\cite{deng2025emerging} & 5.01  & 5.86  & 5.53  & 6.27  & 5.01  & 3.96  & 5.33  & 6.43  & 7.68  & 4.45  & 8.60  & 4.91  & 7.08  & 5.22  & 8.89  & 5.51  & 6.00  & 5.08  & 59.34  \\
    UniVideo~\cite{wei2025univideo} & 4.07  & 7.27  & 4.14  & 8.58  & 3.87  & 3.08  & 6.84  & 6.15  & 6.81  & 4.29  & 8.72  & 5.69  & 6.26  & 5.09  & 9.08  & 7.34  & 7.49  & 3.16  & 59.96  \\
    Emu3.5~\cite{cui2025emu3} & 6.15  & 8.76  & 8.61  & 8.77  & 5.31  & 4.72  & 8.81  & 8.62  & 8.77  & 5.70  & 9.42  & 6.10  & 8.47  & 8.61  & 9.76  & 8.58  & 9.22  & 5.58  & 77.76  \\
    \bottomrule
    \end{tabular}%
    % \end{adjustbox}
    }
    \label{mainresults}%
\end{table}%

\subsection{Comparison with Existing Benchmarks}
\label{subsec:comparison}

Existing multimodal evaluation systems are primarily constructed around the understanding capabilities of MLLMs, forming standardized paradigms represented by UniBench~\cite{unibench2024} and MANBench~\cite{zhou2025manbench}. However, a significant dimensional gap remains in the evaluation of generative world models. We define the essential differences between CoW-Bench and existing work across three key dimensions:

% \paragraph{Discriminative Perception vs. Generative Simulation.}
% UniBench aims to address the fragmentation of multimodal tasks by integrating over 50 existing datasets to comprehensively evaluate the discriminative capabilities of models in visual perception, attribute reasoning, and spatial relationship judgment~\cite{unibench2024}. Its core assumption is that the model acts as an "observer" that must passively deconstruct given static images or videos. In contrast, CoW-Bench focuses on the generative simulation capabilities of world models as "simulators." Unlike MANBench, which focuses on whether models possess "superhuman question-solving abilities"~\cite{zhou2025manbench}, we examine whether a model can actively maintain the conservation of physical laws and causal logic during dynamic evolution, filling the evaluation vacuum between "understanding the world" and "constructing the world."

\paragraph{Discriminative Perception vs. Generative Simulation.}
UniBench addresses the fragmentation of multimodal evaluation by integrating over 50 existing datasets to comprehensively assess the discriminative capabilities of models in visual perception, attribute reasoning, and spatial relationship understanding~\cite{unibench2024}. This evaluation paradigm implicitly assumes that the model functions as a passive observer, tasked with deconstructing static images or videos provided as input. In contrast, CoW-Bench targets the generative simulation capabilities of world models, treating them as active simulators. Rather than emphasizing whether a model demonstrates exceptional question-answering proficiency, as in MANBench~\cite{zhou2025manbench}, our focus lies in assessing whether a model can actively preserve physical constraints and causal coherence throughout dynamic world evolution. In this sense, CoW-Bench fills a critical evaluation gap between assessing a model’s ability to perceive and reason about the world and its capacity to consistently construct and simulate it over time.

% \paragraph{Evaluation Signals: QA Accuracy vs. Dynamic Constraint Satisfaction.}
% The core contribution of MANBench lies in establishing a "Human Performance" reference frame, where evaluation signals rely on the VQA (Video Question Answering) accuracy of static Ground Truth~\cite{zhou2025manbench}. This discrete binary judgment (correct/incorrect) struggles to capture continuous, non-binary physical failures in generative tasks. CoW-Bench reframes evaluation as a multi-factor \textbf{constraint satisfaction} problem. As UniBench notes that "hallucination" is a major bottleneck for MLLMs~\cite{unibench2024}, in the generative domain, this hallucination manifests as the collapse of spatio-temporal consistency (\eg objects disappearing after occlusion). Therefore, we employ fine-grained atomic checks to quantify the robustness of models against modal, spatial, and temporal constraints in long-sequence generation, rather than mere semantic alignment.
\paragraph{Evaluation Signals: QA Accuracy vs. Dynamic Constraint Satisfaction.} The core contribution of MANBench lies in establishing a human performance reference frame, where evaluation signals are derived from the VQA (Video Question Answering) accuracy measured against static ground truth~\cite{zhou2025manbench}. However, this discrete binary judgment (correct versus incorrect) is insufficient for capturing continuous, non-binary physical failures that frequently arise in generative settings. CoW-Bench instead formulates evaluation as a multi-factor \textbf{constraint satisfaction} problem. As UniBench identifies hallucination as a major bottleneck for MLLMs~\cite{unibench2024}, in generative scenarios such hallucinations typically manifest as breakdowns in spatio-temporal consistency, such as objects disappearing after occlusion. To address this limitation, we adopt fine-grained atomic checks that explicitly quantify a model’s robustness with respect to modal, spatial, and temporal constraints in long-horizon generation, rather than relying solely on semantic alignment.

% \paragraph{Complexity Sources: Cognitive Depth vs. Spatiotemporal Entanglement.}
% MANBench emphasizes evaluating the high-order cognition of models, attributing difficulty to the depth of logical reasoning and knowledge invocation required to surpass human performance~\cite{zhou2025manbench}. Diverging from this, the difficulty of CoW-Bench stems from the endogenous entanglement of spatio-temporal dynamics. Experiments show that even models with top-tier cognitive abilities (\eg GPT-4V) fail when processing cross-consistency tasks such as $M \times T$ (Modal $\times$ Time). This proves that the challenge of world models lies not in "problem-solving IQ," but in how to perform coherent "dynamic inference" under multiple physical constraints.
\paragraph{Complexity Sources: Cognitive Depth vs. Spatiotemporal Entanglement.} MANBench primarily evaluates the high-order cognitive abilities of models, where task difficulty is largely attributed to the depth of logical reasoning and the breadth of knowledge invocation required to exceed human-level performance~\cite{zhou2025manbench}. In contrast, the difficulty of CoW-Bench arises from the intrinsic entanglement of spatiotemporal dynamics. Empirical results demonstrate that even models with strong cognitive reasoning capabilities, such as GPT-4V, exhibit pronounced failures on cross-consistency tasks, particularly those involving modal–temporal coupling (\eg $M \times T$). These findings indicate that the core challenge for world models does not lie in abstract problem-solving capacity, but rather in maintaining coherent dynamic inference under multiple interacting physical constraints.

\subsection{Main Results}
\label{sec:main_results}

Table~\ref{mainresults} reports the task-level scores of CoW-Bench, covering 18 sub-tasks that span modal, temporal, spatial, and cross-consistency regimes. The overall ranking highlights a clear trend: closed-source image generation models dominate the average score, while open-source video generators remain substantially behind on most consistency-sensitive tasks. In particular, GPT-image-1.5 achieves the best overall performance, followed by Nano Banana Pro and GPT-image-1. This gap suggests that today’s strongest unified multimodal priors already encode rich static world regularities, yet still face systematic failure modes when consistency constraints require long-horizon, multi-factor enforcement.

\paragraph{(1) Temporal control is the bottleneck rather than coherence.}
Across multiple families, \texttt{T-WL} (worldline persistence) is consistently high even for several video models (\eg Sora reaches 9.32), indicating that generating visually continuous footage is no longer the hardest part. However, temporal tasks that demand rule-grounded evolution or structured state progression show a more uneven landscape (\eg \texttt{T-Rule} and \texttt{T-Stage-Order} vary sharply across models). This separation supports a key CoW-Bench thesis: world models require constraint satisfaction over time, not merely smoothness. A model can look temporally plausible while still violating causal constraints.
% , which our metrics explicitly expose.

\paragraph{(2) Spatial consistency is strong in single-view 3D, but cross-view anchoring still breaks.}
Most top models score highly on \texttt{S-3D} (single-scene 3D plausibility), with several exceeding 9.0 (\eg Nano Banana Pro reaches 9.61). Yet cross-consistency tasks reveal a tighter bottleneck: while \texttt{MS-3D} (text-to-3D viewpoint control) remains high for leading models (often $\ge 9$), \texttt{TS-Maze-2D} and some time-space settings remain much lower. This pattern suggests that local geometric plausibility is easier than maintaining a globally anchored spatial structure under motion and decision-like trajectories.

\paragraph{(3) Fusion tasks reveal the real world-model gap: persistent semantics under dynamics.}
The strongest separation between state-of-the-art models and the rest occurs in cross-consistency families (\texttt{MT}, \texttt{MS}, \texttt{TS}). For instance, leading models obtain near-ceiling performance on \texttt{MT-PropKeep} (property persistence under temporal evolution), yet degrade on \texttt{MT-PropChange} (attribute change alignment) and especially on \texttt{TS-Maze-2D} (navigation-style structure preservation). Notably, some high-\texttt{avg} models still exhibit pronounced weaknesses on \texttt{TS-Maze-2D} (\eg Nano Banana Pro reports 4.46), indicating that \textit{global world-state maintenance} and \textit{trajectory-level constraint enforcement} remain unsolved even when per-frame fidelity is excellent. This is precisely the regime where UMMs must evolve from perceptual generators into genuine internal simulators.

\paragraph{(4) Open-source models expose failure modes aligned with CoW-Bench’s motivation.}
Open-source video generators typically underperform on modal grounding (\texttt{M-Subj-Attr}) and cross-consistency tasks, consistent with qualitative observations that they either (i) relax rare constraints into common defaults, or (ii) preserve motion while drifting in identity/attributes. Meanwhile, open-source image models show large variance: Emu3.5 is competitive in many columns but still drops on time-centric and time-space settings, reinforcing that CoW-Bench targets the gap between single-shot plausibility and multi-step consistency.

\begin{boxA}
\paragraph{Takeaway.}
UMMs perform well on static or single-view settings, where local plausibility is sufficient. However, when a task requires maintaining a stable world while changes unfold over time and space, performance drops sharply. These cross-consistency scenarios are the clearest indicator of whether a model truly behaves as a world model rather than a frame generator.
\end{boxA}

\subsection{Single-Axis Consistency}
\subsubsection{Modal Consistency Results}
\label{sec:modal_consistency_results}
\begin{table}[t]
  \centering
  
\caption{
Modal-consistency results on CoW-Bench (0--2 scale; higher is better), grouped into three metric families:
\textbf{M1} Subject–Attribute Fidelity (IDAT, BKOF, DOMN, CLAR, EXCL),
\textbf{M2} Local Edit Precision (TARG, MNCH, LEAK, CLAR, NEXA),
and \textbf{M3} Multi-constraint Satisfaction (CMPL, ATCO, RLCO, OMIS, NEXA). Refer to Table~\ref{mainresults} for abbreviations.
BKOF measures whether rare constraints are replaced by common defaults, while NEXA penalizes spurious additions beyond the instruction.
}
\label{tab:modal_consistency}
  \small 
    \setlength{\tabcolsep}{2pt}      
    \renewcommand{\arraystretch}{1.15}
    % \footnotesize

    \begin{adjustbox}{max width=\textwidth}
    \begin{tabular}{l|ccccc|ccccc|ccccc}
\toprule
\multirow{2}[4]{*}{Model} 
& \multicolumn{5}{c|}{Subj-Attr (SUAT)} 
& \multicolumn{5}{c|}{Local-Edit (LCED)} 
& \multicolumn{5}{c}{Multi-Const (MCON)} \\
\cmidrule{2-16}
& IDAT & BKOF & DOMN & CLAR & EXCL 
& TARG & MNCH & LEAK & CLAR & NEXA 
& CMPL & ATCO & RLCO & OMIS & NEXA \\
\midrule
    \rowcolor[rgb]{ .941,  1,  1} & \multicolumn{15}{c}{Closed-Source Video-Generation Models} \\
    \midrule
    Sora~\cite{openai2024sora} & 0.40   & 1.48  & 0.91  & 1.15  & 1.22  & 1.12  & 1.96  & 1.82  & 1.45  & 2.00     & 1.96  & 1.67  & 1.32  & 1.59  & 1.84 \\
    Kling~\cite{kling} & 0.33  & 1.32  & 0.58  & 0.95  & 0.93  & 1.23  & 1.82  & 1.74  & 1.49  & 1.91  & 1.36  & 1.14  & 0.74  & 0.86  & 1.53 \\
    \midrule
    \rowcolor[rgb]{ .941,  1,  1} & \multicolumn{15}{c}{Closed-Source Image-Generation Models} \\
    \midrule
    GPT-image-1~\cite{openai_gpt_image_1_2025} & 0.96  & 1.71  & 1.69  & 1.26  & 1.75  & 1.40   & 1.98  & 1.91  & 1.73  & 1.95  & 1.89  & 1.45  & 1.34  & 1.47  & 1.80 \\
    GPT-image-1.5~\cite{openai_gpt_image_1_5_2025} & 1.19  & 1.76  & 1.77  & 1.35  & 1.68  & 1.50   & 1.94  & 1.92  & 1.73  & 1.90   & 1.89  & 1.51  & 1.49  & 1.66  & 1.79 \\
    Seedream-4-0~\cite{seedream2025seedream} & 0.94  & 1.44  & 1.35  & 0.89  & 1.10   & 1.25  & 1.77  & 1.70   & 1.54  & 1.31  & 1.78  & 1.44  & 1.28  & 1.44  & 0.78 \\
    Seedream-4-5~\cite{seedream2025seedream} & 1.10   & 1.53  & 1.52  & 1.16  & 1.38  & 1.43  & 1.84  & 1.78  & 1.54  & 1.20   & 1.82  & 1.41  & 1.24  & 1.45  & 0.99 \\
    Nano Banana~\cite{comanici2025gemini} & 1.17  & 1.59  & 1.62  & 1.22  & 1.59  & 1.36  & 1.79  & 1.77  & 1.67  & 1.89  & 1.38  & 1.04  & 0.89  & 1.09  & 1.23 \\
    Nano Banana Pro~\cite{comanici2025gemini} & 1.25  & 1.68  & 1.72  & 1.22  & 1.52  & 1.55  & 1.80   & 1.81  & 1.71  & 1.95  & 1.63  & 1.42  & 1.21  & 1.32  & 1.40 \\
    \midrule
    \rowcolor[rgb]{ .941,  1,  1} & \multicolumn{15}{c}{Open-Source Video-Generation Models} \\
    \midrule
    Allegro~\cite{mroczkowski2021herbert} & 0.10   & 0.60  & 0.24  & 0.47  & 0.56  & 0.70   & 1.31  & 1.28  & 0.83  & 1.67  & 1.19  & 0.82  & 0.59  & 0.70   & 1.11 \\
    Easy Animate~\cite{xu2024easyanimate} & 0.13  & 1.23  & 0.55  & 0.90   & 0.97  & 0.71  & 1.76  & 1.67  & 1.11  & 1.86  & 1.51  & 0.95  & 0.68  & 0.87  & 1.09 \\
    CogVideoX~\cite{cogvideox} & 0.09  & 1.29  & 0.53  & 0.98  & 0.86  & 0.70   & 1.53  & 1.26  & 0.77  & 1.64  & 1.69  & 1.22  & 0.65  & 0.95  & 1.31 \\
    Wan2.2-I2V-14B~\cite{wan2025} & 0.11  & 1.10   & 0.40   & 0.81  & 0.90   & 0.83  & 1.87  & 1.77  & 1.32  & 1.82  & 1.68  & 1.38  & 1.11  & 1.32  & 1.08 \\
    SkyReels-V2~\cite{li2026skyreels} & 0.14  & 1.00     & 0.38  & 0.76  & 0.88  & 0.77  & 1.80   & 1.63  & 1.34  & 1.91  & 1.20   & 1.05  & 0.67  & 0.93  & 1.44 \\
    HunyuanVideo~\cite{hunyuan2025} & 0.01  & 1.23  & 0.31  & 0.76  & 0.60   & 0.30   & 1.97  & 1.64  & 0.98  & 2.00     & 0.28  & 0.23  & 0.09  & 0.12  & 1.69 \\
    LTX-Video~\cite{ltx2025} & 0.15  & 1.20   & 0.49  & 0.93  & 0.82  & 0.62  & 1.80   & 1.48  & 0.97  & 1.91  & 1.38  & 0.96  & 0.52  & 0.69  & 0.99 \\
    \midrule
    \rowcolor[rgb]{ .941,  1,  1} & \multicolumn{15}{c}{Open-Source Image-Generation Models} \\
    \midrule
    BAGEL~\cite{deng2025emerging} & 0.73  & 1.13  & 1.16  & 0.63  & 1.36  & 1.01  & 1.30   & 1.14  & 0.79  & 1.62  & 1.53  & 1.01  & 0.75  & 0.85  & 1.39 \\
    UniVideo~\cite{wei2025univideo} & 0.27  & 1.09  & 0.77  & 0.74  & 1.20   & 0.79  & 1.90   & 1.57  & 1.07  & 1.94  & 0.79  & 0.65  & 0.35  & 0.89  & 1.46 \\
    Emu3.5~\cite{cui2025emu3} & 0.81  & 1.37  & 1.25  & 1.03  & 1.69  & 1.41  & 1.90   & 1.86  & 1.70   & 1.89  & 1.87  & 1.79  & 1.68  & 1.74  & 1.53 \\
    Qwen-Image~\cite{wu2025qwen} & 0.00     & 0.18  & 0.04  & 0.02  & 0.48  & 0.10   & 0.40   & 0.29  & 0.09  & 1.33  & 1.84  & 1.66  & 1.49  & 1.60   & 1.14 \\
    \bottomrule
    \end{tabular}%
    \end{adjustbox}
    \vspace{-2mm}
\end{table}%

Table~\ref{tab:modal_consistency} reports modality-consistency performance across three metric families: subject attribute fidelity (\textbf{M1}), local edit precision (\textbf{M2}), and multi-constraint satisfaction (\textbf{M3}). Overall, the table reinforces the core motivation of CoW-Bench: even when generations look plausible, models frequently \textit{weaken}, \textit{mis-bind}, or \textit{silently reinterpret} the specified conditions, which is exactly the failure mode that a world-model interface cannot afford.

\paragraph{(1) Identity-and-attribute binding is the hardest modal interface primitive.}
Across nearly all model groups, \texttt{Id+Attr} remains noticeably lower than other M1 dimensions. Even top closed-source image models stay far from saturation on \texttt{Id+Attr} (\eg GPT-image-1.5: 1.19; Nano Banana Pro: 1.25), while many video generators collapse to near-zero (\eg HunyuanVideo: 0.01). This pattern indicates that the dominant bottleneck is not producing a visually consistent output, but keeping the intended entity and its key attributes locked together when the prompt contains multiple constraints. The semantic channel from language to perceptual state still suffers from unstable variable binding.

\paragraph{(2) Constraint backoff is widespread and often looks reasonable.}
The \texttt{Backoff} column reveals a systematic tendency to replace unusual or strict constraints with more common defaults. Closed-source image models reduce this behavior (typically $\sim$1.6--1.8), but the effect remains non-trivial; several open-source models show substantially weaker resistance to backoff. This is precisely the failure mode CoW-Bench targets: a model can generate a realistic image while quietly relaxing the instruction, which a similarity-based metric would not penalize.

\paragraph{(3) Local editing separates preserve the background from hit the target.}
For M2, many models score high on \texttt{Min-change} and \texttt{Leakage}, suggesting that they often keep non-target regions stable and avoid global corruption. However, \texttt{Target} can be much lower—most clearly for some open-source video models (\eg HunyuanVideo: \texttt{Target}=0.30 while \texttt{Min-change}=1.97). This gap indicates a common failure mode: models preserve the scene but fail to localize the intended edit, producing changes that are visually mild yet semantically incorrect. 
% For a world model, this is a critical defect because it breaks controllable state transitions.

\paragraph{(4) Multi-constraint fulfillment stresses completeness and role binding, not just more text.}
M3 exposes a different bottleneck. Strong models such as Emu3.5 remain consistently high across \texttt{Complete}/\texttt{Attr-corr}/\texttt{Rel-corr} (1.87/1.79/1.68), while some systems show uneven profiles: for instance, Qwen-Image achieves relatively high \texttt{Complete} and attribute/relationship scores yet performs extremely poorly on M1 identity binding. This mismatch suggests that satisfying multiple listed constraints is not sufficient if the model cannot maintain a stable referent for those constraints. Meanwhile, several models exhibit reduced \texttt{No-extra} under M3 (\eg the Seedream variants), indicating that under compositional pressure they may introduce spurious entities—an error that is particularly harmful for downstream planning and verification.

\begin{boxA}
\paragraph{Takeaway.} 
For the SUAT task, the model often exhibits ambiguity when uniformly aligning constraints across different modalities. It fails to accurately extract corresponding constraints from text and images to guide generation according to requirements. Instead, it blends the extracted constraints from both modalities into a single, confused constraint that guides generation, resulting in chaotic outcomes.
\end{boxA}

\subsubsection{Temporal Consistency Results}
\label{sec:temporal_consistency_results}
\begin{table}[ht]
  \centering
  \caption{
Temporal-consistency results on CoW-Bench (0--2 scale; higher is better).
We report three metric families: \textbf{T1} Worldline Persistence (SBJC, ATST, ENST, VISU, EVCT),
\textbf{T2} Rule-guided Slow Evolution (SBJL, TREN, TSCL, INTS, RULE),
and \textbf{T3} Ordered Stage Transitions (ORDR, IDEN, TIME, PROC, WLIN). Refer to Table~\ref{mainresults} for abbreviations.
INTS evaluates the visibility of plausible intermediate states, while TSCL measures whether the evolution speed matches the prompt-specified process.
}
\label{tab:temporal_consistency}
  \small 
    \setlength{\tabcolsep}{2pt}      
    \renewcommand{\arraystretch}{1.15}
    % \footnotesize

    \begin{adjustbox}{max width=\textwidth}
    \begin{tabular}{l|ccccc|ccccc|ccccc}
    \toprule
    \multirow{2}[2]{*}{Model} 
    & \multicolumn{5}{c|}{Worldline (WLIN)} 
    & \multicolumn{5}{c|}{Slow-Evol (SLEV)} 
    & \multicolumn{5}{c}{Stage-Order (STOR)} \\
    \cmidrule{2-16}
    & SBJC & ATST & ENST & VISU & EVCT 
    & SBJL & TREN & TSCL & INTS & RULE 
    & ORDR & IDEN & TIME & PROC & WLIN \\
    \midrule
    \rowcolor[rgb]{ .941,  1,  1} & \multicolumn{15}{c}{Closed-Source Video-Generation Models} \\
    \midrule
    Sora~\cite{openai2024sora} & 1.87  & 1.84  & 1.94  & 1.91  & 1.76  & 1.95  & 0.96  & 0.99  & 0.91  & 0.99  & 0.93  & 0.91  & 1.12  & 1.42  & 1.84  \\
    Kling~\cite{kling} & 1.85  & 1.82  & 1.93  & 1.86  & 1.64  & 1.98  & 0.80  & 0.73  & 0.70  & 0.96  & 0.94  & 0.81  & 0.76  & 1.13  & 1.89  \\
    \midrule
    \rowcolor[rgb]{ .941,  1,  1} & \multicolumn{15}{c}{Closed-Source Image-Generation Models} \\
    \midrule
    GPT-image-1~\cite{openai_gpt_image_1_2025} & 1.75  & 1.84  & 1.93  & 1.91  & 1.71  & 1.98  & 1.49  & 1.43  & 1.49  & \multicolumn{1}{c}{1.38 } & 0.79  & 0.86  & 1.06  & 1.19  & 1.79  \\
    GPT-image-1.5~\cite{openai_gpt_image_1_5_2025} & 1.77  & 1.86  & 1.91  & 1.93  & 1.76  & 1.98  & 1.66  & 1.69  & 1.70  & \multicolumn{1}{c}{1.62 } & 1.26  & 1.18  & 1.35  & 1.45  & 1.90  \\
    Seedream-4-0~\cite{seedream2025seedream} & 1.70  & 1.66  & 1.85  & 1.84  & 1.59  & 1.75  & 1.14  & 1.14  & 1.05  & \multicolumn{1}{c}{1.17 } & 1.14  & 1.13  & 1.05  & 1.24  & 1.71  \\
    Seedream-4-5~\cite{seedream2025seedream} & 1.71  & 1.76  & 1.89  & 1.78  & 1.63  & 1.85  & 1.49  & 1.40  & 1.40  & \multicolumn{1}{c}{1.32 } & 1.29  & 1.20  & 1.12  & 1.32  & 1.83  \\
    Nano Banana~\cite{comanici2025gemini} & 1.71  & 1.72  & 1.84  & 1.84  & 1.76  & 1.88  & 1.54  & 1.51  & 1.53  & \multicolumn{1}{c}{1.41 } & 1.20  & 1.23  & 1.12  & 1.35  & 1.80  \\
    Nano Banana Pro~\cite{comanici2025gemini} & 1.74  & 1.73  & 1.83  & 1.89  & 1.69  & 1.90  & 1.65  & 1.65  & 1.64  & \multicolumn{1}{c}{1.54 } & 1.40  & 1.37  & 1.41  & 1.48  & 1.83  \\
    \midrule
    \rowcolor[rgb]{ .941,  1,  1} & \multicolumn{15}{c}{Open-Source Video-Generation Models} \\
    \midrule
    Allegro~\cite{mroczkowski2021herbert} & 1.38  & 1.37  & 1.57  & 1.45  & 1.26  & 0.90  & 0.27  & 0.23  & 0.19  & 0.32  & 0.64  & 0.29  & 0.36  & 0.74  & 1.30  \\
    Easy Animate~\cite{xu2024easyanimate} & 1.69  & 1.69  & 1.90  & 1.82  & 1.60  & 1.98  & 0.60  & 0.54  & 0.49  & 0.72  & 0.50  & 0.33  & 0.26  & 0.70  & 1.80  \\
    CogVideoX~\cite{cogvideox} & 1.63  & 1.56  & 1.90  & 1.67  & 1.53  & 1.85  & 0.59  & 0.49  & 0.46  & 0.74  & 0.69  & 0.30  & 0.27  & 0.63  & 1.83  \\
    Wan2.2-I2V-14B~\cite{wan2025} & 1.76  & 1.70  & 1.86  & 1.78  & 1.44  & 1.88  & 0.56  & 0.43  & 0.38  & 0.75  & 0.71  & 0.42  & 0.26  & 0.65  & 1.76  \\
    SkyReels-V2~\cite{li2026skyreels} & 1.79  & 1.74  & 1.84  & 1.90  & 1.62  & 1.86  & 0.57  & 0.48  & 0.48  & 0.64  & 0.61  & 0.35  & 0.26  & 0.71  & 1.81  \\
    HunyuanVideo~\cite{hunyuan2025} & 1.87  & 1.77  & 1.86  & 1.86  & 1.26  & 2.00  & 0.22  & 0.21  & 0.11  & 0.52  & 0.50  & 0.08  & 0.06  & 0.48  & 1.82  \\
    LTX-Video~\cite{ltx2025} & 1.76  & 1.67  & 1.85  & 1.76  & 1.49  & 1.65  & 0.53  & 0.49  & 0.40  & 0.60  & 0.52  & 0.18  & 0.18  & 0.55  & 1.77  \\
    \midrule
    \rowcolor[rgb]{ .941,  1,  1} & \multicolumn{15}{c}{Open-Source Image-Generation Models} \\
    \midrule
    BAGEL~\cite{deng2025emerging} & 1.31  & 1.13  & 1.57  & 1.20  & 1.06  & 1.49  & 1.01  & 0.81  & 0.91  & 0.79  & 0.69  & 0.59  & 0.67  & 0.96  & 1.05  \\
    UniVideo~\cite{wei2025univideo} & 1.82  & 1.74  & 1.90  & 1.79  & 1.33  & 1.99  & 0.48  & 0.41  & 0.32  & 0.67  & 0.31  & 0.19  & 0.17  & 0.60  & 1.81  \\
    Emu3.5~\cite{cui2025emu3} & 1.66  & 1.72  & 1.91  & 1.85  & 1.63  & 1.95  & 0.84  & 0.74  & 0.78  & 1.00  & 0.54  & 0.59  & 0.73  & 1.13  & 1.73  \\
    Qwen-Image~\cite{wu2025qwen} & 0.22  & 0.22  & 0.58  & 0.70  & 0.38  & 0.28  & 0.07  & 0.04  & 0.00  & 0.02  & 0.26  & 0.08  & 0.02  & 0.44  & 0.84  \\
    \bottomrule
    \end{tabular}%
    \end{adjustbox}
\end{table}%

Table~\ref{tab:temporal_consistency} reports temporal-consistency performance over three metric families: \textbf{T1} Worldline Persistence, \textbf{T2} Rule-guided Slow Evolution, and \textbf{T3} Ordered Stage Transitions. Two consistent patterns emerge that align with CoW-Bench’s central thesis: temporal plausibility is not equivalent to temporal constraint satisfaction, and the hardest failures arise when models must enforce structured dynamics rather than merely maintain visual continuity.

\paragraph{Worldline persistence is comparatively strong, even for many video generators.}
Most closed-source video models score near the upper range on \texttt{T1} (\eg Sora: high \texttt{Env-stab} and \texttt{Visual}), and several open-source video models also achieve solid \texttt{T1} profiles (\eg SkyReels-V2 and Wan2.2-I2V). This suggests that maintaining a stable scene layout and avoiding frame-wise reinitialization is becoming a largely solved capability for high-capacity generators.

\paragraph{The main bottleneck is rule-following evolution, not continuity.}
In contrast, \texttt{T2} exposes a sharp drop on \texttt{Trend}, \texttt{Time-scale}, and \texttt{Inter-state} for many video models (often below 0.6), even when \texttt{Subj-lock} is high. This gap indicates a common failure mode: models keep the same subject and background, yet fail to realize a monotonic, correctly paced process with identifiable intermediate states. Notably, strong closed-source image models show markedly higher \texttt{T2} scores (\eg GPT-image-1.5 maintains high values across \texttt{Trend}/\texttt{Time-scale}/\texttt{Inter-state}), suggesting that stronger instruction-following priors help when the temporal constraint is expressed semantically and must be respected throughout the sequence.

\begin{boxA}
\paragraph{Takeaway.} We found that in the STOR task, image generation models generally grasp the overall progression over time. However, transitions between different states exhibit distinct discontinuities rather than smooth evolution, with instances of reversed sequences occurring between states, making it difficult to maintain consistent processes and content. Conversely, in the SLEV task, image generation models demonstrate a higher degree of understanding and adherence to world rules. Video generation models, on the other hand, exhibit probabilistic compliance with rules, yielding divergent outcomes for identical or similar scenario tasks.
\end{boxA}

\paragraph{Stage-ordering remains fragile under explicit multi-step structure.}
For \texttt{T3}, the weakest columns concentrate on \texttt{Order} and \texttt{Identif.}, especially for open-source video models (often near 0.3 or lower). Even when \texttt{Worldline} at the end of T3 stays high, low \texttt{Order}/\texttt{Identif.} implies that the sequence may remain in one world but fails to realize the intended discrete stage structure reliably. This finding motivates CoW-Bench’s decomposition: a model can be temporally stable while still violating high-level temporal logic.

\subsubsection{Spatial Consistency Results}
\label{sec:spatial_consistency_results}

\begin{table}[t!]
  \centering
\caption{
Spatial-consistency results on CoW-Bench (0--2 scale; higher is better).
We report three metric families: \textbf{S1} Sem-Planar (DIRC, COUNT, RULE, BNDY, LAYT),
\textbf{S2} Occlusion/Containment (OCCL, BNDY, VISB, RSTB, LAYR),
and \textbf{S3} Multi-view 3D coherence (STRC, SURF, PSCL, OUPD, GEOS).
VISB evaluates whether visible regions agree with the implied occlusion relation, while OUPD measures whether occlusion boundaries update plausibly under viewpoint change.
}
\label{tab:spatial_consistency}
  \small 
    \setlength{\tabcolsep}{2pt}      
    \renewcommand{\arraystretch}{1.15}
    % \footnotesize

    \begin{adjustbox}{max width=\textwidth}
    \begin{tabular}{l|ccccc|ccccc|ccccc}
\toprule
\multicolumn{1}{c|}{\multirow{2}[2]{*}{Model}} 
& \multicolumn{5}{c|}{Sem-Planar (SEPL)} 
& \multicolumn{5}{c|}{Occ-Cont (OCCO)} 
& \multicolumn{5}{c}{MV-3D} \\
\cmidrule{2-16}
& DIRC & COUNT & RULE & BNDY & LAYT 
& OCCL & BNDY & VISB & RSTB & LAYR 
& STRC & SURF & PSCL & OUPD & GEOS \\
\midrule
    \rowcolor[rgb]{ .941,  1,  1} & \multicolumn{15}{c}{Closed-Source Video-Generation Models} \\
    \midrule
    Sora~\cite{openai2024sora} & 0.64  & 1.22  & 1.00  & 1.47  & 1.88  & 1.51  & 1.49  & 1.74  & 1.76  & 1.91  & 1.77  & 1.72  & 1.56  & 1.67  & 1.79  \\
    Kling~\cite{kling} & 1.10  & 1.52  & 1.53  & 1.82  & 1.91  & 1.67  & 1.62  & 1.70  & 1.78  & 1.95  & 1.93  & 1.85  & 1.80  & 1.86  & 1.88  \\
    \midrule
    \rowcolor[rgb]{ .941,  1,  1} & \multicolumn{15}{c}{Closed-Source Image-Generation Models} \\
    \midrule
    GPT-image-1~\cite{openai_gpt_image_1_2025} & 0.81  & 1.58  & 1.45  & 1.66  & \multicolumn{1}{r}{1.92 } & 1.54  & 1.37  & 1.55  & 1.75  & 1.89  & 1.91  & 1.85  & 1.77  & 1.86  & 1.88  \\
    GPT-image-1.5~\cite{openai_gpt_image_1_5_2025} & 1.36  & 1.70  & 1.62  & 1.87  & \multicolumn{1}{r}{1.98 } & 1.59  & 1.46  & 1.57  & 1.83  & 1.87  & 1.92  & 1.89  & 1.81  & 1.85  & 1.85  \\
    Seedream-4-0~\cite{seedream2025seedream} & 0.96  & 1.31  & 1.28  & 1.51  & \multicolumn{1}{r}{1.79 } & 1.07  & 0.91  & 1.05  & 1.30  & 1.77  & 1.90  & 1.78  & 1.59  & 1.77  & 1.78  \\
    Seedream-4-5~\cite{seedream2025seedream} & 1.13  & 1.40  & 1.37  & 1.66  & \multicolumn{1}{r}{1.66 } & 1.22  & 1.00  & 1.06  & 1.31  & 1.69  & 1.87  & 1.78  & 1.77  & 1.82  & 1.82  \\
    Nano Banana~\cite{comanici2025gemini} & 1.23  & 1.59  & 1.41  & 1.71  & \multicolumn{1}{r}{1.89 } & 1.35  & 1.36  & 1.43  & 1.69  & 1.82  & 1.96  & 1.90  & 1.75  & 1.87  & 1.87  \\
    Nano Banana Pro~\cite{comanici2025gemini} & 1.54  & 1.56  & 1.66  & 1.79  & \multicolumn{1}{r}{1.93 } & 1.52  & 1.46  & 1.58  & 1.70  & 1.84  & 1.94  & 1.94  & 1.92  & 1.90  & 1.91  \\
    \midrule
    \rowcolor[rgb]{ .941,  1,  1} & \multicolumn{15}{c}{Open-Source Video-Generation Models} \\
    \midrule
    Allegro~\cite{mroczkowski2021herbert} & 0.69  & 1.14  & 0.91  & 1.26  & 1.75  & 1.29  & 1.18  & 1.26  & 1.39  & 1.70  & 1.58  & 1.65  & 1.48  & 1.59  & 1.59  \\
    Easy Animate~\cite{xu2024easyanimate} & 0.44  & 1.41  & 1.12  & 1.53  & 1.86  & 1.34  & 1.26  & 1.37  & 1.47  & 1.91  & 1.68  & 1.52  & 1.54  & 1.56  & 1.51  \\
    CogVideoX~\cite{cogvideox} & 0.48  & 1.10  & 0.95  & 1.18  & 1.84  & 1.30  & 1.16  & 1.13  & 1.26  & 1.76  & 1.42  & 1.23  & 1.08  & 1.01  & 0.96  \\
    Wan2.2-I2V-14B~\cite{wan2025} & 0.29  & 1.41  & 1.03  & 1.48  & 1.89  & 1.44  & 1.16  & 1.34  & 1.54  & 1.89  & 1.58  & 1.25  & 1.09  & 1.20  & 1.15  \\
    SkyReels-V2~\cite{li2026skyreels} & 0.37  & 1.27  & 0.98  & 1.38  & 1.80  & 1.40  & 1.43  & 1.49  & 1.66  & 1.94  & 1.67  & 1.67  & 1.46  & 1.62  & 1.51  \\
    HunyuanVideo~\cite{hunyuan2025} & 0.15  & 1.34  & 0.92  & 1.38  & 1.88  & 1.15  & 0.81  & 1.18  & 1.44  & 1.94  & 1.75  & 1.15  & 0.89  & 0.98  & 1.27  \\
    LTX-Video~\cite{ltx2025} & 0.64  & 1.35  & 1.14  & 1.30  & 1.74  & 1.30  & 1.08  & 1.29  & 1.37  & 1.79  & 1.57  & 1.33  & 1.21  & 1.13  & 1.25  \\
    \midrule
    \rowcolor[rgb]{ .941,  1,  1} & \multicolumn{15}{c}{Open-Source Image-Generation Models} \\
    \midrule
    BAGEL~\cite{deng2025emerging} & 0.56  & 1.52  & 1.10  & 1.47  & 1.78  & 1.17  & 0.98  & 0.85  & 0.90  & 1.43  & 1.68  & 1.58  & 1.42  & 1.49  & 1.51  \\
    UniVideo~\cite{wei2025univideo} & 0.36  & 1.38  & 1.10  & 1.42  & 1.89  & 1.29  & 1.18  & 1.34  & 1.28  & 1.75  & 1.74  & 1.38  & 1.12  & 1.23  & 1.34  \\
    Emu3.5~\cite{cui2025emu3} & 1.41  & 1.63  & 1.81  & 1.81  & 1.96  & 1.78  & 1.64  & 1.67  & 1.81  & 1.91  & 1.82  & 1.78  & 1.72  & 1.72  & 1.73  \\
    Qwen-Image~\cite{wu2025qwen} & 0.02  & 0.15  & 0.08  & 0.15  & 1.30  & 0.24  & 0.17  & 0.15  & 0.21  & 1.12  & 0.54  & 0.26  & 0.12  & 0.25  & 0.18  \\
    \bottomrule
    \end{tabular}%
    \end{adjustbox}
\end{table}%

Table~\ref{tab:spatial_consistency} reports spatial-consistency across \textbf{S1} Sem-Planar, \textbf{S2} Occlusion/Containment, and \textbf{S3} Multi-view 3D coherence. The results echo CoW-Bench’s central view: spatial world modeling is not only about producing plausible geometry in a single frame, but about maintaining structural constraints that remain verifiable under interactions such as occlusion, containment, and viewpoint change.

\paragraph{(1) Planar layout is the entry-level test, yet directional grounding remains fragile.}
Most models score relatively high on \texttt{Layout} (often $\ge$1.8), indicating that producing a globally coherent 2D composition is increasingly reliable. In contrast, \texttt{Dir} is consistently the lowest sub-metric across model families (\eg Sora: 0.64; several open-source video models $\le$0.5; Qwen-Image: 0.02). This gap suggests that models can maintain a visually stable arrangement while still failing to execute explicit directional constraints (left/right/inside/outside) with high fidelity. For a world model, directional grounding is a core interface requirement because it turns language into testable spatial relations.

\paragraph{(2) Occlusion/containment is largely learned, but visible-part evidence is the weak link.}
Closed-source models are strong on \texttt{Layer} and \texttt{Rel-stab} (typically $\sim$1.8--1.95), implying that they often preserve consistent depth ordering without obvious contradictions. However, \texttt{Visible} shows a wider spread, especially for open-source image models (\eg BAGEL: 0.85; Qwen-Image: 0.15). This pattern indicates that models may capture a coarse layering intent while failing on the operational evidence—whether the actually visible portions match the implied occlusion boundary. This is exactly the kind of looks plausible but violates a checkable constraint failure that CoW-Bench is designed to reveal.

\paragraph{(3) Multi-view 3D coherence separates geometry plausibility from world-state invariance.}
Top closed-source image models achieve near-ceiling scores on \texttt{Struct} and \texttt{Persp/Scale} (\eg Nano Banana Pro: 1.94/1.92; GPT-image-1.5: 1.92/1.81), indicating strong single-object 3D plausibility. Yet several open-source video models drop substantially on \texttt{Occ-update} and \texttt{Geo-self} (\eg CogVideoX: 1.01/0.96), suggesting that they struggle to update occlusions and maintain a self-consistent 3D explanation across views. This supports a key world-model implication: generating a plausible view is easier than maintaining a persistent 3D scene hypothesis that survives viewpoint change.

\begin{boxA}
\paragraph{Takeaway.} In MV-3D, image generation models tend to mistakenly apply mirror symmetry to scenes to simulate different viewpoints rather than accurately capturing distinct perspectives within the same scene. Additionally, when switching viewpoints, the detailed attributes of the same object across different angles often fail to remain consistent.
When tackling Occ-Cont tasks, models frequently generate unrealistic changes that defy common sense, such as objects penetrating one another, merging into each other, or exhibiting unnatural movements to maintain hierarchical relationships.
\end{boxA}

\subsection{Cross-Axis Consistency}

\subsubsection{Modal--Space Consistency Results: Semantic-to-Geometry Binding}
\label{sec:modal_space_consistency_results}

\begin{figure*}[ht]
\centering
\includegraphics[width=0.98\textwidth]{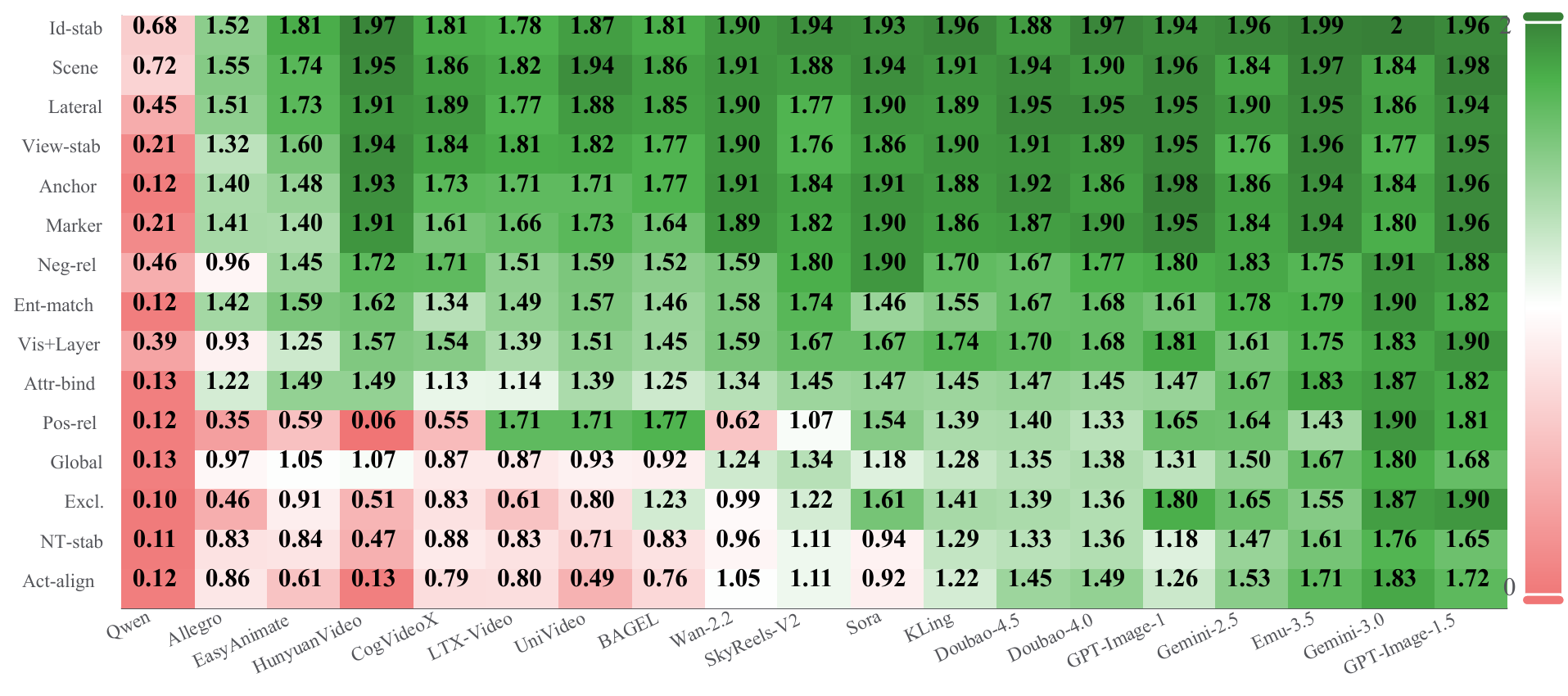}
\caption{Modal--Space consistency on CoW-Bench, shown as a heatmap over sub-metrics (rows) and models (columns), with scores on the 0--2 scale (higher is better). Rows group into \textbf{Sem-Planar} (\texttt{Ent-match}, \texttt{Act-align}, \texttt{NT-stab}, \texttt{Attr-bind}, \texttt{Global}), \textbf{Sem-Hier} (\texttt{Pos-rel}, \texttt{Neg-rel}, \texttt{Excl.}, \texttt{Vis+Layer}, \texttt{Id-stab}), and \textbf{Sem-MV} (\texttt{Anchor}, \texttt{View-stab}, \texttt{Lateral}, \texttt{Scene}, \texttt{Marker}).}
\label{fig:ms_heatmap}
\end{figure*}

Figure~\ref{fig:ms_heatmap} visualizes Modal--Space consistency, where models must map language constraints (entities, attributes, relations) into executable spatial roles and keep them verifiable under layout and viewpoint variation. The figure supports CoW-Bench’s central goal: distinguishing looks plausible from constraint-faithful semantic grounding in space.

\noindent\textbf{Semantic role binding is the dominant bottleneck.}
Across many models, the most consistent performance drops appear on \texttt{Act-align} and \texttt{Pos-rel}. This indicates frequent failures to (i) bind an instructed action/relation to the correct entity and (ii) realize a constructive positive spatial relation precisely. Importantly, these failures can coexist with strong scores on geometry-leaning cues (\eg \texttt{Id-stab}, \texttt{Scene}), producing a characteristic ``plausible-but-misbound'' outcome: the scene is coherent, yet the constraint is attached to the wrong object or only weakly reflected.

\noindent\textbf{Avoiding violations is often easier than constructing exact relations.}
For a broad set of mid-tier models, \texttt{Neg-rel} and \texttt{Excl.} are noticeably stronger than \texttt{Pos-rel}. This asymmetry suggests that models more reliably avoid forbidden configurations than they enforce an exact required placement. For world-model use, this gap matters because planning and verification rely on constructive satisfaction (placing the right entity in the right role), not only on the absence of obvious violations.

\noindent\textbf{Multi-view semantic stability is strong for top models but still reveals tail-risk failures.}
The \texttt{Sem-MV} block (\texttt{Anchor}, \texttt{View-stab}, \texttt{Lateral}, \texttt{Scene}, \texttt{Marker}) is generally high for leading closed-source image models and competitive systems, indicating that stable reference frames and identity markers under viewpoint change are increasingly attainable. However, the heatmap also shows that weaker models can fail catastrophically on these anchors, which makes multi-view stability a sensitive probe of whether a model maintains an invariant scene state rather than redrawing a new world per view.

\begin{boxA}
\noindent{\textbf{Takeaway.} When handling 3D perspective issues, the model also tends to generate mirror-symmetrical results. Additionally, during perspective transitions, non-primary content may retain its original viewpoint, leading to partial object perspective shifts. } 
\end{boxA}

\subsubsection{Modal--Time Consistency Results: Executing a Temporal Program}
\label{sec:modal_time_consistency_results}

\begin{figure*}[h]
\centering
\includegraphics[width=0.98\textwidth]{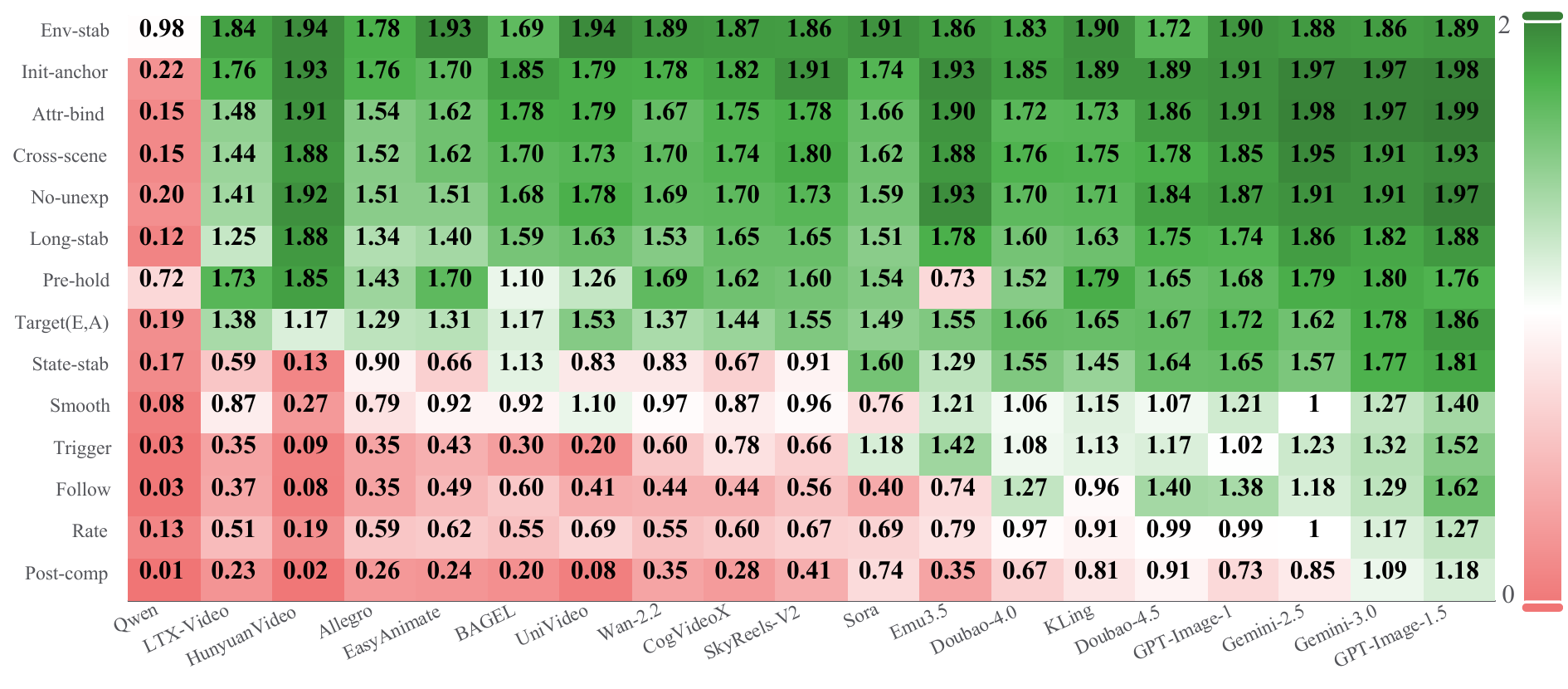}
\caption{Modal--Time consistency on CoW-Bench, shown as a heatmap over sub-metrics (rows) and models (columns), with scores on the 0--2 scale (higher is better). Rows cover \textbf{Long-Horizon} anchoring (\texttt{Init-anchor}, \texttt{Long-stab}, \texttt{Cross-scene}, \texttt{Attr-bind}, \texttt{No-unexp}), \textbf{Attr-Dyn} alignment (\texttt{Target(E,A)}, \texttt{Follow}, \texttt{Smooth}, \texttt{Rate}, \texttt{Env-stab}), and \textbf{Trigger-Event} compliance (\texttt{Pre-hold}, \texttt{Trigger}, \texttt{Post-comp}, \texttt{State-stab}, \texttt{Env-stab}).}
\label{fig:mt_heatmap}
\end{figure*}

Figure~\ref{fig:mt_heatmap} visualizes Modal--Time consistency, where models must (i) keep language-specified anchors stable over long horizons, (ii) execute attribute dynamics specified by the prompt, and (iii) respond to discrete trigger events without breaking the worldline. The heatmap supports CoW-Bench’s central goal: separating visually plausible temporal outputs from \textit{constraint-faithful} temporal execution.

\noindent\textbf{Anchoring is generally strong; the main variance concentrates in the weakest systems.}
The Long-Horizon block is consistently high for leading closed-source image models and remains competitive for many video generators, indicating that persistent identity/attribute anchoring is often attainable once the anchor is observable. The most salient failures appear as isolated low-score columns (\eg very low \texttt{Init-anchor}/\texttt{Long-stab} in the weakest model), which then correlate with downstream temporal-control breakdown.

\noindent\textbf{Dynamics attribute is the primary bottleneck, dominated by instruction-following and rate control.}
Within Attr-Dyn, \texttt{Env-stab} stays near the upper range for most models, while \texttt{Follow} and \texttt{Rate} remain substantially lower—especially for video generators. This pattern indicates a common failure mode: models keep the scene stable but do not execute the instructed evolution reliably (direction/schedule/pacing), yielding sequences that look smooth yet violate semantic temporal commitments.

\noindent\textbf{Triggered events expose timing and post-event persistence failures.}
In Trigger-Event, \texttt{Pre-hold} is often relatively strong, but \texttt{Trigger} and \texttt{Post-comp} degrade noticeably for many video models, revealing two coupled issues: the event is not made salient at the correct time, and the post-trigger state does not persist. These errors are particularly damaging for planning-style use, where discrete events serve as causal checkpoints.

\begin{boxA}
\noindent{\textbf{Takeaway.}  video generation models are more prone to introducing state conditions not specified in text constraints, leading to chaotic variations. In contrast, image generation models demonstrate higher compliance with textual instruction constraints.} 
\end{boxA}

\subsubsection{Time-Space Consistency Results: Navigation Exposes the Missing World State}
\label{sec:time_space_consistency_results}

\begin{figure*}[ht]
\centering
\includegraphics[width=0.98\textwidth]{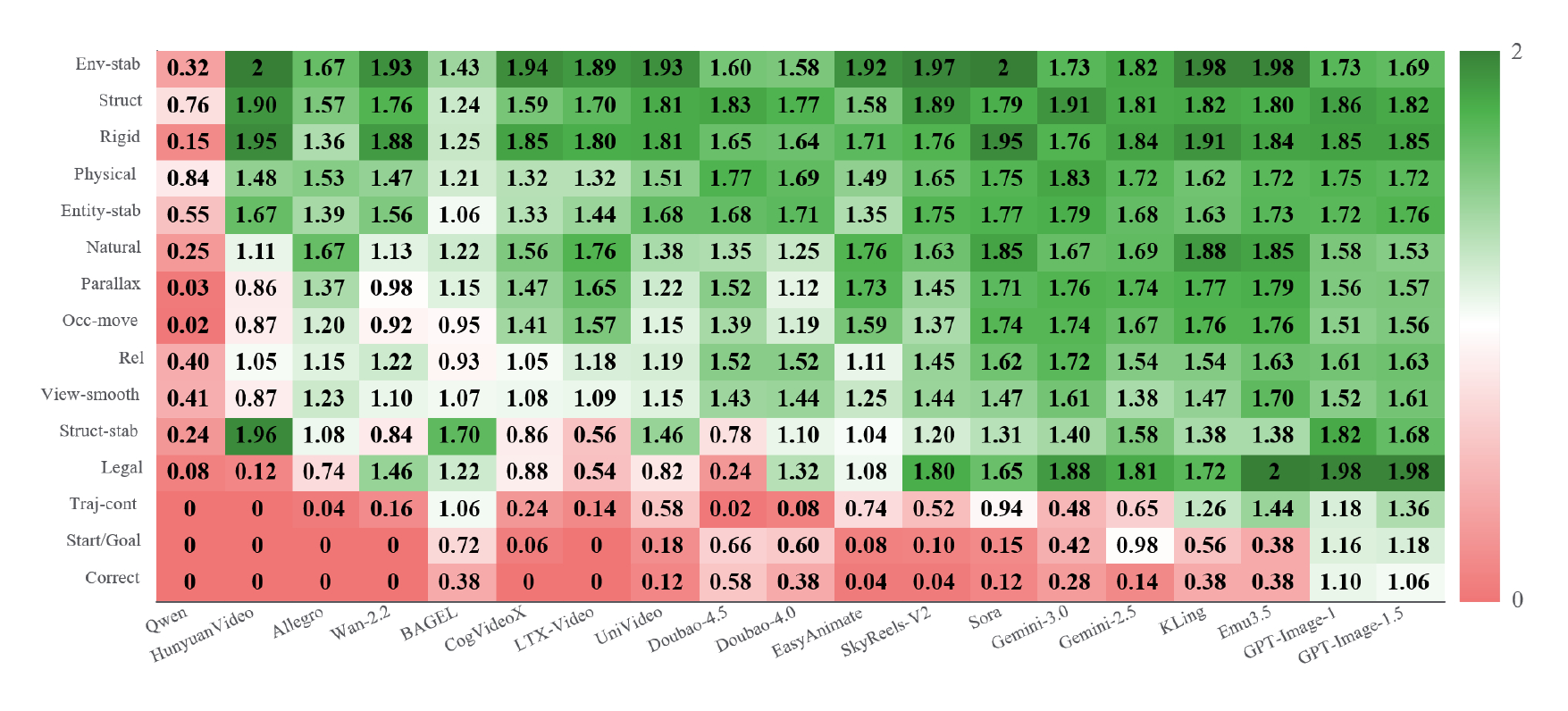}
\caption{Time-Space consistency on CoW-Bench, shown as a heatmap over sub-metrics (rows) and models (columns), with scores on the 0--2 scale (higher is better). Rows cover three metric families: \textbf{ST1} Maze-2D (\texttt{Start/Goal}, \texttt{Traj-cont}, \texttt{Legal}, \texttt{Correct}, \texttt{Struct-stab}), \textbf{ST2} Occlusion Dynamics under Motion (\texttt{Occ-move}, \texttt{Parallax}, \texttt{Rigid}, \texttt{Natural}, \texttt{Env-stab}), and \textbf{ST3} 3D Loop Navigation (\texttt{Struct}, \texttt{Rel}, \texttt{View-smooth}, \texttt{Physical}, \texttt{Entity-stab}).}
\label{fig:ts_heatmap}
\end{figure*}

Time-Space consistency evaluates whether a model maintains an invariant spatial structure while executing temporally extended motion. Figure~\ref{fig:ts_heatmap} visualizes performance across ST1--ST3 and highlights a central message consistent with CoW-Bench’s motivation: models can achieve strong \textit{local} motion plausibility and even stable environments, yet fail when the task requires a persistent, goal-directed world state.

\noindent\textbf{(1) Maze-2D remains the sharpest discriminator.}
In ST1, many video generators score non-trivially on \texttt{Legal} and sometimes on \texttt{Struct-stab}, but still fall to near zero on \texttt{Start/Goal} and \texttt{Correct} (\eg Sora and Kling show low \texttt{Correct} despite moderate \texttt{Legal}). This pattern indicates that the core failure is not producing a plausible maze-like motion, but maintaining a single, identifiable trajectory that starts from the correct anchor and reaches the correct goal without implicit resets or shortcutting. By contrast, stronger image-centric models obtain substantially higher ST1 correctness, suggesting that explicit goal-conditioned state tracking is still the limiting factor for video-style generation.

\noindent\textbf{(2) Occlusion-under-motion is comparatively mature, with remaining errors concentrated on depth-layer updates.}
For ST2, many models achieve high \texttt{Rigid}, \texttt{Natural}, and \texttt{Env-stab}, which suggests that layered motion and global temporal stability are increasingly handled well. The remaining spread concentrates on \texttt{Occ-move} and \texttt{Parallax}, implying that the hardest cases involve consistent depth ordering and visibility updates under motion rather than overall smoothness.

\noindent\textbf{(3) 3D loop navigation stresses viewpoint continuity and relational stability.}
In ST3, leading models maintain strong \texttt{Struct} and \texttt{Entity-stab}, but weaker systems drop on \texttt{View-smooth} and \texttt{Rel}. This is consistent with a ``viewpoint reset” failure mode: geometry can look plausible frame-by-frame, yet the sequence cannot be explained as a single 3D scene traversed along a continuous camera path. CoW-Bench therefore treats loop navigation as a probe of whether models preserve state under transformation, beyond single-view realism.

\begin{boxA}
\noindent{\textbf{Takeaway.} video generation models often yield results with greater freedom and a more pronounced sense of temporal progression, whereas image generation models tend to depict changes over time in a more disjointed manner, with less scope for creative expression.} 
\end{boxA}

\subsection{Sample Analysis}
\begin{figure*}[t]
\centering
\includegraphics[width=0.98\textwidth]{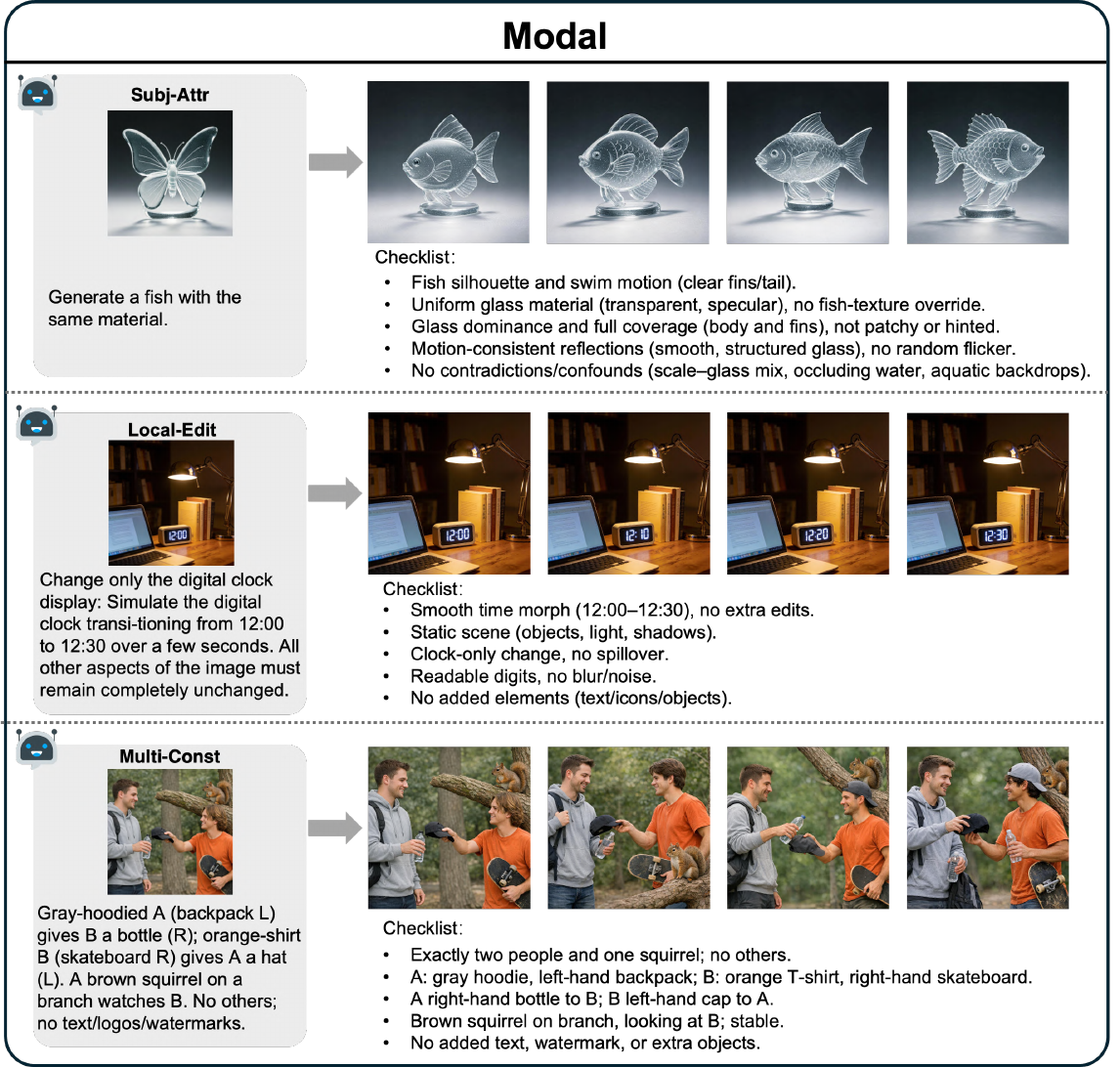}
\caption{Example diagrams of Modal sub-tasks.}
\label{fig:one_analysis-1}
\end{figure*}

\begin{figure*}[t]
\centering
\includegraphics[width=0.98\textwidth]{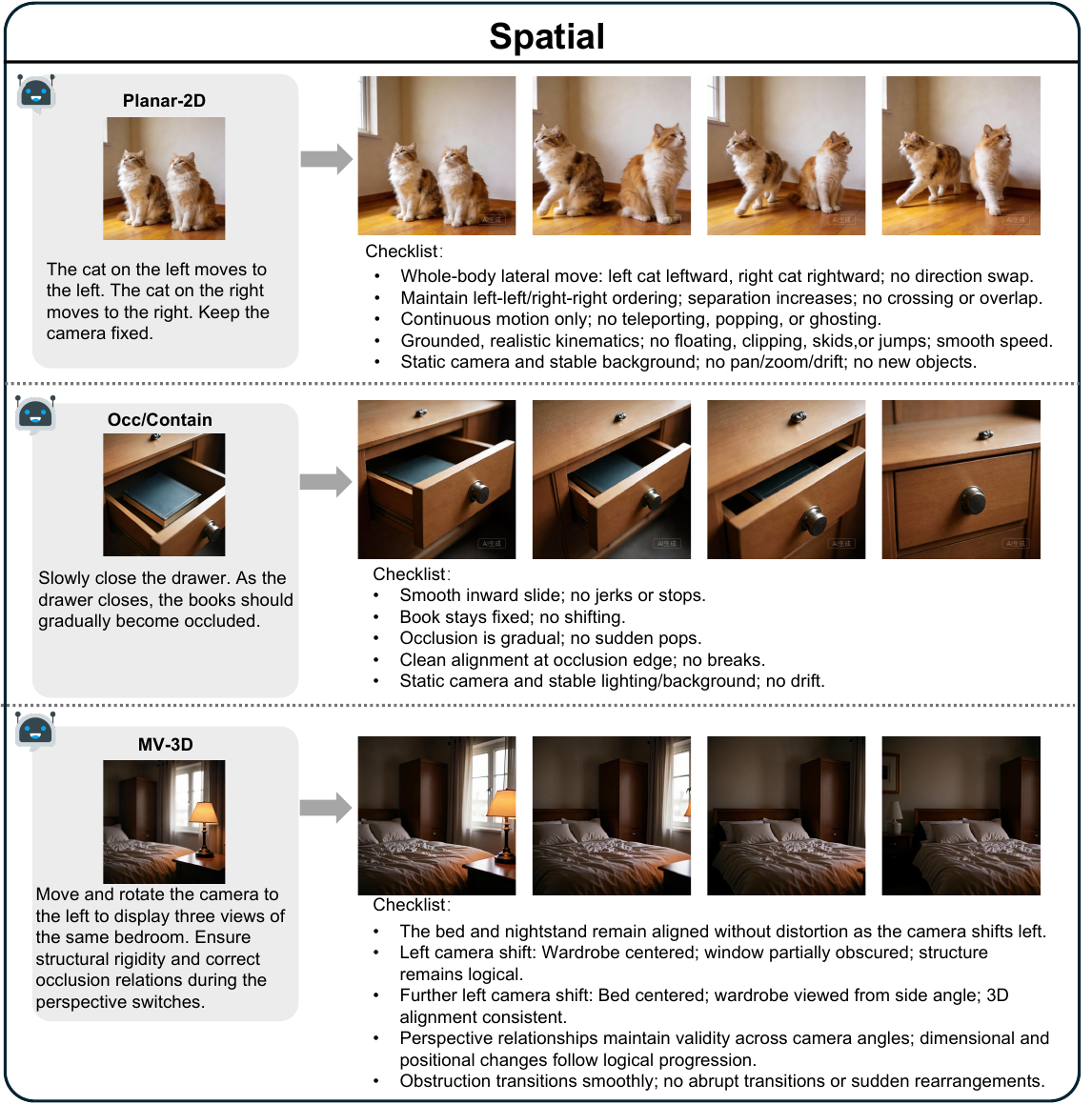}
\caption{Example diagrams of Spatial sub-tasks.}
\label{fig:one_analysis-2}
\end{figure*}

\begin{figure*}[t]
\centering
\includegraphics[width=0.98\textwidth]{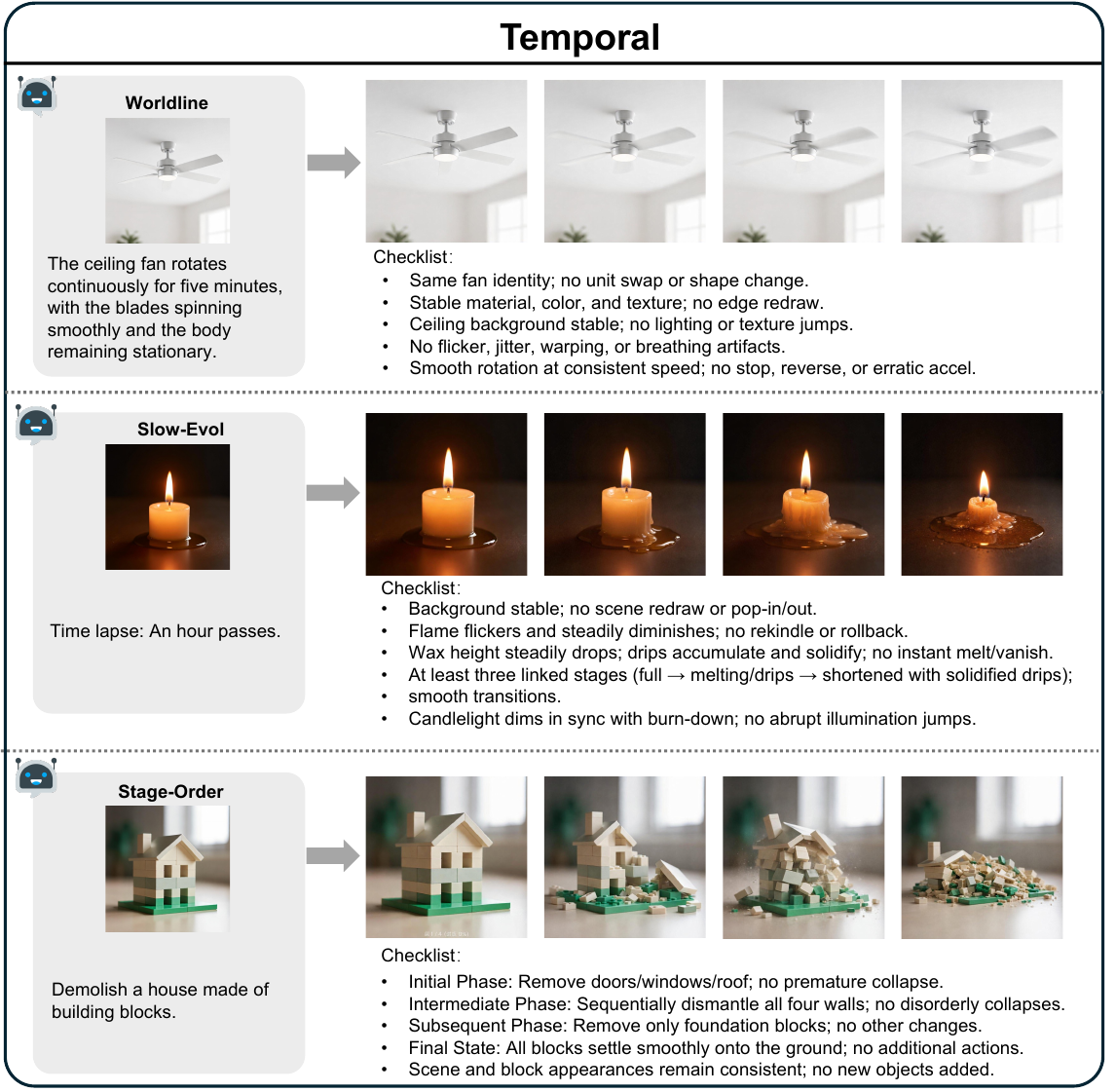}
\caption{Example diagrams of Temporal sub-tasks.}
\label{fig:one_analysis-3}
\end{figure*}

\begin{figure*}[t]
\centering
\includegraphics[width=0.98\textwidth]{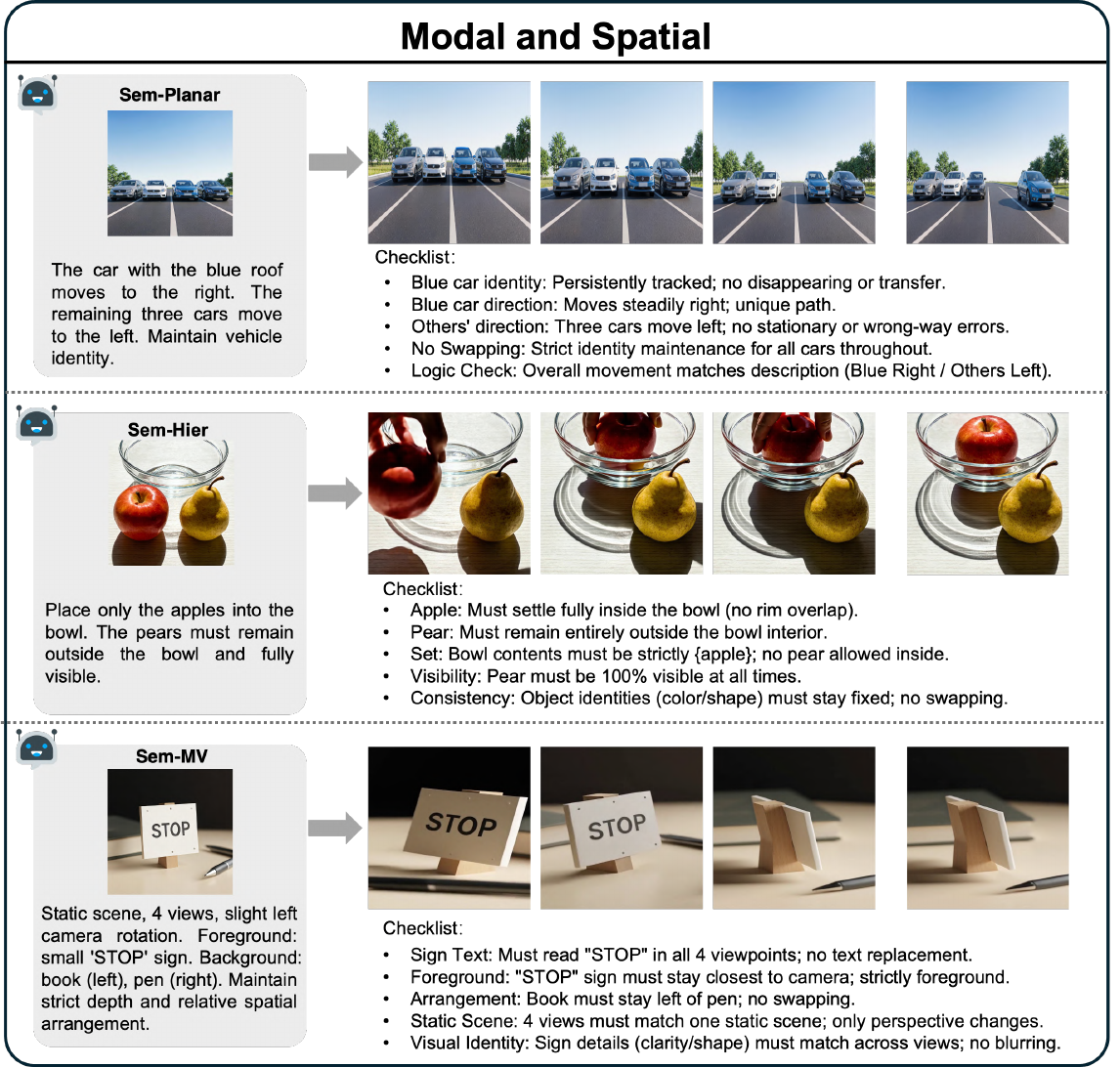}
\caption{Example diagrams of modal and spatial sub-tasks.}
\label{fig:MS_sample}
\end{figure*}

\begin{figure*}[t]
\centering
\includegraphics[width=0.98\textwidth]{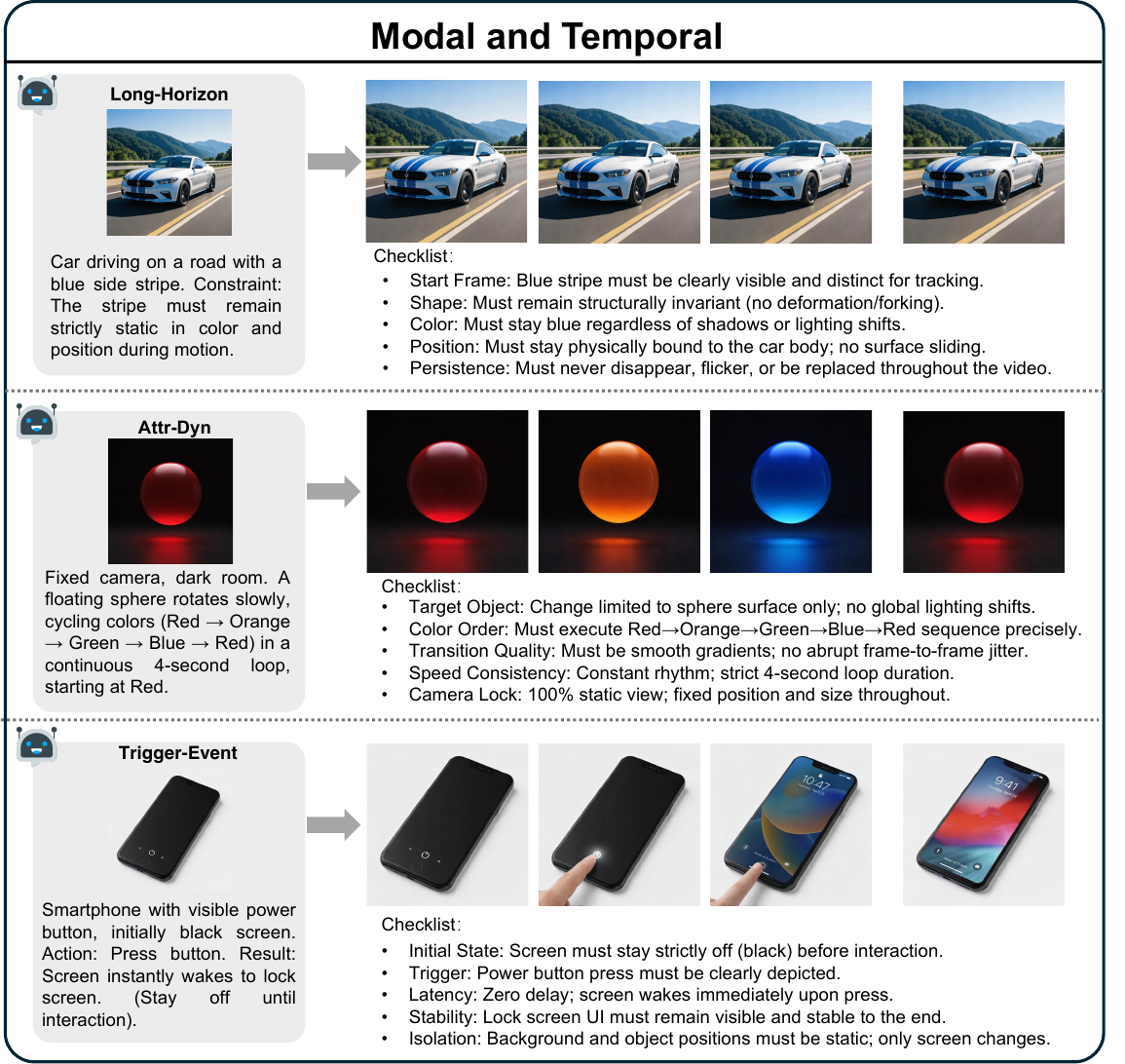}
\caption{Example diagrams of modal and temporal sub-tasks.}
\label{fig:MT_sample}
\end{figure*}

\begin{figure*}[t]
\centering
\includegraphics[width=0.98\textwidth]{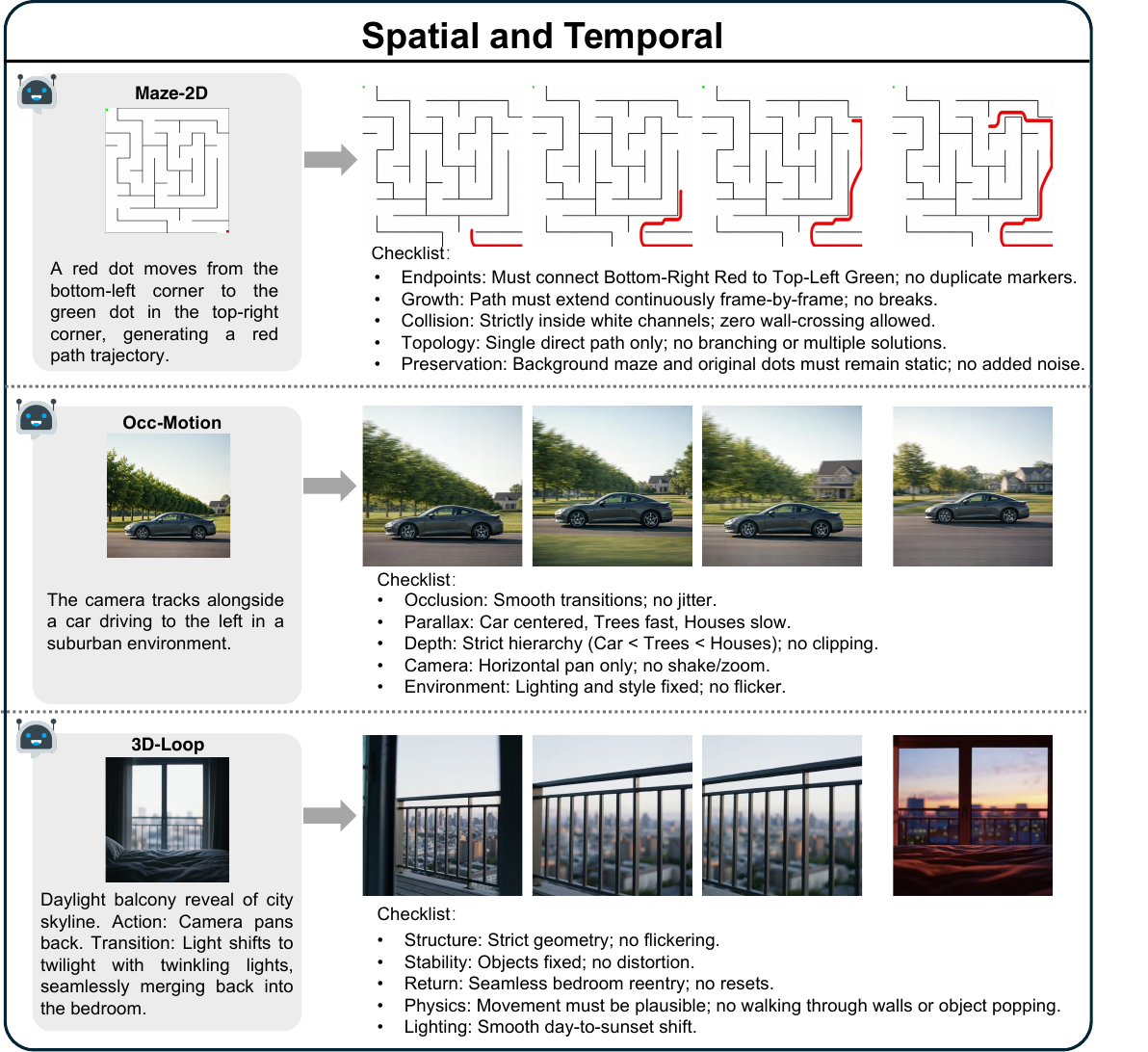}
\caption{Example diagrams of spatial and temporal sub-tasks.}
\label{fig:ST_sample}
\end{figure*}

% \begin{figure*}[t]
% \centering
% \includegraphics[width=0.9\textwidth]{images/Sample Analysis/two_consistency.pdf}
% \caption{Example diagrams of Compound Consistency sub-tasks.}
% \label{fig:two_analysis}
% \end{figure*}

\begin{figure*}[t]
\centering
\includegraphics[width=0.9\textwidth]{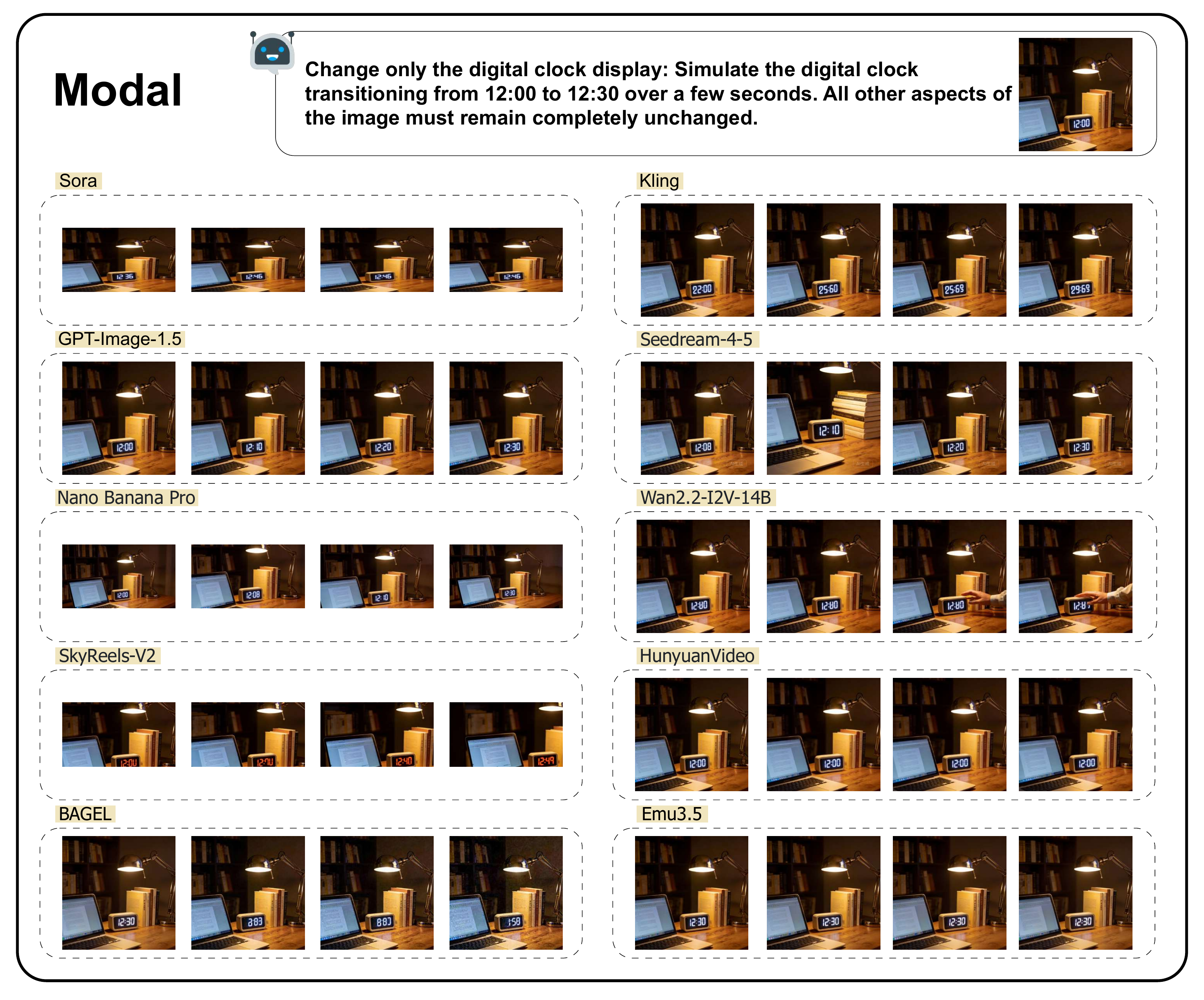}
\caption{Comparison with different models for modal consistency task.}
\label{fig:m_analysis}
\end{figure*}

\begin{figure*}[t]
\centering 
\includegraphics[width=0.9\textwidth]{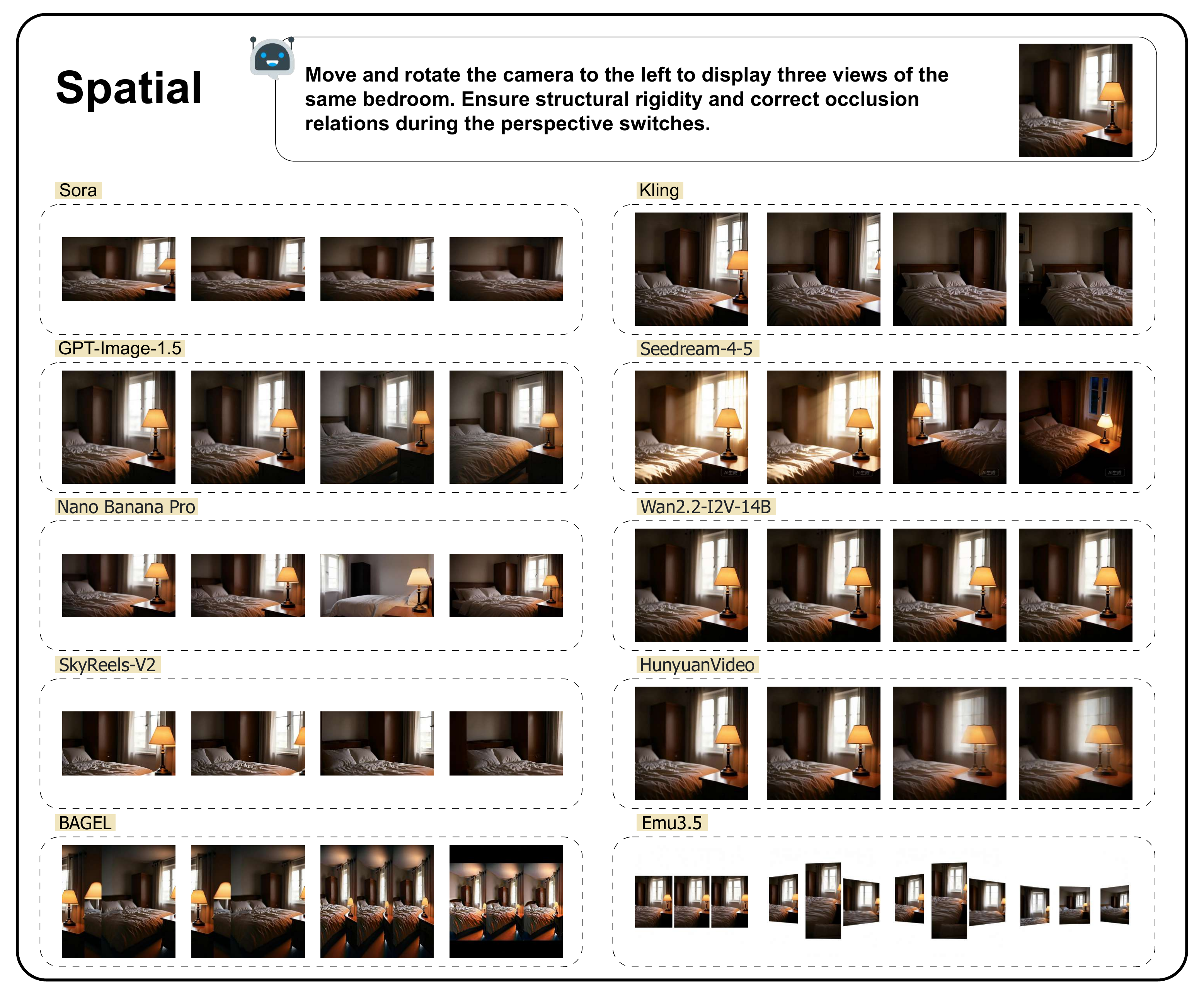}
\caption{Comparison with different models for spatial consistency task.}
\label{fig:s_analysis}
\end{figure*}

\begin{figure*}[t]
\centering
\includegraphics[width=0.9\textwidth]{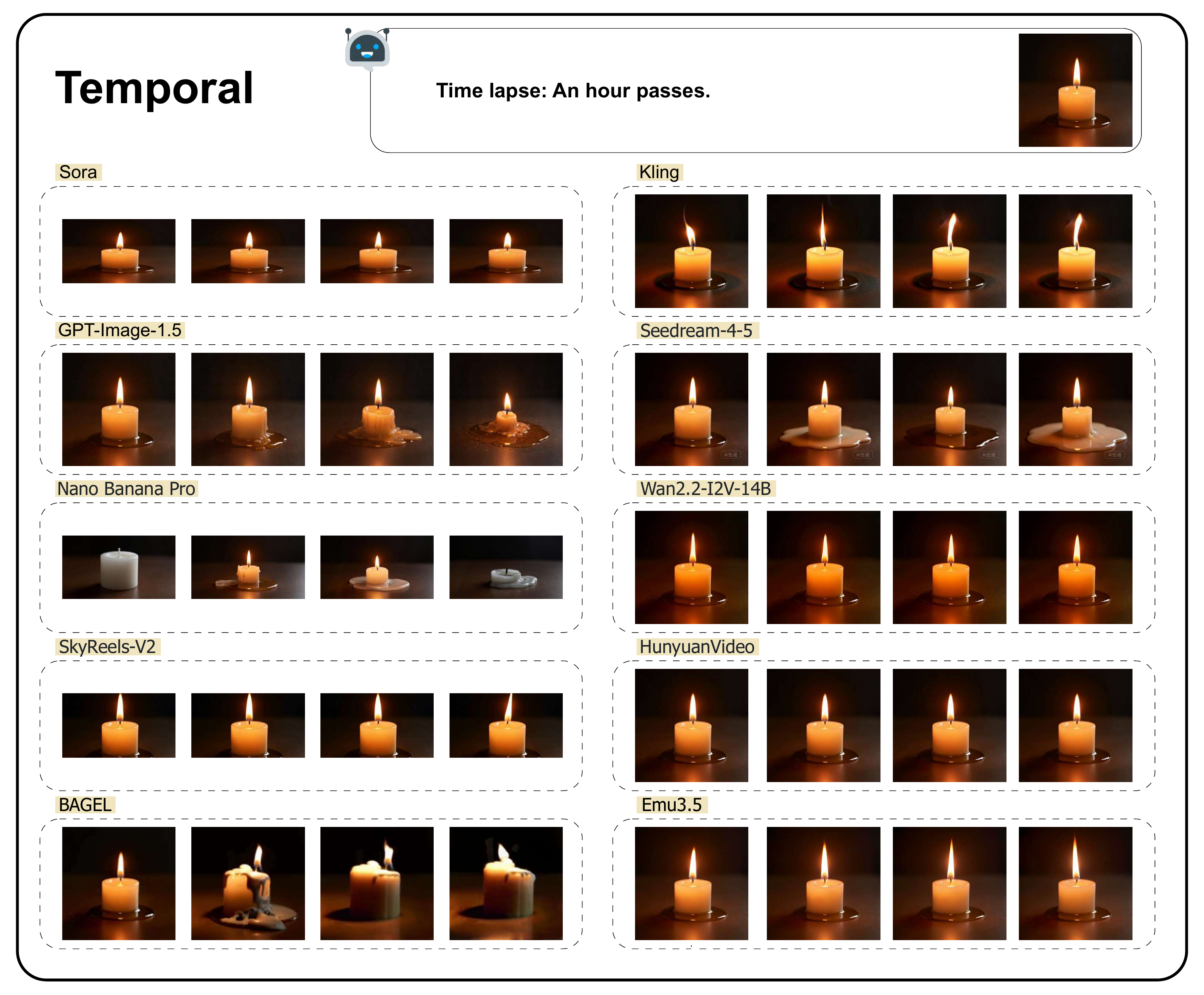}
\caption{Comparison with different models for temporal consistency task.}
\label{fig:t_analysis}
\end{figure*}

\begin{figure*}[t]
\centering
\includegraphics[width=0.9\textwidth]{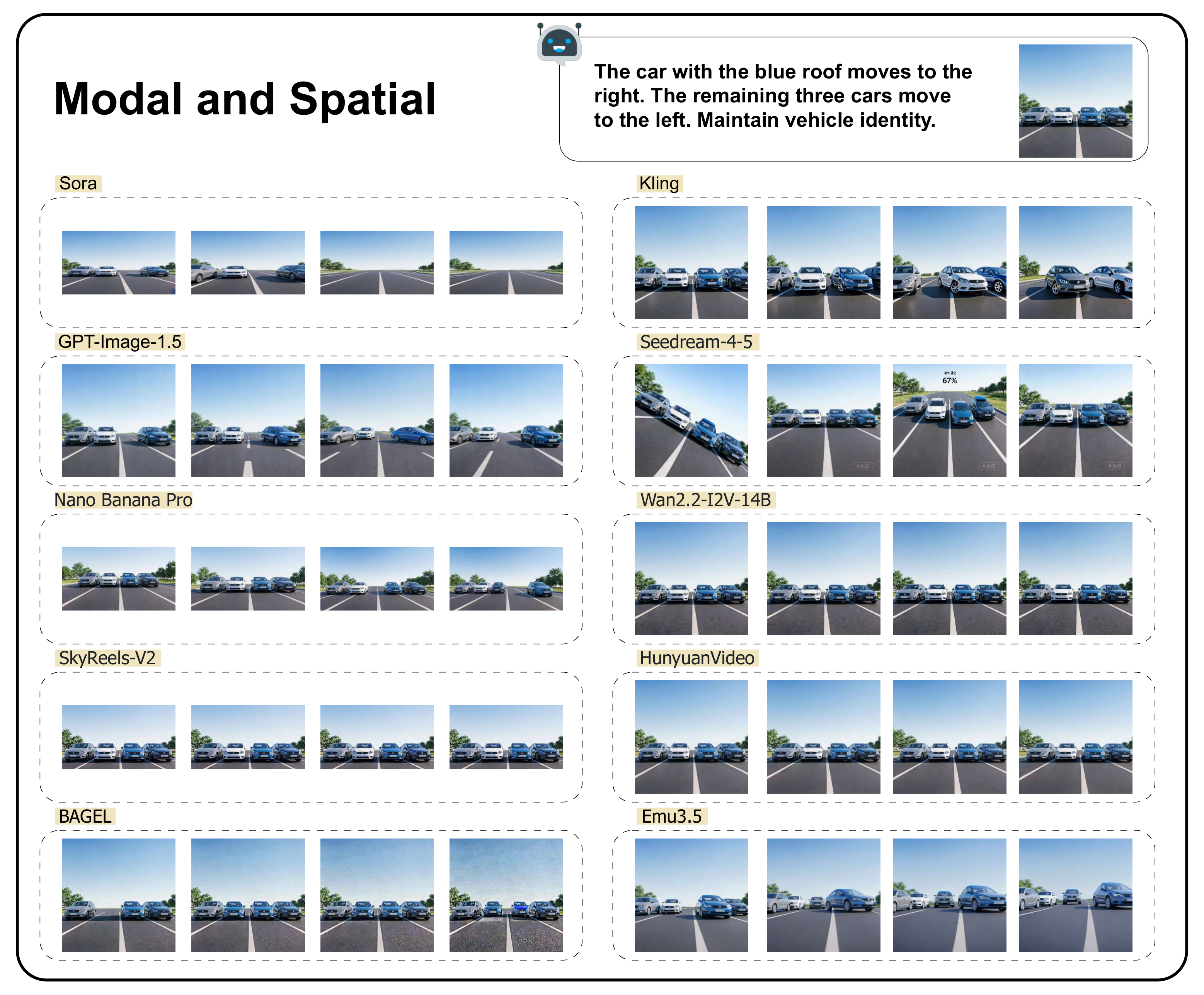}
\caption{Comparison with different models for modal and spatial consistency task.}
\label{fig:ms_analysis}
\end{figure*}

\begin{figure*}[t]
\centering
\includegraphics[width=0.9\textwidth]{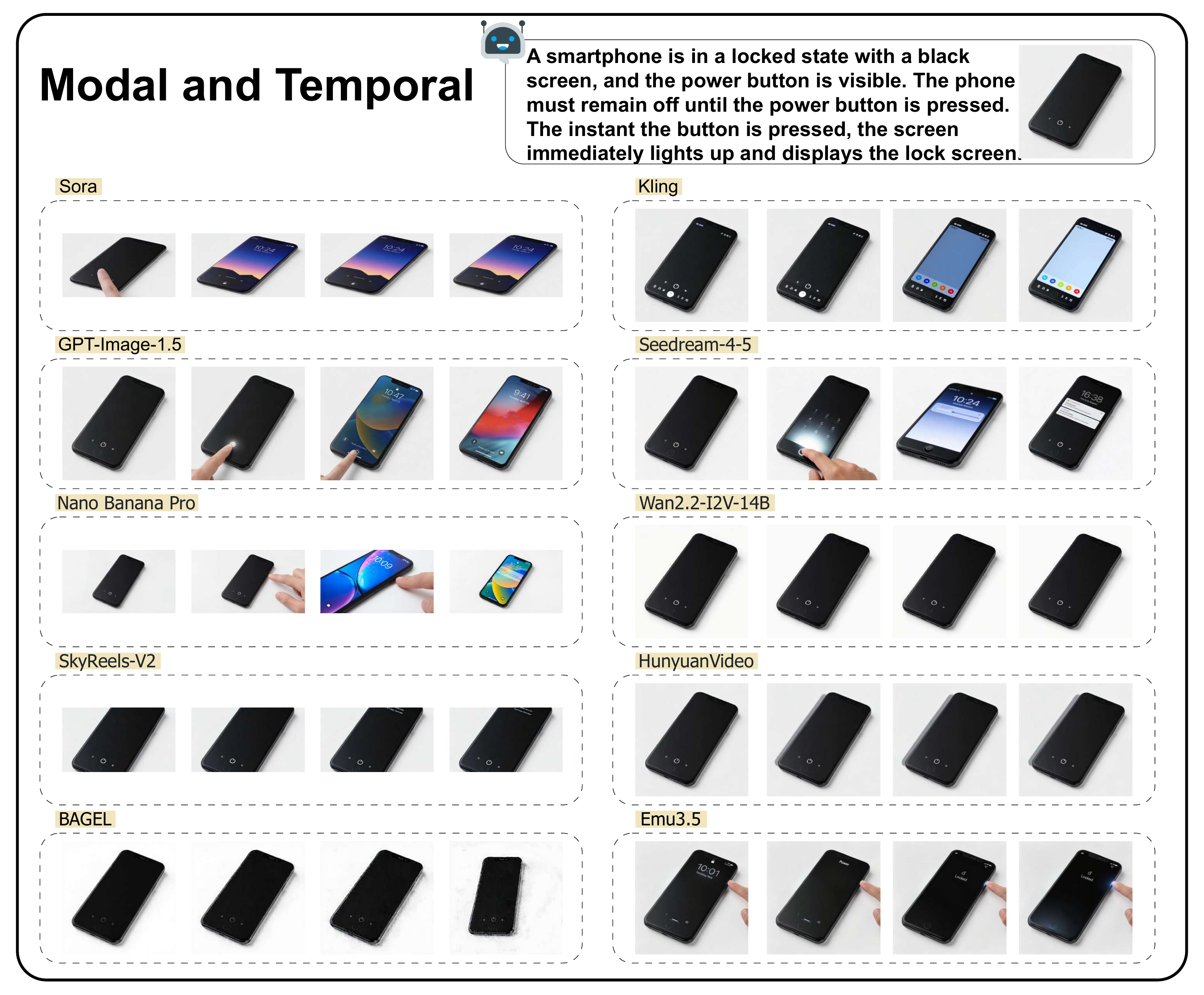}
\caption{Comparison with different models for modal and temporal consistency task.}
\label{fig:mt_analysis}
\end{figure*}

\begin{figure*}[t]
\centering
\includegraphics[width=0.9\textwidth]{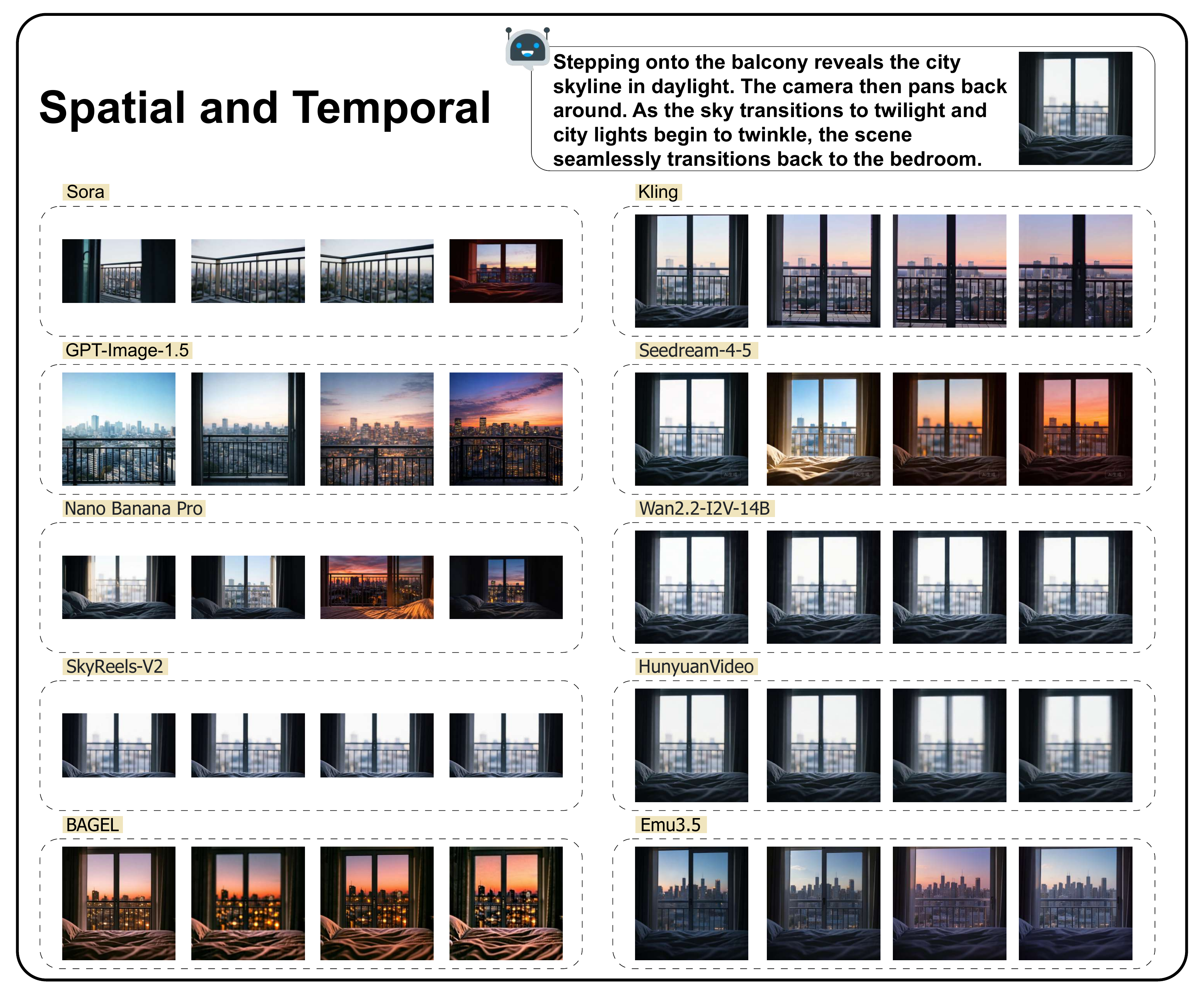}
\caption{Comparison with different models for spatial and temporal consistency task.}
\label{fig:st_analysis}
\end{figure*}

To deeply analyze the model's inference mechanism under multi-dimensional constraints, we have constructed a new set of evaluation criteria in CoW-Bench based on the six core consistency challenges defined previously. Moving beyond traditional evaluations that focus solely on visual quality, this benchmark utilizes frame-by-frame Physical State Ground Truth to precisely quantify the fundamental differences between a "Generator" and a "World Simulator" from the perspectives of dataset construction and model capability boundaries.

\subsubsection{Single Consistency Tasks}

The design philosophy of single consistency tasks is to isolate complex interference and use the purity of simulated data to establish the baseline for a model's foundational reasoning. The specific effects of each sub-task are illustrated in Figures~\ref{fig:one_analysis-1}--\ref{fig:one_analysis-3}.

\paragraph{Modal Consistency Tasks.} This task examines whether a model can clearly distinguish between constraints from different modalities and avoid information blending. In \textit{Subject Attribute Fidelity}, models demonstrate strong feature decoupling capabilities, successfully extracting the texture and material of a "butterfly" from a reference image and mapping it onto the geometric structure of a "fish." The resulting creature possesses distinct scale and luster features without incorporating the butterfly's wing morphology. For finer-grained \textit{Local Edit Precision}, models exhibit precise pixel control in a clock-editing task, modifying only the hour and minute hands according to instructions while the background wall and clock frame remain strictly "locked." Furthermore, under the complex instructions of \textit{Multi-constraint Satisfaction}, models accurately capture the clothing, actions, and positional attributes of multiple characters without incorrect role assignment or attribute leakage, proving their precision in parsing long-text constraints.

\paragraph{Spatial Consistency Tasks.} This task evaluates whether the scenes established by the model are geometrically self-consistent rather than merely "appearing plausible" in 2D images. In \textit{Sem-Planar}, models correctly understand relative positions in non-occluded scenarios, with two cats moving to the left and right respectively without confusing the directional semantics. For more complex \textit{Occlusion/Containment}, as a drawer slowly closes, the model correctly renders the process of internal books gradually moving into darkness; the books follow physical laws of occlusion rather than disappearing abruptly, reflecting an understanding of the "container" concept. In tests of \textit{MV-3D}, as the viewpoint slowly shifts, the desk lamp naturally disappears at the edge of the field of view while the bed layout gradually reveals itself. Throughout this process, the relative positions of objects in the room remain unchanged and the lighting environment stays stable.
% , with no sudden geometric mutations.

\paragraph{Temporal Consistency Tasks.}
We elevate temporal consistency from "visual smoothness" to "rule-governed evolution," examining whether models follow the world's implicit laws. In \textit{Worldline Persistence}, an electric fan maintains the physical integrity of its blades during long-term rotation, with no blade breakage or sudden material mutations. \textit{Rule-guided Slow Evolution} further demonstrates the model's understanding of physical entropy: in a simulated one-hour duration, a candle gradually shortens according to combustion laws rather than staying the same in violation of common sense. In the house-collapse task of \textit{Ordered Stage Transitions}, the model clearly displays a continuous state from structural integrity to ruins. The collapse sequence follows gravitational logic, and the ruins remain static after falling, avoiding non-causal jittering such as "collapsing and then recovering."

\subsubsection{Compound Consistency Tasks}

Compound consistency tasks simulate the complexity of the real world, examining the model's trade-off and reasoning capabilities when multi-dimensional constraints restrict (or even conflict with) each other. The specific effects of each sub-task are illustrated in Figures~\ref{fig:MS_sample}--\ref{fig:ST_sample}.

\noindent\textbf{Modal-Spatial Consistency Tasks} evaluate the model's ability to transform semantic information into executable spatial constraints, achieving "semantic--spatial coupling." The model must not only understand what an object is but also precisely execute geometric instructions regarding where the object is, ensuring accurate grounding of semantic referents in the spatial dimension. As shown in Figure ~\ref{fig:MS_sample}, in \textit{Sem-Planar}, models successfully identify a specific vehicle with a "blue roof" within complex traffic flow and control only that vehicle to move right, achieving precise semantic-spatial binding. In the \textit{Sem-Hier} task, models accurately generate a scene where "apples are in the bowl" and "pears are outside," strictly adhering to the spatial semantics of containment and exclusion. However, \textit{Sem-MV} exposes a current weakness: during viewpoint transitions, while the perspective change of a signpost remains reasonable, the occlusion relationship of books and pens relative to the signpost undergoes erroneous drift (moving from behind to the side), indicating that the model's ability to maintain micro-spatial semantics under dynamic viewpoints still requires improvement.

\noindent\textbf{Modal-Temporal Consistency Tasks} evaluate the model's fidelity to instructions during long-sequence generation. The core is to examine whether the model treats the prompt as a high-priority "temporal constitution," implementing semantic elements and logical constraints throughout the entire video to resist semantic drift and forgetting over time. As shown in Figure ~\ref{fig:MT_sample}, in \textit{Long Horizon} tasks, the body color, patterns, and relative positions of a vehicle remain highly stable during long-distance movement, with no blurring or texture alterations. \textit{Attribute Dynamic} further tests temporal programming capabilities, where models successfully control a sphere's color to change in a complex sequence of "red $\to$ orange $\to$ green $\to$ blue $\to$ red" with clear steps and no color bleeding. In \textit{Trigger Event}, models demonstrate acute capture of causal logic: a phone screen stays black before a button is pressed and lights up instantly only after the action is triggered, aligning exactly to the event's trigger point.

\noindent\textbf{Spatial-Temporal Consistency Tasks} evaluate whether models possess the prototype of a "built-in physical engine" under weak modality constraints. It focuses on whether the model can maintain the self-consistency of spatial topology and motion parallax during dynamic evolution, rather than relying solely on pixel-level smooth interpolation. As shown in Figure~\ref{fig:ST_sample}, in \textit{Maze-2D}, although models maintain the static structure of the maze walls, the subject ultimately fails to correctly plan a path to the goal, suggesting limitations in spatial reasoning. In contrast, \textit{Occlusion Dynamics under Motion} perfectly reproduces motion parallax: near trees move at high speed to create motion blur, while the background moves slowly and the vehicle remains relatively stationary, achieving dual spatio-temporal self-consistency. Finally, \textit{3D Loop Navigation} achieves a closed-loop roaming from a bedroom to a city and back, with smooth structural continuity and no spatial collapse, demonstrating potential reference frame stability over long-term roaming.

To comprehensively verify these mechanisms, we used CoW-Bench to test mainstream world models including Sora~\cite{openai2024sora}, Kling~\cite{kling}, GPT-Image-1.5~\cite{openai_gpt_image_1_5_2025}, Seedream-4-5~\cite{seedream2025seedream}, Nano Banana Pro~\cite{comanici2025gemini}, Wan2.2-I2V-14B~\cite{wan2025} , SkyReels-V2~\cite{li2026skyreels}, HunyuanVideo~\cite{hunyuan2025}, BAGEL~\cite{deng2025emerging}, and Emu3.5~\cite{cui2025emu3}. We display selected results: single consistency comparisons are shown in Figure~\ref{fig:m_analysis} (Modal), Figure~\ref{fig:s_analysis} (Spatial), and Figure~\ref{fig:t_analysis} (Temporal); compound consistency results are shown in Figure~\ref{fig:ms_analysis} (Modal-Spatial), Figure~\ref{fig:mt_analysis} (Modal-Temporal), and Figure~\ref{fig:st_analysis} (Spatial-Temporal).

\begin{boxA}
\noindent\textbf{Takeaway.} 
CoW-Bench establishes a new standard for evaluating world models. While video models outperform image models in instruction following, they fundamentally lack physical evolutionary logic. Heavily relying on pixel-based interpolation rather than genuine reasoning, these models exhibit non-physical distortions and topological collapse.
\end{boxA}

\section{Conclusion}

\label{sec:conclusion}

This survey has re-examined the trajectory of generative AI through the lens of the \textit{Trinity of Consistency}, establishing a general framework for what constitutes a World Model. By deconstructing the capability space into Modality, Spatial, and Temporal dimensions, we argue that true physical understanding does not emerge from single-axis performance but from the robustness of cross-dimensional interactions. Our analysis highlights that the most critical failures in current systems are not visual artifacts, but ruptures in consistency: the inability to bind semantic instructions to geometric roles (Modal-Space), the failure to maintain identity under long-horizon evolution (Modal-Time), and the loss of environmental permanence during navigation (Time-Space).

\begin{figure}[H]
  \centering
  \includegraphics[width=0.98\textwidth]{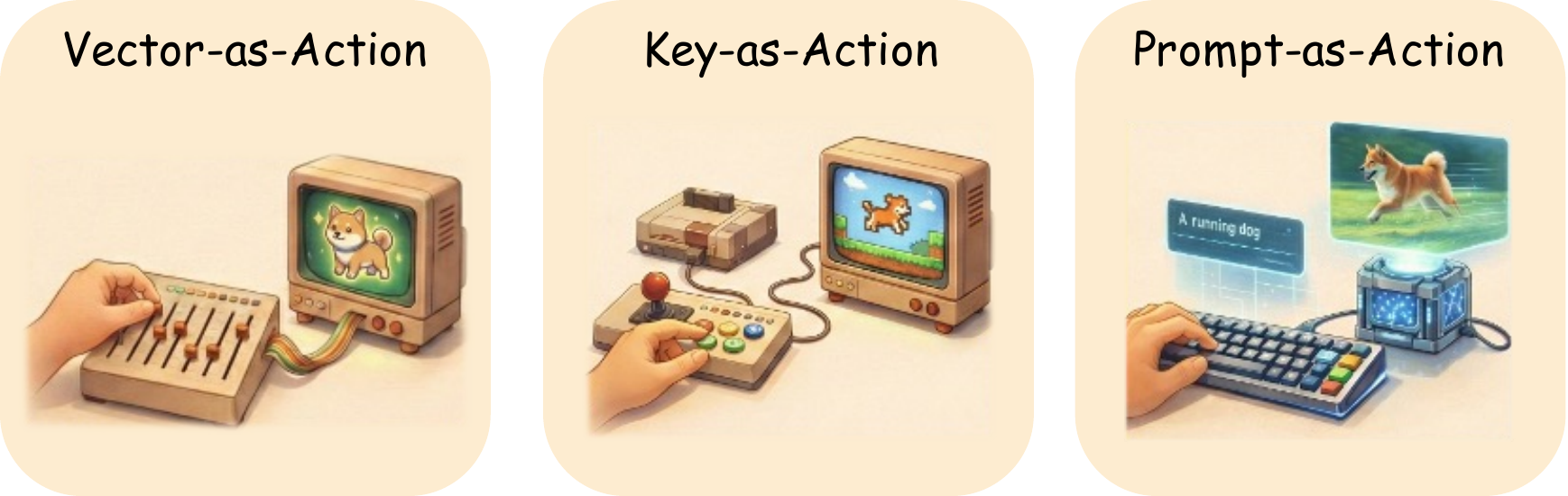}
  \caption{Evolutionary spectrum of World Model paradigms based on interactive action spaces. This figure illustrates the field's trajectory from the early \textbf{``Vector-as-Action''} paradigm (left: \eg JEPA, relying on uninterpretable latent space predictions), through the intermediate \textbf{``Key-as-Action''} paradigm (center: \eg Genie series, constrained by predefined discrete control spaces), and ultimately advancing towards the \textbf{``Prompt-as-Action''} paradigm. In the latter, a semantic compiler translates natural language intents into universal spatiotemporal dynamic simulations.}
  \vspace{-4mm}
  \label{fig:wm_evolution_paradigm}
\end{figure}

To rigorously diagnose these cross-dimensional ruptures, we introduced CoW-Bench, a comprehensive benchmark that unifies the evaluation of mainstream video generation models and UMMs under a shared protocol. CoW-Bench employs a carefully designed multi-frame evaluation protocol derived from human expert reasoning. By analyzing temporally sampled grids against fine-grained atomic checklists, we operationalize consistency as a strict constraint-satisfaction problem. This rigorous approach exposes a pervasive \textit{constraint-backoff} phenomenon, where models generate plausible-looking textures while silently violating logical commitments, thus providing the necessary diagnostic resolution to distinguish between visual mimicry and genuine physical simulation.

Crucially, our findings indicate that constraint backoff is not merely a consequence of insufficient training data or scale, but a structural artifact of how current models represent interaction. When the action space is either uninterpretable or rigidly predefined, models lack the expressive capacity to ground semantic commitments in physical dynamics. Under such constraints, consistency violations become not accidental errors but almost inevitable outcomes. Addressing this limitation therefore demands a paradigm shift in how interaction itself is formalized within world models.

To systematically characterize this transition, we organize the evolution of world model paradigms according to the expressiveness of their interactive action spaces (Figure~\ref{fig:wm_evolution_paradigm}). As illustrated on the left side of the figure, early explorations such as JEPA~\cite{lecun2022path} operate at the \textit{Vector-as-Action} level. While enabling latent-space prediction, their interaction mechanisms remain opaque and lack semantic interpretability. The middle section presents the \textit{Key-as-Action} paradigm, exemplified by the Genie series~\cite{bruce2024genie,parker2024genie,deepmind2025genie3}. Although introducing limited interactivity, these models remain confined to narrow, discrete, and predefined action spaces.

The right side of the figure illustrates a forward-looking paradigm: a \textit{Prompt-as-Action} paradigm in which UMMs with modality consistency and video generation models with spatial–temporal consistency are unified. Equipped with an internal semantic compiler, such models can interpret high-dimensional natural-language prompts and translate them into universal spatiotemporal simulations that adhere to the \textit{Trinity of Consistency}. Recent systems such as PixVerse-R1~\cite{pixverseR1_2026} offer an early glimpse of this direction, demonstrating real-time world modeling that responds instantly to user input and unifies multiple modalities within an autoregressive architecture. By moving beyond predefined action abstractions, this paradigm begins to bridge the gap between human semantic intent and the underlying dynamics of the physical world.

The central conviction of this survey is therefore simple yet uncompromising: consistency is not an optional attribute of a world model—it is its criterion of existence. A system that produces visually compelling pixels but fails to maintain cross-dimensional consistency, regardless of scale, remains fundamentally a texture synthesizer rather than a simulator of the world. \textit{The Trinity of Consistency} thus delineates more than an analytical framework; it marks a boundary—a paradigmatic divide between generating images that resemble the world and constructing models that understand it.

\clearpage
\section{Contributions}

\textbf{Leading Authors}

Jingxuan Wei$^2$, Siyuan Li$^3$, Cheng Tan$^1$

\textbf{Core Contributors}

Yuhang Xu$^2$, Zheng Sun$^2$, Junjie Jiang$^2$, Hexuan Jin$^2$, Caijun Jia$^2$, Honghao He$^2$, Xinglong Xu$^2$, Xi Bai$^2$

\textbf{Other Contributors}

Chang Yu$^3$, Yumou Liu$^5$, Junnan Zhu$^2$, Xuanhe Zhou$^5$, Jintao Chen$^6$, Xiaobin Hu$^4$, Shancheng Pang$^7$, Bihui Yu$^2$, Ran He$^2$, Zhen Lei$^2$, Stan Z. Li$^3$,

\textbf{Corresponding Authors}

Conghui He$^1$, Shuicheng Yan$^4$, Cheng Tan$^1$

\subsection*{Affiliation}

$^1$Shanghai Artificial Intelligence Laboratory

$^2$University of Chinese Academy of Sciences

$^3$Westlake University

$^4$National University of Singapore

$^5$Shanghai Jiaotong University

$^6$Zhejiang University

$^7$China University of Petroleum (East China)

% \author[1]{xxx}
% \author[1]{yyy}
% \author[2]{zzz}

% \affiliation[1]{}
% \affiliation[2]{}

% \section*{Declaration of AI Usage}
% ToDo 1 % 采用AI调研文献、写作、润色文字等
% ToDo 2 % 采用AI绘图
% ToDo 3 % 本文已严格检查参考文献的真实性和综述内容的严谨性

\clearpage

% \bibliographystyle{plainnat}
% \bibliographystyle{IEEEtranN} 
% \bibliographystyle{abbrvnat}
% \bibliographystyle{abbrvnat}
% % \setcitestyle{numbers}
% \setcitestyle{numbers}
% \bibliography{ref}
\bibliographystyle{IEEEtran2}
\setcitestyle{numbers}
\bibliography{ref_after_check}
% \printbibliography
\end{document}